\documentclass{latex/ut-thesis}
\usepackage{graphicx} 
\usepackage{booktabs} 
\usepackage{lipsum} 

\usepackage[round]{natbib}
\usepackage{url}
\usepackage{epsfig}
\usepackage{amssymb}
\usepackage[tbtags]{amsmath}
\usepackage{amsthm}
\usepackage{latexsym}
\usepackage{euscript}
\usepackage{graphics,eepic,epic,psfrag}
\usepackage{bm}
\usepackage{dsfont}
\usepackage{algorithm}
\usepackage{wrapfig}
\usepackage{tikz}
\usetikzlibrary{positioning}
\usetikzlibrary{shapes}
\usepackage{listings}
\usepackage{setspace}
\usepackage{caption}
\usepackage{nicefrac}

\usepackage[utf8]{inputenc} 
\usepackage[T1]{fontenc}    
\usepackage{amsfonts}       
\usepackage{microtype}      
\usepackage{algpseudocode}
\usepackage{siunitx}  
\usepackage{commath}  
\usepackage{mathtools}  
\usepackage{drawmatrix}
\usepackage[english]{babel}
\usepackage{pifont}
\usepackage{multirow}
\usepackage{placeins}  

\usepackage[T1]{fontenc}    
\usepackage{subcaption}
\usepackage{xcolor}

\let\origdegree\degree
\let\degree\relax
\usepackage{mathabx}
\let\degree\origdegree

\usepackage[colorlinks, backref=page]{hyperref}       
\renewcommand*{\backref}[1]{}
\renewcommand*{\backrefalt}[4]{{\footnotesize [%
    \ifcase #1 Not cited.%
	\or Cited on page~#2%
	\else Cited on pages #2%
	\fi%
]}}

\hypersetup{
    colorlinks=true,
    linkcolor=blue,
    filecolor=magenta,      
    urlcolor=cyan,
    citecolor=blue,
}

\usepackage{balance} 

\usepackage{xspace}

\usepackage[framemethod=TikZ]{mdframed}
\newmdenv[
  leftline=true,
  topline=false,
  rightline=false,
  bottomline=false,
  linecolor=gray,
  linewidth=4pt,
  innertopmargin=\baselineskip,
  innerbottommargin=\baselineskip,
  innerrightmargin=20pt,
  innerleftmargin=20pt
]{sidedbox}
\definecolor{mydarkblue}{rgb}{0,0.08,0.45}
\hypersetup{colorlinks=true,
 linkcolor=mydarkblue,
 citecolor=mydarkblue,
 filecolor=mydarkblue,
 urlcolor=mydarkblue}

\newcommand{\argmin}{\mathop{\mathrm{argmin}}\limits}

\newtheorem{theorem}{Theorem}[section]

\newtheorem{lemma}[theorem]{Lemma}

\usepackage{amsmath,amsfonts,bm}

\def\eqref#1{equation~\ref{#1}}

\def\1{\bm{1}}

\def\vx{{\bm{x}}}
\def\vy{{\bm{y}}}

\DeclareMathAlphabet{\mathsfit}{\encodingdefault}{\sfdefault}{m}{sl}
\SetMathAlphabet{\mathsfit}{bold}{\encodingdefault}{\sfdefault}{bx}{n}

\definecolor{mydarkblue}{rgb}{0,0.08,0.45}
\definecolor{mydarkgreen}{rgb}{0.0, .4, 0.0}
\definecolor{mybrown}{rgb}{0.59, 0.29, 0.0}
\hypersetup{
    colorlinks=true,
    linkcolor=mydarkblue,
    citecolor=mydarkblue,
    filecolor=mydarkblue,
    urlcolor=mydarkblue}

\newcommand{\pluseq}{\mathrel{+}=}
\newcommand{\mineq}{\mathrel{-}=}
\newtheorem{thm}{Theorem}  
\newtheorem*{thm*}{Theorem}  
\newtheorem*{lemma*}{Lemma}

\newcommand{\defeq}{\vcentcolon=}

\newcommand{\JL}[1]{{\color{purple}JL: #1}}  

\newcommand{\Real}{\mathbb{R}}  
\newcommand{\ep}{\textbf{w}}  
\newcommand{\hparam}{\boldsymbol{\lambda}}
\newcommand{\hp}{\boldsymbol{\lambda}}
\newcommand{\edim}{m}  
\newcommand{\hdim}{n} 
\newcommand{\edom}{\textbf{W}}  
\newcommand{\xdom}{\mathcal{X}}  
\newcommand{\tdom}{\mathcal{Y}}  
\newcommand{\hdom}{\boldsymbol{\Lambda}} 
\newcommand{\loss}{\mathcal{L}}  
\newcommand{\dataset}{\mathcal{D}}  
\newcommand{\Ltr}{\loss_{\text{T}}}  
\newcommand{\Lval}{\loss_{\text{V}}}  

\newcommand{\hpderiv}[1]{\pd{#1}{\hp}}
\newcommand{\epderiv}[1]{\pd{#1}{\ep}}
\newcommand{\ehpderiv}[1]{\md{#1}{2}{\ep}{}{{\hp^{\transpose}}}{}}
\newcommand{\eepderiv}[1]{\md{#1}{2}{\ep}{}{{\ep^{\transpose}}}{}}

\newcommand{\trainGrad}{\epderiv{\Ltr}}
\newcommand{\trainHess}[1]{\eepderiv{\Ltr}}
\newcommand{\trainMixed}[1]{\ehpderiv{\Ltr}}
\newcommand{\response}{\ep^{*}\left(\hp\right)}
\newcommand{\responseJacobian}{\hpderiv{\response}}
\newcommand{\LvalResponse}{\Lval\left(\hp,\response\right)}
\newcommand{\rpm}{\raisebox{.2ex}{$\scriptstyle\pm$}}

\newcommand{\nn}{neural network\xspace} 
\newcommand{\nns}{neural networks\xspace} 
\newcommand{\ho}{hyperparameter optimization\xspace} 
\newcommand{\HO}{Hyperparameter Optimization\xspace} 
\newcommand{\Ho}{Hyperparameter optimization\xspace} 

\newcommand{\iftf}{\trainGrad}
\newcommand{\iftx}{\hparam}
\newcommand{\ifty}{\ep}
\newcommand{\iftg}{\ifty^*}
\newcommand{\ifthdom}{\hdom}
\newcommand{\iftedom}{\edom}
\newcommand{\ifta}{\iftx'}
\newcommand{\iftb}{\ifty'}

\definecolor{DarkGreen}{RGB}{0,100,0}

\newcommand{\argminSide}{\textnormal{arg\,min}}

\newcommand{\new}[1]{#1}

\newcommand{\eparam}{\textbf{w}}  

\newcommand{\outSymbol}{A}
\newcommand{\inSymbol}{B}

\newcommand{\paramSymbol}{\boldsymbol{\theta}}
\newcommand{\outParam}{\paramSymbol_{\outSymbol}}
\newcommand{\inParam}{\paramSymbol_{\inSymbol}}
\newcommand{\bothParam}{\boldsymbol{\omega}}
\newcommand{\outLoss}{\loss_{\outSymbol}}
\newcommand{\inLoss}{\loss_{\inSymbol}}
\newcommand{\gradSymbol}{\boldsymbol{g}}
\newcommand{\bothGrad}{\hat{\gradSymbol}}
\newcommand{\outGrad}{\gradSymbol_{\outSymbol}}
\newcommand{\inGrad}{\gradSymbol_{\inSymbol}}
\newcommand{\momCoeff}{\beta}
\newcommand{\momVal}{\boldsymbol{\mu}}
\newcommand{\lr}{\alpha}

\newcommand{\spectralRadius}{\rho}
\DeclareMathOperator*{\spectrum}{Sp}

\newcommand{\condition}{\kappa}
\newcommand{\identity}{\boldsymbol{I}}
\newcommand{\fixedPointOp}{\boldsymbol{F}_{\lr, \momCoeff}}
\newcommand{\dynamics}{\boldsymbol{J}}
\newcommand{\dimSymbol}{d}
\newcommand{\inDim}{\dimSymbol_{\inSymbol}}
\newcommand{\outDim}{\dimSymbol_{\outSymbol}}
\newcommand{\bothDim}{\dimSymbol}
\newcommand{\eigval}{\lambda}

\newcommand{\transpose}{\top}

\newcommand\newsubcap[1]{\phantomcaption%
       \caption*{\figurename~\thefigure(\thesubfigure): #1}}

\newcommand{\eqspace}{\,\,\,\,\,}

\newcommand{\resultName}{Corollary}

\usepackage{thmtools}
\usepackage{thm-restate}

\newcommand{\fixedPhase}{\frac{\pi}{\num{16}}}

\newcommand{\ganSize}{\num{16}}
\newcommand{\ytickspace}{0.12\textwidth}

\newcommand{\almostNegMomentumMag}{\num{.986}}
\newcommand{\almostNegMomentumLR}{\num{.75}}
\newcommand{\almostNegRate}{\num{0.9998}}
\newcommand{\almostPosMomentumMag}{\num{.9}}
\newcommand{\almostPosMomentumLR}{\num{.025}}
\newcommand{\almostPosRate}{\num{0.973}}

\newcommand{\smallGANTime}{2}
\newcommand{\bigGANTime}{10}

\newcommand{\shiftHeatmapx}{-1.2cm}

\newcommand{\spectrumVaryWidth}{0.19\linewidth} 
\newcommand{\spectrumAxisWidth}{0.012\linewidth}
\newcommand{\spectrumCbarWidth}{0.03\linewidth}
\newcommand{\spectrumSpacer}{-1.45cm}
\newcommand{\ytickspaceApp}{0.047\textwidth}

\newcommand{\mixtureNum}{\num{8}}
\newcommand{\numHiddenUnits}{\num{256}}
\newcommand{\numIterations}{\num{100000}}
\newcommand{\batchSizeSmallGAN}{\num{100000}}
\newcommand{\dataDim}{\num{2}}
\newcommand{\ganHeatmapWidth}{.4\textwidth}

\newcommand{\localTitle}{Complex Momentum for Optimization in Games}

\newcommand{\hess}{\mathcal{H}}
\newcommand{\bothHess}{\hat{\hess}}
\newcommand{\lyapunovfixedPointOp}{\boldsymbol{F}}
\newcommand{\jacFixedPointOp}{\boldsymbol{J}}
\newcommand*\conj[1]{\bar{#1}}
\newcommand{\cutForACM}[1]{#1}
\newcommand{\BibTeX}{\rm B\kern-.05em{\sc i\kern-.025em b}\kern-.08em\TeX}
\newcommand{\lyap}{\hat{\eigval}}
\newcommand{\displacement}{\boldsymbol{d}}

\newcommand{\prior}[1]{p \left( #1 \right)} 
\newcommand{\param}{\eparam} 
\newcommand{\paramFixed}{\param} 
\newcommand{\hyper}{\hparam} 
\newcommand{\hyperFixed}{\hyper} 
\newcommand{\hyperDist}{\prior{\hyper}} 
\newcommand{\optParam}[1]{\param^{**}}
\newcommand{\optHyper}[1]{\hyper^{*}}
\newcommand{\lossSymbolInner}{\Ltr} 
\newcommand{\lossSymbolOuter}{\Lval} 
\newcommand{\outerLoss}[1]{\lossSymbolOuter  \left( #1 \right)}
\newcommand{\outerUpdateSymbol}{\textrm{hyperopt}} 
\newcommand{\outerUpdate}[1]{\outerUpdateSymbol  \left( #1 \right)} 
\newcommand{\variableData}{\bf{x}} 

\newcommand{\outerIter}{T_{\mathrm{outer}}} 
\newcommand{\innerIter}{T_{\mathrm{inner}}} 
\newcommand{\responseParam}{{\boldsymbol{\phi}}}
\newcommand{\responseParamFixed}{\responseParam} 
\newcommand{\approxResponseSymbol}[1]{\param_{#1}} 
\newcommand{\approxResponse}[2]{\approxResponseSymbol{#2} \left( #1 \right)} 
\newcommand{\sampleRename}[1]{#1} 
\newcommand{\curRename}[1]{\hat{#1}} 
\newcommand{\hyperDistVar}{p \left( \sampleRename{\hyper} | \curRename{\hyper} \right)} 
\newcommand{\lossTrainData}[2]{\lossSymbolInner \left( #1, #2, \variableData \right)} 
\newcommand{\lossValidData}[1]{\lossSymbolOuter \left( #1, \variableData\right)} 

\newcommand{\numModelParams}{\num{7850}} 

\newcommand{\numTrainSmall}{10} 
\newcommand{\numHiddenGlobal}{50} 
\newcommand{\batchSizeGlobal}{2} 
\newcommand{\stepSizeHypernetGlobal}{\num{0.0001}} 
\newcommand{\hyperMeanGlobal}{0} 
\newcommand{\hyperVarGlobal}{\num{1.5}} 
\newcommand{\numHiddenLarge}{10} 
\newcommand{\numTrainVs}{25}
\newcommand{\numValidVs}{\num{10215}}

\author{Jonathan Peter Lorraine}
\title{Scalable Nested Optimization for Deep Learning}
\degree{Doctor of Philosophy}
\department[]{Department of Computer Science}
\gradyear{2023}
\begin{document}
  \frontmatter
    \maketitle
    \begin{abstract}
Gradient-based optimization has been critical to the success of machine learning, updating a single set of parameters to minimize a single loss.
A growing number of applications rely on a generalization of this, where we have a bilevel or nested optimization of which subsets of parameters update on different objectives nested inside each other.
We focus on motivating examples of hyperparameter optimization and generative adversarial networks.
However, na\"ively applying classical methods often fails when we look at solving these nested problems on a large scale.
In this thesis, we build tools for nested optimization that scale to deep learning setups.

\begin{itemize}
    \item In Chapter~\ref{chap:hyperparamWithHypernets}, we provide an explicit, differentiable approximation to how neural network weights best-respond -- or optimize -- for their loss.
    We train a hypernetwork that takes in hyperparameters and outputs neural network weights.
    We use this hypernetwork for hyperparameter optimization.

    \item In Chapter~\ref{chap:hyperparamWithIFT}, we explore an algorithm that implicitly approximates the neural network weights best-response, using the implicit function theorem.
    We apply this to hyperparameter optimization, showing we can tune as many hyperparameters as neural network weights.

    \item In Chapter~\ref{chap:complexMomentum}, we augment simultaneous gradient descent to mitigate rotational dynamics.
    We do this by generalizing gradient descent with momentum to have a complex-valued momentum while retaining real-valued parameter updates.

    \item In Chapter~\ref{chap:lyapunovExponent}, we generalize strategies for finding diverse solutions in single-objective optimization to finding diverse solutions in setups where multiple agents optimize for their own objective.
    We do this by taking the Ridge-rider algorithm and generalizing their branching criteria to occur at bifurcations using connections to Lyapunov exponents.
\end{itemize}

    \end{abstract}
    \begin{acknowledgements}

I extend my deepest gratitude to my advisor, David, whose influence has been nothing short of transformative.
My experience under your mentorship has been educational and a source of personal growth, making my Ph.D. a wonderful journey.
The environment you cultivated was rich with engaging research, unwavering support, and kindness.
You taught me how to be a researcher, maintain high standards, communicate effectively, and everything else.
I will greatly miss brainstorming ideas on your whiteboard.
I am sincerely thankful for your impact on my life and career, and I aspire to provide similar guidance and support to others, just as you have done for me.

A big thank you to Roger Grosse and Nisarg Shah, the members of my supervisory committee, for their guidance and support throughout this journey.
Your advice and wisdom have profoundly shaped my research and broader academic perspective.
Roger, thank you for your exceptional detail-oriented approach, patience, and wide-ranging insights you made through our collaborations.
I also want to thank my external reviewers, Zico Kolter and Murat Erdogdu for their efforts in this process.

I had the great privilege of working with various mentors during my internships.
Jakob Foerster, thank you for supporting my first internship and showing me how to approach complex problems in a way that I will carry to my future endeavors.
Mehadi Hassen, thank you for helping build the foundations of my engineering skills to carry out research in production.
A special thanks to James Lucas for not only fulfilling but greatly surpassing the role of a mentor.
Your job advice, strategic guidance on research agendas, and friendship have been a cornerstone of my success.

I am immensely grateful to all my coauthors for the opportunity to work with brilliant, insightful, and enjoyable colleagues.
My sincere thanks to Paul Vicol for our many fruitful collaborations.
I also want to give a big thanks to my other co-authors: David Acuna, Kevin Xie, Xiaohui Zeng, George Adam, Matthew MacKay, Safwan Hossain, Jack Parker-Holder, Aldo Pacchiano, Luke Metz, Tal Kachman, Aniruddh Raghu, Simon Kornblith, Matthew McDermott, Jack Richter-Powell, Brandon Amos, Fabian Pedregosa, Nihesh Anderson, Chansoo Lee, Quentin De Laroussilhe, Chen-Hsuan Lin, Towaki Takikawa, Nicholas Sharp, Tsung-Yi Lin, Ming-Yu Liu, Derek Lim, Haggai Maron, Marc Law, Michael Zhang, Nishkrit Desai, Juhan Bae, Jimmy Ba, Wu Lin, Nikhil Mehta, Steve Masson, Ramanathan Arunchalam, Zaid Pervaiz Bhat, Arun George Zakhariah, and Dmitry Krass.

I extend my gratitude to the pivotal institutions in my PhD journey.
The University of Toronto, The Vector Institute, Facebook, Google, and NVIDIA - each has provided support, resources, and opportunities that have greatly enriched my academic experience.
In addition, a special thanks to the Toronto AI Lab for creating an exceptional environment.
I extend my heartfelt appreciation to Sanja Fidler for her pivotal role in organizing and leading this vibrant community.

Mom and Dad, you have been the bedrock of my success, consistently promoting the value of education and providing an environment where learning and curiosity could flourish.
Your sacrifices, guidance, and encouragement have been the driving forces behind my achievements.
To my mother, thank you for your endless patience, nurturing spirit, and the confidence you instilled in me.
To my father, thank you for your wisdom, resilience, and the invaluable life lessons that have shaped my character.
You both have ensured my success and taught me the importance of perseverance and curiosity.
I am forever grateful for everything you have done and continue to do.

\newpage

To all of my other wonderful friends in Toronto, I cannot express enough gratitude for the support and joy you brought to my life during my PhD journey.
Thank you for being the anchor that kept me sane amid the whirlwind of academia.
Your presence has been a constant reminder of the world beyond research.
Thank you to Moeen, Hashir, Marius, Jack, Mousa, Aly, Jaideep, Andy, Carson, Ewan, Shoaib, Calypso, Peter, Dan, Gareth, Luis, Samuel, Tahsin, and Mahrukh.

Alex, you were with me on my very first day of university, and I deeply wish you could have seen this journey to its end.
The void that you left is irreplaceable.
Thank you, Alex, for the shared moments; I will always cherish our memories.
    \end{acknowledgements}
    \tableofcontents
    \listoffigures
    \listoftables
    \listofalgorithms
  \mainmatter
    \chapter{Introduction}

\vspace{-0.02\textheight}
\noindent
\textbf{Motivating single-objective optimization in machine learning:}
In recent years, machine learning has emerged as a transformative force in numerous scientific and industrial domains, profoundly impacting everything from computer-vision, to natural language processing, to personalized medicine.
Large neural networks have become a foundational workhorse for using machine learning to advance these fields.
Central to this revolution has been the advancement of gradient-based optimization techniques, which have effectively trained increasingly large and complex neural network architectures.
However, the traditional focus on optimizing a single set of parameters for a singular objective increasingly gives way to more nuanced paradigms.
\newline

\noindent
\textbf{Motivating the generalized nested optimization paradigm in machine learning:}
A growing number of applications require learning with subsets of parameters updating on different objectives.
Important examples are hyperparameter optimization~\citep{maclaurin2015gradient, andrychowicz2016learning, fu2016drmad, shaban2018truncated}, GANs~\citep{goodfellow2014generative}, actor-critic models~\citep{pfau2016connecting}, curriculum learning~\citep{baker2019emergent, balduzzi2019open, sukhbaatar2017intrinsic}, adversarial examples~\citep{bose2020adversarial, yuan2019adversarial}, learning models~\citep{rajeswaran2020game, abachi2020policy, nikishin2021control}, domain adversarial adaptation~\citep{acuna2021fdomainadversarial}, neural architecture search~\citep{elsken2019neural}, and meta-learning~\citep{finn2017model, ren2018meta, ren2020flexible}.
This thesis focuses on motivating examples of hyperparameter optimization and GANs.
Furthermore, we will call a setup with more than one objective a \emph{game}, call optimization here \emph{learning in games}, and each subset of parameters with their respective losses will be called a \emph{player}.
\newline

            

\noindent
\textbf{Introducing nested optimization more formally:}
Consider $\num{2}$-player games\footnote{See our non-thesis research in \citet{raghu2020teaching} for an example with more players.} ---with players denoted by $A$ and $B$---where each player minimizes their loss $\loss_A, \loss_B$ with their parameters $\paramSymbol_A, \paramSymbol_B$.
We work with setups where the objectives are nested inside of each other -- so-called Stackelberg games or bilevel optimization~\citep{von1952theory} -- whose solutions can be defined as:\vspace{-0.01\textheight}
\begin{align}\label{eq:main_bilevel_opt}
    \begin{split}
        \outParam^{*} &= \argmin_{\outParam} \outLoss\left(\outParam, \inParam^{*}\left(\outParam\right)\right),\\
        \inParam^{*}\left(\outParam\right) &= \argmin_{\inParam} \inLoss\left(\outParam, \inParam\right)
    \end{split}
\end{align}
\newline

Here, $\inParam^*\left(\outParam\right)$ denotes player $\inSymbol$'s best-response function, which says how their optimal parameters change as a function of the other players' parameters.
In general, the inner and outer optimization can have multiple solutions -- see \citet{vicol2020implicit} -- giving rise to a best-response set instead of a function. However, for most of the thesis, we focus on the setup with unique solutions, except in Chapter~\ref{chap:lyapunovExponent}, where we look at methods to find multiple valid solutions.
\newline

\vspace{-0.015\textheight}
\noindent
\textbf{Existing nested optimization approaches and their limitations:}
The overarching goal of this thesis is to efficiently find solutions $(\outParam^{*}, \inParam^{*})$ when our parameters $\outParam$ and $\inParam$ are approaching the size of modern neural network parameters.
We may be able to approximately find $\outParam^{*}$ efficiently if we can do gradient-based optimization on:\vspace{-0.01\textheight}
\begin{equation}
    \outLoss^*\left(\outParam\right) = \outLoss\left(\outParam, \inParam^{*}(\outParam)\right)\vspace{-0.015\textheight}
\end{equation}
Optimizing this objective would require computing the \emph{hypergradient} $\pd{\outLoss^*}{\outParam}$:\vspace{-0.01\textheight}
\begin{align}\label{intro_eq:gradient-decomp}
    \begin{split}
        &\underbrace{\pd{\outLoss^*(\outParam)}{\outParam}}_{\text{\tiny{hypergradient}}}
        = \left. \left( \pd{\outLoss}{\outParam} + \pd{\outLoss}{\inParam} {\pd{\inParam^*}{\outParam}} \right) \right|_{\outParam, \inParam^*(\outParam)} =\\ 
        &{\!\!\!\!\!\! \underbrace{\pd{\outLoss\left(\outParam, \inParam^*\left(\outParam\right)\right)}{\outParam}}_{\text{\tiny{direct gradient}}}} \hphantom{A} + \hphantom{A}
        \overbrace{\!\!
            \pd{\outLoss\left(\outParam, \inParam^*\left(\outParam\right)\right)}{\inParam^*\left(\outParam\right)}
            \hphantom{a} \times \!\!\!\!
            {\underbrace{\pd{\inParam^*\left(\outParam\right)}{\outParam} \vphantom{\pd{A^A}{A_A}}}_{\text{\tiny{best-response Jacobian}}}}
        \!\!\!\!\!\!\!\!\!\!\!\!\!\!}^{\text{\tiny{indirect gradient}}}
    \end{split}
\end{align}
Unfortunately, this often requires $\pd{\inParam^{*}}{\outParam}$, but $\inParam^{*}(\outParam)$ and its Jacobian are typically intractable, as they require exact evaluation and differentiation through optimization.
An alternative strategy is simultaneous/alternating gradient descent, where each player does gradient descent on their own objectives -- $\pd{\outLoss}{\outParam}$ and $\pd{\inLoss}{\inParam}$ respectively.
However, strategies like this can fail.
For example, because a gradient is identically zero as in hyperparameter optimization or because of rotational dynamics~\citep{berard2019closer} as in GANs.
Black-box methods like random-search or Bayesian Optimization~\citep{movckus1975bayesian, snoek2012practical} circumvent gradient calculations but are ineffective for high-dimensional problems -- e.g., greater than \num{100} dimensions.
Specifically, we want the ability to optimize high-dimensional problems prevalent in machine learning.
\newline

\vspace{-0.01\textheight}
\section{Thesis Outline}
\vspace{-0.005\textheight}
    In the Chapters~\ref{chap:hyperparamWithHypernets} and \ref{chap:hyperparamWithIFT}, we explore approximations for the best-response $\inParam^*(\outParam)$ and its Jacobian $\pd{\inParam^{*}}{\outParam}$, which are applied to hyperparameter optimization.
    In Chapter~\ref{chap:complexMomentum}, we instead look at augmenting the gradient dynamics to avoid rotational dynamics, without approximating $\pd{\inParam^{*}}{\outParam}$, which is applied to GANs.
    Finally, we look at generalizing a branching single-objective optimization algorithm that finds multiple solutions to set-ups with multiple objectives.
    \begin{itemize}
        \item In Chapter~\ref{chap:hyperparamWithHypernets}, we explicitly approximate the best-response function $\inParam^{*}(\outParam)$ with a hypernetwork which we differentiate through for optimization.
        Our method collapses the nested optimization of model weights and hyperparameters into a joint stochastic optimization 
        We do this via amortized optimization with hypernetworks, which output approximately optimal weights as a function of hyperparameters.
        We compare this with standard hyperparameter optimization and demonstrate its use in tuning hundreds to thousands of hyperparameters.
    
        \item In Chapter~\ref{chap:hyperparamWithIFT} we approximate the best-response Jacobian $\pd{\inParam^{*}}{\outParam}$ using the implicit function theorem (IFT), which we then use for optimization.
        We propose an inexpensive gradient-based hyperparameter optimization algorithm that combines the IFT with efficient inverse Hessian approximations.
        We present results on the relationship between IFT and differentiation through optimization, which motivates our algorithm.
        We use the proposed approach to train modern network architectures with millions of weights and \textit{millions of hyperparameters}.
        We learn a data-augmentation network---where every weight is a hyperparameter tuned for validation performance---that outputs augmented training examples; we learn a distilled dataset where every feature in each data point is a hyperparameter; and we tune millions of regularization hyperparameters.
        Jointly tuning weights and hyperparameters with our approach is only a few times more costly in memory and compute than standard training.
    
        \item In Chapter~\ref{chap:complexMomentum}, we augment simultaneous gradient descent to mitigate the rotational dynamics.
        We generalize gradient descent with momentum for optimization in differentiable games to have complex-valued momentum.
        We give theoretical motivation for our method by proving convergence on bilinear zero-sum games for simultaneous and alternating updates.
        Our method gives real-valued parameter updates, making it a drop-in replacement for standard optimizers.
        We empirically demonstrate that complex-valued momentum can improve convergence in realistic adversarial games---like generative adversarial networks---by showing that we find better solutions with an almost identical computational cost.
        We also show a practical complex-valued Adam variant, which we use to train BigGAN to improve inception scores on CIFAR-10.
    
        \item Ridge Rider (RR) is an algorithm to find diverse solutions in optimization problems by following eigenvectors of the Hessian (``ridges'').
        RR is designed for conservative gradient systems (i.e., settings involving a single loss function), where it branches at saddles --- easy-to-find bifurcation points.
        In Chapter~\ref{chap:lyapunovExponent}, we generalize this idea to nonconservative dynamics from gradient optimization on multiple objectives by proposing a method -- denoted Generalized Ridge Rider (GRR) -- for finding arbitrary bifurcation points.
        We give theoretical motivation for our method by leveraging machinery from the field of dynamical systems.
        We construct novel toy problems where we visualize new phenomena while giving insight into high-dimensional problems of interest.
        Finally, we evaluate our method by finding diverse solutions to the iterated prisoners' dilemma and relevant machine learning problems, including generative adversarial networks.
    \end{itemize}

\section{Summary of Publications}

    \subsection{Research Used in Thesis}
        The contents of Chapter~\ref{chap:hyperparamWithHypernets} consist of research ideas and results taken from:

        \begin{sidedbox}
            Lorraine, Jonathan, and David Duvenaud. ``Stochastic hyperparameter optimization through hypernetworks'' 31st Conference on Neural Information Processing Systems (NIPS 2017), Workshop on Meta-learning. Long Beach, USA. 2017.
        \end{sidedbox}

        \hphantom{a}\newline
        \noindent
        The contents of Chapter~\ref{chap:hyperparamWithIFT} consist of research ideas and results taken from:

        \begin{sidedbox}
            Lorraine, Jonathan, Paul Vicol, and David Duvenaud. ``Optimizing millions of hyperparameters by implicit differentiation.'' International Conference on Artificial Intelligence and Statistics (AISTATS). PMLR, 2020.
        \end{sidedbox}

        \noindent
        The contents of Chapter~\ref{chap:complexMomentum} consist of research ideas and results taken from:

        \begin{sidedbox}
            Lorraine, Jonathan, David Acuna, Paul Vicol, and David Duvenaud. ``Complex momentum for optimization in games.'' In International Conference on Artificial Intelligence and Statistics (AISTATS). PMLR, 2022.
        \end{sidedbox}

        \noindent
        The contents of Chapter~\ref{chap:lyapunovExponent} consist of research ideas and results taken from:
        
        \begin{sidedbox}
            Lorraine, Jonathan, Paul Vicol, Jack Parker-Holder, Tal Kachman, Luke Metz, and Jakob Foerster. ``Lyapunov Exponents for Diversity in Differentiable Games.'' In Proceedings of the 21st International Conference on Autonomous Agents and Multiagent Systems (AAMAS). 2022.
        \end{sidedbox}

    \subsection{Non-thesis Research}
        Publications I have worked during my Ph.D. and are excluded from this thesis include \citet{mehta2024fmsARXIV, mehta2024fms, bae2024tda, xie2024latte3d, lim2024graph, lim2024graphARXIV, zhang2023using, zhang2023usingARXIV, lorraine2023att3d, lorraine2023att3dARXIV, lorraine2022task, vicol2020implicit, vicol2020implicitARXIV, vicol2020implicitWORKSHOP, richter2021input, raghu2021meta, raghu2021metaARXIV, lorraine2019jacnet, mackay2019self, mackay2019selfARXIV, adam2019understanding}.
        \newline

        \begin{sidedbox}
            Mehta, Nikhil, Jonathan Lorraine, Steve Masson, Ramanathan Arunachalam, Zaid Pervaiz Bhat, James Lucas, Arun George Zachariah. ``Improving Hyperparameter Optimization with Checkpointed Model Weights.'' In Submission to Advances in Neural Information Processing Systems (NeurIPS), 2024.
            \newline

            \noindent
            Bae, Juhan, Wu Lin, Jonathan Lorraine, Roger Grosse. ``Training Data Attribution via Approximate Unrolled Differentiation.'' In Submission to Advances in Neural Information Processing Systems (NeurIPS), 2024.
            \newline

            \noindent
            Xie, Kevin, Jonathan Lorraine, Tianshi Cao, Jun Gao, James Lucas, Antonio Torralba, Sanja Fidler, Xiaohui Zeng. ``LATTE3D: Large-scale Amortized Text-To-Enhanced3D Synthesis.'' In Submission to The European Conference on Computer Vision (ECCV), 2024.
            \newline

            \noindent
            Lim, Derek, Haggai Maron, Marc T. Law, Jonathan Lorraine, and James Lucas. ``Graph Metanetworks for Processing Diverse Neural Architectures.'' The International Conference on Learning Representations (ICLR), 2024.
            \newline

            \newpage
            \noindent
            Zhang, Michael, Nishkrit Desai, Juhan Bae, Jonathan Lorraine, and Jimmy Ba. ``Using Large Language Models for Hyperparameter Optimization.'' Advances in Neural Information Processing Systems (NeurIPS), 2023 Foundation Models for Decision Making Workshop.
            \newline

            \noindent
            Lorraine, Jonathan, Kevin Xie, Xiaohui Zeng, Chen-Hsuan Lin, Towaki Takikawa, Nicholas Sharp, Tsung-Yi Lin, Ming-Yu Liu, Sanja Fidler, and James Lucas. ``ATT3D: Amortized Text-to-3D Object Synthesis.'' In Proceedings of the International Conference on Computer Vision (ICCV), 2023.
            \newline

            \noindent
            Lorraine, Jonathan, Nihesh Anderson, Chansoo Lee, Quentin De Laroussilhe, and Mehadi Hassen. ``Task Selection for AutoML System Evaluation.'' arXiv:2208.12754 (2022).
            \newline

            \noindent
            Vicol, Paul, Jonathan Lorraine, Fabian Pedregosa, David Duvenaud, and Roger B. Grosse. ``On Implicit Bias in Overparameterized Bilevel Optimization.'' In International Conference on Machine Learning (ICML), pp. 22234-22259. PMLR, 2022.
            \newline

            \noindent
            Richter-Powell, Jack, Jonathan Lorraine, and Brandon Amos. ``Input Convex Gradient Networks.'' Advances in Neural Information Processing Systems (NeurIPS) Optimal Transport and Machine Learning (OTML) Workshop, November 2021.
            \newline

            \noindent
            Raghu, Aniruddh, Jonathan Lorraine, Simon Kornblith, Matthew McDermott, and David K. Duvenaud. ``Meta-Learning to Improve Pre-Training.'' Advances in Neural Information Processing Systems 34 (NeurIPS) 2021: 23231-23244.
            \newline

            \noindent
            Lorraine, Jonathan, and Safwan Hossain. ``JacNet: Learning Functions with Structured Jacobians.'' In International Conference on Machine Learning Invertible (ICML), 2019 Neural Nets and Normalizing Flows (INNF) Workshop.
            \newline

            \noindent
            Mackay, Matthew, Paul Vicol, Jonathan Lorraine, David Duvenaud, and Roger Grosse. ``Self-Tuning Networks: Bilevel Optimization of Hyperparameters using Structured Best-Response Functions.'' In International Conference on Learning Representations (ICLR). 2018.
            \newline

            \noindent
            Adam, George, and Jonathan Lorraine. ``Understanding Neural Architecture Search Techniques.'' arXiv:1904.00438 (2019).
        \end{sidedbox}

  
    \chapter{Hyperparameter Optimization Through Hypernetworks}\label{chap:hyperparamWithHypernets}
Machine learning models often nest optimization of model weights within the optimization of hyperparameters.
We give a method to collapse this nested optimization into joint stochastic optimization of weights and hyperparameters.
We train a neural (hyper)network to output optimized weights as a function of hyperparameters.
We compare this method to standard hyperparameter optimization methods and demonstrate its effectiveness in tuning thousands of hyperparameters.

\subsubsection*{Context for this Work}
This was my first paper, guided to a workshop submission and then a technical report by David.
All material in this chapter comes from \citet{lorraine2018stochastic}.
We had a follow-up paper, Self-Tuning Networks~\citep{mackay2019self}, led by Matthew MacKay and Paul Vicol, which improves many aspects of this work.
There, we build more scalable hypernetwork architectures, a better parameterization of the distribution of hyperparameters for training the hypernetwork, tuning hyperparameters of larger-scale networks in varying modalities, a wide variety of hyperparameters are optimized, and we demonstrate practical benefits from the method.
Further improvements are in the follow-up of $\Delta$-STN~\citep{bae2020delta}.

\section{Introduction}
Model selection and hyperparameter tuning are significant bottlenecks in the design of predictive models.
Hyperparameter optimization is a nested optimization, which can be viewed as a special case of Equation~\ref{eq:main_bilevel_opt}.
Here, the inner optimization finds the neural network's weights $\eparam$ that minimize the training loss $\Ltr$ given hyperparameters $\hparam$.
The outer optimization chooses $\hyper$ to reduce the validation loss with the best-responding weights $\Lval^*$:
\begin{figure}
\centering
\begin{tikzpicture}
    \node[shape=circle,draw=black] (optParams) at (-0.5,0.1) {$\alpha$};
    \node[shape=circle,draw=black, fill=gray] (x) at (-0.5,1) {$\textbf{x}$};
    \node[shape=circle,draw=black, fill=gray] (t) at (-0.5,2) {$\textbf{t}$};
    \node[shape=circle,draw=black] (lambda) at (1,-0.7) {$\hparam$};
    \node[shape=rectangle,draw=black] (train) at (1,0.1) {Training};
    \node[shape=circle,draw=black] (w) at (1,1) {$\eparam^{*}$};
    \node[shape=circle,draw=black] (y) at (1,2) {$\textbf{y}$};
    \node[shape=rectangle,draw=black] (L) at (1,2.8) {$\Lval$};
    
    \path [->] (optParams) edge node {} (train);
    \path [->] (lambda) edge node {} (train);
    \path [->] (train) edge node {} (w);
    \path [->] (x) edge node {} (y);
    \path [->] (w) edge node {} (y);
    \path [->] (t) edge node {} (L);
    \path [->] (y) edge node {} (L);
    
    \node[shape=circle,draw=black] (phi) at (2.5,0.1) {{\color{red}$\responseParam$}};
    \node[shape=circle,draw=black, fill=gray] (xNEW) at (2.5,1) {$\textbf{x}$};
    \node[shape=circle,draw=black, fill=gray] (tNEW) at (2.5,2) {$\textbf{t}$};
    \node[shape=circle,draw=black] (lambdaNEW) at (4.5,-0.7) {$\hparam$};
    \node[shape=rectangle,draw=black] (hypernet) at (4.5,0.1) {{\color{red}Hypernetwork}};
    \node[shape=circle,draw=black] (wNEW) at (4.5,1) {$\eparam^{*}$};
    \node[shape=circle,draw=black] (yNEW) at (4.5,2) {$\textbf{y}$};
    \node[shape=rectangle,draw=black] (LNEW) at (4.5,2.8) {$\Lval$};
    
    \path [->] (phi) edge node {} (hypernet);
    \path [->] (lambdaNEW) edge node {} (hypernet);
    \path [->] (hypernet) edge node {} (wNEW);
    \path [->] (xNEW) edge node {} (yNEW);
    \path [->] (wNEW) edge node {} (yNEW);
    \path [->] (tNEW) edge node {} (LNEW);
    \path [->] (yNEW) edge node {} (LNEW);
    \node[above] at (current bounding box.north) {\,\,\,\,\,\,\,\,Cross-validation \,\,\,\,\,\,\,\,\,\,\,\,\,\,\,\,\,\,\,\,\,Hyper-training};
\end{tikzpicture}
\caption[Compute graph for hyperparameter optimization with hypernetworks]{
\emph{Left:} A typical computational graph for cross-validation, where $\alpha$ are the optimizer parameters, and $\hyper$ are training loss hyperparameters.
It is expensive to differentiate throughout the training procedure.
\emph{Right:} The proposed computational graph with our changes in {\color{red}red}, where $\responseParam$ are the hypernetwork parameters.
We can differentiate cheaply through the hypernetwork to optimize the validation loss $\Lval$ with respect to hyperparameters $\hyper$.
We use $\textbf{x}$, $\textbf{t}$, and $\textbf{y}$ to refer to a data point, a label, and a prediction.}
\end{figure}
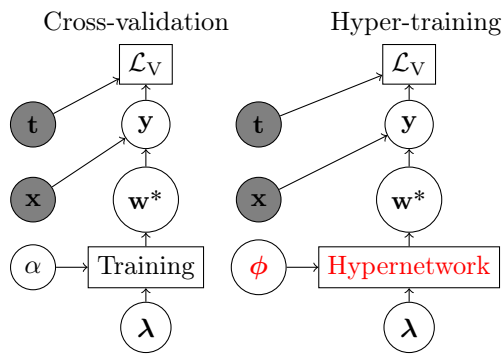
\begin{figure}
	\centering
        \begin{tikzpicture}
            \centering
            \node (img){\includegraphics[trim={1.25cm 1.25cm 0.15cm 0cm},clip, width=.75\linewidth]{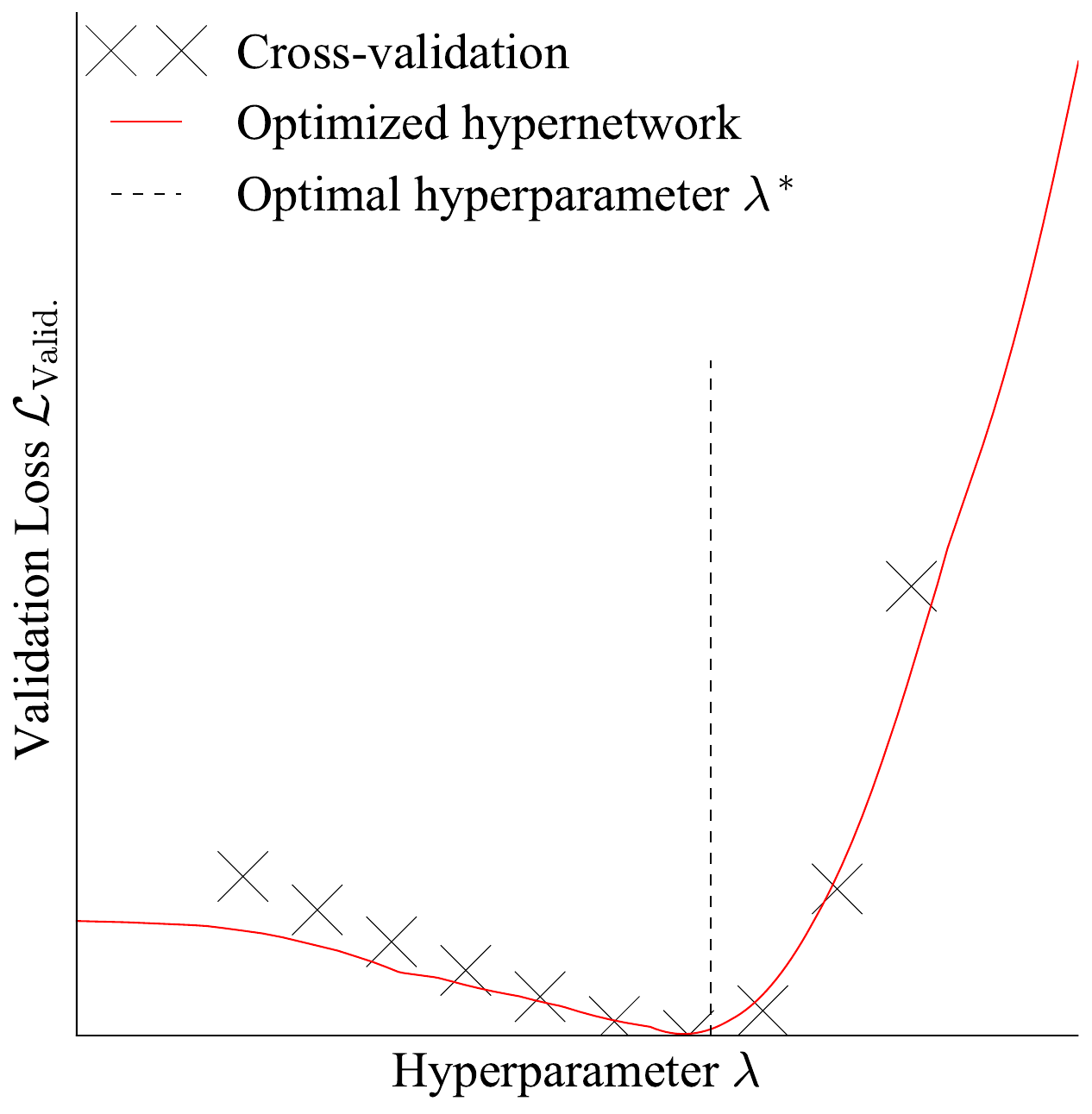}};
            \node[left=of img, node distance=0cm, rotate=90, xshift=2.75cm, yshift=-.90cm, font=\color{black}] {Best-responding Validation Loss $\Lval^{*}$};
            \node[below=of img, node distance=0cm, yshift=1.25cm,font=\color{black}] {Hyperparameter $\hparam$};
        \end{tikzpicture}

	\caption[Hypernetwork validation loss visualization]{
	The validation loss of a neural net is estimated by cross-validation (crosses) or a hypernetwork (line), which outputs \numModelParams-dimensional network weights.
	Cross-validation requires optimizing from scratch each time.
	The hypernetwork can be used as a proxy to cheaply evaluate the best-responding validation loss $\Lval^{*}$.
	\label{hypernet_fig:exp1}
	}
\end{figure}
%
\begin{align} \label{hypernet_eq nested}
    \hp^* &= \argmin_{\hp} \Lval^*\left(\hp\right) \text{ where } \\
    \Lval^*\left(\hp\right) &= \LvalResponse \text{ and }
    \response = \argmin_{\ep} \Ltr\left(\hp,\ep\right)
\end{align}
Standard practice in machine learning solves Equation~\ref{hypernet_eq nested} by gradient-free optimization of hyperparameters, such as grid or random search.
Each set of hyperparameters is evaluated by reinitializing the weights and training the model to completion.
Retraining a model from scratch is wasteful if the hyperparameters change by a small amount.
Some approaches, such as Hyperband~\citep{li2016hyperband} and freeze-thaw Bayesian optimization~\citep{swersky2014freeze}, resume model training but often scale poorly beyond $10$ to $20$ dimensions.
%
How can we avoid retraining from scratch each time?
Well, the optimal parameters $\param$ are a function of the hyperparameters $\hyper$:
\vspace{-0.01\textheight}
\begin{align}
\eparam^{*}\left(\hparam\right) = \argmin_{\eparam} \Ltr\left(\eparam, \hparam\right)
\label{hypernet_best response equation}
\end{align}
We propose to \emph{learn this function}.
\footnote{This notation implies $\param$ has a unique solution, which is used for simplicity but is not generally true.
See \citet{vicol2020implicit} for an investigation of the consequences of this in hyperparameter optimization.}
Specifically, we train a neural network that takes hyperparameters as input and outputs an approximately optimal set of weights.
This formulation provides two main benefits.
First, we can train the hypernetwork to convergence using stochastic gradient descent without training any particular model to completion.
Second, differentiating through the hypernetwork allows us to optimize hyperparameters with stochastic gradient-based optimization.
%

\begin{figure*}
	\centering
	\begin{minipage}{1.0\textwidth}
	\begin{tabular}{c l}
\phantom{aaaaaaaaaaaaaa} Training loss surface & \phantom{aaaaaaaa} Validation loss surface\\
\includegraphics[width=0.48\textwidth, clip, trim=10mm 0mm 0mm 15mm]{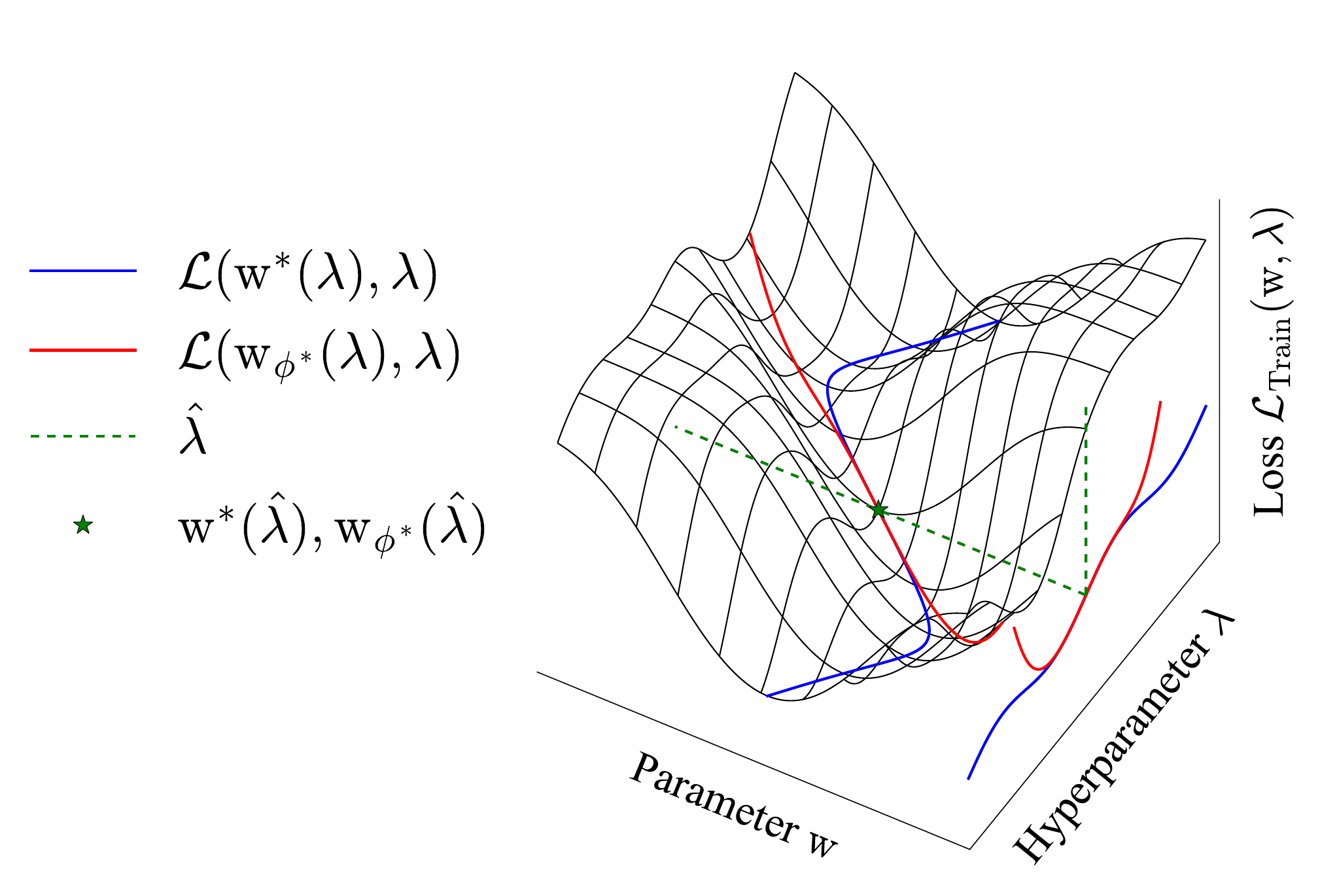} &
\includegraphics[width=0.48\textwidth, clip, trim=35mm 0mm 0mm 18mm]{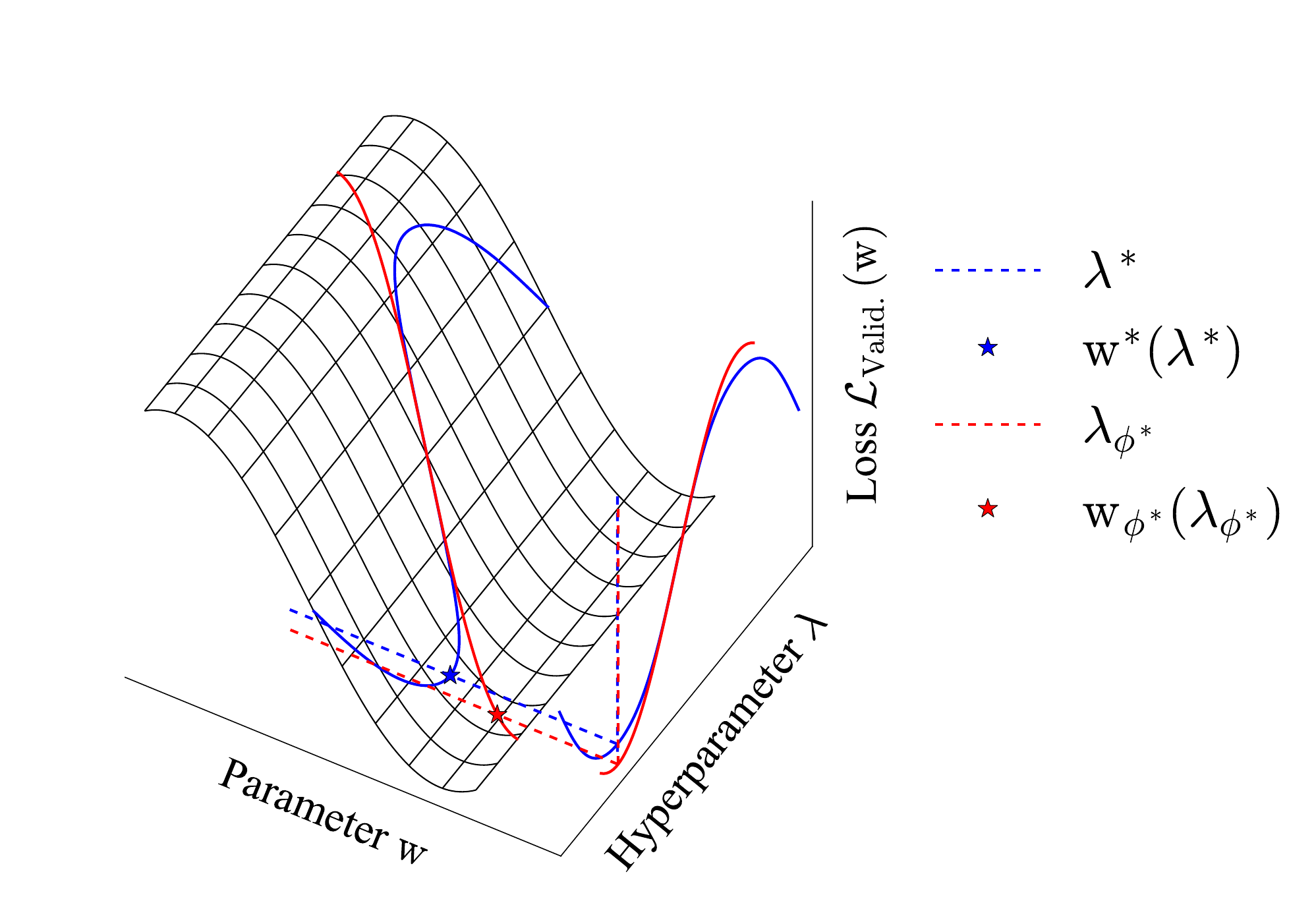}
\end{tabular}
\end{minipage}
\caption[Inner and outer loss approximation visualization]{
A visualization of exact (blue) and approximate (red) optimal weights as a function of hyperparameters.
The approximately optimal weights $\param_{\phi^{*}}$ are produced by a linear model fit at $\curRename{\lambda}$.
The true optimal hyperparameter is $\lambda^{*}$, while the hyperparameter minimizing the hypernetwork-approximated validation loss is $\lambda_{\phi^{*}}$.
\label{hypernet_fig:theory1}
}
\end{figure*}

\section{Training a network to output optimal weights}
\label{hypernet_sec.new_formulation}
Given some hyperparameters, how can we teach a hypernetwork~\citep{ha2016hypernetworks} to produce approximately optimal weights for another neural network?
At each iteration, we ask a hypernetwork with parameters $\responseParam$ to input hyperparameters $\hparam$ and to output a set of weights:
$\approxResponse{\hyperFixed}{\responseParamFixed}$.
Instead of updating the weights $\param$ using the training loss gradient $\nicefrac{\partial \Ltr (\param)}{\partial \param}$, we update the hypernetwork weights $\responseParamFixed$ using the chain rule: $\frac{\partial \Ltr(\param_\responseParamFixed)}{\partial \param_\responseParamFixed} \frac{\partial \param_\responseParamFixed}{\partial \responseParamFixed}$.
At convergence, we want our hypernetwork to match the best-response function closely: $\approxResponse{\hyperFixed}{\responseParamFixed} \approx \response$.
This formulation allows us to optimize the hyperparameters $\hyper$ using our hypernetwork-learned function as a surrogate best-response.
We call this method \emph{hyper-training} and contrast it with standard training methods.
%
%

Our method is related to the concurrent work of \citet{brock2017smash}, whose SMASH algorithm also approximates optimal weights as a function of model architectures to perform a gradient-free search over discrete model structures.
They only estimate the performance of various model architectures while we additionally compute gradients for the hypernetwork input.

We further extend this idea by formulating an algorithm to jointly optimize the hypernetwork and hyperparameters.
Joint optimization of parameters and hyperparameters addresses one of SMASH's main weaknesses: the hypernetwork must be very large to learn approximately optimal weights for many different settings.
During joint optimization, the hypernetwork only needs to model approximately optimal weights for the neighborhood around the current hyperparameters, allowing us to use even linear hypernetworks.


\subsection{Advantages of hypernetwork-based optimization}
Hyper-training is a method to learn a mapping from hyperparameters to validation loss that is differentiable and inexpensive to evaluate.
Alternatively, model-based hyperparameter schemes, like Bayesian optimization (e.g., \citet{snoek2012practical}) build a model of the validation loss as a function of hyperparameters.
This approach has several disadvantages compared to hyper-training.

First, obtaining data for standard Bayesian optimization requires optimizing models from initialization for each set of hyperparameters.
In contrast, hyper-training never needs to optimize any one model fully, removing choices like how many models to train and for how long.
Second, standard Bayesian optimization treats the best-responding validation loss as a black-box function. 
In contrast, hyper-training takes advantage of the fact that validation loss is a known, differentiable function, which can be evaluated stochastically by sampling points from the validation set.
%
This can have a better generalization for learning hyperparameter to validation loss than directly fitting the loss using a Gaussian process (e.g., \citet{rasmussen2006gaussian}), as in Figure~\ref{hypernet_fig:exp5}.

\subsection{Limitations of hypernetwork-based optimization}
We apply this method to unconstrained continuous bilevel optimization problems with a differentiable inner and outer loss function.
But what kind of parameters can be optimized by our approach?
Hyperparameters typically fall into two broad categories:
(1) parameters that change the set of locally optimal points, like regularization parameters or architectural choices, and
(2) parameters that affect the choice of locally optimal point and the rate we converge to them, like optimization hyperparameters such as learning rates.
Hyper-training does not have inner optimization parameters because there is no internal training loop, so we cannot optimize these.
We must still choose the optimization parameters for the fused optimization loop.
In principle, hyper-training can handle continuous relaxations of discrete hyperparameters, which we further explore in non-thesis research of \citet{mackay2019self}.

A clear difficulty of this approach is that hypernetworks can require several times as many parameters as the original model.
For example, training a fully connected hypernetwork with $1$ hidden layer of $H$ units to output $D$ parameters requires at least $D \times H$ hypernetwork parameters.
To partially address this problem, in Section~\ref{hypernet_joint optimization}, we propose an algorithm that trains only a linear model that maps hyperparameters to model weights.
We further explore better alternatives in non-thesis research of \citet{mackay2019self}.

Another limitation is that our approach only proposes making local changes to the hyperparameters and does not do global optimization or uncertainty-based exploration.
Finally, choosing the training distribution of hyperparameters $p(\hyper)$ is not obvious.
If we do not sample a sufficient range of hyperparameters, the estimated gradient of the validation loss with respect to the hyperparameters may be inaccurate.
We discuss several approaches to this problem in Section~\ref{hypernet_joint optimization}, and further explore this in non-thesis research of \citet{mackay2019self}.

%
\begin{algorithm}[H]
\caption{Standard cross-validation with stochastic optimization}
\label{hypernet_alg1}
\begin{algorithmic}[1]
\For{$i = 1, \dots, \outerIter$}
    \State Initialize elementary parameters $\paramFixed$
    \State $\hyperFixed = \outerUpdate{\hyperFixed^{(1:i)}, \outerLoss{\hyperFixed^{(1:i)}, \paramFixed^{(1:i)}}}$
    \While{$\innerIter$ steps}
        \State $\variableData \sim \textnormal{Training data}$
        \State $\paramFixed \mathrel{-}= \alpha \nabla_{\param} \lossTrainData{\hyper}{\param}$
    \EndWhile
    \State $\hyper^{i}, \param^{i} = \hyper, \param$
\EndFor
\State $i = \argmin_{i} \lossValidData{\hyper^{(i)}, \param^{(i)}}{}$
\State \Return $\hyperFixed^{(i)}, \param^{(i)}$, potentially re-training $\param$ with $\hyperFixed^{(i)}$ on all data
\end{algorithmic}
\end{algorithm}

\vspace{-0.03\textheight}
\begin{algorithm}[H]
\caption{Optimization of hypernetwork, then hyperparameters}
\label{hypernet_algGlobal}
\begin{algorithmic}[1]
\State Initialize hypernetwork $\textcolor{black}{\responseParamFixed}$
\State $\textcolor{black}{\text{Initialize hyperparameters } \curRename{\hyper}}$
\For{$T_{\text{{\color{black}hypernetwork}}}$ steps}
    \State $\variableData \sim \text{Training data}\textcolor{black}{, \hyperFixed \sim \hyperDist}$
    \State $\textcolor{black}{\responseParamFixed} \mathrel{-}= \alpha \nabla_{\textcolor{black}{\responseParam}}\Ltr\left(\hyper, \approxResponseSymbol{\textcolor{black}{\responseParam}} \textcolor{black}{ ( \hyperFixed ), \boldsymbol{x}}\right)$
\EndFor
\For{$\textcolor{black}{T_{\text{hyperparameter}}}$ steps}
    \State $\textcolor{black}{\variableData \sim \text{Validation data}}$
    \State $\textcolor{black}{\curRename{\hyperFixed} \mathrel{-}= \beta \nabla_{\curRename{\hyper}} \lossValidData{\curRename{\hyperFixed}, \approxResponse{\curRename{\hyper}}{\responseParamFixed}}}$
\EndFor
\State \Return $\curRename{\hyperFixed}, \approxResponseSymbol{\textcolor{black}{\responseParam}} \textcolor{black}{ \left( \curRename{\hyperFixed} \right)}$
\end{algorithmic}
\end{algorithm}
\vspace{-0.03\textheight}
\begin{algorithm}
\caption{Joint optimization of hypernetwork and hyperparameters}\label{hypernet_algLocal}
\begin{algorithmic}[1]
\State Initialize hypernetwork $\responseParamFixed$
\State Initialize hyperparameters $\curRename{\hyperFixed}$
\Loop
    \State $\variableData \sim \textnormal{Training data}$, $\sampleRename{\hyper} \sim p \left( \sampleRename{\hyper} {| \curRename{\hyper}} \right)$
    \State $\responseParamFixed \mathrel{-}= \alpha \nabla_{\responseParam} \Ltr\left(\hyper, \approxResponse{\hyperFixed}{\responseParamFixed}, \variableData \right)$
    \State $\variableData \sim \textnormal{Validation data}$
    \State $\curRename{\hyperFixed} \mathrel{-}= \beta \nabla_{\curRename{\hyper}} \lossValidData{\curRename{\hyper}, \approxResponse{\curRename{\hyper}}{\responseParamFixed}}$
\EndLoop
\State \Return $\curRename{\hyperFixed}, \approxResponse{\curRename{\hyperFixed}}{\responseParamFixed}$
\end{algorithmic}
\end{algorithm}


In Algorithms~\ref{hypernet_alg1}, \ref{hypernet_algGlobal}, and \ref{hypernet_algLocal}, we compare standard hyperparameter optimization, our global algorithm, and our joint algorithm, respectively.
Here, $\outerUpdateSymbol$ refers to a generic hyperparameter optimization, and we abuse notation, making the loss dependence on sampled data $\boldsymbol{x}$ explicit.
Notably, instead of updating the weights $\param$ using the loss gradient $\nabla_{\param} \loss$, we update the hypernetwork $\responseParamFixed$ and hyperparameters $\hyper$, which we use for gradient-based hyperparameter optimization.

\begin{figure}[ht]
\centering
\begin{tikzpicture}
    \centering
    \node (img){\includegraphics[trim={0cm 0cm 0cm 9cm},clip, width=.45\linewidth]{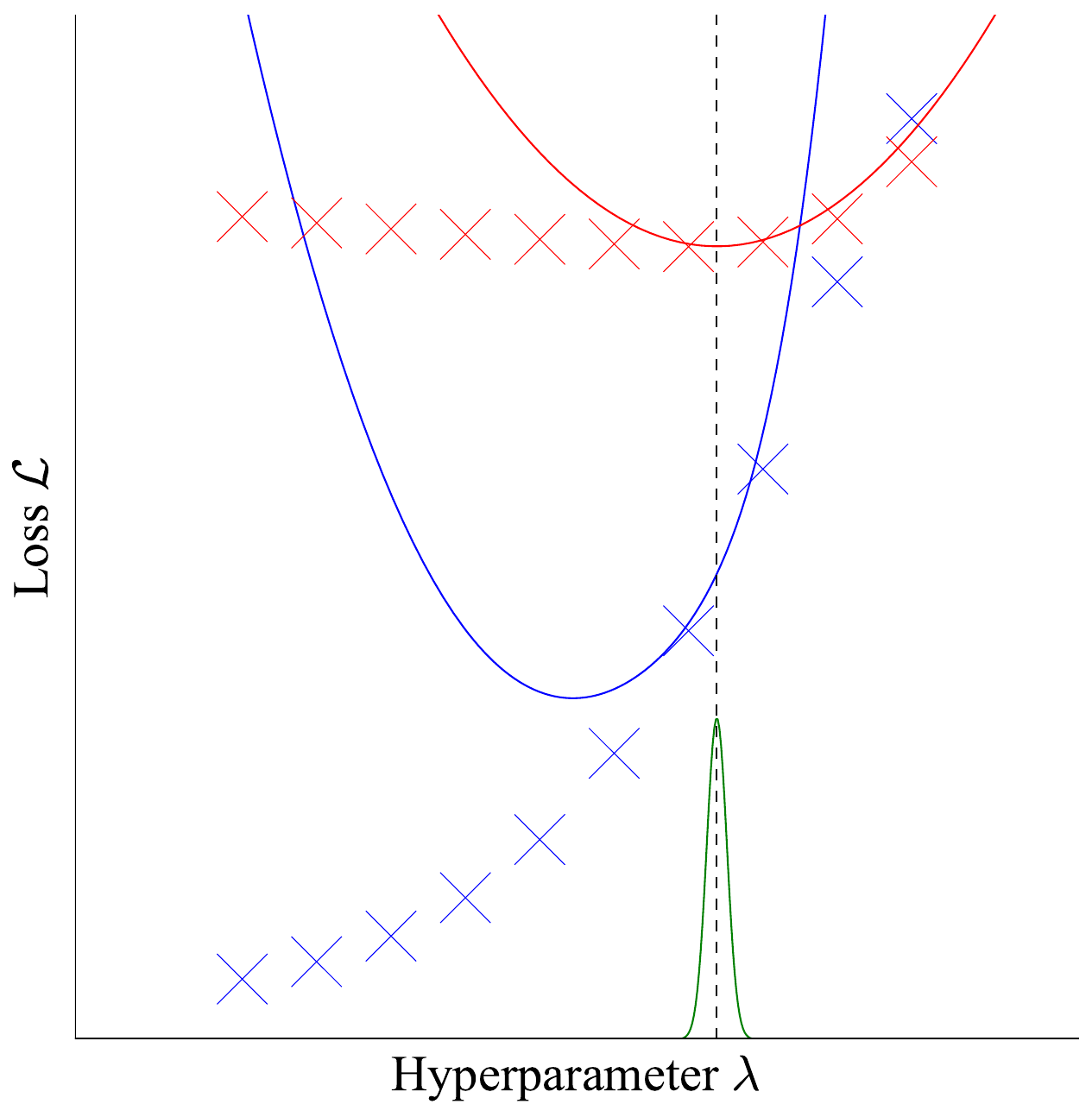}};
    \node[right=of img, xshift=-1.5cm] (img2){\includegraphics[trim={0cm 22.0cm 0cm 0cm},clip, width=.5\linewidth]{latex/hyperparamsWithHypernets/hypernets_local_small.pdf}};
\end{tikzpicture}

\caption[Trained hypernetwork on locally sampled hyperparameters]{
Training and validation losses of a neural network are estimated by cross-validation (crosses) or a linear hypernetwork (lines).
The hypernetwork's limited capacity makes it only accurate where the hyperparameter distribution puts mass.
	\label{hypernet_fig:exp2}}
\vspace{-0.02\textheight}
\end{figure}

\subsection{Jointly training parameters and hyperparameters}
\label{hypernet_joint optimization}

In practice, we should choose a distribution of hyperparameters for training the hypernet $\hyperDist$ that places most of its mass on promising hyperparameter values.
It may not be possible to learn a best-response for all hyperparameters due to limited hypernetwork capacity.
Thus, we propose Algorithm~\ref{hypernet_algLocal}, which only tries to match a best-response locally.
We introduce a ``current'' hyperparameter $\curRename{\hyperFixed}$, updated each iteration.
We define a conditional hyperparameter distribution, $\hyperDistVar$, which only puts the mass close to $\curRename{\hyperFixed}$.

Algorithm~\ref{hypernet_algLocal} combines both phases of Algorithm~\ref{hypernet_algGlobal}.
Instead of training a hypernetwork to output weights for any hyperparameter and then optimizing the hyperparameters, Algorithm~\ref{hypernet_algLocal} only samples hyperparameters near the current $\curRename{\hyperFixed}$.
So, the hypernetwork only needs to estimate weights for a concentrated set of hyperparameters.
There is an extra cost to retrain the hypernetwork on each hyperparameter $\curRename{\hyperFixed}$ update.
The locally trained hypernetwork is used for gradients to update $\curRename{\hyperFixed}$.

How simple can we make the hypernetwork and still obtain useful gradients to optimize hyperparameters?
Consider the case in our experiments where the hypernetwork is a linear function of the hyperparameters and the conditional hyperparameter distribution is $\hyperDistVar = \mathcal{N} \left( \curRename{\hyper}, \sigma \identity \right)$ for some small $\sigma$.
This hypernetwork learns a tangent hyperplane to a best-response function and only needs minor adjustments at each step if the hyperparameter updates are small.
We can further restrict the capacity of a linear hypernetwork by factorizing its weights, effectively adding a bottleneck layer with a linear activation and a small number of hidden units.

\section{Related Work}
Here, we briefly review the gradient-free and gradient-based methods we compare to in our experiments.
Our work complements the SMASH algorithm of \citet{brock2017smash}, with Section~\ref{hypernet_sec.new_formulation} discussing our differences.
We are both special cases of amortized optimization methods, which use learning to predict solutions when we repeatedly solve instances of the same problem~\citep{amos2022tutorial}.
The idea of weight generation given a context vector has been used in multiple areas~\citep{requeima2019fast, ratzlaff2019hypergan, pilault2020conditionally, tay2021hypergrid, rusu2018meta}.


\textbf{Gradient-free Hyperparameter Optimization}:
Various gradient-free hyperparameter optimization methods -- not scalable to this regime -- are detailed in Appendix Section~\ref{ift_app:related_work}.
%
Model-free approaches
use only trial and error to explore the hyperparameter space.
Simple model-free approaches applied to hyperparameter optimization include grid and random search~\citep{bergstra2012random}, or more sophisticated methods like Hyperband~\citep{li2016hyperband} combine bandit approaches with modeling the learning procedure.
%
Model-based approaches
build a surrogate function, like Bayesian optimization~\citep{movckus1975bayesian, snoek2012practical}, which we use as a baseline in our experiments, implemented by \citet{Spearmint2019}.
Evolutionary methods~\citep{alexandropoulos2019multi} are similar to our method, such as population-based training~\citep{jaderberg2017population}, which maintains a set of networks instead of using a hypernetwork.

\textbf{Gradient-based Hyperparameter Optimization}:
Notably, we compare to differentiating through optimization, which \citet{domke2012generic} proposes, \citet{maclaurin2015gradient} scales to differentiating through learning procedures in deep learning, and is implemented by \citet{franceschi2017forward}.
More gradient-based methods are discussed in Chapter~\ref{chap:hyperparamWithIFT}, Section~\ref{ift_sec:related-work} where we compare with them.

\begin{figure}[ht!]
\vspace{-0.005\textheight}
\centering
\includegraphics[width=0.49\textwidth]{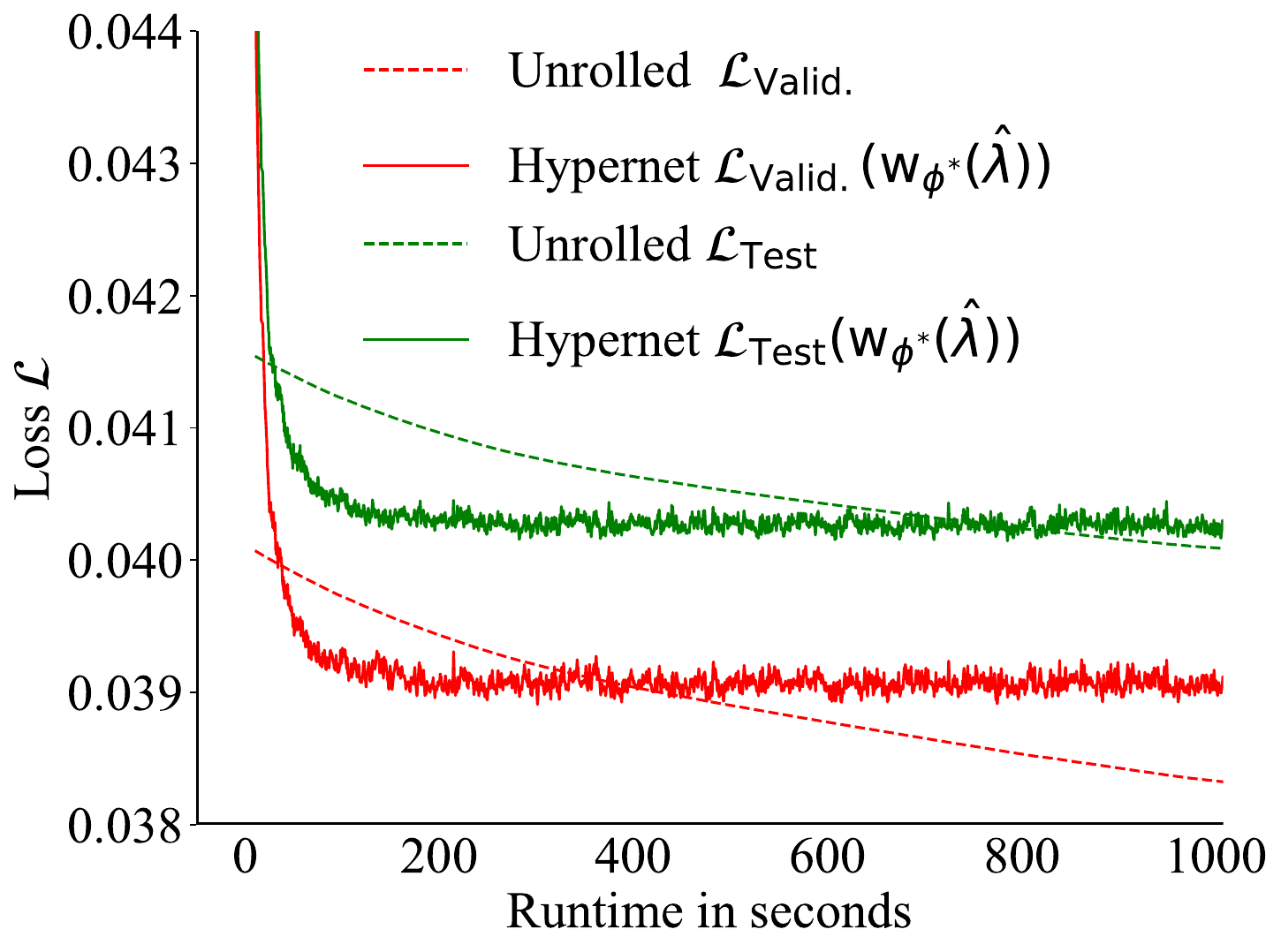}
\includegraphics[width=0.49\textwidth]{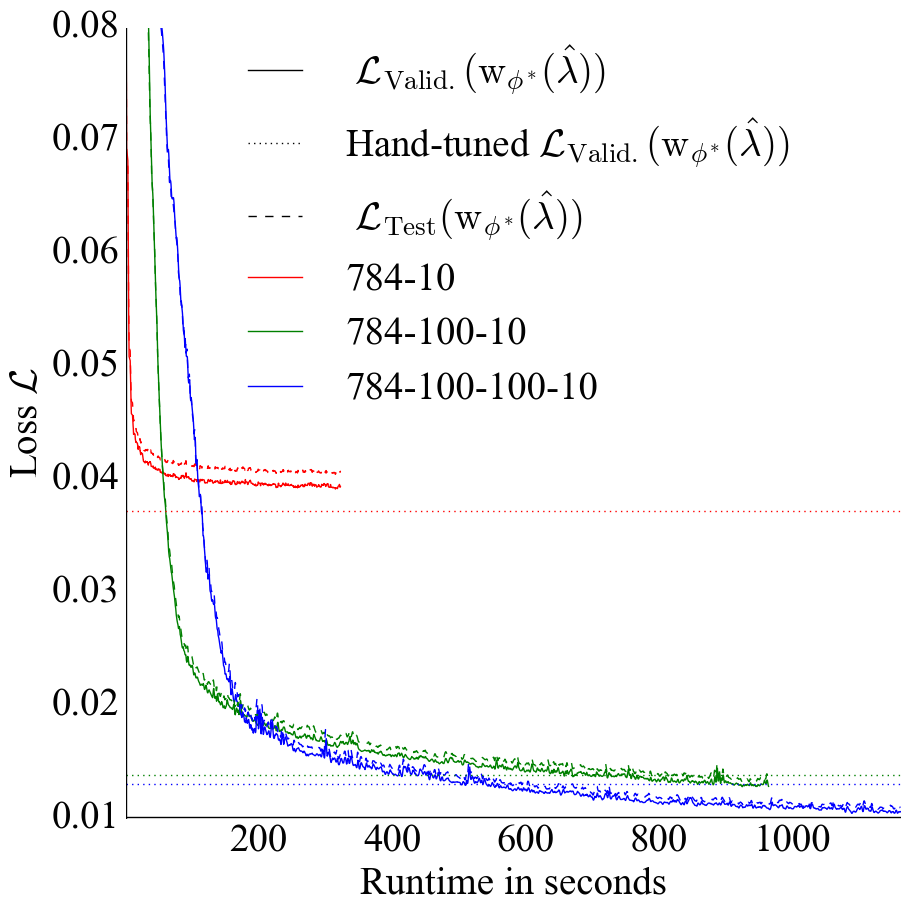}
\caption[Hypernetwork optimization curves]{
Validation and test losses during hyperparameter optimization with a separate $\ell_{2}$ weight decay applied to each weight in the model.
Thus, models with more parameters have more hyperparameters.
\emph{Left:} We solve the $\num{7850}$-dimensional hyperparameter optimization problem with a linear network and multiple algorithms.
Hypernetwork-based optimization converges to a suboptimal solution faster than unrolled optimization from \citet{maclaurin2015gradient}.
\emph{Right:} Hyper-training is applied to different layer configurations in the model.
\label{hypernet_fig:exp3}
}
\vspace{-0.02\textheight}
\end{figure}

\section{Experiments}
\label{hypernet_sec.experiments}
Our experiments examine the standard example of stochastic gradient-based optimization of neural networks with a weight regularization penalty.
%
%
In all experiments, Algorithms~\ref{hypernet_algGlobal} or \ref{hypernet_algLocal} are used to optimize the weights on MNIST~\citep{lecun1998gradient} with a $\ell_{2}$ weight decay penalty weighted by $\exp(\hyper)$.
Unless otherwise specified, all hidden units in the hypernetwork have ReLU activation~\citep{nair2010rectified}.
Autograd~\citep{maclaurin2015autograd} was used to compute all derivatives.
The mini-batch samples $\batchSizeGlobal$ sets of hyperparameters and up to $\num{1000}$ training data points for each experiment.
We use Adam~\citep{kingma2014adam} to train the hypernetwork and hyperparameters with a step size of $\stepSizeHypernetGlobal$, and $\left(\beta_1, \beta_2\right) = \left(.9, .999\right)$.

\subsection{Learning a global best-response}
Our first experiment, shown in Figure~\ref{hypernet_fig:exp1}, demonstrates learning a global approximation to a best-response function using Algorithm~\ref{hypernet_algGlobal}.
To simplify visualization of the regularization loss, we use $\numTrainSmall$ training data points to exacerbate overfitting.
We compare the performance of the weights output by the hypernetwork with those trained by standard cross-validation (Algorithm~\ref{hypernet_alg1}).
The elementary weights were randomly initialized for each hyperparameter choice.
When training the hypernetwork, the hyperparameters were sampled from a broad Gaussian distribution: $\hyperDist = \mathcal{N} ( \hyperMeanGlobal, \hyperVarGlobal)$.
The hypernetwork has $\numHiddenGlobal$ hidden units.
%
\textbf{Takeaway}:
The minimum of the best-response in Figure~\ref{hypernet_fig:exp1} is close to the real minimum of the validation loss, which shows that a hypernetwork can approximate a global best-response function in small problems.

\subsection{Learning a local best-response}
Figure~\ref{hypernet_fig:exp2} shows the same experiment as Figure~\ref{hypernet_fig:exp1}, except using Algorithm~\ref{hypernet_algLocal}.
The fused updates result in finding a best-response approximation whose minimum is the actual minimum faster than in the prior experiment.
The conditional hyperparameter distribution is given by $\hyperDistVar = \mathcal{N} \left( \curRename{\hyperFixed}, \num{0.00001} \identity \right)$.
Here, the hypernetwork is a linear model.
%
Again, the minimum of the best-response at the end of training minimizes the validation loss.
\textbf{Takeaway}:
Using only a locally trained linear best-response function can give sufficient gradient information to optimize hyperparameters while being less computationally expensive than learning a global best-response.

\begin{figure}[t!]
\vspace{-0.005\textheight}
	\begin{tabular}{@{\hskip3pt}c @{\hskip3pt}c @{\hskip3pt}c @{\hskip3pt}c} 
	&\small{GP mean on $\numTrainVs$ tuples $\left(\hparam, \eparam^*\right)$}&\small{Hyper-training, $\numTrainVs$ fixed $\hparam$}&\small{Hyper-training, many sampled $\hparam$}\\
	\rotatebox{90}{\,\,\,\,\,\,\,\,\,\,\,\, \hspace{0.08\textwidth} Inferred loss}\hspace{-0.02\textwidth}&\includegraphics[width=0.31\textwidth, trim={0 0 0 2.5cm}, clip]{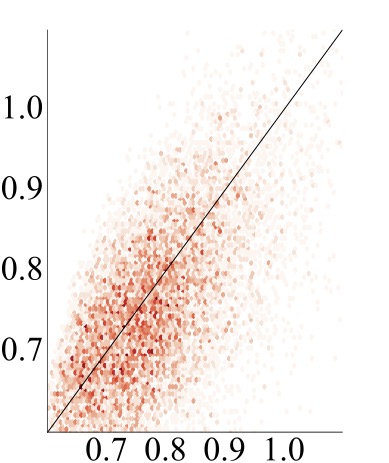}&\includegraphics[width=0.31\textwidth, trim={0 0 0 2.5cm}, clip]{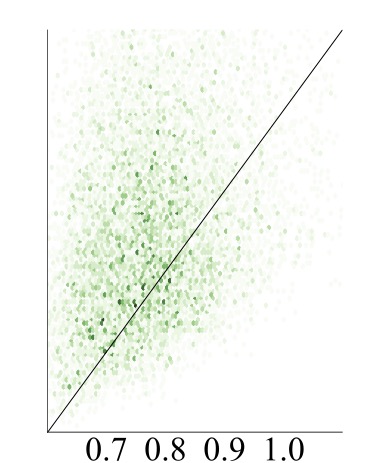}&\includegraphics[width=0.31\textwidth, trim={0 0 0 2.5cm}, clip]{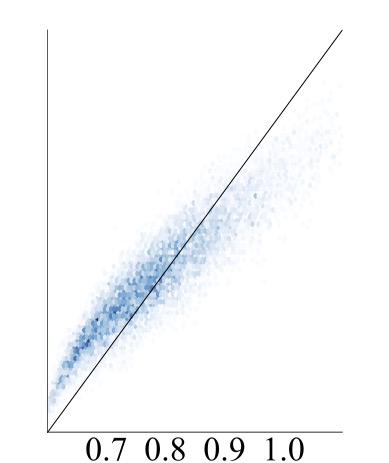}\\
	&&True loss&\\
	\rotatebox{90}{\,\,\,\,\,\,\,\,\,\,\,\,\,\,\, \hspace{0.08\textwidth} Frequency}&\includegraphics[width=0.27\textwidth, trim={0 0 0 1.1cm}, clip]{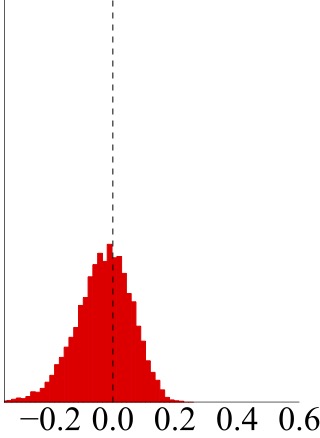}&\includegraphics[width=0.27\textwidth, trim={0 0 0 1.1cm}, clip]{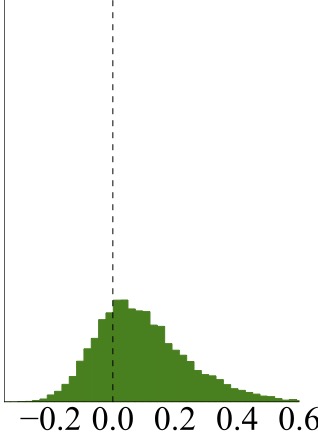}&\includegraphics[width=0.27\textwidth, trim={0 0 0 1.1cm}, clip]{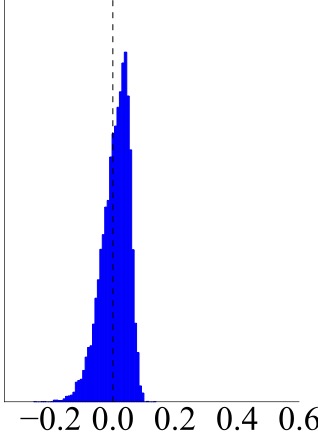}\\
	&&Inferred - true loss&\\
	\end{tabular}
	\caption[Validation loss approximation distributions]{
	Comparing three approaches to inferring the best-responding validation loss.
	\emph{First column:}
	{\color{red}A Gaussian process}, fit to $\numTrainVs$ hyperparameters, and the corresponding validation losses.
	\emph{Second column:}
	{\color{green}A hypernetwork that fits the same $\numTrainVs$ hyperparameters} and the corresponding optimized weights.
	\emph{Third column:}
	Our proposed method, {\color{blue}a hypernetwork trained with stochastically sampled hyperparameters}.
	\emph{Top row:}
	The distribution of inferred and true losses.
	The diagonal black line is where the predicted loss equals the true loss.
	\emph{Bottom row:}
	The distribution of differences between inferred and true losses.
	The Gaussian process more often under-predicts the true loss, while the hypernetwork trained on the same data tends to over-predict it.
        \textbf{Takeaway:}
        {\color{blue}Our method, in blue} has a distribution more concentrated around $0$, showing that it estimates the best-responding validation loss most accurately, and this performance is from seeing many more different hyperparameters during optimization.
        In particular, {\color{blue}our method} uses the same amount of compute as training the $\numTrainVs$ models used to fit the Gaussian process.
	\label{hypernet_fig:exp5}
	}
\vspace{-0.01\textheight}
\end{figure}

\subsection{Hyper-training and unrolled optimization}
We train models with a separate $\ell_{2}$ weight decay applied to each weight in a one-layer model to compare hyper-training with other gradient-based hyperparameter optimization methods.
The conditional hyperparameter distribution and optimizer for the hypernetwork and hyperparameters are the same as in the prior experiments.
Here, we use a hypernetwork with $\numHiddenLarge$ hidden units.
%
Figure~\ref{hypernet_fig:exp3}, top, shows that Algorithm~\ref{hypernet_algLocal} converges more quickly than the unrolled reverse-mode optimization implemented by \citet{franceschi2017forward}.
Hyper-training overfits validation data less than unrolling but reaches sub-optimal solutions -- perhaps because of limitations on how many hyperparameters can be sampled for each update.
Standard Bayesian optimization cannot be scaled to this many hyperparameters.
\textbf{Takeaway}:
Algorithm~\ref{hypernet_algLocal} can efficiently partially optimize thousands of hyperparameters.
But, unrolled optimization performed better asymptotically.

\subsection{Optimizing with deeper networks}
To see if we can optimize deeper networks with hyper-training, we optimize models with $1$, $2$, and $3$ layers and a separate $\ell_{2}$ weight decay applied to each weight.
The conditional hyperparameter distribution and optimizer for the hypernetwork and hyperparameters are the same as in the prior experiment.
Again, we select a hypernetwork with $\numHiddenLarge$ hidden units.
%
Figure~\ref{hypernet_fig:exp3}, bottom, shows that Algorithm~\ref{hypernet_algLocal} can scale to networks with multiple hidden layers and outperform hand-tuned settings.
Adding more layers improves model performance with lower training, validation, and test losses, indicating that using a separate weight decay on each weight could be useful for generalization.
The follow-up work of \citet{mackay2019self} compares other architectures, such as recurrent or convolutional networks, on various other hyperparameter choices.
%
%
%

\subsection{Estimating weights versus estimating loss}
Our approach differs from Bayesian optimization, which attempts to model the validation loss of optimized weights directly, where we try to learn to predict optimal weights.
In this experiment, we begin to unravel the reason for the better performance of our method.
Is it because of a better inductive bias or because our way can see more hyperparameter settings during optimization?

First, we constructed a hyper-training set: We optimized $\numTrainVs$ sets of weights to completion, given randomly sampled hyperparameters.
We chose $\numTrainVs$ samples because that is the regime in which we expect Gaussian process-based approaches to have the largest advantage.
We constructed an unseen set of $\numValidVs$ (optimized weight, hyperparameter) tuples generated similarly.
{\color{red}We then fit a Gaussian process (GP)} regression model with an RBF kernel from sklearn on the validation loss data.
{\color{green}A hypernetwork is fit to the same set of hyperparameters and data}.
Finally, {\color{blue}we optimize another hypernetwork using Algorithm~\ref{hypernet_algGlobal}} for the same amount of time as building the GP training set.
The two hypernetworks are linear models with the same optimizer parameters as prior experiments.

Figure~\ref{hypernet_fig:exp5} shows the distribution of prediction errors for these three models.
The Gaussian process tends to underestimate the loss.
The hypernetwork trained with the same small fixed set of examples tends to overestimate loss.
We conjecture that this is due to the hypernetwork producing bad weights in regions without enough training data.
Because the hypernetwork must provide actual weights to predict the validation loss, poorly fit regions can overestimate the loss $\Lval^{*}$.
Finally, the hypernetwork trained with Algorithm~\ref{hypernet_algGlobal} produces errors tightly centered around 0.
\textbf{Takeaway}:
A hypernetwork can learn more accurate surrogate functions than a GP for equal compute budgets because it views (noisy) evaluations of many more points.
%



\section{Conclusions and Future Work}
This chapter addressed the tuning of hyperparameters using gradient-based optimization by replacing the training optimization loop with a differentiable hypernetwork.
We also presented a simple and more scalable method that jointly optimizes hyperparameters and hypernetwork weights, allowing our method to work with manageable-sized hypernetworks.
Experimentally, we showed that hypernetworks could provide a better inductive bias for hyperparameter optimization than Gaussian processes that fit the validation loss.
%
There are many ways to extend the proposed methods, with more examples in Section~\ref{conclusion_sec:future_and_limit}.
For example, the hypernetwork could consist of several optimization iterations as an easily differentiated fine-tuning step.
Hypernetworks could be incorporated into meta-learning schemes, such as MAML~\citep{finn2017model}, which finds weights that perform various tasks after unrolling gradient descent.
We also note that optimizing thousands of hyperparameters raises the question of \emph{hyper-regularization}, or regularization of hyperparameters.

A key limitation of this work is that it does not scale to ultra-large hyperparameter regimes, where we have as many hyperparameters as neural network parameters for large networks.
In the following chapter, we look at methods that scale to this regime.
%


\section*{Acknowledgments}
We thank Matthew MacKay, Dougal Maclaurin, Daniel Flam-Shepard, Daniel Roy, and Jack Klys for helpful discussions.

\subsection*{My Contributions Towards this Paper As it Pertains to the Thesis}
    This was my first paper and was shepherded into a submission under the close guidance of David Duvenaud, for which I will forever be grateful.
    I proposed the idea while David helped code up initial versions of this in Autograd, which I fleshed out into our experiments.
    I also wrote most of the paper with much assistance from David.
    This paper has an updated arXiv version~\citep{lorraine2018stochasticARXIV} and an older workshop version~\citep{lorraine2018stochastic}.

    \chapter{Optimizing Millions of Hyperparameters by Implicit Differentiation}\label{chap:hyperparamWithIFT}
        We propose an inexpensive, gradient-based hyperparameter optimization algorithm that combines the implicit function theorem (IFT) with efficient inverse Hessian approximations.
        We present results on the relationship between IFT and differentiation through optimization, motivating our algorithm.
        We use our approach to train modern network architectures with millions of weights and \textit{millions of hyperparameters}.
        Specifically, we learn a data-augmentation network---where every weight is a hyperparameter tuned for validation performance---that outputs augmented training examples; we learn a distilled dataset where every feature in each data point is a hyperparameter; and we tune millions of regularization hyperparameters.
        Jointly tuning weights and hyperparameters with our approach is only a few times more costly in memory and compute than standard training.
    \section{Introduction}\label{ift_sec:introduction}
        Neural network generalization to unseen data crucially depends on hyperparameter choice.
        Hyperparameter optimization (HO) has a rich history~\citep{schmidhuber1987evolutionary, bengio2000gradient}, and has recently achieved successful scaling due to gradient-based optimizers~\citep{domke2012generic, maclaurin2015gradient, franceschi2017forward, franceschi2018bilevel, shaban2018truncated, finn2017model, rajeswaran2019meta, liu2018darts, grefenstette2019meta, mehra2019penalty}.
        There are dozens of regularization techniques to combine in deep learning, and each may have multiple hyperparameters~\citep{kukavcka2017regularization}.
        If we can scale \ho to have as many---or more---hyperparameters as parameters, various exciting regularization strategies exist to investigate.
        For example, we could learn a distilled dataset with a hyperparameter for every feature of each input~\citep{maclaurin2015gradient, wang2018dataset}, weights on each loss term~\citep{ren2018learning, kim2018screenernet, zhang2019autoassist}, or augmentation on each input~\citep{cubuk2018autoaugment, xie2019unsupervised}.
        \newline
        \newline

        When the hyperparameters are low-dimensional---e.g., \num{1}-\num{5} dimensions---simple methods, like random search, work; however, these break down for medium-dimensional \ho---e.g., \num{5}-\num{100} dimensions.
        We may use more scalable algorithms like Bayesian optimization~\citep{movckus1975bayesian, snoek2012practical, kandasamy2019tuning}, but this often breaks down for high-dimensional \ho---e.g., >\num{100} dimensions.
        Hypernetwork-based approaches, as in Chapter~\ref{chap:hyperparamWithHypernets}, scale further ($\sim\num{1000}$ dimensions), but break down when approaching the scale of modern neural networks.
        We can solve high-dimensional \ho problems locally with gradient-based optimizers, but this is difficult because we must differentiate through the optimized weights as a function of the hyperparameters.
        Formally, we must approximate the Jacobian of the best-response function of the parameters to the hyperparameters.
        
        We leverage the Implicit Function Theorem (IFT) to compute the optimized validation loss gradient with respect to the hyperparameters, hereafter denoted the \textit{hypergradient}.
        The IFT requires inverting the training Hessian for the \nn weights, which is infeasible for modern, deep networks.
        Thus, we propose an approximate inverse, motivated by a link to unrolled differentiation~\citep{domke2012generic} that scales to Hessians of large neural networks, is more stable than conjugate gradient~\citep{liao2018reviving, shaban2018truncated}, and only requires a constant amount of memory.
        
        Finally, when fitting many parameters, the amount of data can limit generalization.
        There are \textit{ad hoc} rules for partitioning data into training and validation sets---e.g., using \num{10}\% for validation.
        Practitioners often re-train their models from scratch on the combined training and validation partitions with optimized hyperparameters, which can provide marginal test-time performance increases.
        We verify empirically that standard partitioning and retraining procedures perform well when fitting a few hyperparameters but break down when fitting many.
        When fitting many hyperparameters, we may need a large validation partition, which makes retraining our model with optimized hyperparameters vital for strong test performance.
        
        \begin{figure*}[ht!]
            \centering
            \begin{tikzpicture}
                \centering
                \node (img)[xshift=-1cm]{\includegraphics[trim={1.5cm, .35cm, .75cm, 3.1cm},clip, width=.48\linewidth]{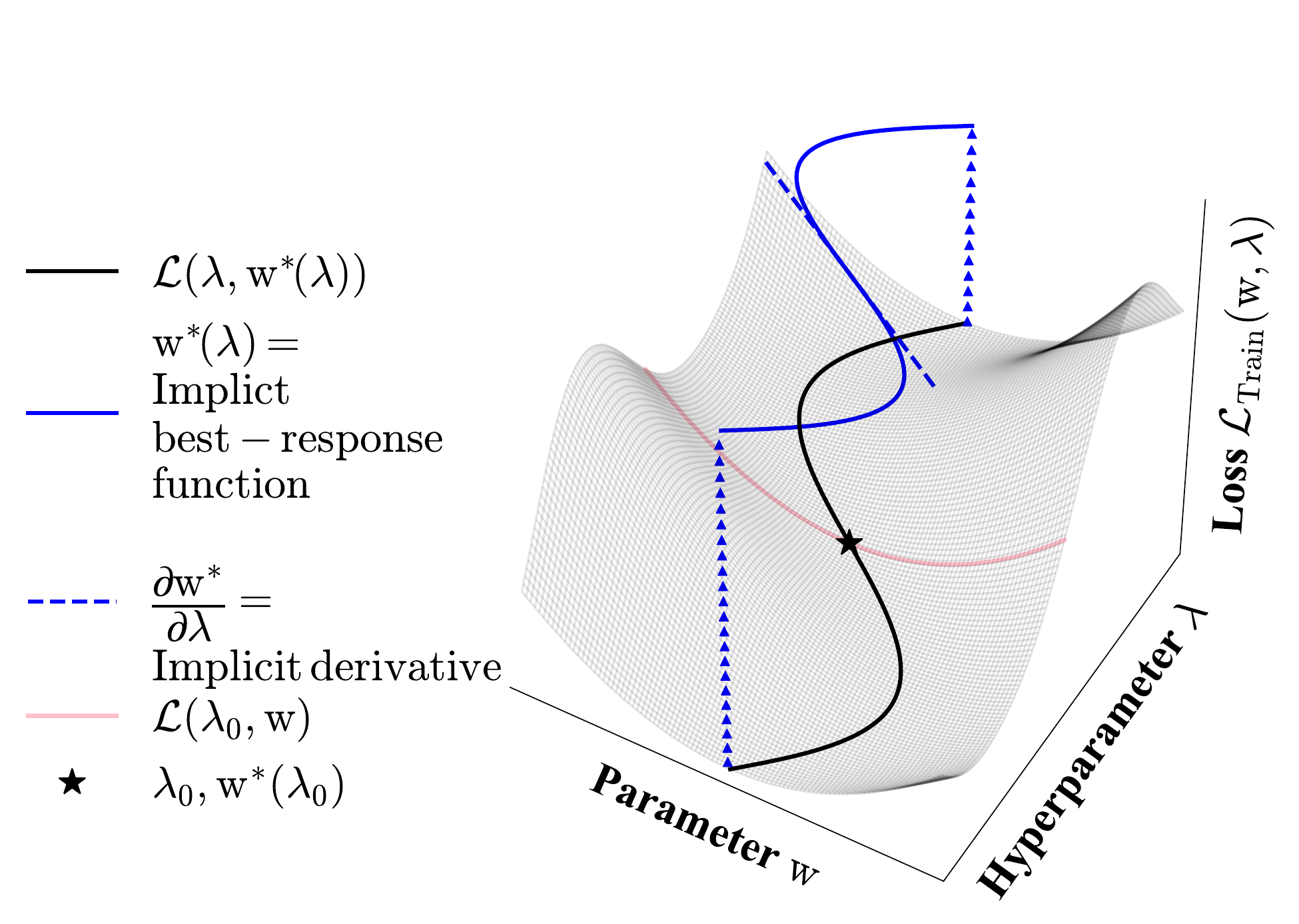}};
                
                \node (img2)[right=of img, xshift=-1.cm]{\includegraphics[trim={3.7cm, .35cm, 1.2cm, 2.6cm},clip, width=.48\linewidth]{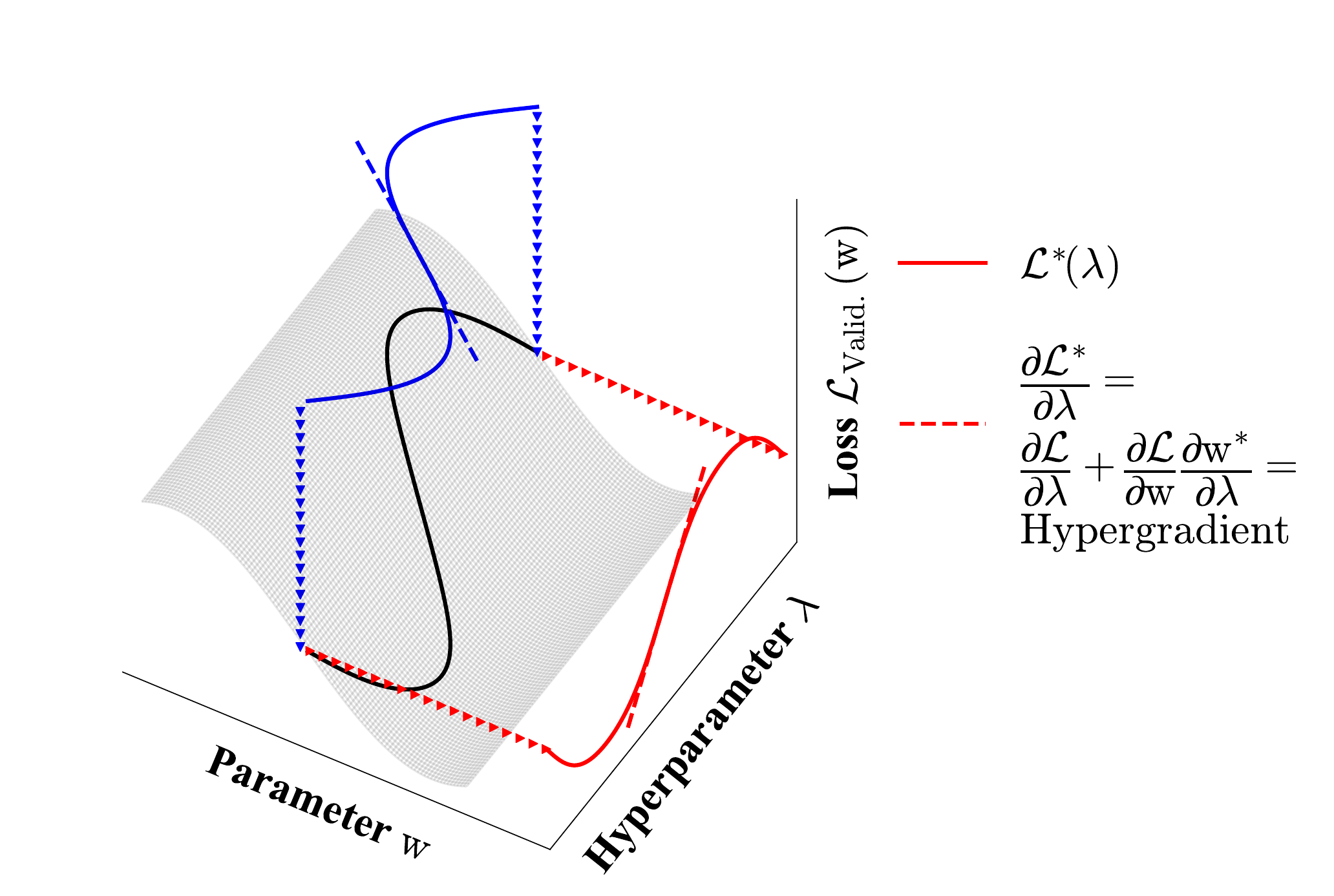}};
            \end{tikzpicture}
            \caption[Gradient-based hyperparameter optimization overview]{
               Overview of gradient-based hyperparameter optimization.
               \emph{Left:} a training loss manifold; \emph{Right:} a validation loss manifold.
               {{The implicit function $\ep^*(\hp)$ is the best-response of the weights to the hyperparameters}} and shown in {\color{blue}blue} projected onto the $(\hp, \ep)$-plane.
               We obtain our desired objective function {\color{red}$\Lval^{*}(\hp)$} when the best-response gives the network weights in the validation loss, shown projected on the hyperparameter axis in {\color{red}red}.
               The validation loss does not depend directly on the hyperparameters, as is typical in hyperparameter optimization.
               Instead, the hyperparameters only affect the validation loss by changing the weights' response.
               We show the best-response Jacobian in {\color{blue}blue} and the hypergradient in {\color{red}red}.
            }
            \label{ift_fig:train_and_val_manifolds}
        \end{figure*}
    %
    \subsection*{Contributions}
        \begin{itemize}
            \item We motivate existing inverse Hessian approximation algorithms by connecting them to iterative optimization algorithms.
            \item We scale IFT-based hyperparameter optimization to large neural architectures, including AlexNet and LSTM-based language models.
            \item We demonstrate several uses for fitting hyperparameters almost as easily as weights, including per-parameter regularization, data distillation, and learned data augmentation methods.
            \item We explore how training-validation splits should change when tuning many hyperparameters.
        \end{itemize}
    \newcommand{\sectionSymbol}{Section}
    
    \section{Overview of Proposed Algorithm}
    \label{ift_sec:background}
        There are four essential components to understanding our proposed algorithm.
        Further background is provided in Appendix~\ref{ift_app:background}, and the notation is shown in Table~\ref{ift_tab:TableOfNotation}.
        
        \textbf{1. \Ho is nested optimization:}
            $\Ltr$ and $\Lval$ denote the training and validation losses, $\ep$ the \nn weights, and $\hp$ the hyperparameters.
            We aim to find optimal hyperparameters $\hp^*$ such that the \nn minimizes the validation loss after training:
            \begin{align} \label{ift_eqn:main-equilibrium}
                \hp^* &= \argmin_{\hp} {\color{red} \Lval^*(\hp)} \text{ where } \\
                {\color{red}\Lval^*(\hp)} &= \LvalResponse \text{ and }
                \response = \argmin_{\ep} \Ltr(\hp,\ep)
            \end{align}
            As in Chapter~\ref{chap:hyperparamWithHypernets}, our implicit function $\response$ is the \emph{best-response} of the weights to the hyperparameters, and our desired objective is the best-responding validation loss {\color{red}$\Lval^{*}$ in red}.
            Again, for simplicity, we assume unique solutions to $\argmin$, and refer to non-thesis research of \citet{vicol2020implicit} for analysis in the non-unique setup.
        \newline

        \textbf{2. Hypergradients have two terms:}
            For gradient-based \ho we want the hypergradient $\hpderiv{\Lval^*(\hp)}$, which decomposes into:
            \begin{align}\label{ift_eqn:gradient-decomp}
                \begin{split}
                    &\underbrace{\hpderiv{{\color{red} \Lval^*(\hp)}}}_{\text{\tiny{hypergradient}}}
                    = \left. \left( {\color{mydarkgreen}\hpderiv{\Lval}} + \epderiv{\Lval} {\color{blue}\hpderiv{\ep^*}} \right) \right|_{\hp, \ep^*(\hp)} =\\ 
                    &{\!\!\!\!\!\! \color{mydarkgreen}\underbrace{\hpderiv{\LvalResponse}}_{\text{\tiny{hyperparam direct grad.}}}} \!\!\!+ \hphantom{A}
                    \overbrace{\!\!\!\!\!\!\!\!
                        \underbrace{\pd{\LvalResponse}{\response}}_{\text{\tiny{parameter direct grad.}}}
                        \times \!\!\!\!
                        {\color{blue}\underbrace{\responseJacobian \vphantom{\pd{A^A}{A_A}}}_{\text{\tiny{best-response Jacobian}}}}
                    \!\!\!\!\!\!\!\!\!\!\!\!\!\!}^{\text{\tiny{hyperparam indirect grad.}}}
                \end{split}
            \end{align}
        The {\color{mydarkgreen}direct gradient} is easy to compute.
        However, the indirect gradient is difficult to compute because we must account for how the optimal weights change for the hyperparameters (i.e., {\color{blue}$\responseJacobian$}).
        In \ho the {\color{mydarkgreen}direct gradient} is often identically $\textbf{\num{0}}$, necessitating an approximation of the indirect gradient to make any progress (visualized in Figure~\ref{ift_fig:train_and_val_manifolds}).
        In Chapter~\ref{chap:complexMomentum}, we look at algorithms that do not require {\color{blue}response Jacobians} for setups where the {\color{mydarkgreen}direct gradient} is non-zero.
        \newline

        \textbf{3. We can estimate the implicit best-response with the IFT:}
            We approximate the {\color{blue}best-response Jacobian}---how the optimal weights change with respect to the hyperparameters---using the IFT (Theorem~\ref{ift_thm:IFT}).
            We present the complete statement in Appendix~\ref{ift_app:IFTComplete}, but highlight the key assumptions and results here.
            \begin{thm}[Cauchy, Implicit Function Theorem]
                If for some \textnormal{$\left(\hp', \ep'\right), \left. \trainGrad \right|_{\hp', \ep'} = 0$} and regularity conditions are satisfied, then surrounding \textnormal{$\left(\hp', \ep'\right)$} there is a function \textnormal{$\ep^{*}\left(\hp\right)$} such that \textnormal{$\left. \trainGrad \right|_{\hp, \response} = 0$} and we have:
                \textnormal{
                    \begin{equation}\label{ift_eqn:response-approximation}
                        \!\!\left. {\color{blue} \hpderiv{\ep^*}} \!\right|_{\hp'}\!
                        = \!-{\color{magenta} \Big[ \!\!\! \!\!\! \underbrace{\trainHess{\ep}}_{\text{\tiny training Hessian}} \!\!\! \!\!\! \Big]^{-1}}  \times \!\!\!\!\!\!
                        \underbrace{\trainMixed{\ep}}_{\text{training mixed partials}} \!\!\!\!\!\!\!\!\!\!\!\! \Big|_{\hp' \!, \ep^*\!(\hp')}
                    \end{equation}
                }
                \label{ift_thm:IFT}
            \end{thm}
            The condition $\left. \trainGrad \right|_{\hp'  ,  \ep'} = 0$ is equivalent to $\hp', \ep'$ being a fixed point of the training gradient field.
            Since $\ep^*\left(\hp'\right)$ is a fixed point of the training gradient field, we can leverage the IFT to evaluate the {\color{blue}best-response Jacobian} locally.
            We only have access to an approximation of the true best-response---denoted $\widehat{\ep^*}$---which we can find with gradient descent.
            \newline
        
        \textbf{4. Tractable inverse Hessian approximations:}
            To exactly invert a general $m \times m$ Hessian, we often require $\mathcal{O}(m^3)$ operations, which is intractable for the matrix in Equation~\ref{ift_eqn:response-approximation} for modern \nns.
            We can efficiently approximate the inverse with the Neumann series:
            \begin{equation}\label{ift_eq:neumann}
                {\color{magenta}\left[ \trainHess{} \right]^{-1}} = {\color{magenta}\lim_{i \to \infty} \alpha \sum_{j = 0}^{i} \left[ \identity - \alpha \trainHess{\ep} \right]^j}
            \end{equation}
            In our exposition, we assume that $\alpha=1$ for simplicity, but in practice, we reuse the learning rate for the inner optimization.
            In \sectionSymbol~\ref{ift_sec:theory}, we show that unrolling differentiation for $i$ steps around locally optimal weights $\ep^*$ is equivalent to approximating the inverse with the first $i$ terms in the Neumann series.
            We then show how to make this approximation without instantiating any matrices using efficient vector-Hessian products~\citep{pearlmutter1994fast}.
            \begin{figure}
                \vspace{-0.02\textheight}
                \begin{align*}
                    \overbrace{\drawmatrix[fill=red, height=.1, width=.5]\!}^{\hpderiv{{\color{red}\Lval^*}}}
                    =& \,\,\,\,\, \,\,
                    \overbrace{\drawmatrix[height=.1, width=.5, fill=mydarkgreen]\!}^{\color{mydarkgreen}\hpderiv{\Lval}} \,\,\,\,\,\,\,\, + \,\,\,\,\,\,\,
                    \overbrace{\drawmatrix[height=.1]\!}^{\epderiv{\Lval}} \; \,\,\,\,\,\,
                    \overbrace{\drawmatrix[width=.5, fill=blue]\!}^{\color{blue} \smash{\hpderiv{\ep^*}}\vphantom{A_{A_{A_A}}}} 
                    \\
                    =& \,\,\,\,\,\,\,\,
                    \overbrace{\drawmatrix[height=.1, width=.5, fill=mydarkgreen]\!}^{\color{mydarkgreen}\hpderiv{\Lval}} \,\,\,\,\,\,\, +
                    {\color{orange}\underbrace{
                        \overbrace{\drawmatrix[height=.1]\!}^{\color{black}\epderiv{\Lval}} \;
                        \overbrace{\drawmatrix[fill=magenta]\!}^{\color{magenta}\hphantom{a} -\big[\! \smash{\trainHess{\ep}} \!\big]^{\!-\!1}} \;
                    \!\!\!\!\!\!}_{\text{vector-inverse-Hessian product}}}\,\,\,\,
                    \overbrace{\drawmatrix[width=.5]\!}^{\smash{\trainMixed{\ep}}\vphantom{A_{A_A}}}
                    \\
                    =& \,\,\,\,\,\,\,\,
                    \overbrace{\drawmatrix[height=.1, width=.5, fill=mydarkgreen]\!}^{\color{mydarkgreen}\hpderiv{\Lval}} \,\,\,\,\,\,\, + \,\,\,\,\,\,\,\,\,\,\,\,\,\,\,\,\,\,\,\,\,\,\,\,\,\,\,\,
                    \underbrace{\!\!\!\!\!\!
                        {\color{orange} \overbrace{\drawmatrix[height=.1, fill=orange]\!}^{\,\,\,\,\epderiv{\Lval}\times -\big[\! \trainHess{\ep} \!\big]^{\!-\!1}}} \;
                        \overbrace{\drawmatrix[width=.5]\!}^{\smash{\trainMixed{\ep}}\vphantom{A_{A_A}}}
                    \!\!\!}_{\text{vector-Jacobian product}}
                \end{align*}
                \vspace{-0.01\textheight}
                \caption[Hypergradient computation overview]{
                    {We visualize the hypergradient computation, which can be performed efficiently using vector-Jacobian products if provided a cheap approximation to the {\color{orange}vector-inverse-Hessian product}.}
                }
                \label{ift_fig:equations}
                \vspace{-0.01\textheight}
            \end{figure}
            \begin{algorithm}
                \caption{Gradient-based \ho for $\hp^*, \ep^*\left(\hp^*\right)$}
                \label{ift_alg:joint_train}
                \begin{algorithmic}[1]
                    \State Initialize hyperparameters $\hp'$ and weights $\ep'$ 
                    \While{\text{not converged}}
                        \For{$k = 1 \dots N$} 
                            \State $\ep' \mineq \left. \alpha \cdot \trainGrad \right|_{\hp', \ep'}$ 
                        \EndFor
                        \State $\hp' \mineq {\color{red}\texttt{hypergradient}\left(\Lval, \Ltr, \hp', \ep'\right)}$ 
                    \EndWhile
                    \State \Return $\hp', \ep'$ \Comment{$\hp^*, \ep^*\left(\hp^*\right)$ from \textbf{Equation~\ref{ift_eqn:main-equilibrium}}}
                \end{algorithmic}
            \end{algorithm}
            \begin{algorithm}
                \caption[Hypergradient computation]{{\color{red}\texttt{hypergradient}$\left(\Lval, \Ltr, \hp', \ep'\right)$}}
                \label{ift_alg:hyper_gradient}
                \begin{algorithmic}[1]
                    \State $\textbf{v}_1 = \left. \epderiv{\Lval} \right|_{\hp', \ep'}$ 
                    \State {\color{orange}$\textbf{v}_2 = \texttt{approxInverseHVP} \left(\textbf{v}_1, \epderiv{\Ltr}\right)$} 
                    \State $\textbf{v}_3 = \texttt{grad}\left(\epderiv{\Ltr}, \hp, \texttt{grad\_outputs}=\textbf{v}_2\right)$
                    \State \Return $\left. \color{mydarkgreen}\hpderiv{\Lval} \right|_{\hp', \ep'} - \textbf{v}_3$
                    \Comment{Return to \textbf{Algorithm~\ref{ift_alg:joint_train}}}
                \end{algorithmic}
            \end{algorithm}
            \begin{algorithm}[h!]
                \caption[Neumann approximation of vector-inverse-Hessian product]{{\color{orange}\texttt{approxInverseHVP}$\left(\textbf{v}, \textbf{f}\right)$}: Neumann approximation of vector-inverse-Hessian  $\textbf{v}\left[\epderiv{\textbf{f}}\right]^{-1}$}\label{ift_alg:approxInverseHVP}
                \begin{algorithmic}[1]
                    \State Initialize sum $\textbf{p} = \textbf{v}$
                    \For{$j = 1 \dots i$}
                        \State $\textbf{v} \mineq \alpha \cdot \texttt{grad}\left(\textbf{f}, \ep, \texttt{grad\_outputs}=\textbf{v}\right)$ 
                        \State $\textbf{p} \pluseq \textbf{v}$
                    \EndFor
                    \State \Return $\alpha \textbf{p}$  \Comment{Return to \textbf{Algorithm~\ref{ift_alg:hyper_gradient}}}.
                \end{algorithmic}
            \end{algorithm}
            
            \vspace{-0.2cm}
            \subsection{Proposed Algorithms}
                We outline our method in Algorithms~\ref{ift_alg:joint_train},~\ref{ift_alg:hyper_gradient}, and~\ref{ift_alg:approxInverseHVP}, where $\alpha$ denotes the learning rate.
                Algorithm~\ref{ift_alg:approxInverseHVP} is also shown in \citet{liao2018reviving} and is a special case of algorithms from \cite{christianson1998reverse}.
                We visualize the hypergradient computation in Figure~\ref{ift_fig:equations}.
            
    \vspace{-0.3cm}
    \section{Related Work}\label{ift_sec:related-work}
    \vspace{-0.1cm}
        \begin{table*}
            \vspace{-0.01\textheight}
            \footnotesize
            \begin{center}
                \begin{tabular}{c | c | c | r}
                    \textbf{Method} \vphantom{\Big |}
                    & \textbf{Steps}
                    & \textbf{Eval.}
                    & \textbf{Hypergradient Approximation\hphantom{AAAAAA}} \\
                \hline
                    Exact IFT
                    &$\infty$
                    &$\ep^*(\hp)$
                    &${\color{mydarkgreen}\hpderiv{\Lval}} - \epderiv{\Lval}\times$\hfill
                        $\left. {\color{magenta}\left[ \trainHess{\ep} \right]^{-1}} \trainMixed{\ep} \right|_{\response}$\,\,\,\\
                    
                    Unrolled Diff.
                    &$i$
                    &$\ep_0$
                    &${\color{mydarkgreen}\hpderiv{\Lval}} - \epderiv{\Lval}\times$\hfill
                        $\sum_{j\leq i}  \left[  \prod_{k<j} \identity -  \left. \trainHess{\ep_{i-k}(\hp)} \right|_{\ep_{i-k}} \! \right] \! \left. \trainMixed{\ep_{i-j}(\hp)} \vphantom{A^{A^{A^A}}}\right|_{\ep_{i-j}\hphantom{\,\,}}$\\
                    
                    $L$-Step Truncated Unrolled Diff.
                    &$i$
                    &$\ep_L$
                    &${\color{mydarkgreen}\hpderiv{\Lval}} - \epderiv{\Lval}\times$\hfill
                        $\sum_{L \leq j\leq i} \! \left[ \! \prod_{k<j} \identity -  \left. \trainHess{\ep_{i-k}(\hp)} \right|_{\ep_{i-k}} \! \right] \! \left. \trainMixed{\ep_{i-j}(\hp)} \vphantom{A^{A^{A^A}}} \right|_{\ep_{i-j}\hphantom{\,\,}}$\\
                    
                    \citet{larsen1996design}
                    &$\infty$
                    &$\widehat{\ep^*}\!(\hp)$
                    &${\color{mydarkgreen}\hpderiv{\Lval}} - \epderiv{\Lval}\times$\hfill
                        $\left. {\color{magenta}\left[\! \trainGrad \! \trainGrad^{\! \! \transpose} \!\right]^{-1}} \trainMixed{\ep} \right|_{\widehat{\ep^*}(\hp)}$\\
                    
                    \citet{bengio2000gradient}
                    &$\infty$
                    &$\widehat{\ep^*}\!(\hp)$
                    &${\color{mydarkgreen}\hpderiv{\Lval}} - \epderiv{\Lval}\times$\hfill
                        $\left. {\color{magenta}\left[ \trainHess{\ep} \right]^{-1}} \trainMixed{\ep} \right|_{\widehat{\ep^*}(\hp)}$\\
                    
                    $T1 - T2$
                    &$1$
                    &$\widehat{\ep^*}\!(\hp)$
                    &${\color{mydarkgreen}\hpderiv{\Lval}} - \epderiv{\Lval}\times$\hfill
                    $\left.{\color{magenta}\left[\identity\right]^{-1}} \trainMixed{\ep} \right|_{\widehat{\ep^*}(\hp)}$\\
                    
                    \textbf{Ours}
                    &$i$
                    &$\widehat{\ep^*}\!(\hp)$
                    &${\color{mydarkgreen}\hpderiv{\Lval}} - \epderiv{\Lval}\times$\hfill
                        $\left. {\color{magenta} \left( \sum_{j \leq i} \left[\identity - \trainHess{\ep(\hp)}\right]^j \right)} \trainMixed{\ep(\hp)} \right|_{\widehat{\ep^*}(\hp)}$\\
                    
                    Conjugate Gradient (CG) $\approx$
                    &-
                    &$\widehat{\ep^*}\!(\hp)$
                    &${\color{mydarkgreen}\hpderiv{\Lval}} \,\,-$\hfill
                        $\left. {\color{orange}\left(\argmin_{\textbf{x}}\| \textbf{x} \trainHess{\ep} - \epderiv{\Lval} \|\right)} \trainMixed{\ep(\hp)} \right|_{\widehat{\ep^*}(\hp)}$\\
                    
                    Hypernetwork
                    &-
                    &-
                    &${\color{mydarkgreen}\hpderiv{\Lval}} + \epderiv{\Lval}\times$\hfill
                        ${\color{blue}\hpderiv{\ep_{\phi}^*}}$ where $\ep_{\phi}^*(\hp) = \argmin_{\phi} \Ltr(\hp, \ep_{\phi}(\hp))$\\
                    
                    Bayesian Optimization
                    &-
                    &-
                    &$\hpderiv{\mathbb{E}\left[{\color{red}\Lval^{*}}\right]}$ where ${\color{red}\Lval^{*}} \sim \textnormal{Gaussian-Process}(\{\hp_i, \Lval(\hp_i, \ep^{*}(\hp_i))\})$\\
                \end{tabular}
            \end{center}
            \caption[Overviewing methods to approximate hypergradients]{
                {An overview of methods to approximate hypergradients.}
                Some methods can be viewed as using an approximate inverse in the IFT or differentiating through optimization at an evaluation point.
                Here, $\widehat{\ep^*}(\hp)$ approximates the best-response at a fixed $\hp$, often found with gradient descent.
            }
            \label{ift_tab:response_approxes}
        \end{table*}
        \begin{table}[h!]
            \begin{center}
                \begin{tabular}{c|c}
                    \textbf{Method} & \textbf{Memory Cost} \\
                    \hline
                    Differentiation through Optimization~\citep{domke2012generic, maclaurin2015gradient}
                        & $\mathcal{O}\left(PI + H\right)$ \\
                    Linear Hypernetwork~\citep{lorraine2018stochastic}
                        & $\mathcal{O}\left(PH\right)$ \\
                    Self-Tuning Nets (STN)~\citep{mackay2019self}
                        & $\mathcal{O}\left(\left(P+H\right)K\right)$ \\
                    Neumann/CG IFT
                        & $\mathcal{O}\left(P + H\right)$
                \end{tabular}
            \end{center}
            \caption[Gradient based hyperparameter optimization cost comparison]{
                {Gradient-based methods for \ho.}
                Differentiation through optimization scales with the number of unrolled iterations $I$; the STN scales with bottleneck size $K$, while our method only scales with the weight and hyperparameter sizes $P$ and $H$.
            }
            \label{ift_tab:hyper_methods}
        \end{table}
        \textbf{Implicit Function Theorem.}
            The IFT has been used for optimization in nested optimization problems~\citep{ochs2015bilevel, anonymous2019ridge, lee2019meta}, backpropagating through arbitrarily long RNNs~\citep{liao2018reviving}, $k$-fold cross-validation~\citep{beirami2017optimal}, and influence functions~\citep{koh2017understanding}.
            Early work applied the IFT to regularization by explicitly computing the Hessian (or Gauss-Newton) inverse~\citep{larsen1996design, bengio2000gradient}.
            In \cite{luketina2016scalable}, the identity matrix approximates the IFT's inverse Hessian.
            HOAG~\citep{pedregosa2016hyperparameter} uses conjugate gradient (CG) to approximately invert the Hessian and provides convergence results given tolerances on the optimal parameter and inverse.
            In iMAML~\citep{rajeswaran2019meta}, a center to the weights is fit to perform on multiple tasks, where we fit to perform on the validation loss.
            In DEQ~\citep{bai2019deep}, implicit differentiation is used to add differentiable fixed-point methods to \nn architectures.
            We use a Neumann approximation for the inverse Hessian instead of CG~\citep{pedregosa2016hyperparameter,rajeswaran2019meta} or the identity.
        
        \textbf{Approximate inversion algorithms.}
            CG is difficult to scale to modern, deep \nns.
            We use a Neumann inverse approximation, which is a more stable alternative to CG in \nns~\citep{liao2018reviving, shaban2018truncated} and useful in stochastic settings~\citep{agarwal2017second}.
            The stability is motivated by connections between the Neumann series and unrolled differentiation~\citep{shaban2018truncated}.
            Alternatively, we could use prior knowledge about the \nn structure to aid in the inversion---e.g., as with KFAC~\citep{martens2015optimizing}.
            Approximating the Hessian with the Gauss-Newton matrix or the Fisher information matrix~\citep{larsen1996design} is possible.
            Various works use an identity approximation to the inverse, which is equivalent to \num{1}-step unrolled differentiation~\citep{luketina2016scalable, ren2018learning, balaji2018metareg, liu2018darts, finn2017model, shavitt2018regularization, nichol2018first, mescheder2017numerics}.
        \newline
        \newline
        
        \textbf{Unrolled differentiation for \ho.}
            A key difficulty in nested optimization is approximating how the optimized inner parameters (i.e., \nn weights) change with respect to the outer parameters (i.e., hyperparameters).
            We often optimize the inner parameters with gradient descent so that we can differentiate through this optimization.
            Differentiation through optimization has been applied to nested optimization problems by \cite{domke2012generic}, was scaled to \ho for \nns by \cite{maclaurin2015gradient}, and has been applied to various applications, such as learning optimizers~\citep{andrychowicz2016learning}.
            \cite{franceschi2018bilevel} provides convergence results for this class of algorithms, while \cite{franceschi2017forward} discusses forward- and reverse-mode variants.
            
            As the number of gradient steps we backpropagate through increases, so does the memory and computational cost.
            Often, gradient descent does not exactly minimize our objective after a finite number of steps, but only approaches a local minimum.
            Thus, we may have to unroll the optimization infeasibly far to see how the hyperparameters affect the local minima.
            Unrolling a small number of steps can be crucial for performance but may induce bias~\citep{wu2018understanding}.
            \cite{shaban2018truncated} discusses connections between unrolling and the IFT and proposes to unroll only the last $L$-steps.
            DrMAD~\citep{fu2016drmad} proposes an interpolation scheme for the optimization to save memory.
        
        We compare the hypergradient approximations in Table~\ref{ift_tab:response_approxes} and the memory costs of gradient-based \ho methods in Table~\ref{ift_tab:hyper_methods}.
        We survey gradient-free \ho in Appendix~\ref{ift_app:related_work}.
    
    \section{Method}\label{ift_sec:theory}
        In this section, we discuss how \ho is a uniquely challenging nested optimization problem and how to combine the benefits of the IFT and unrolled differentiation.
        
        \subsection{Hyperparameter Optimization is Pure-Response}
            Equation~\ref{ift_eqn:gradient-decomp} shows that the hypergradient decomposes into a \textit{\color{mydarkgreen}direct} and \textit{indirect gradient}.
            The bottleneck in hypergradient computation is usually finding the indirect gradient because we must consider how the optimized parameters vary for the hyperparameters.
            A simple optimization approach is to neglect the indirect gradient and only use the {\color{mydarkgreen}direct gradient}.
            This can be useful in zero-sum games like GANs~\citep{goodfellow2014generative} because they always have a non-zero direct term.
            
            However, using only the {\color{mydarkgreen}direct gradient} does not work in general games~\citep{balduzzi2018mechanics}.
            In particular, it does not work for \ho because the {\color{mydarkgreen}direct gradient} is identically \num{0} when the hyperparameters $\hp$ only influence the validation loss by changing the optimized weights $\ep^*(\hp)$.
            For example, if we use regularization like weight decay when computing the training loss but not the validation loss, then the {\color{mydarkgreen}direct gradient} is always \num{0}.
            
            If the {\color{mydarkgreen}direct gradient} is identically \num{0}, we call the game \textit{pure-response}.
            Pure-response games are difficult nested optimization problems for gradient-based methods because we cannot simply use the {\color{mydarkgreen}direct gradient} like in simultaneous SGD.
            So, we must approximate the indirect gradient.
        
        \newcommand{\smallSpace}{\hspace{0.03\textwidth}}
        \newcommand{\medSpace}{\hspace{0.09\textwidth}}
        \newcommand{\epA}{\ep_{ \infty}}
        \subsection{The Relationship Between Unrolled Optimization and the IFT}
            Here, we (a) introduce the recurrence relation that arises when we unroll SGD optimization, (b) give a formula for the derivative of the recurrence, and (c) establish conditions for the recurrence to converge.
            Notably, we show that the fixed points of the recurrence recover the IFT solution.
            We use these results to motivate a computationally tractable approximation scheme to the IFT solution.
            Proofs of all the results are in Appendix~\ref{ift_app:proofs}.
            Unrolling SGD optimization at an initialization $\ep_0$ gives us the recurrence:
            \begin{equation}\label{ift_eqn:sgd_recurrence}
                \ep_{i+1}(\hp) = \lyapunovfixedPointOp(\hp, \ep_{i}) = \ep_{i}(\hp) - \alpha \epderiv{\Ltr(\hp, \ep_{i}(\hp))}
            \end{equation}
            We provide a formula for the derivative of the recurrence, to show that it converges to the IFT under some conditions.
            \begin{lemma*}\label{ift_lemma:unrolled_grad}
                Given the recurrence from unrolling SGD optimization in Equation~\ref{ift_eqn:sgd_recurrence}, we have:
                \textnormal{
                    \begin{equation}
                        \hpderiv{\ep_{i+1}} = -\alpha \sum_{j\leq i}  \left[  \prod_{k<j}  \identity  -  \left. \alpha \trainHess{\ep_{i-k}(\hp)} \right|_{\hp, \ep_{i-k}(\hp)}  \right]  \left. \trainMixed{\ep_{i-j}(\hp)} \right|_{\hp, \ep_{i-j}(\hp)}
                    \end{equation}
                }
            \end{lemma*}
            This recurrence converges to a fixed point if the Jacobian of the fixed-point operator $\epderiv{\lyapunovfixedPointOp} = \identity - \alpha \trainHess{\ep}$ is contractive by the Banach Fixed-Point Theorem~\citep{banach1922operations}.
            Theorem~\ref{ift_thm:unrolled_to_IFT} shows that the recurrence converges to the IFT if we start with locally optimal weights $\ep_0 = \ep^*(\hparam)$, and the Jacobian of the fixed-point operator $\epderiv{\lyapunovfixedPointOp}$ is contractive.
            We leverage that if an operator $U$ is contractive, then the Neumann series $\sum_{i = 0}^{\infty}\boldsymbol{U}^k = (\identity - \boldsymbol{U})^{-1}$.
            \begin{thm}[Neumann-SGD]
                Given the recurrence from unrolling SGD optimization in Equation~\ref{ift_eqn:sgd_recurrence}, if \textnormal{$\ep_0 = \ep^*(\hp)$}:
                \textnormal{
                    \begin{equation}
                        \hpderiv{\ep_{i+1}} = -\left. {\color{magenta}\left( \alpha \sum_{j \leq i} \left[\identity -\alpha \trainHess{\ep(\hp)}\right]^j \right)} \trainMixed{\ep(\hp)} \right|_{\ep^*(\hp)}
                    \end{equation}
                }
                and if \textnormal{$\identity - \alpha \trainHess{\ep(\hp)}$} is contractive:
                \textnormal{
                    \begin{equation}
                        \lim_{i\to\infty} \hpderiv{\ep_{i+1}} = -\left. {\color{magenta}\left[ \trainHess{\ep(\hp)} \right]^{-1}} \trainMixed{\ep(\hp)} \right|_{\ep^*(\hp)}
                    \end{equation}
                }
                \label{ift_thm:unrolled_to_IFT}
            \end{thm}
            \vspace{-0.02\textheight}
            This result is also shown in~\cite{shaban2018truncated}, but they use a different approximation for computing the hypergradient---see Table~\ref{ift_tab:response_approxes}.
            Instead, we use the following best-response Jacobian approximation, where $i$ controls the trade-off between computation and error bounds:
            \begin{align}\label{ift_eqn:inv_approx}
                \hpderiv{\ep^*} 
                &\approx \left. -{\color{magenta}\left( \alpha \sum_{j \leq i} \left[\identity - 
                \alpha \trainHess{\ep(\hp)}\right]^j \right)} \trainMixed{\ep(\hp)} \right|_{\ep^*(\hp)}
            \end{align}
            \citet{shaban2018truncated} use an approximation that scales memory linearly in $i$, while ours is constant.
            We save memory because we reuse last $\ep$ $i$ times, while \citet{shaban2018truncated} needs the last $i$ $\ep$'s.
            Scaling the Hessian by the learning rate $\alpha$ is key for convergence.
            Our algorithm has the following main advantages relative to other approaches:
            \begin{itemize}
                \item It requires a constant amount of memory, unlike other unrolled differentiation methods~\citep{maclaurin2015gradient, shaban2018truncated}.
                \item It is more stable than conjugate gradient~\citep{liao2018reviving}.
            \end{itemize}

        \subsection{Scope and Limitations}
            The assumptions necessary to apply the IFT in our setup are as follows:
            (1) $\Lval: \hdom \times \edom \to \Real$ is differentiable, 
            (2) $\Ltr: \hdom \times \edom \to \Real$ is twice differentiable with an invertible Hessian at $\response$, and
            (3) $\ep^*: \hdom \to\edom$ is differentiable.
            
            We need continuous hyperparameters to use gradient-based optimization, but many discrete hyperparameters (e.g., number of hidden units) have continuous relaxations~\citep{maddison2016concrete, jang2016categorical}.
            Also, we can only optimize hyperparameters that change the loss manifold, so our approach is not straightforwardly applicable to optimizer hyperparameters.

            To exactly compute hypergradients, we must find $\left(\hp',\ep'\right)$ such that $\left. \trainGrad\right|_{\hp', \ep'}=0$, which we can only solve to a tolerance with an approximate solution denoted $\widehat{\ep^*}\left(\hp\right)$.
            \cite{pedregosa2016hyperparameter} shows results for how an error in the approximate solution affects the inversion.

    \section{Experiments}\label{ift_sec:experiments}
        We first compare the properties of Neumann inverse approximations and conjugate gradient with experiments similar to \cite{liao2018reviving, maclaurin2015gradient, shaban2018truncated, pedregosa2016hyperparameter}.
        Then, we demonstrate that our proposed approach can {overfit} the validation data with small training and validation sets.
        Finally, we apply our approach to high-dimensional \ho tasks: (1) {dataset distillation}; (2) learning a data augmentation network; and (3) tuning regularization parameters for an LSTM language model.
        
        \Ho algorithms that are not based on implicit differentiation or differentiation through optimization---such as \citet{jaderberg2017population, jamieson2016non, bergstra2012random, kumar2018parallel, li2016hyperband, snoek2012practical}---do not scale to the high-dimensional hyperparameters we use.
        Thus, we cannot sensibly compare to them for these high-dimensional problems.
        
        \subsection{Approximate Inversion Algorithms}
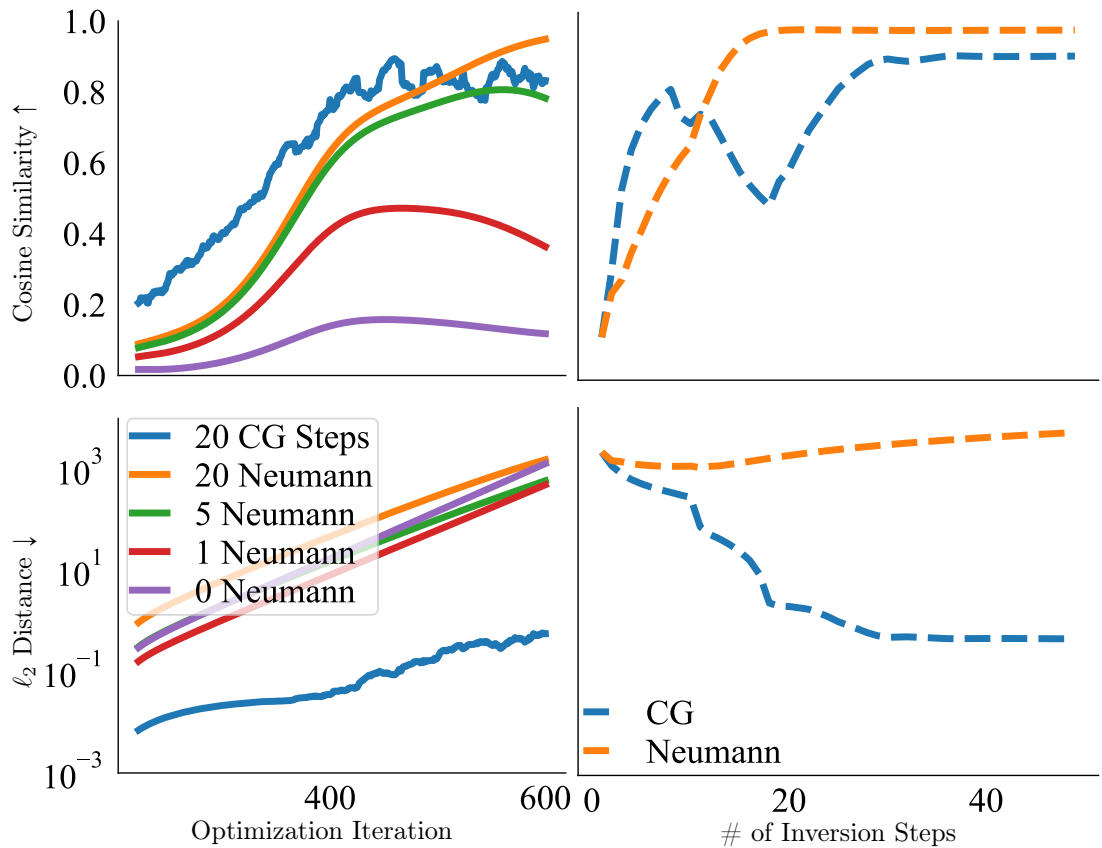
\begin{figure}[ht!]
                \centering
                \begin{tikzpicture}
                    \centering
                    \node (img){\includegraphics[trim={.25cm .25cm 0.25cm .25cm},clip, width=.4625\linewidth]{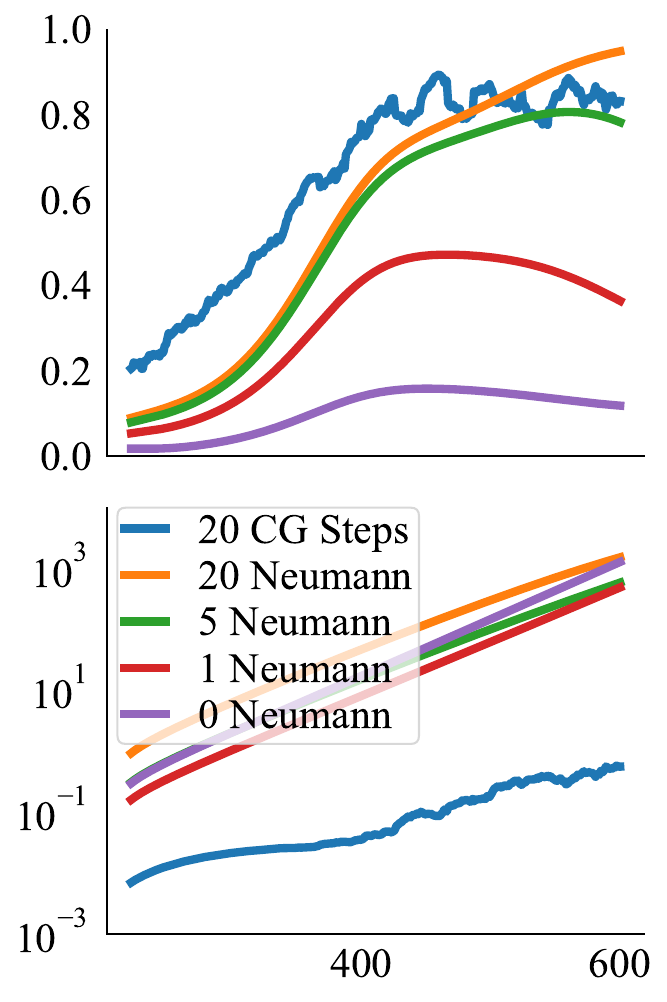}};
                    \node[left=of img, node distance=0cm, rotate=90, xshift=4.25cm, yshift=-.9cm, font=\color{black}] {Cosine Similarity $\uparrow$};
                    \node[left=of img, node distance=0cm, rotate=90, xshift=-1.5cm, yshift=-.9cm, font=\color{black}] {$\ell_2$ Distance $\downarrow$};
                    \node[below=of img, node distance=0cm, yshift=1.2cm, xshift=.2cm, font=\color{black}] {Optimization Iteration};
                    
                    \node (img2)[right=of img, xshift=-1.25cm]{\includegraphics[trim={.25cm .25cm 0.25cm .25cm},clip, width=.465\linewidth]{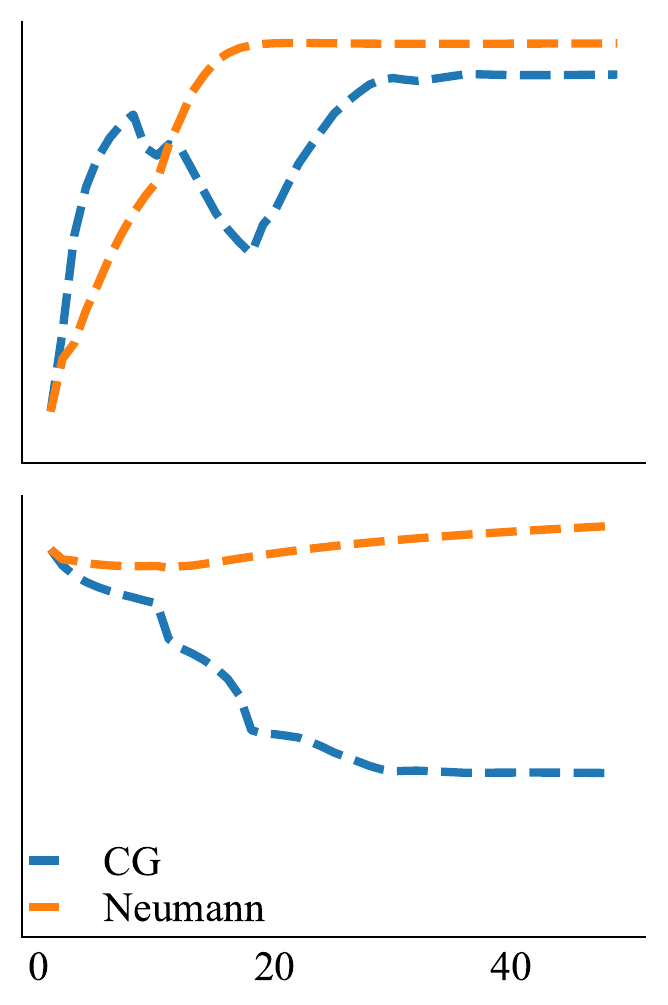}};
                    \node[below=of img2, node distance=0cm, yshift=1.2cm,font=\color{black}] {\# of Inversion Steps};
                \end{tikzpicture}
                \caption[Neumann matrix inverse approximations quantitative comparison]{
                    {Comparing approximate hypergradients for inverse Hessian approximations to true hypergradients.}
                    The Neumann scheme can achieve a greater cosine similarity than CG but a larger $\ell_2$ distance for equal steps.
                }
                \label{ift_fig:inverse_error}
            \end{figure}

            In Figure~\ref{ift_fig:inverse_error}, we investigate how close various approximations are to the true inverse.
            We calculate the distance between the approximate hypergradient and the true hypergradient.
            We can only do this for small-scale problems because we need the exact inverse for the true hypergradient.
            We use a linear network on the Boston housing dataset~\citep{harrison1978hedonic}, which makes finding the best-response $\ep^*$ and inverse training Hessian feasible.
            
            We measure the cosine similarity, which tells us how accurate the direction is, and the $\ell_2$ (Euclidean) distance between the approximate and true hypergradients.
            The Neumann approximation performs better than CG in cosine similarity if we take enough \ho steps, while CG always performs better for $\ell_2$ distance.

\begin{figure}[ht!]
                \centering
                \begin{tikzpicture}
                    \centering
                    \node[draw,inner sep=1pt] (img){\includegraphics[trim={3.5cm 1.4cm 4.2cm 2.0cm},clip, width=.285\linewidth]{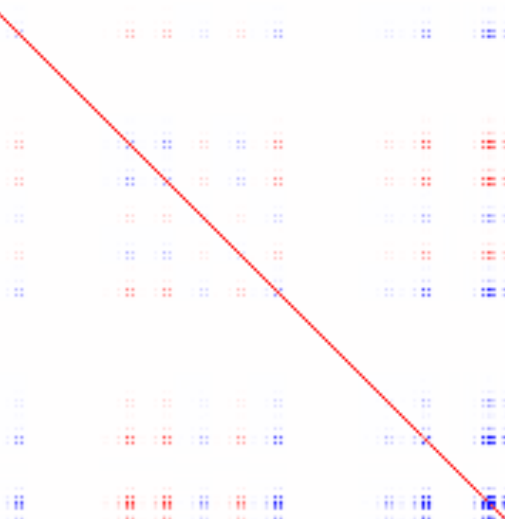}};
                    \node[below=of img, node distance=0cm, yshift=1.0cm, xshift=-.1cm,font=\color{black}] {1 Neumann};
                    
                    \node[draw,inner sep=1pt] (img2)[right=of img, xshift=-.98cm]{\includegraphics[trim={3.5cm 1.4cm 4.20cm 2.0cm},clip, width=.285\linewidth]{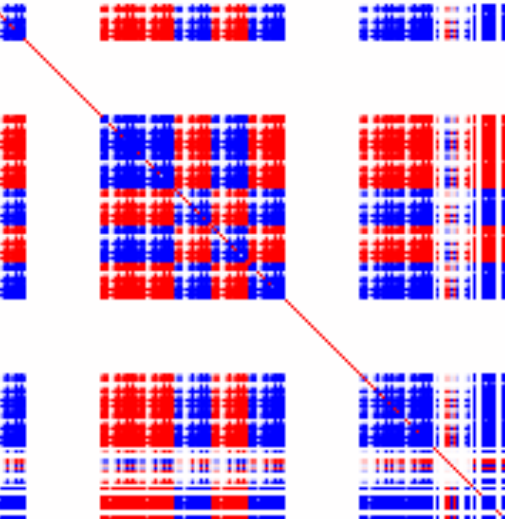}};
                    \node[below=of img2, node distance=0cm, yshift=1.cm,xshift=.1cm,font=\color{black}] {5 Neumann};
                    
                    \node[draw,inner sep=1pt] (img3)[right=of img2, xshift=-.98cm]{\includegraphics[trim={3.5cm 1.4cm 4.20cm 2.0cm},clip, width=.285\linewidth]{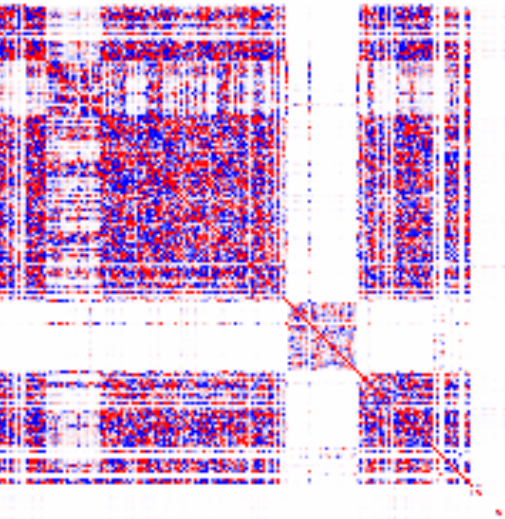}};
                    \node[below=of img3, node distance=0cm, yshift=1.cm,font=\color{black}] {True Inverse};
                    
                    \node (img4)[right=of img3, xshift=-1.1cm, yshift=-.025cm]{\includegraphics[trim={12.7cm 1.cm 1.05cm 1.2cm},clip, width=.082\linewidth]{latex/hyperparamsWithIFT/hessian_true_inv.pdf}};
                \end{tikzpicture}
                \caption[Inverse Hessian approximation visualizations]{
                    {Inverse Hessian approximations} for a 1-layer, fully-connected \nn on the Boston housing dataset as in \cite{zhang2017noisy}.
                }
                \label{ift_fig:hessian_visual}
            \end{figure}

            In Figure~\ref{ift_fig:hessian_visual}, we show the inverse Hessian for a fully connected $1$-layer \nn on the Boston housing dataset.
            The true inverse has a dominant diagonal, motivating identity approximations, while using more Neumann terms yields a structure closer to the true inverse.
        
        \newcommand{\numOverfitTrain}{\num{50}}
        \newcommand{\numOverfitValid}{\num{50}}
        \subsection{Overfitting a Small Validation Set}
\begin{figure}[ht!]
                \vspace{-0.01\textheight}
                \centering
                \begin{tikzpicture}
                    \centering
                    \node (img){\includegraphics[trim={0.15cm 0cm 0.15cm 0cm},clip, width=.68\linewidth]{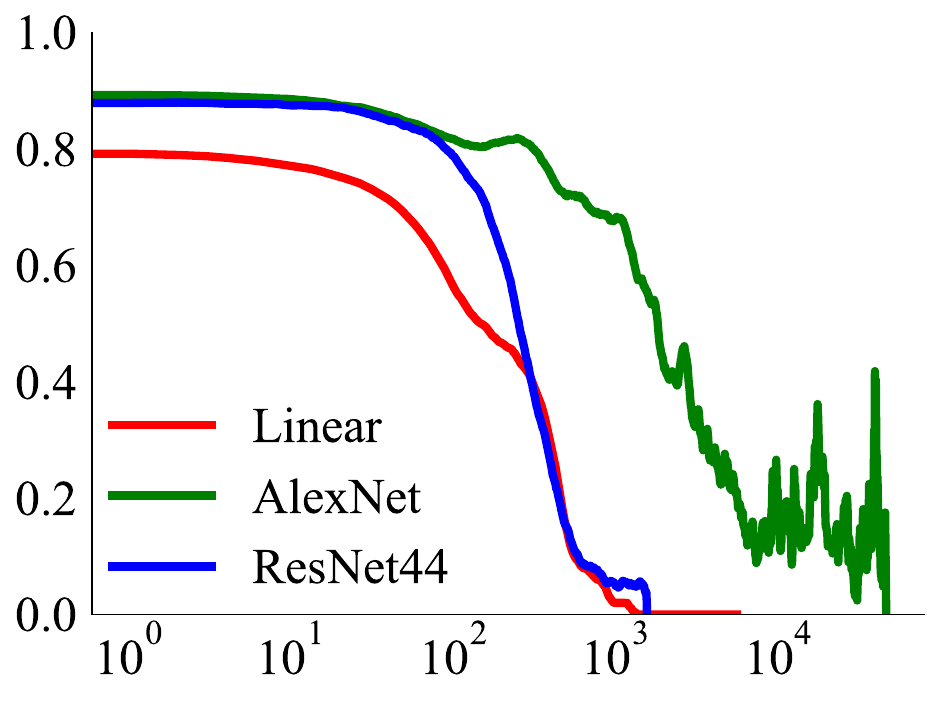}};
                    \node[left=of img, node distance=0cm, rotate=90, xshift=2.25cm, yshift=-.90cm, font=\color{black}] {1.0 - Validation Accuracy};
                    \node[below=of img, node distance=0cm, yshift=1.35cm,font=\color{black}] {Iteration};
                \end{tikzpicture}
                \caption[Overfitting CIFAR-10 with many hyperparameters]{
                    {Algorithm~\ref{ift_alg:joint_train} can overfit a small validation set on CIFAR-\num{10}.}
                    It optimizes for loss and achieves \SI{100}{\percent} validation accuracy for standard, large models. 
                }
                \label{ift_fig:overfit_valid}
                \vspace{-0.025\textheight}
            \end{figure}

            In Figure~\ref{ift_fig:overfit_valid}, we check the capacity of our \ho algorithm to overfit the validation data.
            We use the same restricted dataset as in \cite{franceschi2017forward, franceschi2018bilevel} of $\numOverfitTrain$ training and validation examples, which allows us to assess \ho performance easily.
            We tune a separate weight decay hyperparameter for each \nn parameter as in \cite{balaji2018metareg, maclaurin2015gradient}.
            We show the performance with a linear classifier, AlexNet~\citep{krizhevsky2012imagenet}, and ResNet\num{44}~\citep{he2016deep}.
            For AlexNet, this yields $>\num{50000000}$ hyperparameters, so we can perfectly classify our validation data by optimizing the hyperparameters.
            
            Algorithm~\ref{ift_alg:joint_train} achieves \SI{100}{\percent} accuracy on the training and validation sets with significantly lower accuracy on the test set (Appendix~\ref{ift_app:experiments}, Figure~\ref{ift_fig:overfit_valid_all}), showing that we have a powerful \ho algorithm that can overfit both the training and validation data.
            The same optimizer is used for weights and hyperparameters in all cases.
            %
        
        \newcommand{\numDistillValid}{300}
        \subsection{Dataset Distillation}
            
            \begin{figure*}
                \vspace{-0.015\textheight}
                \centering
                \begin{tikzpicture}
                    \centering
                    \node (img){\includegraphics[trim={3.15cm 0cm 2.5cm 0cm},clip, width=.98\linewidth,height=1.5cm]{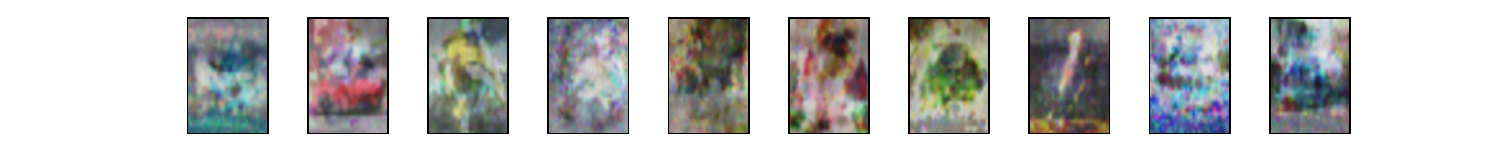}};
                    \node[above=of img, node distance=0cm, yshift=-.95cm,font=\color{black}] {{CIFAR-10 Distillation}};
                    \node(img_c1)[above=of img, node distance=0cm, yshift=-1.35cm, xshift=-7.0cm, font=\color{black}] {Plane};
                    \node(img_c2)[right=of img_c1, node distance=0cm, xshift=-.4cm, font=\color{black}] {Car};
                    \node(img_c3)[right=of img_c2, node distance=0cm, xshift=-.35cm, font=\color{black}] {Bird};
                    \node(img_c4)[right=of img_c3, node distance=0cm, xshift=-.30cm, font=\color{black}] {Cat};
                    \node(img_c5)[right=of img_c4, node distance=0cm, xshift=-.35cm, font=\color{black}] {Deer};
                    \node(img_c6)[right=of img_c5, node distance=0cm, xshift=-.35cm, font=\color{black}] {Dog};
                    \node(img_c7)[right=of img_c6, node distance=0cm, xshift=-.35cm, font=\color{black}] {Frog};
                    \node(img_c8)[right=of img_c7, node distance=0cm, xshift=-.45cm, font=\color{black}] {Horse};
                    \node(img_c9)[right=of img_c8, node distance=0cm, xshift=-.55cm, font=\color{black}] {Ship};
                    \node(img_c10)[right=of img_c9, node distance=0cm, xshift=-.50cm, font=\color{black}] {Truck};

                    \node (img2)[below=of img, yshift=.65cm]{\includegraphics[trim={3.15cm .5cm 2.5cm .5cm},clip, width=.98\linewidth,height=1.2cm]{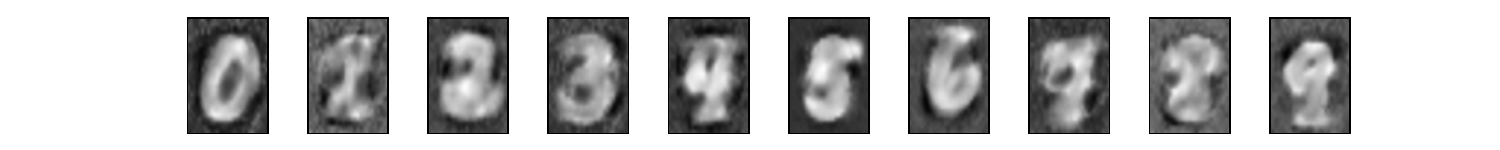}};
                    \node[above=of img2, node distance=0cm, yshift=-1.cm,font=\color{black}] {{MNIST Distillation}};
                    
                    \node (img3)[below=of img2, yshift=.2cm]{\includegraphics[trim={3.15cm 18.5cm 2.5cm 3cm},clip, width=.98\linewidth]{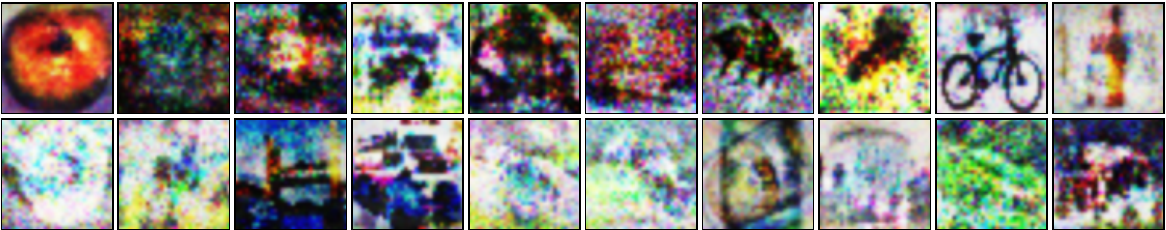}};
                    \node[above=of img3, node distance=0cm, yshift=-0.8cm,font=\color{black}] {{CIFAR-100 Distillation}};
                    \node(img_d1)[above=of img3, node distance=0cm, yshift=-1.15cm, xshift=-6.75cm, font=\color{black}] {Apple};
                    \node(img_d2)[right=of img_d1, node distance=0cm, xshift=-.50cm, font=\color{black}] {Fish};
                    \node(img_d3)[right=of img_d2, node distance=0cm, xshift=-.50cm, font=\color{black}] {Baby};
                    \node(img_d4)[right=of img_d3, node distance=0cm, xshift=-.60cm, font=\color{black}] {Bear};
                    \node(img_d5)[right=of img_d4, node distance=0cm, xshift=-.60cm, font=\color{black}] {Beaver};
                    \node(img_d6)[right=of img_d5, node distance=0cm, xshift=-.60cm, font=\color{black}] {Bed};
                    \node(img_d7)[right=of img_d6, node distance=0cm, xshift=-.45cm, font=\color{black}] {Bee};
                    \node(img_d8)[right=of img_d7, node distance=0cm, xshift=-.50cm, font=\color{black}] {Beetle};
                    \node(img_d9)[right=of img_d8, node distance=0cm, xshift=-.70cm, font=\color{black}] {Bicycle};
                    \node(img_d10)[right=of img_d9, node distance=0cm, xshift=-.85cm, font=\color{black}] {Bottle};
                    
                    \node(img_e1)[below=of img3, node distance=0cm, yshift=1.15cm, xshift=-6.75cm, font=\color{black}] {Bowl};
                    \node(img_e2)[right=of img_e1, node distance=0cm, xshift=-.45cm, font=\color{black}] {Boy};
                    \node(img_e3)[right=of img_e2, node distance=0cm, xshift=-.55cm, font=\color{black}] {Bridge};
                    \node(img_e4)[right=of img_e3, node distance=0cm, xshift=-.65cm, font=\color{black}] {Bus};
                    \node(img_e5)[right=of img_e4, node distance=0cm, xshift=-.65cm, font=\color{black}] {Butterfly};
                    \node(img_e6)[right=of img_e5, node distance=0cm, xshift=-.95cm, font=\color{black}] {Camel};
                    \node(img_e7)[right=of img_e6, node distance=0cm, xshift=-.65cm, font=\color{black}] {Can};
                    \node(img_e8)[right=of img_e7, node distance=0cm, xshift=-.5cm, font=\color{black}] {Castle};
                    \node(img_e9)[right=of img_e8, node distance=0cm, xshift=-1.0cm, font=\color{black}] {Caterpillar};
                    \node(img_e10)[right=of img_e9, node distance=0cm, xshift=-1.0cm, font=\color{black}] {Cattle};
                \end{tikzpicture}
                \caption[Distilled datasets for CIFAR-10, MNIST, and CIFAR-100]{
                    {Distilled datasets for CIFAR-10, MNIST, and CIFAR-100.}
                    For CIFAR-100, we show the first $20$ classes---the rest are in Appendix Figure~\ref{ift_app:cifar-100-distillation}.
                    We learn one distilled image per class. So, after training a logistic regression classifier on the distillation, it generalizes to the rest of the data.
                }
                \label{ift_fig:distilled_cifar}
            \end{figure*}
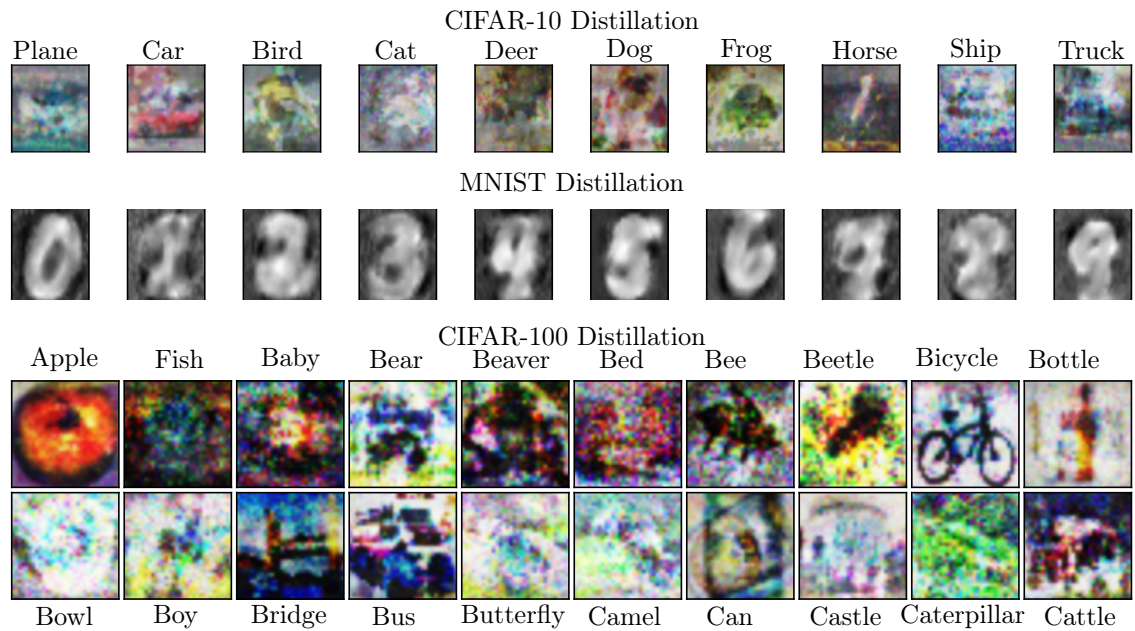
            
            Dataset distillation~\citep{maclaurin2015gradient, wang2018dataset} aims to learn a small, synthetic training dataset from scratch that condenses the knowledge in the original full-sized training set.
            The goal is for a model trained on the synthetic data to generalize to the original validation and test sets.
            Distillation is an interesting benchmark for \ho as it allows us to introduce tens of thousands of hyperparameters and visually inspect what is learned. Here, every pixel value in each synthetic training example is a hyperparameter.
            We distill MNIST and CIFAR-\num{10}/\num{100}~\citep{krizhevsky2009learning}, yielding $\num{28} \times \num{28} \times \num{10} = \num{7840}$, $\num{32} \times \num{32} \times \num{3} \times \num{10} = \num{30720}$, and $\num{32} \times \num{32} \times \num{3} \times \num{100} = \num{300720}$ hyperparameters, respectively.
            All labeled data are in our validation set for these experiments, while our distilled data are in the training set.
            We visualize the distilled images for each class in Figure~\ref{ift_fig:distilled_cifar}, recovering recognizable digits for MNIST and reasonable color averages for CIFAR-\num{10}/\num{100}.
        
        \newcommand{\augmentFinalValAcc}{\SI{94}{\percent}}
        \newcommand{\augmentFinalTestAcc}{\SI{94}{\percent}}
        \newcommand{\numUnetParam}{\num{6659}}
        \subsection{Learned Data Augmentation}\label{ift_sec:exp_data_augment}
            Data augmentation is a simple way to introduce invariances to a model---such as scale or contrast invariance---that improve generalization~\citep{cubuk2018autoaugment,xie2019unsupervised}.
            Taking advantage of the ability to optimize many hyperparameters, we learn data augmentation {from scratch} (Figure~\ref{ift_fig:augment}).
            
            Specifically, we learn a data augmentation network $\mathbf{\tilde{x}} = \mathbf{f}_{\hp}(\mathbf{x}, \boldsymbol{\epsilon})$ that takes a training example $\mathbf{x}$ and noise $\boldsymbol{\epsilon} \sim \mathcal{N}(\textbf{0}, \mathbf{I})$, and outputs an augmented example $\mathbf{\tilde{x}}$.
            The noise $\boldsymbol{\epsilon}$ allows us to learn {stochastic} augmentations.
            We parameterize $\mathbf{f}$ as a U-Net~\citep{ronneberger2015u} with a residual connection from the input to the output, making identity mapping easy to learn.
            The parameters of the U-Net, $\hp$, are hyperparameters tuned for the validation loss ---- therefore, we have $\numUnetParam$ hyperparameters.
            We trained a ResNet\num{18}~\citep{he2016deep} on CIFAR-10 with augmented examples produced by the U-Net, which is trained simultaneously on the validation set.

            The results for the identity and the Neumann inverse approximations are shown in Table~\ref{ift_tab:neumann}.
            We omit CG because it performed no better than the identity.
            Using the data augmentation network improves the validation and test accuracy by $2-3\%$ and produces a smaller variance between multiple random restarts.
            In \citet{mounsaveng2019adversarial}, a different augmentation network architecture is learned with adversarial training.
            \newcommand{\augWidth}{0.23\textwidth} 
            \newcommand{\augHeight}{\augWidth}
            \newcommand{\augXShift}{-0.0675\textwidth}
            \newcommand{\augYShift}{0.0675\textwidth}
            \begin{figure}
                \centering
                \begin{tikzpicture}
                    \centering
                    \node (img11){\includegraphics[trim={0cm 0cm 0cm 0cm},clip, width=\augWidth,height=\augHeight]{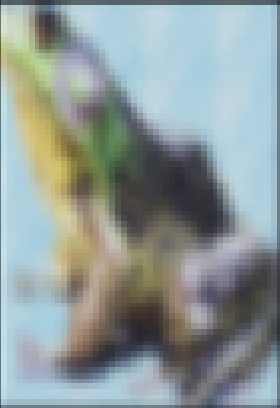}};
                    \node (img12)[right=of img11, xshift=\augXShift]{\includegraphics[trim={0cm 0cm 0cm 0cm},clip, width=\augWidth,height=\augHeight]{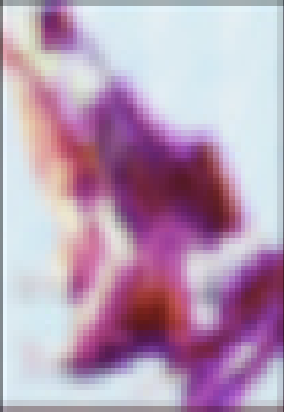}};
                    \node (img13)[right=of img12, xshift=\augXShift]{\includegraphics[trim={0cm 0cm 0cm 0cm},clip, width=\augWidth,height=\augHeight]{{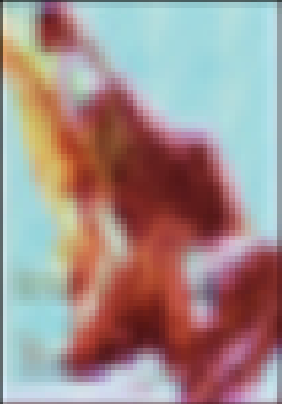}}};
                    \node (img14)[right=of img13, xshift=\augXShift]{\includegraphics[trim={0cm 0cm 0cm 0cm},clip, width=\augWidth,height=\augHeight]{{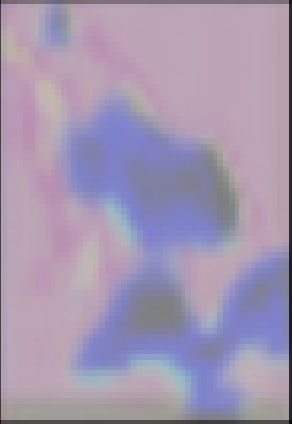}}};
                    
                    \node[above=of img11, node distance=0cm, yshift=-1.15cm,font=\color{black}] {{Original}};
                    \node[above=of img12, node distance=0cm, yshift=-1.15cm,font=\color{black}] {{Sample 1}};
                    \node[above=of img13, node distance=0cm, yshift=-1.15cm,font=\color{black}] {{Sample 2}};between
                    \node[above=of img14, node distance=0cm, yshift=-1.08cm,font=\color{black}] {{Pixel Std.}};
                    
                    \node (img21)[below=of img11, yshift=\augYShift]{\includegraphics[trim={0cm 0cm 0cm 0cm},clip, width=\augWidth,height=\augHeight]{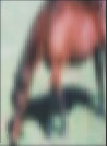}};
                    \node (img22)[right=of img21, xshift=\augXShift]{\includegraphics[trim={0cm 0cm 0cm 0cm},clip, width=\augWidth,height=\augHeight]{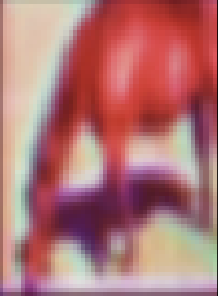}};
                    \node (img23)[right=of img22, xshift=\augXShift]{\includegraphics[trim={0cm 0cm 0cm 0cm},clip, width=\augWidth,height=\augHeight]{{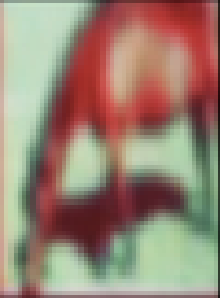}}};
                    \node (img24)[right=of img23, xshift=\augXShift]{\includegraphics[trim={0cm 0cm 0cm 0cm},clip, width=\augWidth,height=\augHeight]{{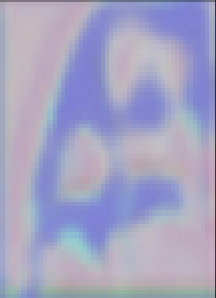}}};
                    
                    \node (img31)[below=of img21, yshift=\augYShift]{\includegraphics[trim={0cm 0cm 0cm 0cm},clip, width=\augWidth,height=\augHeight]{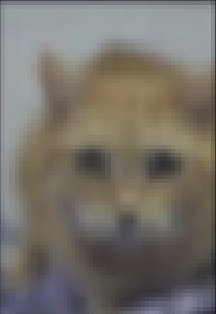}};
                    \node (img32)[right=of img31, xshift=\augXShift]{\includegraphics[trim={0cm 0cm 0cm 0cm},clip, width=\augWidth,height=\augHeight]{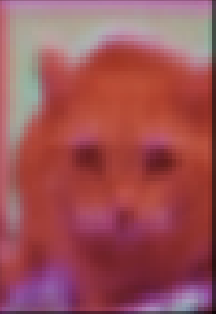}};
                    \node (img33)[right=of img32, xshift=\augXShift]{\includegraphics[trim={0cm 0cm 0cm 0cm},clip, width=\augWidth,height=\augHeight]{{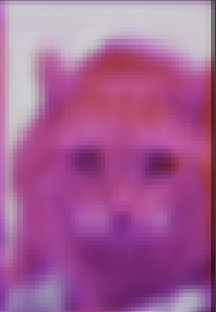}}};
                    \node (img34)[right=of img33, xshift=\augXShift]{\includegraphics[trim={0cm 0cm 0cm 0cm},clip, width=\augWidth,height=\augHeight]{{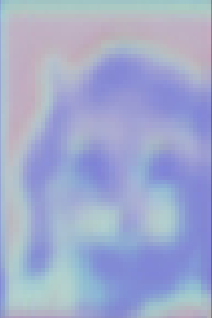}}};
                \end{tikzpicture}
                \caption[Learned data augmentations]{
                    {Learned data augmentations.}
                    The original image is on the left, followed by two augmented samples and the standard deviation of the pixel intensities from the augmentation distribution.
                }
                \label{ift_fig:augment}
            \end{figure}
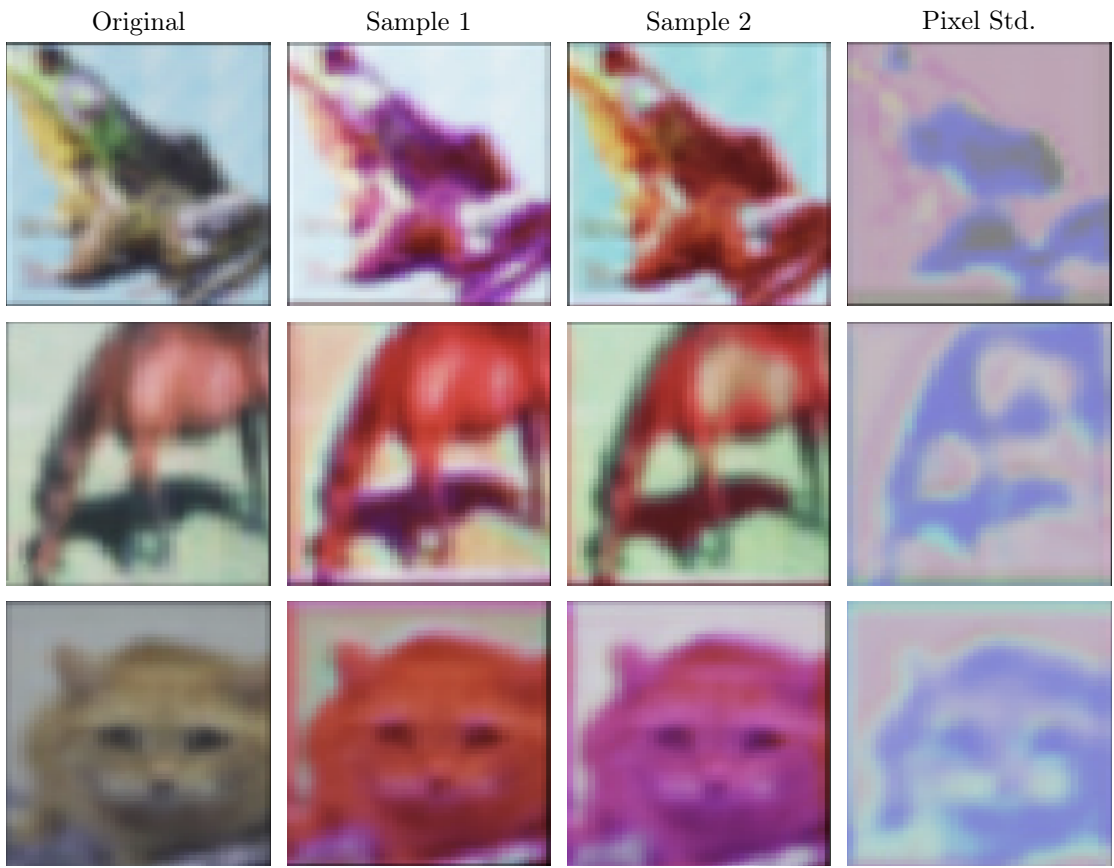
            
            \begin{table}
            	\centering
            	\begin{tabular}{ccc}
            		{Inverse Approximation} & {Validation Accuracy $\uparrow$} & {Test Accuracy $\uparrow$}
            		\\
            		\midrule
            	    \num{0} 
            	        & 92.5 \rpm 0.21
            	        & 92.6 \rpm 0.17
            	        \\
            		\num{3} Neumann steps
            		    & \textbf{95.1} \rpm 0.02
            		    & 94.6 \rpm 0.01
            		    \\
            		\num{3} steps unrolled differentiation
            		    & 95.0 \rpm 0.02
            		    & \textbf{94.7} \rpm 0.01 
            		    \\
            	    $\identity$
            		    & 94.6 \rpm 0.02
            		    & 94.1 \rpm 0.02
            		  \\
            	\end{tabular}
                \caption[Accuracy of different inverse approximations]{
                    {Accuracy of different inverse approximations.}
                    Using $\num{0}$ means no \ho occurs, and the augmentation is initially the identity.
                    The Neumann approach performs similarly to unrolled differentiation~\citep{maclaurin2015gradient, shaban2018truncated} with equal steps and less memory.
                    Using more terms does better than the identity, and the identity performed better than CG (not shown), which was unstable.
                }
                \label{ift_tab:neumann}
            \end{table}
            %
        
    \subsection{RNN Hyperparameter Optimization}
    \label{ift_sec:exp_rnn}
        We also used our proposed algorithm to tune regularization hyperparameters for an LSTM~\citep{hochreiter1997long} trained on the Penn TreeBank (PTB) corpus~\citep{marcus1993building}.
        As in~\citet{gal2016theoretically}, we used a \num{2}-layer LSTM with \num{650} hidden units per layer and \num{650}-dimensional word embeddings.
        Additional details are provided in Appendix~\ref{ift_app:exp_rnn}.
        \newline
        
\begin{figure}
                \centering
                \begin{tikzpicture}
                    \centering
                    \node (img){\includegraphics[trim={0.7cm .8cm .55cm .55cm},clip, width=.73\linewidth]{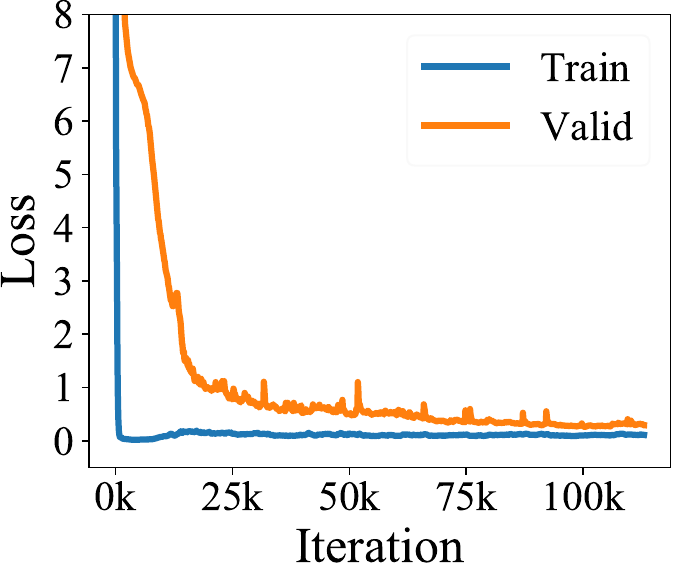}};
                    \node[left=of img, node distance=0cm, rotate=90, xshift=.5cm, yshift=-.90cm, font=\color{black}] {Loss};
                    \node[below=of img, node distance=0cm, yshift=1.25cm,font=\color{black}] {Iteration};
                \end{tikzpicture}
                \caption[Overfitting hyperparameters on LSTMs]{
                    {Algorithm~\ref{ift_alg:joint_train} can overfit a small validation set with an LSTM on PTB.}
                }
                \label{ift_fig:rnn-overfitting}
            \end{figure}

        \textbf{Overfitting Validation Data.}
            We first verify that our algorithm can overfit the validation set in a small-data setting with $10$ training and $10$ validation sequences (Figure~\ref{ift_fig:rnn-overfitting}).
            The LSTM architecture we use has \num{13280400} weights, and we tune a separate weight-decay hyperparameter per weight.
            We overfit the validation set, reaching nearly 0 validation loss.

        \textbf{Large-Scale \HO.}
            There are various forms of regularization used for the training of RNNs, including {variational dropout}~\citep{kingma2015variational} on the input, hidden state, and output; {embedding dropout} that sets rows of the embedding matrix to $0$, removing tokens from all sequences in a mini-batch; DropConnect~\citep{wan2013regularization} on hidden-to-hidden weights; and activation and temporal activation regularization.
            We tune these $\num{7}$ hyperparameters simultaneously.
            Additionally, we experiment with tuning separate dropout/DropConnect rates for each activation/weight, giving $\num{1691951}$ total hyperparameters.
            To allow gradient-based optimization of dropout rates, we use concrete dropout~\citep{gal2017concrete}.
            
            Instead of using the small dropout initialization as in \citet{mackay2019self}, we use a larger initialization of $\num{0.5}$.
            The results for our new initialization without \ho, our method tuning the same hyperparameters as \citet{mackay2019self} (``Ours''), and our method tuning many more hyperparameters (``Ours, Many'') are shown in Table~\ref{ift_tab:rnn_compare}.
            We can tune hyperparameters more quickly and achieve better perplexities than the alternatives.
            \begin{table}
            	\centering
            	\begin{tabular}{cccc}
            		Method & Validation Loss $\Lval^{*} \downarrow$ & Test Loss $\downarrow$ & Time(s)\\
            		\midrule
            	    Grid Search
            	        & 97.32
            	        & 94.58
            	        & 100k\\
            		Random Search
            		    & 84.81
            		    & 81.46
            		    & 100k\\
            	    Bayesian Opt.
            		    & 72.13
            		    & 69.29
            		    & 100k\\
            		STN
            		    & 70.30
            		    & 67.68
            		    & 25k\\
            		\hline
            		\textbf{No \ho}
            		    & 75.72
        		        & 71.91
        		        & 18.5k\\
        		    \textbf{Ours}
        		        & 69.22
        		        & 66.40
        		        & 18.5k\\
            		\textbf{Ours, Many}
            		    & \textbf{68.18}
            		    & \textbf{66.14}
            		    & 18.5k\\
                \end{tabular}
                \caption[Comparing hyperparameter optimization methods for LSTM training]{
                    {Comparing \HO methods for LSTM training on PTB.}
                    We tune millions of hyperparameters faster and with memory comparable to competitors that tune a handful.
                    Our method competitively optimizes the same \num{7} hyperparameters as the baselines of \cite{mackay2019self} (first four rows).
                    We show a performance boost by tuning millions of hyperparameters introduced with per-unit/weight dropout and DropConnect.
                    ``No hyperparameter optimization'' shows how our augmented hyperparameter initialization affects training.
                }
                \label{ift_tab:rnn_compare}
            \end{table}
    
    \subsection{Effects of Many Hyperparameters}
        Given the ability to tune high-dimensional hyperparameters and the potential to overfit the validation set, should we reconsider how our training and validation splits are structured?
        Do the same heuristics apply as with low-dimensional hyperparameters (e.g., use $\sim 10\%$ of the data for validation)?

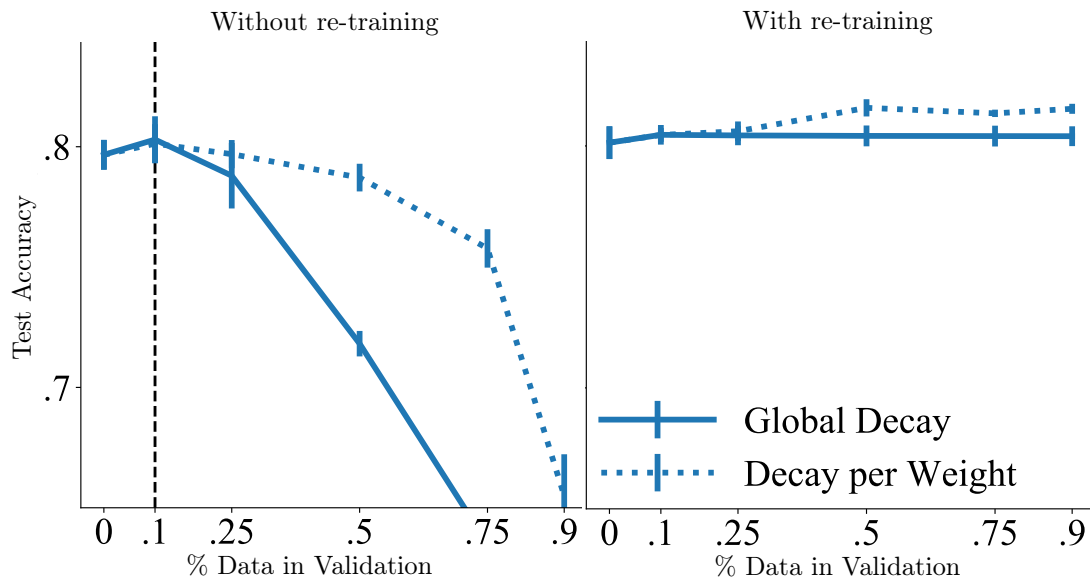
\begin{figure}
            \begin{tikzpicture}
                \centering
                \node (img){\includegraphics[trim={1.25cm 1.35cm .6cm 1.3cm},clip, width=.475\linewidth]{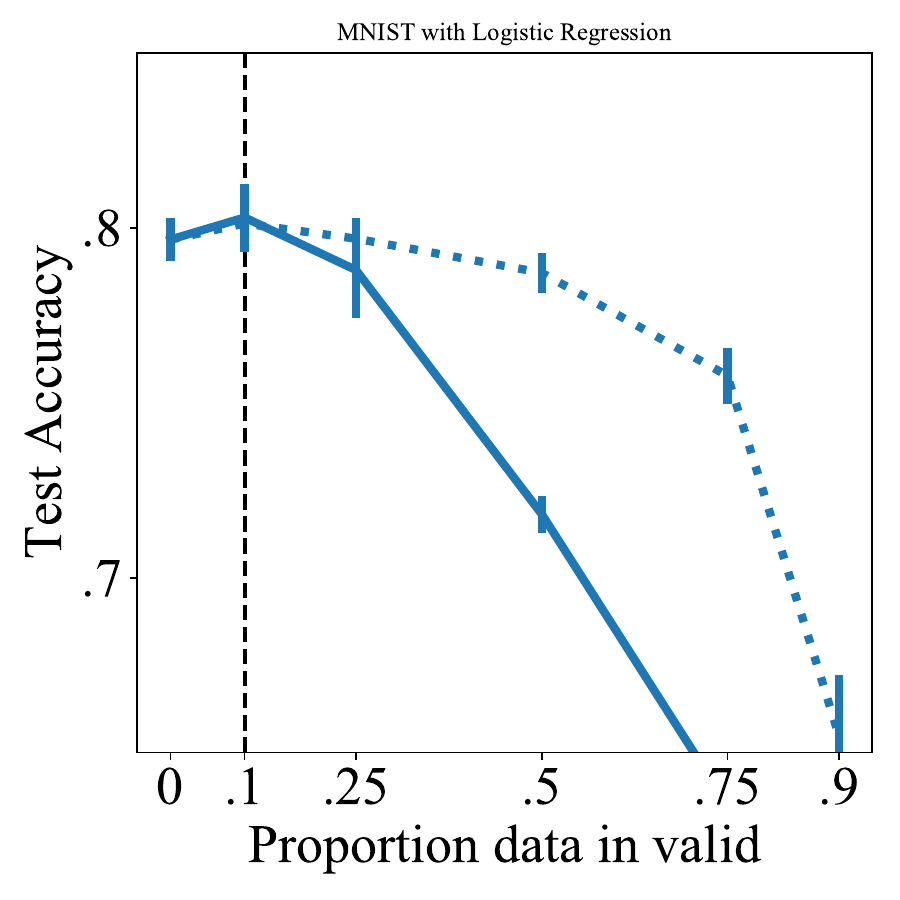}};
                \node[left=of img, node distance=0cm, rotate=90, xshift=1.5cm, yshift=-.95cm, font=\color{black}] {Test Accuracy};
                \node[above=of img, node distance=0cm, yshift=-1.15cm, xshift=.2cm, font=\color{black}] {{Without re-training}};
                \node[below=of img, node distance=0cm, yshift=1.25cm,font=\color{black}] {\% Data in Validation};
                
                \node (img2)[right=of img, xshift=-1.2cm]{\includegraphics[trim={2.3cm 1.35cm .6cm 1.3cm},clip, width=.44\linewidth]{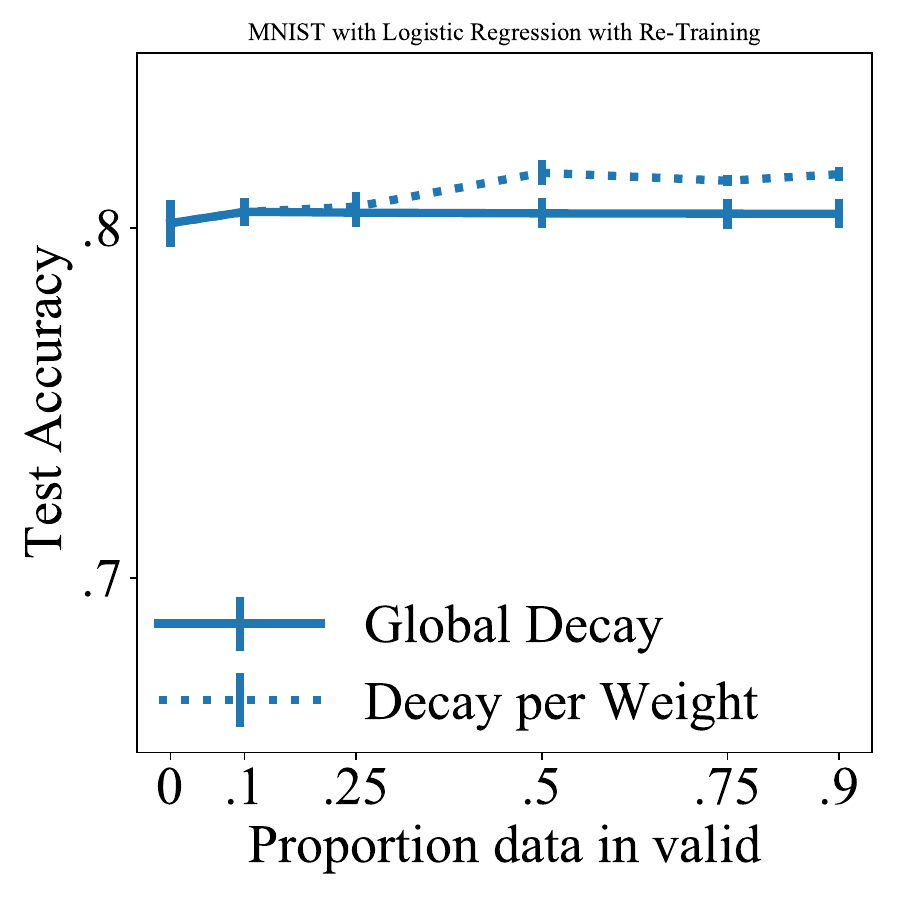}};
                \node[above=of img2, node distance=0cm, yshift=-1.15cm,font=\color{black}] {{With re-training}};
                \node[below=of img2, node distance=0cm, yshift=1.25cm,font=\color{black}] {\% Data in Validation};
            \end{tikzpicture}
            \caption[Validation split sizes and overfitting capacity visualization]{
                {Test accuracy of logistic regression on MNIST, with different size validation splits.}
                Solid lines correspond to a single global weight decay ($\num{1}$ hyperparameter), while dotted lines correspond to a separate weight decay per weight (many hyperparameters).
                The best validation proportion for test performance differs after retraining for many hyperparameters but is similar for few hyperparameters.
            }
            \label{ift_fig:valid_prop_comparison}
        \end{figure}

        In Figure~\ref{ift_fig:valid_prop_comparison}, we see how splitting our data into training and validation sets of different ratios affects test performance.
        We show the results of jointly optimizing the \nn weights and hyperparameters, as well as the results of fixing the final optimized hyperparameters and retraining the \nn weights from scratch, which is a common technique for boosting performance~\citep{Goodfellow-et-al-2016}.
        
        We evaluate a high-dimensional regime with a separate weight decay hyperparameter per \nn parameter and a low-dimensional regime with a single global weight decay.
        We observe that: 
        (1) for a single weight decay, the optimal combination of validation data and weight decay has similar test performance with and without retraining because the optimal amount of validation data is small and
        (2) for many weight decays, the optimal combination of validation data and weight decay is significantly affected by retraining because the optimal amount of validation data needs to be large to fit our hyperparameters effectively.
        
        For few hyperparameters, our results agree with the standard practice of using \num{10}\% of the data for validation and the other \num{90}\% for training.
        For many hyperparameters, our example shows that we may need larger validation partitions for \ho.
        Retraining our model with all data can be critical if we use a large validation partition to fit the hyperparameters.

    \section{Conclusion}
        We present a gradient-based hyperparameter optimization algorithm that scales to high-dimensional hyperparameters for modern, deep \nns.
        We use the implicit function theorem to formulate the hypergradient as a matrix equation whose bottleneck is inverting the Hessian of the training loss with respect to the \nn parameters.
        We scale the hypergradient computation to large \nns by approximately inverting the Hessian, leveraging a relationship with unrolled differentiation.
        We believe algorithms of this nature provide a path for practical nested optimization, where we have Hessians with a known structure.
        Examples of this include GANs~\citep{goodfellow2014generative}, and other multi-agent games~\citep{foerster2018learning, letcher2018stable}.
        However, these games have a non-zero direct gradient, which we leverage in the next chapter to create a robust method that circumvents any best-response approximations and scales to (at the time) SOTA architectures.
    
    \section*{Acknowledgements}
        We thank Chris Pal for recommending we check re-training with all the data, Haoping Xu for discussing experiments on inverse-approximation variants, Roger Grosse for guidance, Cem Anil \& Chris Cremer for their feedback on this work, and Xuanyi Dong for noting corrections in the algorithms.
        Paul Vicol was supported by a JP Morgan AI Fellowship.
        We also thank anonymous reviewers for their suggestions and everyone else at Vector for helpful discussions and feedback.

    \subsection*{My Contributions Towards this Paper As it Pertains to the Thesis}
        The ideas underlying this project were a collaborative effort from all authors.
        I ran many of the experiments with much help from Paul Vicol.
        I also did most of the writing for the paper.
        This paper has a conference version~\citep{lorraine2019optimizing} and an arXiv version~\citep{lorraine2019optimizingARXIV}.

    \chapter{Complex Momentum for Optimization in Games}\label{chap:complexMomentum}
        We generalize gradient descent with momentum for optimization in differentiable games to have complex-valued momentum.
        We give theoretical motivation for our method by proving convergence on bilinear zero-sum games for simultaneous and alternating updates.
        Our method gives real-valued parameter updates, making it a drop-in replacement for standard optimizers.
        We empirically demonstrate that complex-valued momentum can improve convergence in realistic adversarial games---like generative adversarial networks---by showing that we find better solutions with an almost identical computational cost.
        We also show a practical complex-valued Adam variant, which we use to train BigGAN to improve inception scores on CIFAR-10.
    \section{Introduction}\label{cm_sec:introduction}
        Gradient-based optimization has been critical for the success of machine learning, updating a single set of parameters to minimize a single loss.
        A growing number of applications require learning with multiple players, each with parameters and objectives.
        Common examples are GANs~\citep{goodfellow2014generative}, hyperparameter optimization~\citep{maclaurin2015gradient, andrychowicz2016learning, fu2016drmad, shaban2018truncated}, actor-critic models~\citep{pfau2016connecting}, curriculum learning~\citep{baker2019emergent, balduzzi2019open, sukhbaatar2017intrinsic}, adversarial examples~\citep{bose2020adversarial, yuan2019adversarial}, learning models~\citep{rajeswaran2020game, abachi2020policy, nikishin2021control}, domain adversarial adaptation~\citep{acuna2021fdomainadversarial}, neural architecture search~\citep{elsken2019neural}, and meta-learning~\citep{finn2017model, ren2018meta, ren2020flexible}.

        
        We often want solutions where no player gains from changing their strategy unilaterally, e.g., Nash equilibria~\citep{morgenstern1953theory} or Stackelberg equilibria~\citep{von2010market}.
        Classical gradient-based learning often fails to find these equilibria due to rotational dynamics~\citep{berard2019closer}, or identically zero direct gradients as in Chapters~\ref{chap:hyperparamWithHypernets} and \ref{chap:hyperparamWithIFT}.
        %
        There are numerous saddle point finding algorithms for zero-sum games~\citep{arrow1958studies, freund1999adaptive}.
        \newline
        \newline
        \newline
        
        \cite{gidel2018negative} generalize GD with momentum to games, showing that we can use a negative momentum to converge if the eigenvalues of the Jacobian of the gradient vector field have a large imaginary part.
        We use terminology in \cite{gidel2018negative} and say \textit{(purely) cooperative or adversarial games} for games with (purely) real or imaginary eigenvalues.
        Setups like GANs are not purely adversarial but have both \emph{purely cooperative and adversarial eigenspaces} -- i.e., eigenspaces with purely real or imaginary eigenvalues.
        More generally, we say that an eigenspace is cooperative when the real part of the eigenvalue is larger than the imaginary part and is otherwise adversarial.
        In cooperative eigenspaces, classical optimization methods perform best, while in adversarial eigenspaces, methods customized for games work best -- see Figure~\ref{cm_fig:partial_coop_phases}.
        
        We want a method that converges with simultaneous and alternating updates in purely adversarial games -- a setup where existing momentum methods fail.
        Also, we want a method that robustly converges with different mixtures of adversarial and cooperative eigenspaces -- see Figure~\ref{cm_fig:partial_coop_phases_robustness} -- because finding all eigenspaces depends on an eigendecomposition that can be intractable.
        To solve this, we unify and generalize existing momentum methods~\citep{lucas2018aggregated, gidel2018negative} to recurrently linked momentum -- a setup with multiple recurrently linked momentum buffers with potentially negative coefficients -- see Figure~\ref{cm_fig:recurrent_comp_graph}.
        
        Selecting two of these recurrently linked buffers with appropriate momentum coefficients can be interpreted as the real and imaginary parts of a single \emph{complex buffer} and \emph{complex momentum coefficient} -- see Appendix Figure~\ref{cm_fig:cm_comp_graph_complex}.
        This setup (a) allows us to converge in adversarial games with simultaneous updates, (b) only introduces one new optimizer parameter -- the phase or $\arg$ of our momentum, (c) allows us to gain intuitions via complex analysis, (d) is trivial to implement in libraries supporting complex arithmetic, and (e) robustly converges for different eigenspace mixtures.
        
        
        Intuitively, our complex buffer stores historical gradient information, oscillating between adding or subtracting at a frequency dictated by the momentum coefficient.
        Classical momentum only adds gradients, and negative momentum changes between adding or subtracting each iteration while we oscillate at an arbitrary (fixed) frequency -- see Figure~\ref{cm_fig:frequency_vis}.
        This reduces rotational dynamics during training by canceling out opposing updates.
        
        
        \hphantom{a}\newline
        Our contributions include:
        \begin{itemize}
        	\item Providing generalizations and variants of classical~\citep{polyak1964some}, negative~\citep{gidel2018negative}, and aggregated~\citep{lucas2018aggregated} momentum for learning in differentiable games.
        	\item Showing our method converges on adversarial games, including bilinear zero-sum games and a Dirac-GAN, with simultaneous and alternating updates.
        	\item Illustrating a robustness during optimization, converging faster and over a larger range of mixtures of cooperative and adversarial games than existing first-order methods.
        	\item Giving a practical extension of our method to a complex-valued Adam~\citep{kingma2014adam} variant, which we use to train a BigGAN~\citep{brock2018large} on CIFAR-10, improving their inception scores.
        \end{itemize}
        %
    \begin{figure}
        \centering
        \begin{minipage}{.55\textwidth}
            \centering
            \lstset{
            language=Python,
            basicstyle=\footnotesize\ttfamily,
            alsoletter={., +},
            keywords=[2]{jnp.real, .3j},
            keywordstyle=[2]{\color{DarkGreen}}
            }
            \textbf{\hspace{-1.25cm}Actual JAX implementation: changes in {\color{DarkGreen}green}}
            \captionsetup{type=figure}
            \begin{lstlisting}
mass = .8 + .3j
def momentum(step_size, mass):
 ...
 def update(i, g, state):
  x, velocity = state
  velocity = mass * velocity + g
  x=x-jnp.real(step_size(i)*velocity)
  return x, velocity
 ...
            \end{lstlisting}
        \end{minipage}
        \hspace{1cm}
        \caption[Complex momentum in Jax]{
            How to modify JAX's \ref{cm_eq:single_objective_gd} with momentum \href{https://jax.readthedocs.io/en/latest/_modules/jax/experimental/optimizers.html}{{\color{blue}here}} to use complex momentum.
            The only changes are in {\color{DarkGreen}green}.
            $\texttt{jnp.real}$ gets the real part of \texttt{step\_size} times the momentum buffer (called $\texttt{velocity}$ here).
            In this case, we use a complex \texttt{mass} for our method $\momCoeff = |\momCoeff|\exp\left(i\arg\left(\momCoeff\right)\right) = \num{0.9}\exp\left(i\nicefrac{\pi}{\num{8}}\right) \approx .8 + .3 i$.
        }
        \label{cm_fig:jax_code_change}
        \vspace{-0.01\textheight}
    \end{figure}
    
    
    \section{Background}\label{cm_sec:background}
            Appendix Table~\ref{cm_tab:TableOfNotation} summarizes our notation.
            Consider the single-objective optimization problem:
            \begin{equation}\label{cm_eq:single_objective_opt} 
                    \paramSymbol^{*} = \argmin_{\paramSymbol} \loss\left(\paramSymbol\right) 
            \end{equation}
            %
            We can find local minima of loss $\loss$ using (stochastic) gradient descent with step size $\lr$.
            We denote the loss gradient at parameters $\paramSymbol^{j}$ by $\gradSymbol^j = \gradSymbol\left(\paramSymbol^j\right) = \left. \nabla_{\paramSymbol} \loss(\paramSymbol) \right|_{\paramSymbol^{j}}$.
            %
            \begin{equation}\label{cm_eq:single_objective_gd}\tag{SGD}
                \paramSymbol^{j+1} = \paramSymbol^{j} - \alpha \gradSymbol^j
            \end{equation}
             Momentum can generalize \ref{cm_eq:single_objective_gd}.
            For example, Polyak's Heavy Ball~\citep{polyak1964some}:
            \begin{align}\label{cm_eq:single_objective_polyak}
                \paramSymbol^{j+1} = \paramSymbol^{j} - \lr \gradSymbol^j + \momCoeff \left(\paramSymbol^{j} - \paramSymbol^{j-1}\right)
            \end{align}
            This can be equivalently written with momentum buffer $\momVal^{j} = \nicefrac{\left(\paramSymbol^{j} - \paramSymbol^{j-1}\right)}{\lr}$.
            %
            \begin{equation}\label{cm_eq:single_objective_sgdm}\tag{SGDm}
                \momVal^{j+1} = \momCoeff \momVal^{j} - \gradSymbol^j,\quad\quad
                \paramSymbol^{j+1} = \paramSymbol^{j} + \lr \momVal^{j+1}
            \end{equation}
            \ref{cm_eq:single_objective_sgdm} generalizes to aggregated momentum~\citep{lucas2018aggregated}, shown in Appendix Algorithm~\ref{cm_alg:aggmo}.
        %
        \subsection{Game Formulations}
            Another class of problems is learning in \textit{games}, which includes problems like generative adversarial networks (GANs).
            As in Chapters~\ref{chap:hyperparamWithHypernets} and \ref{chap:hyperparamWithIFT}, consider $\num{2}$-player games -- with players denoted by $A$ and $B$ -- where each player minimizes their loss $\loss_A, \loss_B$ with their parameters $\paramSymbol_A, \paramSymbol_B$.
            If $\outLoss$ and $\inLoss$ are differentiable in $\outParam$ and $\inParam$, we say the game is differentiable.
            In deep learning, losses are non-convex with many parameters, so we focus on finding local solutions.
            %
            If $\inParam^*(\outParam)$ denotes player $\inSymbol$'s best-response function, then solutions can be defined as:
            \begin{align}\label{cm_eq:stackelberg_equilibrium}
                \begin{split}
                    \outParam^{*} &= \argmin_{\outParam} \outLoss\left(\outParam, \inParam^{*}\left(\outParam\right)\right),\\
                    \inParam^{*}\left(\outParam\right) &= \argmin_{\inParam} \inLoss\left(\outParam, \inParam\right)
                \end{split}
            \end{align}

            \noindent
            We may be able to approximately find $\outParam^{*}$ efficiently if we can do \ref{cm_eq:single_objective_gd} on:
            \begin{equation}
                \outLoss^*\left(\outParam\right) = \outLoss\left(\outParam, \inParam^{*}\left(\outParam\right)\right)
            \end{equation}
            SGD requires computing $\pd{\outLoss^*}{\outParam}$, which often requires $\pd{\inParam^{*}}{\outParam}$, which we seek to circumvent computing in this chapter.
            A common optimization algorithm to analyze for finding solutions is simultaneous SGD (\ref{cm_eq:multi_objective_gd_long}) -- sometimes called gradient descent ascent for zero-sum games -- where $\outGrad^j = \outGrad\left(\outParam^j, \inParam^j\right)$ and $\inGrad^j = \inGrad\left(\outParam^j, \inParam^j\right)$ are estimators for $\left. \nabla_{\outParam} \outLoss\right|_{\outParam^j, \inParam^j}$ and $\left. \nabla_{\inParam} \inLoss \right|_{\outParam^j, \inParam^j}$: 
            %
            \begin{align}\label{cm_eq:multi_objective_gd_long}\tag{SimSGD}
                \outParam^{j+1} = \outParam^{j} - \lr \outGrad^j,\quad
                \inParam^{j+1} = \inParam^{j} - \lr \inGrad^j
            \end{align}
            We simplify the notation with the concatenated or joint-parameters $\bothParam = \left[\outParam, \inParam\right] \in \mathbb{R}^{\bothDim}$ and the joint-gradient vector field $\bothGrad: \mathbb{R}^{\bothDim} \to \mathbb{R}^{\bothDim}$, which at the $j^{\textnormal{th}}$ iteration is the joint-gradient denoted:
            \begin{equation}
                \bothGrad^j = \bothGrad\left(\bothParam^j\right) = \left[\outGrad\left(\bothParam^j\right), \inGrad\left(\bothParam^j\right)\right] = \left[\outGrad^j, \inGrad^j\right]
            \end{equation}
            We extend to $n$-player games by treating $\bothParam$ and $\bothGrad$ as concatenations of the players' parameters and loss gradients, allowing for a concise expression of the \ref{cm_eq:multi_objective_gd_long} update with momentum:
            \begin{align}\label{cm_eq:multi_objective_sgdm}\tag{SimSGDm}
                \momVal^{j+1} = \momCoeff \momVal^{j} - \bothGrad^j,
                \bothParam^{j+1} = \bothParam^{j} + \lr \momVal^{j+1}
            \end{align}
            \cite{gidel2018negative} show classical momentum choices of $\momCoeff \in [\num{0}, \num{1})$ do not improve convergence rate over \ref{cm_eq:multi_objective_gd_long} in some games.
            Specifically, any non-negative momentum and step size will not converge for purely adversarial games with imaginary eigenvalues.
            In contrast, negative momentum helps if the Jacobian of the joint-gradient vector field $\nabla_{\bothParam} \bothGrad$ has complex eigenvalues.
            For purely cooperative games -- ex., single-objective minimization -- $\nabla_{\bothParam} \bothGrad$ has real eigenvalues and (if symmetric) can be viewed as the Hessian of a loss, so classical momentum works well.
            %
            %

            \subsection{Limitations of Existing Methods}\label{cm_sec:limitation_existing}

                \textbf{Higher-order:} Methods using higher-order gradients are often harder to parallelize across GPUs~\citep{osawa2019large}, get attracted to bad saddle points~\citep{mescheder2017numerics}, require estimators for inverse Hessians~\citep{schafer2019competitive, wang2019solving}, are complicated to implement, have numerous optimizer parameters, and can be more expensive in iteration and memory cost~\citep{hemmat2020lead, wang2019solving, schafer2019competitive, schafer2020competitive, czarnecki2020real, zhang2020newton}.
                Instead, we focus on first-order methods.

                \noindent
                \textbf{First-order:} Some first-order methods such as extragradient~\citep{korpelevich1976extragradient} require a second, costly, gradient evaluation per step.
                Similarly, methods alternating player updates must wait until after the first player's gradient is used to evaluate the second player's gradient.
                However, many deep learning setups can parallelize the computation of both players' gradients, making alternating updates effectively cost another gradient evaluation.
                We are interested in convergence speed in the number of gradient queries, so we naturally looked at methods that update with the effective cost of one gradient evaluation -- see Appendix Figure~\ref{cm_fig:det_minmax_expAvg_alt_phase_tune}.
                Additionally, simultaneous updates are a standard choice in some settings~\citep{acuna2021fdomainadversarial}.
                \newline

                \noindent
                \textbf{Robust convergence:} We want our method to converge in purely adversarial games with simultaneous updates -- a setup where existing momentum methods fail~\citep{gidel2018negative}.
                Furthermore, computing a games eigendecomposition is often infeasibly expensive, so we want methods that robustly converge over different mixtures of adversarial and cooperative eigenspaces -- see Figure~\ref{cm_fig:partial_coop_phases_robustness}.
                We are interested in relevant eigenspace mixtures during GAN training -- see Figures~\ref{cm_fig:gan_spectrum_main} and \ref{cm_fig:gan_decomp}.
            
            \subsection{Coming up with our Method}
                \textbf{Combining existing methods:}
                Given the preceding limitations, we would like a robust first-order method using a single, simultaneous gradient evaluation.
                We looked at combining aggregated~\citep{lucas2018aggregated} with negative~\citep{gidel2018negative} momentum by allowing negative coefficients because these methods are first-order and use a single gradient evaluation -- see Figure~\ref{cm_fig:aggmo_comp_graph}.
                Additionally, aggregated momentum provides robustness during optimization by quickly converging on problems with a wide range of conditioning, while negative momentum works in adversarial setups.
                We hoped to combine their benefits, gaining robustness to different mixtures of adversarial and cooperative eigenspaces.
                However, with this setup, we could not find solutions that converged with simultaneous updates in purely adversarial games.

                \noindent
                \textbf{Generalize to allow solutions:}
                We generalized the setup to allow recurrent connections between momentum buffers with potentially negative coefficients -- see Figure~\ref{cm_fig:recurrent_comp_graph} and Appendix Algorithm~\ref{cm_alg:recurrent_momentum}.
                Optimizer parameters exist that allow convergence with simultaneous updates in purely adversarial games while being first-order with a single gradient evaluation -- see \resultName~\ref{cm_thm:theorem_existence}.
                However, this setup could introduce many optimizer parameters, have unintuitive behavior, and be unsuitable for analysis in general.
                So, we use a special case of this method to help solve these problems.

                \noindent
                \textbf{A simple solution:} With two momentum buffers and correctly chosen recurrent weights, we can conveniently interpret our buffers as the real and imaginary parts of one complex buffer -- see Appendix Figure~\ref{cm_fig:cm_comp_graph_complex}.
                This method is (a) capable of converging in purely adversarial games with simultaneous updates -- \resultName~\ref{cm_thm:theorem_existence}, (b) only introduces one new optimizer parameter -- the phase of the momentum coefficient, (c) is tractable to analyze and have intuitions for with Euler's formula -- ex., Equation~\ref{cm_eq:polar_complex_update}, (d) is trivial to implement in libraries supporting complex arithmetic -- see Figure~\ref{cm_fig:jax_code_change}, and (e) can be robust to games with different mixtures of cooperative and adversarial eigenspaces -- see Figures~\ref{cm_fig:partial_coop_phases} and \ref{cm_fig:partial_coop_phases_robustness}.

        \section{Complex Momentum}\label{cm_sec:theory}
            We describe our proposed method, where the momentum coefficient $\momCoeff \in \mathbb{C}$, step size $\lr \in \mathbb{R}$, momentum buffer $\momVal \in \mathbb{C}^{\bothDim}$, and player parameters $\bothParam \in \mathbb{R}^{\bothDim}$.
            The simultaneous (or Jacobi) update is:
            \begin{align*}\label{cm_eq:simul_update}\tag{SimCM}
                \momVal^{j+1} = \momCoeff \momVal^{j} - \bothGrad^j,\eqspace
                \bothParam^{j+1} = \bothParam^j + \Re\left(\lr \momVal^{j+1}\right)
            \end{align*}
            There are many ways to get a real-valued update from $\momVal \in \mathbb{C}$, but we only consider updates equivalent to classical momentum when $\momCoeff \in \mathbb{R}$.
            The update can use the imaginary part of the momentum buffer, which also works and could yield better solutions.
            However, we only use the real component of the momentum -- $\Re(\momVal)$ -- because this is the simplest setup and is sufficient to work.
            
            %
            \begin{figure}
                    \begin{algorithm}[H]
                        \caption[Simultaneous update complex momentum]{\ref{cm_eq:simul_update} Momentum} \label{cm_alg:simultaneous_complex}
                        \begin{algorithmic}[1]
                            \State $\momCoeff, \lr \in \mathbb{C}, \momVal^{0} \in \mathbb{C}^{\bothDim}, \bothParam^{\num{0}} \in \mathbb{R}^{\bothDim}$

                            \For{$j = \num{1} \dots N$ $\vphantom{\outGrad^j}$}
                                \State $\momVal^{j+1} = \momCoeff \momVal^j - \bothGrad^j$ 
                                \State $\bothParam^{j+1} = \bothParam^{j} + \Re\left(\lr \momVal^{j+1}\right)$ 
                            \EndFor
                            \State \Return $\bothParam^N$
                        \end{algorithmic}
                    \end{algorithm}
            \end{figure}
            We show the \ref{cm_eq:simul_update} update in Algorithm~\ref{cm_alg:simultaneous_complex} and visualize it in Appendix Figure~\ref{cm_fig:cm_comp_graph_complex}.
            We also show the alternating (or Gauss-Seidel) update, which is common for GAN training:

            \begin{align*}\label{cm_eq:alt_update}\tag{AltCM}
                \momVal^{j+1}_{\outSymbol} &= \momCoeff \momVal^{j}_{\outSymbol} - \bothGrad_{\outSymbol}\left(\bothParam^j\right),
                \outParam^{j+1} = \outParam^j + \Re\left(\lr \momVal^{j+1}_{\outSymbol}\right)\\
                \momVal^{j+1}_{\inSymbol} &= \momCoeff \momVal^{j}_{\inSymbol} - \bothGrad_{\inSymbol}\left(\outParam^{j+1}, \inParam^{j}\right),
                \inParam^{j+1} = \inParam^j + \Re\left(\lr \momVal^{j+1}_{\inSymbol}\right)
            \end{align*}

            \paragraph{Generalizing negative momentum:} Consider the negative momentum of \cite{gidel2018negative}: $\bothParam^{j+1} = \bothParam^j - \lr \bothGrad^j + \momCoeff \left(\bothParam^j - \bothParam^{j-1}\right)$.
            %
            Expanding \ref{cm_eq:simul_update} with $\momVal^j = \nicefrac{\left(\bothParam^j - \bothParam^{j-1}\right)}{\lr}$ for real momentum shows the negative momentum method of \cite{gidel2018negative} is a special case of our method:
            \begin{align}
                \bothParam^{j+1} &= \bothParam^j + \Re\left(\lr \left(\momCoeff\nicefrac{\left(\bothParam^j - \bothParam^{j-1}\right)}{\lr} - \bothGrad^j\right)\right)\\
                &= \bothParam^j - \lr \bothGrad^j + \momCoeff (\bothParam^j - \bothParam^{j-1})
            \end{align}
            %
            
            \subsection{Dynamics of Complex Momentum}\label{cm_sec:dynamics}
\begin{figure}
                    \centering
                        \begin{tikzpicture}
                            \centering
                            \node (img1){\includegraphics[trim={1.0cm .6cm .97cm .9cm},clip,width=.8\linewidth]{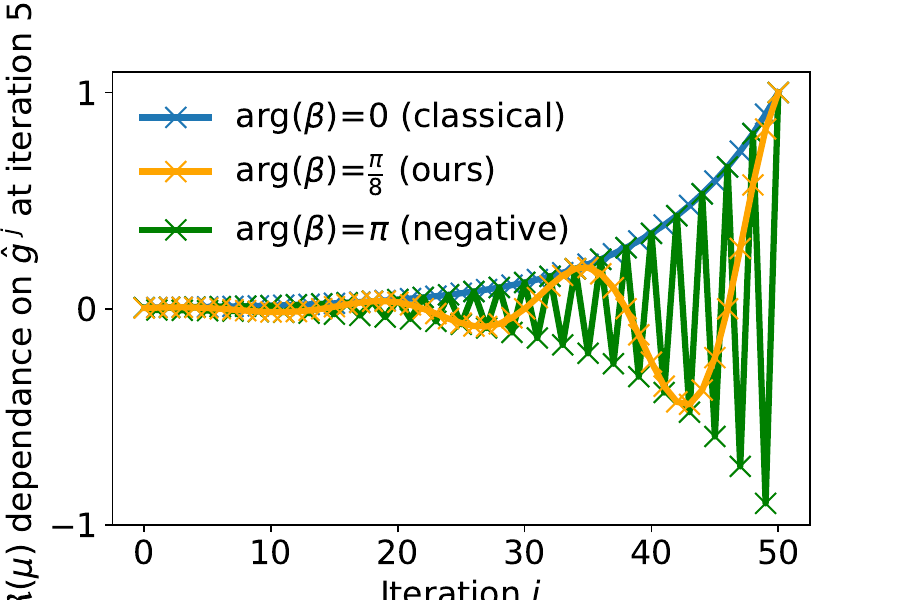}};
                            \node[left=of img1, node distance=0cm, rotate=90, xshift=2.0cm, yshift=-.8cm, font=\color{black}] {$\Re(\momVal^{50})$ dependence on $\bothGrad^{k}$};
                            \node[below=of img1, node distance=0cm, xshift=-.1cm, yshift=1.1cm,font=\color{black}] {Iteration $k$};
                        \end{tikzpicture}
                        \caption[The complex buffers dependence on gradient history]{
                            We show the real part of our momentum buffer -- which dictates the parameter update -- at the $50^{\textnormal{th}}$ iteration $\Re\left(\momVal^{50}\right)$ dependence on past gradients $\bothGrad^{k}$ for $k = 1 \dots 50$.
                            The magnitude of the momentum is fixed at $ | \momCoeff | = \num{.9}$ as in Figure~\ref{cm_fig:trajectories_limited}.
                            Euler's formula is used in Equation~\ref{cm_eq:polar_complex_update} to find the dependence or coefficient of $\bothGrad^{k}$ via $\Re\left(\momVal^{50}\right) = -\sum_{k=0}^{k=50}|\momCoeff|^{k} \cos\left(k \arg\left(\momCoeff\right)\right)\bothGrad^{j-k}$.
                            Complex momentum allows for smooth changes in the buffer's dependence on past gradients.
                        }
                        \label{cm_fig:frequency_vis}
                \end{figure}

                For simplicity, we assume NumPy-style~\citep{numpy} component-wise broadcasting for operations like taking the real-part $\Re(\boldsymbol{z})$ of vector $\boldsymbol{z} = [z_1, \dots, z_n] \in \mathbb{C}^{n}$, with proofs in the Appendix.
                Expanding the buffer updates with the polar components of $\momCoeff$ gives intuition for complex momentum:
                \begin{align}\label{cm_eq:polar_complex_update}
                    \begin{split}
                        \momVal^{j+1} = \momCoeff \momVal^{j} - \bothGrad^j &\iff
                        \momVal^{j+1} = \momCoeff \left(\momCoeff \left(\cdots\right) - \bothGrad^{j-1}\right) - \bothGrad^j \iff \\
                        \momVal^{j+1} = -&\sum_{k=0}^{k=j}\momCoeff^{k}\bothGrad^{j-k} \iff\\
                        \Re\left(\momVal^{j+1}\right) = -&\sum_{k=0}^{k=j} \cos\left(k \arg\left(\momCoeff\right)\right) \,\, |\momCoeff|^{k} \bothGrad^{j-k},\\
                        \Im\left(\momVal^{j+1}\right) = -&\sum_{k=0}^{k=j} \sin\left(k \arg\left(\momCoeff\right)\right) \,\, |\momCoeff|^{k} \bothGrad^{j-k} 
                    \end{split}
                \end{align}
                The final line is simply by Euler's formula (Equation~\ref{cm_eq:euler_formula}).
                Equation~\ref{cm_eq:polar_complex_update} shows how $\momCoeff$ controls the momentum buffer $\momVal$ by having $|\momCoeff|$ dictate the decay rates of the prior gradient, while $\arg(\momCoeff)$ controls the oscillation frequency between the addition and subtraction of the prior gradients, which we visualize in Figure~\ref{cm_fig:frequency_vis}.
                \newline
                \begin{figure}
                    \centering
                        \begin{tikzpicture}
                            \centering
                            \node (img1){\includegraphics[trim={1.0cm .9cm .97cm .9cm},clip,width=.70\linewidth]{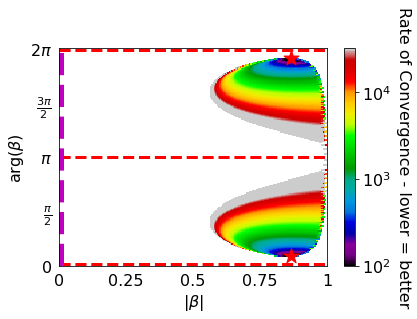}};
                            \node[left=of img1, node distance=0cm, rotate=90, xshift=2.0cm, yshift=-.9cm, font=\color{black}] {Momentum phase $\arg(\momCoeff)$};
                            \node[below=of img1, node distance=0cm, xshift=-.1cm, yshift=1.3cm,font=\color{black}] {Momentum magnitude $| \momCoeff |$};
                            \node[right=of img1, node distance=0cm, rotate=270, xshift=-2.1cm, yshift=-.9cm, font=\color{black}] {Number of steps to converge};
                        \end{tikzpicture}
                        \caption[Heatmap of convergence rate over optimizer parameters for complex momentum]{
                            We show how many steps simultaneous complex momentum on a Dirac-GAN takes for a set solution distance.
                            We fixed the step size $\lr = \num{0.1}$ as in Figure~\ref{cm_fig:trajectories_limited}, while varying the phase and magnitude of our momentum $\momCoeff  = |\momCoeff| \exp\left(i \arg\left(\momCoeff\right)\right)$.
                            There is a {\color{red}red} star at the minimum, dashed {\color{red}red} lines at real $\momCoeff$, and a {\color{magenta}magenta} line for simultaneous gradient descent.
                            No real-valued $\momCoeff$ converges for this -- or any -- $\lr$ with simultaneous updates ~\citep{gidel2018negative}.
                            Appendix Figure~\ref{cm_fig:det_minmax_expAvg_alt_phase_tune} compares this with alternating updates (\ref{cm_eq:alt_update}), where negative momentum converges.
                        }
                        \label{cm_fig:det_minmax_expAvg_simul_phase_tune}
                \end{figure}
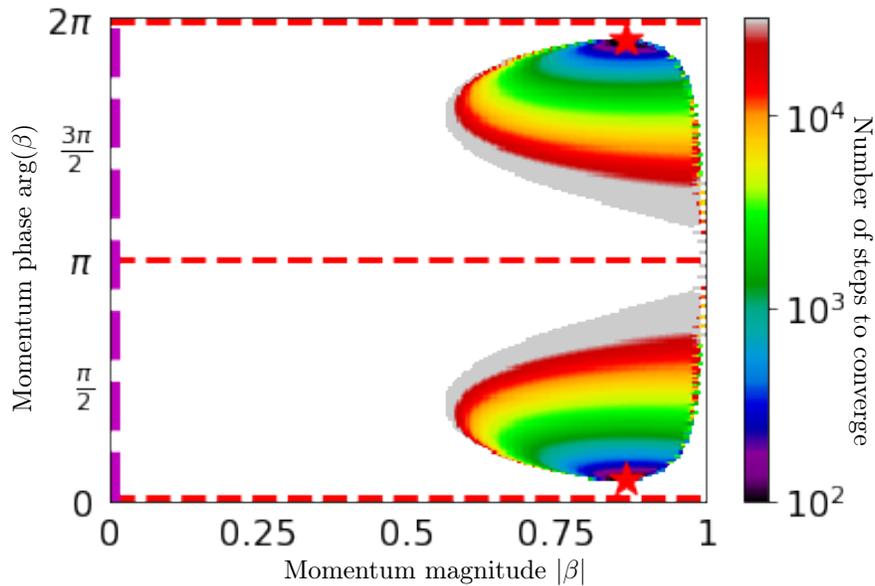
                
                Expanding the parameter updates with the Cartesian components of $\lr$ and $\momCoeff$ is key for Theorem~\ref{cm_thm:theorem}, which characterizes the convergence rate:
                \vspace{-0.01\textheight}
                \begin{align}\label{cm_eq:cartesian_complex_update}
                    \begin{split}
                        \momVal^{j+1} &= \momCoeff \momVal^{j} - \bothGrad^j \iff \\
                        \Re\left(\momVal^{j+1}\right) &= \Re(\momCoeff)  \Re\left(\momVal^j\right) - \Im\left(\momCoeff\right)  \Im\left(\momVal^j\right)  - \Re\left(\bothGrad^j\right),\\
                        \Im\left(\momVal^{j+1}\right) &= \Im\left(\momCoeff\right)  \Re\left(\momVal^{j}\right) + \Re\left(\momCoeff\right)  \Im\left(\momVal^{j}\right) 
                    \end{split}
                \end{align}
                \vspace{-0.02\textheight}
                \begin{align}\label{cm_eq:parameter_dynamics}
                        \bothParam^{j+1} = \bothParam^j + \Re\left(\lr \momVal^{j+1}\right) \iff
                        \bothParam^{j+1} = \bothParam^j - \lr\bothGrad^j + \Re\left(\lr \momCoeff\right)  \Re\left(\momVal^j\right) - \Im\left(\lr \momCoeff\right)\Im\left(\momVal^j\right)
                \end{align}
                So, we can write the next iterate with a fixed-point operator:\vspace{-0.01\textheight}
                \begin{equation}\label{cm_eq:fixed_point_operator}
            	    \left[\Re\left(\momVal^{j+1}\right), \Im\left(\momVal^{j+1}\right), \bothParam^{j+1}\right] = \fixedPointOp \left(\left[\Re\left(\momVal^{j}\right), \Im\left(\momVal^{j}\right), \bothParam^{j}\right]\right)
            	\end{equation}
                Equations~\ref{cm_eq:cartesian_complex_update} and \ref{cm_eq:parameter_dynamics} allow us to write the Jacobian of $\fixedPointOp$, which can be used to bound convergence rates near fixed points, which we call the Jacobian of the augmented dynamics of buffer $\momVal$ and joint-parameters $\bothParam$ and denote with:
                \vspace{-0.01\textheight}
                \begin{equation}
            	    \dynamics = 
            	    \nabla_{[\momVal, \bothParam]}\fixedPointOp = \begin{bmatrix}
                        \Re(\momCoeff) \identity & -\Im(\momCoeff) \identity & -\nabla_{\bothParam} \bothGrad\\
                        \Im(\momCoeff) \identity & \Re(\momCoeff) \identity & 0\\
                        \Re(\lr \momCoeff) \identity & -\Im(\lr \momCoeff) \identity & \identity - \lr\nabla_{\bothParam}\bothGrad\\
                        \end{bmatrix}
            	\end{equation}
            	So, for quadratic losses our parameters evolve via:
            	\begin{equation}\label{cm_eq:dyn_system}
            	    \left[\Re\left(\momVal^{j+1}\right), \Im\left(\momVal^{j+1}\right), \bothParam^{j+1}\right]^{\transpose} = \dynamics \, \left[\Re\left(\momVal^{j}\right), \Im\left(\momVal^{j}\right), \bothParam^{j}\right]^{\transpose}
            	\end{equation}
            	We bound convergence rates near fixed points by using the spectrum of $\dynamics$ with Theorem~\ref{cm_thm:theorem}.
                \begin{restatable}[Consequence of Prop. 4.4.1 ~\citet{bertsekas2008nonlinear}]{thm}{mainThm}
                    \label{cm_thm:theorem}
                    Complex momentum's convergence rate:
                    If the spectral radius $\spectralRadius\left(\nabla \fixedPointOp\left(\momVal^{*}, \bothParam^{*}\right)\right) < 1$, then, for $\left[\momVal, \bothParam\right]$ in a neighborhood of $\left[\momVal^*, \bothParam^*\right]$, the distance of $\left[\momVal^{j}, \bothParam^{j}\right]$ to the stationary point $\left[\momVal^*, \bothParam^*\right]$ converges at a linear rate $\mathcal{O}\left(\left(\spectralRadius\left(\dynamics\right) + \epsilon\right)^j\right)\!\!, \forall \epsilon > 0$.
                \end{restatable}
                %
                %
                Here, linear convergence means $\lim_{j \to \infty}  \nicefrac{\| \bothParam^{j+1} - \bothParam^*\|}{\| \bothParam^{j} - \bothParam^*\|} \in (\num{0}, \num{1})$, where $\bothParam^*$ is a fixed point.
                To converge, we select the optimizer parameters $\lr, \momCoeff$ so that the augmented dynamics spectral radius $\spectralRadius \left(\dynamics\left(\lr, \momCoeff\right)\right) < 1$---with the dependence on $\lr$ and $\momCoeff$ now explicit.
                To easily find convergent parameters, we express $\spectrum\left(\dynamics\left(\lr, \momCoeff\right)\right)$ in terms of the spectrum $\spectrum\left(\nabla_{\bothParam}\bothGrad\right)$, as in Theorem 3 in \cite{gidel2018negative}:
                %
                \begin{equation}\label{cm_eq:spectrum_desired}
                    \boldsymbol{f}\left(\spectrum\left(\nabla_{\bothParam}\bothGrad\right), \lr, \momCoeff\right) = \spectrum\left(\dynamics\left(\lr, \momCoeff\right)\right)
                \end{equation}
                We provide a Mathematica command in Appendix~\ref{cm_sec:poly} for a cubic polynomial characterizing $\boldsymbol{f}$ with coefficients that are functions of $\lr, \momCoeff$ and $\eigval \in \spectrum\left(\nabla_{\bothParam}\bothGrad\right)$, whose roots are eigenvalues of $\dynamics$, which we use in subsequent results.
                \cite{o2015adaptive} and \cite{lucas2018aggregated} mention that we often do not know the condition number, eigenvalues -- or the mixture of cooperative and adversarial eigenspaces  -- of a set of functions that we are optimizing, so we try to design algorithms which work over a large range.
                Sharing this motivation, we consider convergence behavior in games ranging from purely adversarial to cooperative.
                
                In Section~\ref{cm_sec:exp_spectrum} at every non-real $\momCoeff$ we could select $\lr$ and $| \momCoeff |$ so Algorithm~\ref{cm_alg:simultaneous_complex} converges.
                We define \emph{almost-positive} to mean $\arg(\momCoeff) = \epsilon$ for small $\epsilon$, and show there are almost-positive $\momCoeff$ which converge.
                \begin{restatable}[Convergence of Complex Momentum]{result}{resultExist}
                    \label{cm_thm:theorem_existence}
                         There exist $\lr \in \mathbb{R}, \momCoeff \in \mathbb{C}$ so Algorithm~\ref{cm_alg:simultaneous_complex} converges for bilinear zero-sum games.
                         More-so, for small $\epsilon$ (we show for $\epsilon = \frac{\pi}{\num{16}}$), if $\arg(\momCoeff) = \epsilon$ (i.e., almost-positive) or $\arg(\momCoeff) = \pi - \epsilon$ (i.e., almost-negative), then we can select $\lr, |\momCoeff|$ to converge.
                \end{restatable}
                %
                \textbf{Why show this?}
                Our result complements \cite{gidel2018negative} who show that for all real $\lr, \momCoeff$ Algorithm~\ref{cm_alg:simultaneous_complex} \emph{does not} converge.
                We include the proof for bilinear zero-sum games, but the result generalizes to some purely adversarial games near fixed points, such as Dirac GANs~\citep{mescheder2017numerics}.
                The second part of the result shows evidence that there is a sense in which the only $\momCoeff$ that do not converge are real (with simultaneous updates on purely adversarial games).
                It also suggests a form of robustness because almost-positive $\momCoeff$ can approach acceleration in cooperative eigenspaces while converging in adversarial eigenspaces, so almost-positive $\momCoeff$ may be useful when our games have an uncertain or variable mixture of real and imaginary eigenvalues like GANs.
                Sections \ref{cm_sec:exp_spectrum}, \ref{cm_sec:train_small_gan}, and \ref{cm_sec:train_big_gan} investigate this further.
                

               \begin{figure}[h]
                    \centering
                        \begin{tikzpicture}
                            \centering
                            \node (img1){\includegraphics[trim={.1cm 1.cm 1.0cm .35cm},clip,width=.45\linewidth]{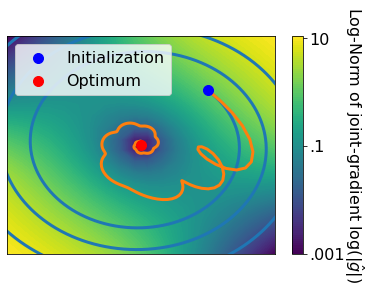}};
                            \node[left=of img1, rotate=90, node distance=0cm, xshift=.9cm, yshift=-.9cm, font=\color{black}] {Discriminator};
                            \node[below=of img1, node distance=0cm, xshift=-.7cm, yshift=1.2cm,font=\color{black}] {Generator};
                            \node[right=of img1, rotate=270, node distance=0cm, xshift=-2.1cm, yshift=-1.2cm, font=\color{black}] {Norm of joint-gradient $\|\bothGrad\|$};

                            \node (img3)[right=of img1, node distance=0cm, xshift=-.5cm, yshift=-0.25cm, font=\color{black}]{\includegraphics[trim={.8cm .8cm .25cm .2cm},clip,width=.45\linewidth]{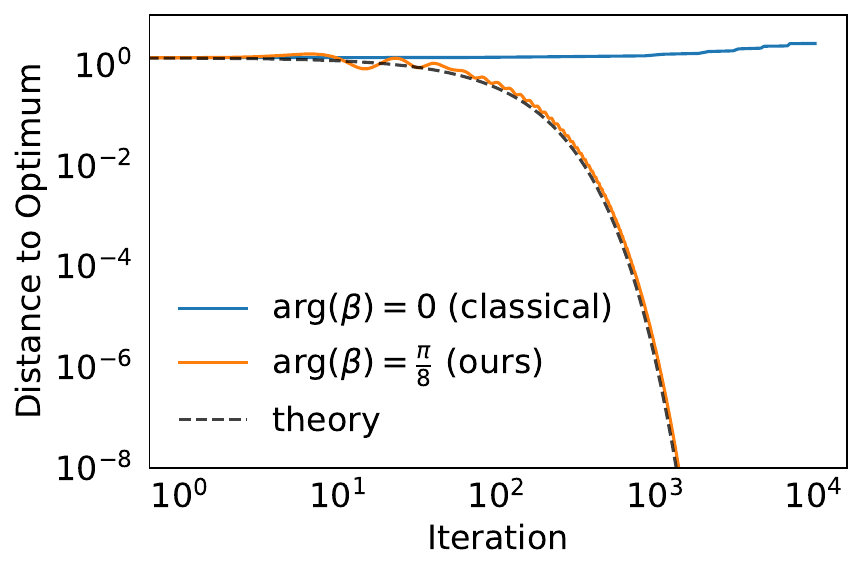}};
                            \node[left=of img3, rotate=90, node distance=0cm, xshift=1.75cm, yshift=-1cm, font=\color{black}] {Distance to optimum};
                            \node[below=of img3, node distance=0cm, xshift=.15cm, yshift=1.25cm,font=\color{black}] {Iterations};
                        \end{tikzpicture}
                        \caption[Complex momentum on a Dirac GAN]{
                            Complex momentum helps correct rotational dynamics when training a Dirac-GAN~\citep{mescheder2018training}.
                            \emph{Top:} Parameter trajectories with step size $\lr = \num{0.1}$ and momentum $\momCoeff = \num{0.9} \exp\left(i \nicefrac{\pi}{8}\right)$.
                            We include the classical, real, and positive momentum, which diverges for any $\lr$.
                            \emph{Bottom:} The distance to the optimum, which has a convergence rate that matches our prediction with Theorem~\ref{cm_thm:theorem} and Equation~\ref{cm_eq:spectrum_desired}.
                        }
                        \vspace{-0.01\textheight}
                        \label{cm_fig:trajectories_limited}
                \end{figure}
            \subsection{What about Acceleration?}\label{cm_sec:acceleration}
            
                With classical momentum, finding the step size $\lr$ and the momentum $\momCoeff$ to optimize the convergence rate is tractable if $\num{0} < l \leq L$ and $\spectrum\left(\nabla_{\bothParam}\bothGrad\right) \in \left[l, L\right]^{\bothDim}$~\citep{goh2017momentum} -- that is, we have a $l$-strongly convex loss and $L$-Lipschitz loss.
                The conditioning $\condition = \nicefrac{L}{l}$ can characterize the problem's difficulty.
                Gradient descent with an appropriate $\lr$ can achieve a convergence rate of $\frac{\condition - \num{1}}{\condition + \num{1}}$, but using momentum with an appropriate $\left(\lr^*, \momCoeff^*\right)$ can achieve an \emph{accelerated} rate of $\spectralRadius^* = \frac{\sqrt{\condition} - \num{1}}{\sqrt{\condition} + \num{1}}$.
                However, there is no consensus on constraining $\spectrum\left(\nabla_{\bothParam}\bothGrad\right)$ in games to obtain tractable and useful results.
                Candidate constraints include monotonic vector fields generalizing notions of convexity or vector fields with bounded eigenvalue norms that capture a kind of sensitivity~\citep{azizian2019tight}.
                %
                Figure~\ref{cm_fig:gan_spectrum_main} shows $\spectrum\left(\nabla_{\bothParam}\bothGrad\right)$ for a GAN, motivating the varying of $\lr$ and $\momCoeff$ for each player as in Section~\ref{cm_sec:train_big_gan}.

            \subsection{Implementing Complex Momentum}
                Complex momentum is trivial to implement with libraries supporting complex arithmetic like JAX~\citep{jax2018github} or PyTorch~\citep{paszke2017automatic}.
                Given an SGD implementation, we often only need to change a few lines of code; see Figure~\ref{cm_fig:jax_code_change}.
                In addition, Equations~\ref{cm_eq:cartesian_complex_update} and \ref{cm_eq:parameter_dynamics} can be easily used to implement Algorithm~\ref{cm_alg:simultaneous_complex} in a library without complex arithmetic.
                More sophisticated optimizers like Adam can trivially support complex optimizer parameters with real-valued updates, which we explore in Section~\ref{cm_sec:train_big_gan}.
                
            \subsection{Scope and Limitations}
                For some games, we need higher-than-first-order information to converge -- e.g., pure-response games -- because the first-order information for a player is identically zero.
                So, momentum methods that only use first-order info will generally not converge.
                However, we can combine the methods with second-order information, as in the prior chapters, using momentum algorithms.
                The computational cost of complex momentum is almost identical to classical and negative momentum, except we now have a buffer with twice as many real parameters.
                We require one more optimization hyperparameter than the classical momentum, for which we provide an initial guess in Section~\ref{cm_sec:init_guess}.
    \vspace{-0.01\textheight}
    \section{Experiments}\label{cm_sec:experiments}
        We investigate complex momentum's performance in training GANs and games with different cooperative and adversarial eigenspaces mixtures, showing improvements over standard baselines.
        
        \textbf{Overview:} We start with a purely adversarial Dirac-GAN and zero-sum games with known solutions $\bothParam^* = \left(\outParam^*, \inParam^*\right)$ and spectrums $\spectrum\left(\nabla_{\bothParam} \bothGrad\right)$, so we can assess convergence rates.
        Next, we evaluate GANs that generate $\num{2}$D distributions because they are simple enough to train with a plain, alternating SGD.
        Finally, we look at scaling to larger-scale GANs on images with brittle optimization, which requires optimizers like Adam.
        Complex momentum provides benefits in each setup.
        We only compare to first-order optimization methods, despite there being various second-order methods due to limitations discussed in Section~\ref{cm_sec:limitation_existing}.
        \vspace{-0.01\textheight}
        \subsection{Optimization in Purely Adversarial Games}\label{cm_sec:single_param_det}
                Here, we consider optimizing the Dirac-GAN objective, which is surprisingly hard and where many classical optimization methods fail, because $\spectrum(\nabla_{\bothParam} \bothGrad)$ is imaginary near solutions:
                \vspace{-0.01\textheight}
                \begin{equation}\label{cm_eq:min_max_xy}
                    \min_x \max_y -\log\left(1 + \exp\left(-xy\right)\right) - \log(2)
                \end{equation}
                %
                Figure~\ref{cm_fig:trajectories_limited} empirically verifies the convergence rates given by Theorem~\ref{cm_thm:theorem} with Equation~\ref{cm_eq:spectrum_desired}, by showing the optimization trajectories with simultaneous updates.
                
                Figure~\ref{cm_fig:det_minmax_expAvg_simul_phase_tune} shows how the momentum components $\momCoeff$ affect the convergence rates with simultaneous updates and fixed step size.
                The best $\momCoeff$ was almost positive (that is, $\arg(\momCoeff) = \epsilon$ for small $\epsilon$).
                We repeat this experiment with alternating updates in Appendix Figure~\ref{cm_fig:det_minmax_expAvg_alt_phase_tune}, standard in GAN training.
                There, almost-positive momentum is best (but negative momentum also converges), and the benefit of alternating updates depends on whether we can parallelize player gradient evaluations.
        
        \subsection{Adversarialnesses Effect on Convergence}\label{cm_sec:exp_spectrum}
            %
            We compare optimization with first-order methods for purely adversarial, cooperative, and mixed games.
            We use the following game to easily interpolate between these regimes, where if $\boldsymbol{\gamma} = \identity$, it is purely adversarial, while if $\boldsymbol{\gamma} = \boldsymbol{0}$ it is purely cooperative:
            \vspace{-0.01\textheight}
            \begin{equation}\label{cm_eq:bilinear_adversarial_interpolate}
                    \min_{\boldsymbol{x}} \max_{\boldsymbol{y}}  \boldsymbol{x}^\transpose \left(\boldsymbol{\gamma} \boldsymbol{A}\right) \boldsymbol{y} +
                    \boldsymbol{x}^\transpose \left(\left(\identity - \boldsymbol{\gamma}\right)\boldsymbol{B}_1\right) \boldsymbol{x} - \boldsymbol{y}^\transpose \left(\left(\identity - \boldsymbol{\gamma}\right)\boldsymbol{B}_2\right) \boldsymbol{y}
            \end{equation}
            \vspace{-0.02\textheight}
            
            \begin{figure}[t]
                \vspace{-0.015\textheight}
                \centering
                \begin{tikzpicture}
                    \centering
                    \node (img1){\includegraphics[trim={.0cm .0cm .0cm .0cm},clip,width=.7\linewidth]{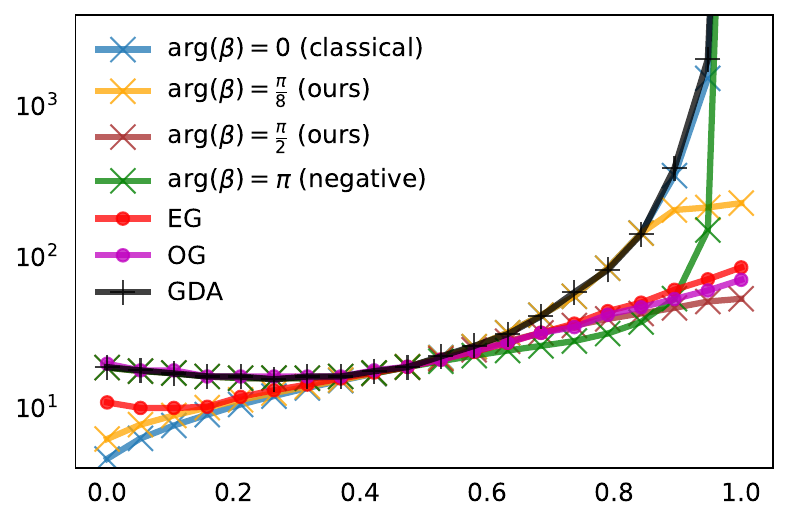}};
                    \node[left=of img1, node distance=0cm, rotate=90, xshift=2.75cm, yshift=-1.0cm, font=\color{black}] {\# gradient evaluations to converge};
                    \node[below=of img1, node distance=0cm, xshift=.75cm, yshift=1.3cm,font=\color{black}]{Max adversarialness $\gamma_{max}$};
                \end{tikzpicture}
                \caption[Complex momentum bilinear convergence rates]{
                    We compare first-order methods convergence rates on the game in Equation~\ref{cm_eq:bilinear_adversarial_interpolate}, with $\boldsymbol{A} = \boldsymbol{B}_1 = \boldsymbol{B}_2$ diagonal and entries linearly spaced in $\left[\nicefrac{\num{1}}{\num{4}}, \num{4}\right]$.
                    We interpolate from purely cooperative to a mixture of purely cooperative and adversarial eigenspaces in $\spectrum\left(\nabla_{\bothParam}\bothGrad\right)$ by making $\boldsymbol{\gamma}$ diagonal with $\gamma_j \sim U\left[0, \gamma_{max}\right]$, inducing the $j^{\textnormal{th}}$ eigenvalue pair to have $\arg\left(\eigval_j\right) \approx \pm \gamma_j \frac{\pi}{2}$.
                    Therefore, $\gamma_{max}$ controls the largest possible eigenvalue $\arg$ or \emph{max adversarialness}.
                    Every method generalizes gradient descent-ascent (GDA) by adding an optimizer parameter, tuned via grid search.
                    {\color{blue}Positive momentum} and {\color{green}negative momentum} do not converge if there are purely adversarial eigenspaces (i.e., $\gamma_{max} = 1$).
                    {\color{orange}Almost-positive momentum $\arg\left(\momCoeff\right) = \epsilon$} $> 0$ like ${\color{orange}\nicefrac{\pi}{8}}$ allows us to approach the acceleration of positive momentum if sufficiently cooperative (i.e., $\gamma_{max} < \num{.5}$), while still converging if there are purely adversarial eigenspaces (i.e., $\gamma_{max} = 1$).
                    Tuning $\arg\left(\momCoeff\right)$ with complex momentum performs competitively with {\color{red}extragradient (EG)}, {\color{magenta}optimistic gradient (OG)} for any adversarialness -- ex., ${\color{brown}\arg\left(\momCoeff\right) = \nicefrac{\pi}{2}}$ does well if there are purely adversarial eigenspaces (i.e., $\gamma_{max} = 1$).
                }
                \vspace{-0.015\textheight}
                \label{cm_fig:partial_coop_phases}
            \end{figure}
    %
    \begin{figure}[h!]
            \centering
            \begin{tikzpicture}
                \centering
                \node (img1){\includegraphics[trim={.0cm .0cm .0cm .0cm},clip,width=.8\linewidth]{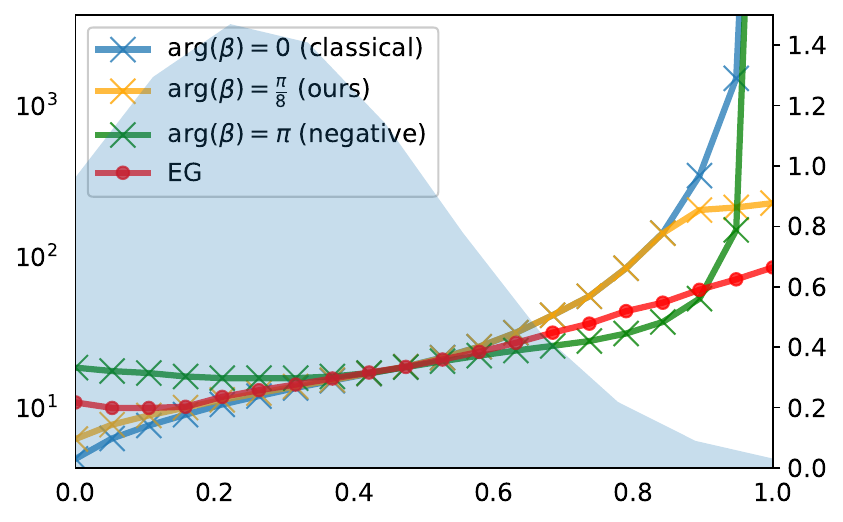}};
                \node[left=of img1, node distance=0cm, rotate=90, xshift=2.75cm, yshift=-1.0cm, font=\color{black}] {\# gradient evaluations to converge};
                \node[right=of img1, node distance=0cm, rotate=270, xshift=-2.75cm, yshift=-1.0cm, font=\color{black}] {Probability of game with this $\gamma_{max}$};
                \node[below=of img1, node distance=0cm, xshift=-.1cm, yshift=1.3cm,font=\color{black}]{Max adversarialness $\gamma_{max}$};
            \end{tikzpicture}
            \caption[Complex momentum convergence in eigenspaces]{
                We show a setup where almost-positive, complex momentum with ${\color{orange}\arg(\momCoeff) = \nicefrac{\pi}{8}}$ minimizes risk over gradient evaluations by robustly converging for the distribution of games shown in light blue.
                We compare with the methods in Figure~\ref{cm_fig:partial_coop_phases}.
                In particular, we believe that this distribution of games may be a useful proxy for what occurs during GAN training.
                The distribution was calculated during the GAN training in Figure~\ref{cm_fig:gan_spectrum_main} by approximating the distribution of $\gamma_{max}$ in the last $\num{1000}$ training steps with a truncated Gaussian using the empirical mean and variance.
                We approximated $\gamma_{max}$ by looking at the subset of eigenspaces in which the parameters lie and filtering any eigenspaces with a negative real part or where the component of the parameters in that eigenspace was smaller than $10^{-5}$ for simple visualization.
            }
            \label{cm_fig:partial_coop_phases_robustness}
        \end{figure}
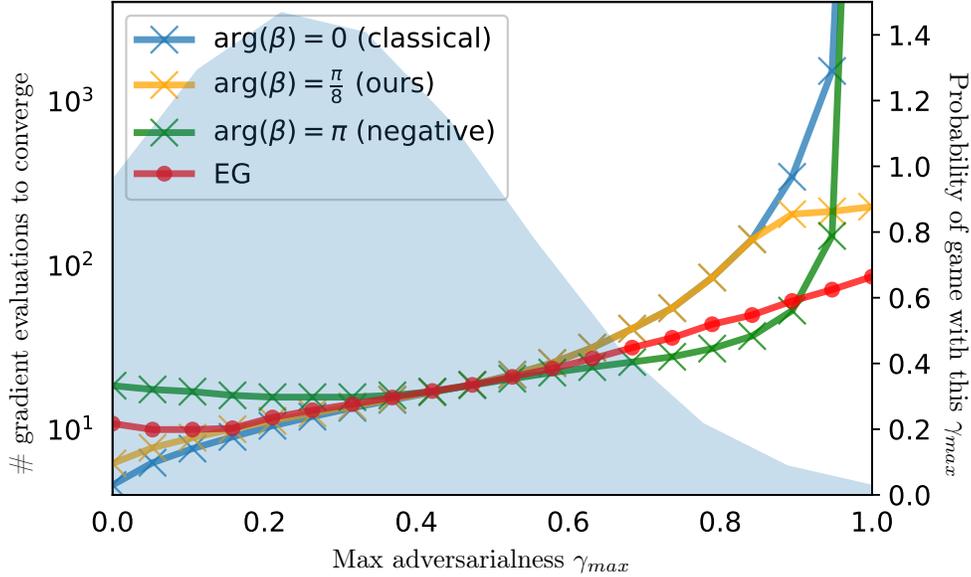

            \noindent
            Appendix Figure~\ref{cm_fig:spectrum_vary_phase} explores $\spectrum(\dynamics)$ in purely adversarial games for various $\lr, \momCoeff$, generalizing Figure 4 in \cite{gidel2018negative}.
            At every non-real $\momCoeff$---i.e., $\arg(\momCoeff) \neq \pi$ or $\num{0}$---we can select convergent $\lr, |\momCoeff|$.
            
            Figure~\ref{cm_fig:partial_coop_phases} compares the first-order algorithms as we interpolate from purely cooperative games (i.e., minimization) to mixtures of purely adversarial and cooperative eigenspaces because this setup range can occur during GAN training -- see Figure~\ref{cm_fig:gan_spectrum_main}.
            Our baselines are simultaneous SGD (or gradient descent-ascent (GDA)), extragradient (EG) \citep{korpelevich1976extragradient}, optimistic gradient (OG) \citep{chiang2012online, rakhlin2013optimization, daskalakis2017training}, and momentum variants.
            We additionally \emph{tuned the extrapolation parameters for EG and OG separately} -- a nonstandard modification so that EG and OG are competitive with momentum in cooperative eigenspaces; see Appendix Section~\ref{cm_sec:app_spectrum}.
            We show how many gradient evaluations for a set solution distance, and EG costs two evaluations per update.
            We optimize convergence rates for each game and method by grid search, as is common for optimization parameters in deep learning.

            The $x$-axis in Figures~\ref{cm_fig:partial_coop_phases} and \ref{cm_fig:partial_coop_phases_robustness} is the \emph{max adversarialness}.
            Intuitively, this is just a rectangle's (normalized) width bounding the spectrum in polar coordinates.

            \textbf{Takeaway:} In the regime where all eigenspaces are cooperative -- i.e., $\gamma_{max} < .5$ or\\ $\max_{\eigval \in \spectrum(\nabla_{\bothParam}\bothGrad)} \nicefrac{|\Im(\eigval)|}{|\Re(\eigval)|} < 1$ -- the best method is classical, positive momentum.
            Otherwise, we will benefit from a method to learn in games.
            If we have purely adversarial eigenspaces -- i.e.,$\gamma_{max} = 1$ -- then GDA, positive and negative momentum fail to converge, while EG, OG, and complex momentum can converge.
            Choosing any non-real momentum $\momCoeff$ allows robust convergence for every eigenspace mixture.
            More so, almost-positive momentum $\momCoeff$ allows us to approach acceleration when cooperative while still converging if there are purely adversarial eigenspaces.
            
            In games like GANs, our eigendecomposition is infeasible to compute and changes during training -- see Appendix Figure~\ref{cm_fig:gan_decomp} -- so we want an optimizer that converges robustly for any potential $\gamma_{max}$.
            A natural goal is to minimize the risk over the possible games of the number of gradient evaluations to converge.
            Figure~\ref{cm_fig:partial_coop_phases_robustness} displays some methods from Figure~\ref{cm_fig:partial_coop_phases} superimposed with a distribution of potential games.
            \textbf{Takeaway:} The best method is complex momentum because most of the game distribution is in cooperative eigenspaces (i.e., $\gamma_{max} < .5$) but could have purely adversarial eigenspaces (i.e., $\gamma_{max} = 1$).
            Also, we believe that the displayed game distribution could be a reasonable proxy for what is encountered in GAN training.

        \subsection{Training GANs on 2D Distributions}\label{cm_sec:train_small_gan}
            Here, we investigate improving GAN training using alternating gradient descent updates with complex momentum.
            We examine alternating updates because they are standard in GAN training~\citep{goodfellow2014generative, brock2018large, wu2019logan}.
            We focus on comparisons to positive and negative momentum, the strongest baselines.
            Note that EG and OG do not have obvious generalizations to alternating updates.
            We train to generate a $\num{2}$D mixture of Gaussians because more complicated distributions require more complicated optimizers than SGD.
            Figure~\ref{cm_fig:jax_code_change} shows all the changes necessary to use the JAX momentum optimizer for our updates, with complete details in Appendix~\ref{cm_sec:app_2d_gan}.
            We evaluated the log-likelihood of GAN samples under the mixture as an imperfect proxy for matching.
            
            Appendix Figure~\ref{cm_fig:gan_heatmaps} shows heatmaps for tuning $\arg(\momCoeff)$ and $|\momCoeff|$ with select step sizes.
            \textbf{Takeaway:}
            The best momentum was found at the almost-positive $\momCoeff \approx \num{0.7} \exp(i \nicefrac{\pi}{\num{8}})$ with step size $\lr = 0.03$, and for each $\lr$ we tested a broad range of non-real $\momCoeff$ outperformed any real $\momCoeff$.
            This suggests we can often improve GAN training with alternating updates and complex momentum.
            
        \subsection{Training BigGAN with a Complex Adam}\label{cm_sec:train_big_gan}
            Here, we investigate improving larger-scale GAN training with complex momentum.
            However, larger-scale GANs train with more complicated optimizers than gradient descent -- like Adam~\citep{kingma2014adam} -- and have notoriously brittle optimization.
            We looked at training BigGAN~\citep{brock2018large} on CIFAR-10~\citep{krizhevsky2009learning} but were unable to succeed with optimizers other than \citep{brock2018large}-supplied setups due to brittle optimization.
            So, we attempted to change the procedure minimally by taking \citep{brock2018large}-supplied code \href{https://github.com/ajbrock/BigGAN-PyTorch}{{\color{blue}here}} which was trained with Adam and making only the $\momCoeff_1$ parameter -- analogous to momentum -- complex.
            The modified complex Adam is shown in Algorithm~\ref{cm_alg:complex_adam}, where the momentum bias correction is removed to match our theory better.
            It is an open question of how best to carry over the design of Adam (or other optimizers) to the complex setting.
            Training each BigGAN took $\num{10}$ hours on an NVIDIA T4 GPU, so Figure~\ref{cm_fig:biggan_inception_heatmap} and Table~\ref{cm_tab:biggan} took about $\num{1000}$ and $\num{600}$ GPU hours, respectively.
            
            Figure~\ref{cm_fig:biggan_inception_heatmap} shows a grid search on $\arg\left(\momCoeff_1\right)$ and $|\momCoeff_1|$ for a BigGAN trained with Algorithm~\ref{cm_alg:complex_adam}.
            We only changed $\momCoeff_1$ for the discriminator's optimizer.
            \textbf{Takeaway:}
            The best momentum was at the almost-positive $\momCoeff_1 \approx \num{0.8} \exp\left(i \nicefrac{\pi}{\num{8}}\right)$, whose samples are in Appendix Figure~\ref{cm_fig:biggan_samples}.

            \begin{figure}[h!]
                \centering
                    \begin{algorithm}[H]
                        \caption[Complex Adam variant without momentum bias-correction]{Complex Adam variant without momentum bias-correction} \label{cm_alg:complex_adam}
                        \begin{algorithmic}[1]
                            \State $\momCoeff_1 \in \mathbb{C}, \momCoeff_2 \in [\num{0}, \num{1})$
                            \State $\lr \in \mathbb{R}^{+}, \epsilon \in \mathbb{R}^{+}$
                            \For{$j = 1 \dots N$}
                                \State $\!\!\!\!\!\momVal^{j+1} = \momCoeff_1 \momVal^{j} - \gradSymbol^j$
                                \State $\!\!\!\!\!\boldsymbol{v}^{j+1} = \momCoeff_2 \boldsymbol{v}^{j} + \left(1-\momCoeff_2\right) \left(\gradSymbol^j\right)^{\num{2}}$
                                \State $\!\!\!\!\!\hat{\boldsymbol{v}}^{j+1} = \frac{\boldsymbol{v}^{j+1}}{1 - \left(\momCoeff_2\right)^j}$
                                \State $\!\!\!\!\!\bothParam^{j+1} = \bothParam^{j} + \lr \frac{\Re\left(\momVal^j\right)}{\sqrt{\hat{\boldsymbol{v}}^{j+1}} + \epsilon}$
                            \EndFor
                            \State \Return $\bothParam^N$
                        \end{algorithmic}
                    \end{algorithm}
        \end{figure}
        \begin{figure}
                    \vspace{-0.01\textheight}
                    \begin{table}[H]
                        \vspace{-0.02\textheight}
                    	\centering
                    	\begin{tabular}{lllll}  
                    		&\multicolumn{4}{c}{CIFAR-10 BigGAN IS for \num{10} seeds} \\
                    		\cmidrule(lr){2-5}
                    		Discriminator $\momCoeff_1$ & Max & Mean & Median & Std.\\  
                    		\cmidrule(lr){1-5}
                    		$\num{0}$, BigGAN default & $\num{9.10}$ & $\num{8.93}$ & $\num{8.89}$ & $\num{0.076}$ \\
                    		$.8\exp(i \nicefrac{\pi}{\num{8}})$, ours & $\num{9.25} ({\color{green}+.15})$ & $\num{8.97} ({\color{green}+.04})$ & $\num{8.97} ({\color{green}.08})$ & $\num{0.079} ({\color{red}.003})$ \\
                    		$.8$ & $\num{9.05}({\color{red}-.05})$ & $\num{7.19}({\color{red}-1.7})$ & $\num{8.82}({\color{red}.07})$ & $\num{2.753} ({\color{red}2.67})$\\
                    	\end{tabular}
                    	\caption[BigGAN inception scores with complex momentum]{
                    	    We display the inception scores (IS) found over $\num{10}$ runs for training BigGAN on CIFAR-10 with various optimizer settings.
                    	    We use a complex Adam variant described in Algorithm~\ref{cm_alg:complex_adam}, where we only tuned $\momCoeff_1$ for the discriminator.
                    	    The best parameters found in Appendix Figure~\ref{cm_fig:biggan_inception_heatmap} were $\momCoeff_1 = \num{.8}\exp\left(i \nicefrac{\pi}{\num{8}}\right)$, which improved the maximum IS in our runs, as well as the final mean and median IS of the BigGAN authors baseline, which was the SoTA optimizer in this setting to the best of our knowledge.
                    	    We tested $\momCoeff_1 = \num{.8}$ to see if the gain was only due to tuning $|\momCoeff_1|$, which occasionally failed and decreased the IS.
                    	 }
                    	\label{cm_tab:biggan}
                    \end{table}
                \vspace{-0.035\textheight}
            \end{figure}
            %
            We tested the best momentum value over $\num{10}$ seeds against the author-provided baseline in Appendix Figure~\ref{cm_fig:biggan_inception_runs}, with the results summarized in Table~\ref{cm_tab:biggan}.
            \textbf{Takeaway:} Our method improves the mean inception score (IS) with a $t$-test significance of $\num{0.071}$, which shows the desired phenomena -- i.e., complex momentum improving training -- with a reasonable significance.
            Furthermore, the complex momentum improves the best IS found with $\num{9.25} ({\color{green}+.15} \text{ over author code}, {\color{green}+.03} \text{ author reported})$.
            \cite{brock2018large} reported a single inception score on CIFAR-10 of $\num{9.22}$, but the best we could reproduce on the seeds with the provided PyTorch code and settings was $\num{9.10}$.

            We trained a real momentum $|\momCoeff_1| = \num{0.8}$ to see if the improvement was solely due to the adjustment of the momentum magnitude.
            This occasionally failed to train and decreased the best IS over re-runs, showing that we benefit from a non-zero $\arg\left(\momCoeff_1\right)$.

            %
    \vspace{-0.0125\textheight}
    \subsection{A Practical Initial Guess for the Momentum's Argument}\label{cm_sec:init_guess}
        We propose a practical initial guess for our new hyperparameter $\arg(\momCoeff)$.
        \resultName~\ref{cm_thm:theorem_existence} shows that we can use almost-real momentum coefficients (i.e., $\arg(\momCoeff)$ is near, but not equal to $\num{0}$).
        Figure~\ref{cm_fig:partial_coop_phases} shows almost-positive $\momCoeff$ approach acceleration in cooperative eigenspaces while converging in all eigenspaces.
        Figure~\ref{cm_fig:gan_spectrum_main} shows that GANs can have cooperative and adversarial eigenspaces.
        Figure~\ref{cm_fig:partial_coop_phases_robustness} shows a distribution of games -- from GAN training -- where almost-positive $\momCoeff$ robustly converges and minimizes the risk of gradient evaluations.
        Figures~\ref{cm_fig:gan_heatmaps} and \ref{cm_fig:biggan_inception_heatmap} perform a grid search over $\arg(\momCoeff)$ for GANs, finding that almost-positive $\arg(\momCoeff) \approx \nicefrac{\pi}{\num{8}}$ works in both cases.
        Also, by minimally changing $\arg(\momCoeff)$ from $0$ to a small $\epsilon$, we can minimally change other hyperparameters in our model, which is useful to adapt existing, brittle setups like in GANs.
        Based on this, we propose an initial guess of $\arg(\momCoeff) = \epsilon$ for a small $\epsilon > 0$, where $\epsilon = \nicefrac{\pi}{\num{8}}$ worked in our GAN experiments.

  \vspace{-0.0125\textheight}
  \section{Related Work}\label{cm_sec:
 related-work}
        \textbf{Accelerated first-order methods}:
            A broad body of work exists using momentum-type methods~\citep{polyak1964some, nesterov1983method, nesterov2013introductory, maddison2018hamiltonian}, with a recent focus on deep learning~\citep{sutskever2013importance, zhang2017yellowfin, choi2019empirical, zhang2019lookahead, chen2020self}.
            But these focus on momentum for minimization as opposed to in games.
        
        \textbf{Learning in games}:
            Various works approximate response-gradients - some by differentiating through optimization~\citep{foerster2018learning, mescheder2017numerics, maclaurin2015gradient} -
            or leverage game eigenstructure during optimization~\citep{letcher2019differentiable, nagarajan2020chaos, omidshafiei2020navigating, czarnecki2020real, gidel2020minimax, perolat2020poincar}.
        
        \textbf{First-order methods in games}:
            \cite{zhang2021don, zhang2020unified, ibrahim2020linear, bailey2020finite, jin2020local, azizian2019tight, nouiehed2019solving, zhang2020optimality, zhang2023deep} characterize convergence with various first-order methods.
            \cite{gidel2018negative} is the closest work to ours; showing a negative momentum can help in some games.
            \cite{zhang2020suboptimality} note the suboptimality of negative momentum in a class of games.
            \cite{azizian2020accelerating, domingo2020average} investigate acceleration in some games.
        
        \textbf{Bilinear zero-sum games}:
            \cite{zhangconvergence} study the convergence of gradient methods in bilinear zero-sum games, extending \citet{gidel2018negative}, showing we can achieve faster convergence with separate step sizes and momentum for each player or tuning the extragradient step size.
            \cite{loizou2020stochastic} provide convergence guarantees for games that satisfy a \emph{sufficiently bilinear} condition.
        
        \textbf{Learning in GANs}:
            Various works make GAN training easier with methods that leverage the structure of the game~\citep{Liu2020Towards, peng2020training, albuquerque2019multi, wu2019logan, hsieh2019finding}.
            \cite{metz2016unrolled} approximate the discriminator's response function by differentiating through optimization.
            \cite{mescheder2017numerics} find solutions by minimizing the norm of the players' updates.
            These methods and various others~\citep{qin2020training, schafer2019implicit, jolicoeur2019connections} require higher-order information.
            \cite{daskalakis2017training, gidel2018variational, chavdarova2019reducing} look at first-order methods.
            \cite{mescheder2018training} explore problems for GAN training convergence and \cite{berard2019closer} show that GANs have significant rotations in learning.
        
    \section{Conclusion}\label{cm_sec:conclusion}
        In this chapter, we generalized existing momentum methods for learning in differentiable games by allowing complex momentum with real-valued updates.
        Our method robustly converges in games with a different range of cooperative and adversarial eigenspace mixtures than existing methods.
        We also presented a practical generalization of our method to the Adam optimizer, which we used to improve BigGAN training.
        More generally, we highlight and lay the groundwork for investigating optimizers that work well with various mixtures of cooperative and competitive dynamics in games.

        
    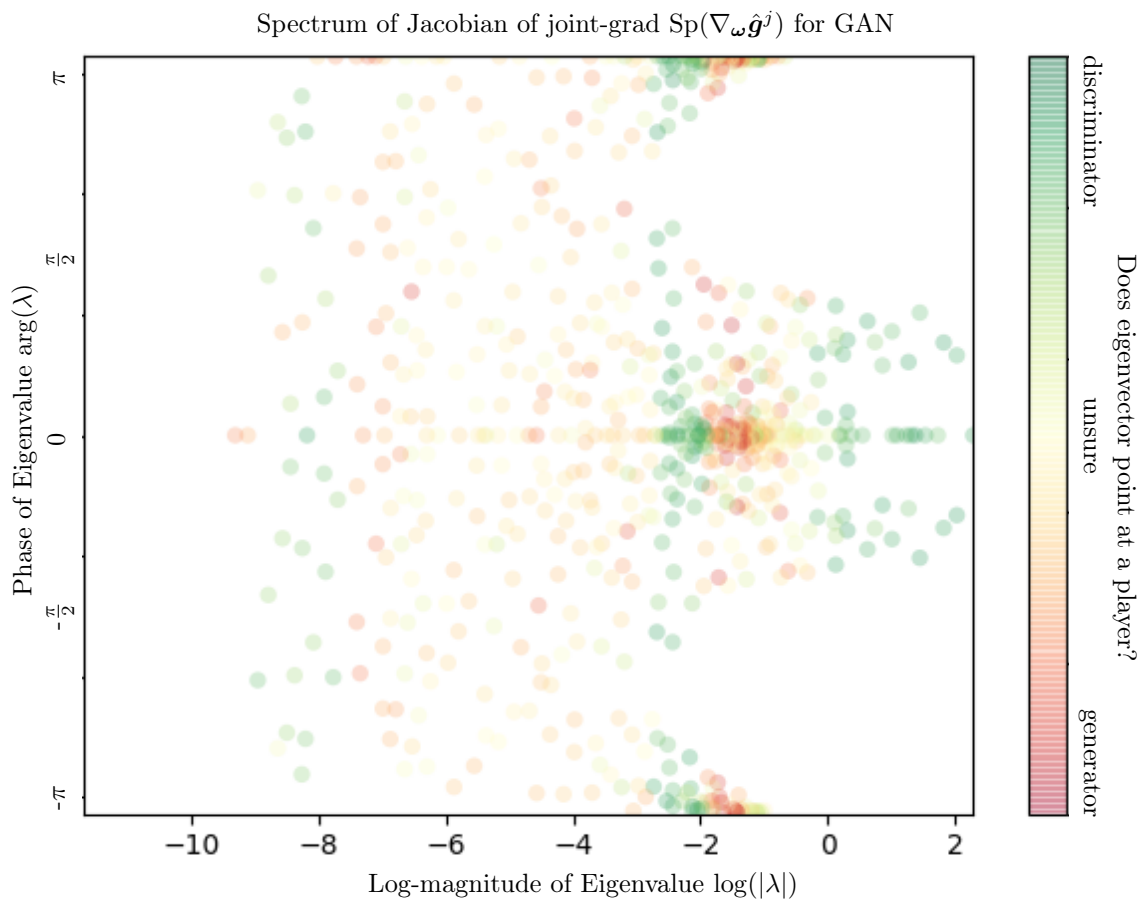
\begin{figure}
        \centering
        \begin{tikzpicture}
            \centering
            \node (img1){\includegraphics[trim={1.3cm .5cm .9cm .475cm},clip,width=.86\linewidth]{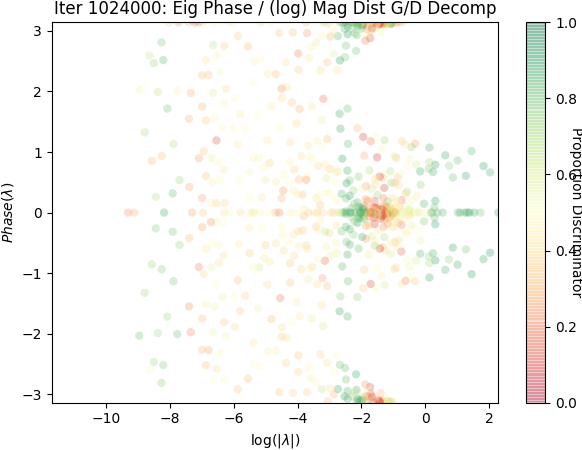}};
            \node (img1ytick)[left=of img1, node distance=0cm, rotate=90, xshift=5.3cm, yshift=-0.825cm, font=\color{black}] {-$\pi$ \hspace{\ytickspace} -$\frac{\pi}{2}$ \hspace{\ytickspace} $\num{0}$ \hspace{\ytickspace} \hphantom{-}$\frac{\pi}{2}$ \hspace{\ytickspace} \hphantom{-}$\pi$};
            \node[left=of img1ytick, node distance=0cm, rotate=90, xshift=7.25cm, yshift=-0.53cm, font=\color{black}] {Phase of Eigenvalue $\arg(\eigval)$};
            
            \node[above=of img1, node distance=0cm, xshift=-.0cm, yshift=-1.1cm,font=\color{black}] {Spectrum of Jacobian of joint-grad $\spectrum(\nabla_{\bothParam}\bothGrad^{j})$ for GAN};
            \node[below=of img1, node distance=0cm, xshift=.1cm, yshift=1.2cm,font=\color{black}] {Log-magnitude of Eigenvalue $\log(|\eigval|)$};
            
            \node (img1colortick)[right=of img1, node distance=0cm, rotate=270, xshift=-5.5cm, yshift=-0.83cm, font=\color{black}] {discriminator \hspace{0.15\textwidth} unsure \hspace{0.19\textwidth} generator};
            \node[right=of img1colortick, node distance=0cm, rotate=270, xshift=-7.75cm, yshift=-.55cm, font=\color{black}] {Does eigenvector point at a player?};
        \end{tikzpicture}
        \caption[Log-polar eigenspace visualization for GANs]{
            A log-polar visualization reveals structure in the spectrum for a GAN at the end of the training on a $\num{2}$D mixture of Gaussians with a $\num{1}$-layer discriminator and generator, so the joint-parameters $\bothParam \in \mathbb{R}^{\num{723}}$.
            Appendix Figure~\ref{cm_fig:gan_decomp} shows the spectrum through training.
            There is a mixture of many cooperatives (that is, real or $\arg(\eigval) \approx \num{0}, \pm \pi$) and some adversarial (i.e., imaginary or $\arg(\eigval) \approx \pm \frac{\pi}{\num{2}}$) eigenvalues, so -- contrary to what the name may suggest -- generative adversarial networks are not purely adversarial.
            We may benefit from optimizers that leverage this structure, such as complex momentum.
            \newline
            Eigenvalues are colored if the associated eigenvector is mostly in one player's part of the joint-parameter space -- see Appendix Figure~\ref{cm_fig:gan_decomp} for details on this.
            Many eigenvectors lie mostly in the space of (or point at) one player.
            The structure of the set of eigenvalues for the discriminator ({\color{green}green}) is different from the generator ({\color{red}red}), but further investigation of this is an open problem.
            In particular, this may motivate separate optimizer choices for each player, as in Section~\ref{cm_sec:train_big_gan}.
        }
        \label{cm_fig:gan_spectrum_main}
    \end{figure}
    %
    \section*{Acknowledgements}
        Resources used in preparing this research were provided, in part, by the Province of Ontario, the Canadian government through CIFAR, and companies sponsoring the Vector Institute.
        Paul Vicol was supported by an NSERC PGS-D Scholarship.
        We thank Guodong Zhang, Guojun Zhang, James Lucas, Romina Abachi, Jonah Phillion, Will Grathwohl, Jakob Foerster, Murat Erdogdu, Ken Jackson, Ioannis Mitliagkis, and Barbara Norton for feedback and discussion.

    \subsection*{My Contributions Towards this Paper As it Pertains to the Thesis}
        I contributed the initial idea to this project, ran all small-scale experiments, collaborated on full-scale experiments, did all the theoretical work, and did most of the writing.
        This paper was also submitted to arXiv~\citep{lorraine2021complexARXIV}.

    \chapter{Lyapunov Exponents for Diversity in Differentiable Games}\label{chap:lyapunovExponent}
    \vspace{-0.01\textheight}
    Ridge Rider (RR) is a method to find diverse solutions in optimization problems by following eigenvectors of the Hessian (``ridges'').
    RR is designed for conservative gradient systems (i.e., settings with a single loss/potential function), where it branches at saddles --- easy-to-find bifurcations.
    We generalize RR to nonconservative, multi-agent gradient systems by proposing the Generalized Ridge Rider (GRR) method to find arbitrary bifurcations.
    We give theoretical motivation for our method by leveraging tools from the field of dynamical systems.
    We construct novel toy problems where we visualize new phenomena while giving insight into high-dimensional problems of interest.
    Finally, we empirically evaluate our method by finding diverse solutions to the iterated prisoners' dilemma and machine learning problems, including generative adversarial networks.



%


         


    

    
    \vspace{-0.01\textheight}
    \section{Introduction}\label{lyapunov_sec:introduction}
        Selecting solutions with desirable properties that an arbitrary (global or local) minimum might not have is often useful.
        For example, finding solutions in image classification using shapes that generalize more effectively than textures.
        Important examples of this in single-objective minimization are seeking solutions that generalize to unseen data in supervised learning~\citep{geirhos2018imagenet, geirhos2020shortcut}, in policy optimization~\citep{cully2015robots}, and generative models \citep{song2020score}.
        Many real-world systems are not so simple and instead involve multiple agents, each using a different subset of parameters to minimize their own objective.
        Some examples are generative adversarial networks (GANs)~\citep{goodfellow2014generative,pfau2016connecting},
        actor-critic models~\citep{pfau2016connecting},
        curriculum learning~\citep{bengio2009curriculum, baker2019emergent, balduzzi2019open, sukhbaatar2017intrinsic},
        hyperparameter optimization~\citep{domke2012generic,maclaurin2015gradient,lorraine2018stochastic, mackay2019self, raghu2020teaching, lorraine2020optimizing, raghu2021meta}, 
        adversarial examples~\citep{bose2020adversarial, yuan2019adversarial}, 
        learning models~\citep{rajeswaran2020game, baconlagrangian, nikishin2021control},
        domain adversarial adaptation~\citep{acuna2021fdomainadversarial},
        neural architecture search~\citep{zoph2016neural,real2019regularized,liu2018darts,grathwohl2018gradient, adam2019understanding},
        multi-agent settings~\citep{foerster2018learning}
        and meta-learning~\citep{ren2018meta, rajeswaran2019meta,ren2020flexible}.
        In these settings, the aim is to find one equilibrium -- of potentially many equilibria -- where the agents exhibit some desired behavior.
        
        For example, in the iterated prisoners' dilemma (Section \ref{lyapunov_subsec:new_problems}), solutions favoring reciprocity over unconditional defection result in higher returns for all agents.
        In GANs, solutions often generate a subset of the modes of the target distribution~\citep{arjovsky2017wasserstein}, and in Hanabi, some solutions coordinate better with humans~\citep{hu2020other}.
        Existing methods often find solutions in small subspaces -- even after many random restarts, as shown in Table~\ref{lyapunov_tab:results}.
        By finding a diverse set of equilibria, we may be able to (a) find solutions with a better joint outcome, (b) develop stronger generative models in adversarial learning, or (c) find solutions that coordinate better with humans.
         
        Recently, Ridge Rider (RR)~\citep{parker2020ridge} proposed a method to find diverse solutions in \emph{single-objective} optimization.
        RR is a branching tree search, which starts at a stationary point and then follows different \emph{eigenvectors of the Hessian} (``ridges'') with negative eigenvalues, moving downhill from saddle point to saddle point.
        In settings where multiple agents each have their own objective (i.e., games), the relevant generalization of the Hessian --- the \emph{game Hessian}~\citep{balduzzi2018mechanics} in Equation~\ref{lyapunov_eq:game_hess} --- is not symmetric. 
        Thus, the game Hessian generally has complex eigenvalues (EVals) with associated complex eigenvectors, making RR not directly applicable.
        
        In this chapter, we generalize RR to multiagent settings using machinery from \emph{dynamical systems}.
        We connect RR with methods for finding \emph{bifurcation} points, i.e., points where small changes in the initial parameters lead to different optimization dynamics and learning outcomes.
        We propose novel metrics inspired by \emph{Lyapunov exponents}~\citep{katok1997introduction} that measure the speed with which learning trajectories separate.
        \cutForACM{These metrics generalize the branching criterion from RR, allowing us to locate a broad class of potential bifurcations and find more diverse behavior.}
        \cutForACM{Starting points with rapid trajectory separation can be found via gradient-based optimization by differentiating through the exponent calculation, which is implemented efficiently for large models with Jacobian-vector products.}
        %
        %
        Our contributions include:
        \begin{itemize}
            \item Connections between finding diverse solutions and Lyapunov exponents, allowing us to leverage a broad body of work in dynamical systems.
            \item Proposing Generalized Ridge Rider (GRR), scaling Ridge Rider (RR) to differentiable games.
            \cutForACM{\item Presenting novel diagnostic problem settings based on high dimensional games like the iterated prisoners dilemma (IPD) and GANs to study various bifurcation types.} 
            \item Compared to existing methods, GRR finds diverse solutions in the IPD, spanning cooperation, defection, and reciprocity.
            \item Lastly, we present larger-scale experiments on GANs --- a model class of interest to the machine learning community.
        \end{itemize}
    \newpage

    \section{Background}\label{lyapunov_sec:background}
        Appendix Table~\ref{lyapunov_tab:TableOfNotation} summarizes our notation.
        As before, consider the single-objective problem:
        \begin{equation}\label{lyapunov_eq:single_objective_opt}
                \paramSymbol^{*} \in \argmin_{\paramSymbol} \loss\left(\paramSymbol\right) 
        \end{equation}
        Again, we denote the gradient of the loss at parameters $\paramSymbol^{j}$ by $\gradSymbol^j \defeq\ \gradSymbol\left(\paramSymbol^j\right) \defeq \left. \nabla_{\paramSymbol} \loss\left(\paramSymbol\right) \right|_{\paramSymbol^{j}}$.
        We can locally minimize the loss $\loss$ using gradient descent with step size $\lr$:
        \begin{equation}\label{lyapunov_eq:single_objective_gd}
            \paramSymbol^{j+1} = \paramSymbol^{j} - \alpha \gradSymbol^j
        \end{equation}
        Due to the potential non-convexity of $\loss$, multiple stationary points can exist, and gradient descent will only find a particular solution based on the initialization $\paramSymbol^{0}$.
        \subsection{Ridge Rider (RR)}
            Ridge Rider (RR)~\citep{parker2020ridge} finds diverse solutions in single-objective minimization problems.
            The method first finds a saddle point, either analytically as in tabular reinforcement learning or by minimizing the gradient norm.
            
            Then, RR branches the optimization procedure following different directions (or ``ridges'') given by the eigenvectors of the Hessian $\hess = \nabla_{\paramSymbol} \gradSymbol = \nabla_{\paramSymbol} \left(\nabla_{\paramSymbol} \loss\right)$. 
            The complete computation of the eigendecomposition of $\hess$, i.e., its eigenvectors and eigenvalues, is often prohibitively expensive.
            However, we can efficiently access some of the eigenspaces via Hessian-vector products $\hess \boldsymbol{v} = \nabla_{\paramSymbol} \left(\left(\nabla_{\paramSymbol} \loss\right) \boldsymbol{v}\right)$ ~\citep{pearlmutter1994fast, tensorflow2015-whitepaper, paszke2017automatic, jax2018github}.
        
        \subsection{Optimization in Games}
            Instead of simply optimizing a single loss, optimization in games involves multiple agents, each with a loss function that can depend on other agents.
            Again, we look at $\num{2}$-player games with players (denoted by $A$ and $B$) who want to minimize their loss -- $\loss_A\left(\paramSymbol_A, \paramSymbol_B\right)$ or  $\loss_B\left(\paramSymbol_A, \paramSymbol_B\right)$ --  with their parameters -- $\paramSymbol_A$ or $\paramSymbol_B$.
            \begin{align}\label{lyapunov_eq:nash_equilibrium}
                \outParam^{*} \in \argmin_{\outParam} \outLoss\left(\outParam, \inParam^{*}\right),
                \inParam^{*} \in \argmin_{\inParam} \inLoss\left(\outParam^{*}, \inParam\right)
            \end{align}
            %
            %
            One of the simplest optimization methods is to find local solutions simply by following the players' gradients, but this is often unstable~\citep{bailey2020finite}.
            Here, $\outGrad^j = \outGrad\left(\outParam^j, \inParam^j\right)$ is an estimator for $ \nabla_{\outParam} \outLoss |_{\outParam^j, \inParam^j}$ with $\inGrad^j$ defined analogously, and the simultaneous gradient update is:
            \begin{align}\label{lyapunov_eq:multi_objective_gd_long}\tag{SimSGD}
                \outParam^{j+1} = \outParam^{j} - \lr \outGrad^j,\quad\quad
                \inParam^{j+1} = \inParam^{j} - \lr \inGrad^j
            \end{align}
            Again, we denote the concatenation of all players' parameters (or joint-parameters) $\bothParam = \left[\outParam, \inParam\right] \in \mathbb{R}^{\bothDim}$ and the joint-gradient vector field $\bothGrad: \mathbb{R}^{\bothDim} \to \mathbb{R}^{\bothDim}$, denoted at the $j^{\textnormal{th}}$ iteration by:
            \begin{equation}
                \bothGrad^j = \bothGrad(\bothParam^j) = \left[\outGrad\left(\bothParam^j\right), \inGrad\left(\bothParam^j\right)\right] = \left[\outGrad^j, \inGrad^j\right]
            \end{equation}
            We write the next iterate in \ref{lyapunov_eq:multi_objective_gd_long} with the fixed-point operator $\lyapunovfixedPointOp$:\vspace{-0.01\textheight}
            \begin{equation}\label{lyapunov_eq:fixed_point_operator}
                \bothParam^{j+1} = \lyapunovfixedPointOp_{SGD} \left(\bothParam^{j}\right) = \bothParam^{j} - \lr \bothGrad^{j}
            \end{equation}
            As in Chapter~\ref{chap:complexMomentum}, the Jacobian of the fixed-point operator $\lyapunovfixedPointOp$ -- denoted $\jacFixedPointOp$ -- is useful for analysis, including bounding convergence rates near fixed-points~\citep{bertsekas2008nonlinear} and finding points where local changes to parameters cause convergence to qualitatively different solutions~\citep{hale2012dynamics}.
            The Jacobian of the fixed-point operator crucially depends on the Jacobian of the joint-gradient $\bothGrad$, which is called the \emph{game Hessian}~\citep{letcher2019differentiable} because it generalizes the Hessian:
            \begin{equation}\label{lyapunov_eq:game_hess}
                \bothHess \defeq \nabla_{\bothParam} \bothGrad = \begin{bmatrix}
                    \nabla_{\outParam}^2 \outLoss & \nabla_{\outParam} \nabla_{\inParam} \outLoss \\
                    \nabla_{\inParam} \nabla_{\outParam} \inLoss^\top & \nabla_{\inParam}^2 \inLoss
                \end{bmatrix}
            \end{equation}
            \begin{equation}\label{lyapunov_eq:jac)fixed_point_operator}
                \jacFixedPointOp_{SGD} \defeq \nabla_{\bothParam} \lyapunovfixedPointOp_{SGD} \left(\bothParam\right) = \identity - \lr \bothHess
            \end{equation}
            Figure~\ref{lyapunov_fig:fig_new_problems_baseline} shows a game with a solution we can only converge to using an appropriate optimizer.
            Thus, we must incorporate information about the optimizer to generalize RR, which we do by using the (largest) eigenvalues/vectors of $\jacFixedPointOp$ instead of the (most negative) eigenvalues/vectors of $\bothHess$.
            
            The crucial difference between optimization with a single and multiple objectives is as follows:
            In single-objective optimization, following the gradient forms trajectories from a conservative vector field because $\bothHess = \hess$ is the Hessian of the loss, which is symmetric and has real eigenvalues.
            However, in games with multiple objectives, $\bothHess$ can be non-symmetric and have complex eigenvalues, resulting in a nonconservative vector field from optimization, opening the door to many new phenomena.

    \vspace{-0.01\textheight}
    \section{Methods for Generalizing RR}\label{lyapunov_sec:theory}
        Here, we present two key contributions for generalizing RR to games.
        We first connect diversity in optimization to the concept of bifurcations, where a small change in the parameters causes a large change in the optimization trajectories.
        Second, we introduce Lyapunov exponents~\citep{katok1997introduction} and easy-to-optimize variants to find these bifurcations. 
        
        \vspace{-0.01\textheight}
        \subsection{Connecting Diversity and Bifurcations}
            In dynamical systems, \emph{bifurcations} are regions of the parameter space where small perturbations result in qualitatively different optimization trajectories.
            In general, a dynamical system can contain various bifurcation types.
            In contrast, saddle points (and their separatrix) are the only relevant bifurcation type in conservative gradient vector fields. 
            Consequently, their eigenvectors play a key role in the shape of the phase portraits, which are geometric representations of the underlying dynamics.
            In particular, the negative eigenvectors are orthogonal to separatrices~\citep{tabor1989chaos}, boundaries between regions in our system with different dynamical behavior, thus providing directions to move in for finding different solutions.
            This perspective provides a novel view of RR, which branches at saddle points, the relevant class of bifurcation points in single-loss optimization. 
            
            However, in the literature on dynamical systems, many bifurcation types have been studied~\citep{katok1997introduction}.
            This inspires generalizing RR to nonconservative gradient fields (e.g., multi-agent settings) where various bifurcations occur. 
            See Figure~\ref{lyapunov_fig:fig_new_problems_baseline} for a \emph{Hopf bifurcation}~\citep{hale2012dynamics} or Figure~\ref{lyapunov_fig:complicated_toy} for various others.
            
        \subsection{Lyapunov Exponents for Bifurcations}\label{lyapunov_sec:lyap_for_bifurcation}
\begin{figure}[t!]
                \vspace{-0.03\textheight}
                \centering
                \begin{tikzpicture}
                    \centering
                    \node (img){\includegraphics[trim={.0cm .0cm .0cm .0cm},clip, width=.97\linewidth]{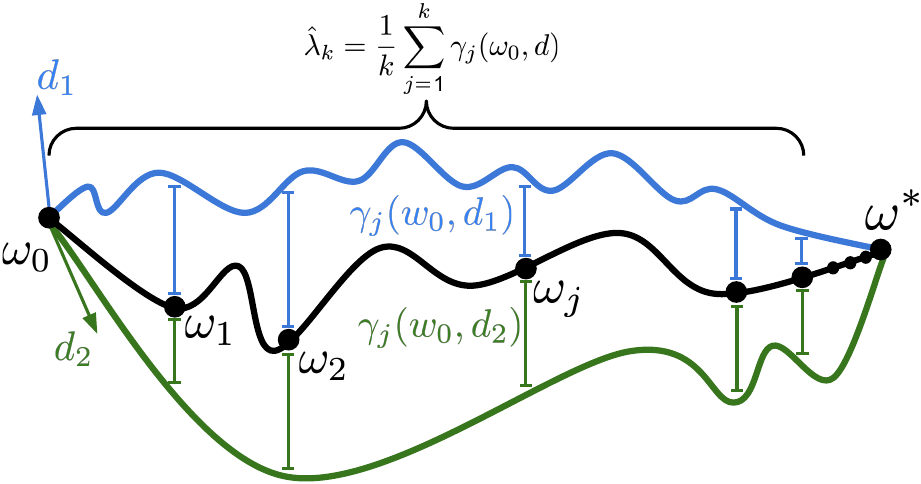}};
                \end{tikzpicture}
                \caption[Lyapunov exponent visualization]{
                    Visualization of the components to a Lyapunov exponent $\lyap_k\left(\bothParam_0, \displacement\right)$ described in Equations~\ref{lyapunov_eq:lyap_calc2},~\ref{lyapunov_eq:lyap_calc}, which measures how quickly trajectories separate starting at a point $\bothParam_0$ in direction $\displacement$.
                    Here, the optimization trajectory iterates $\bothParam_j$ accumulate at a fixed-point $\bothParam^{*}$.
                    We show two displacements -- $\color{blue}\displacement_1$ and ${\color{green}\displacement_2}$ -- resulting in separate ``perturbed'' trajectories as shown in {\color{blue}blue} and {\color{green}green}.
                    We measure the separation rate between the true and perturbed trajectories at the $j^{\textnormal{th}}$ optimizer iteration with the Lyapunov term $\gamma_j\left(\bothParam_0, \displacement\right)$.
                    The exponent $\lyap_k\left(\bothParam_0, \displacement\right)$ is the average of the first $k$ terms.
                    See Figure~\ref{lyapunov_fig:fig_mix} for actual trajectories on a toy problem used in an exponent calculation.
                }\label{lyapunov_fig:lyap_calc}
                \vspace{-0.01\textheight}
            \end{figure}

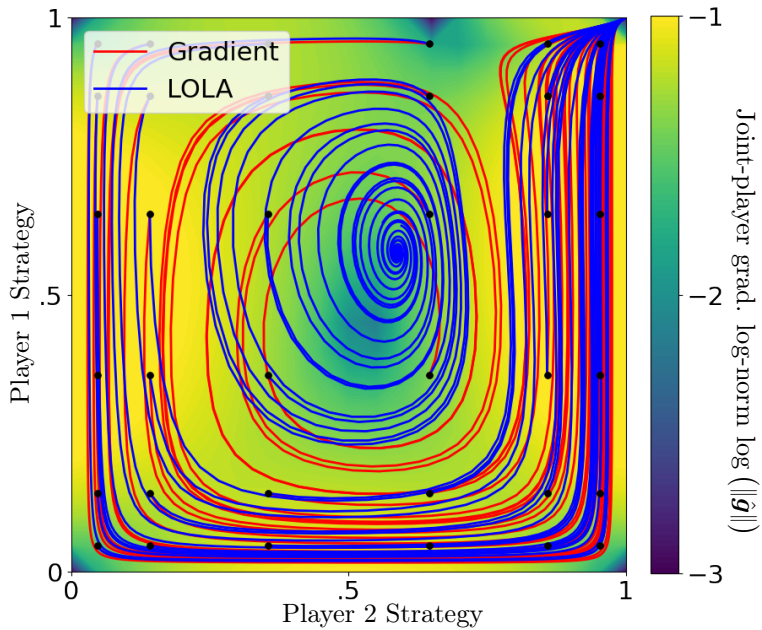
\begin{figure}
            \centering
            \begin{tikzpicture}
                \centering
                \node (img){\includegraphics[trim={.25cm .25cm 1.05cm .25cm},clip, width=.6\linewidth]{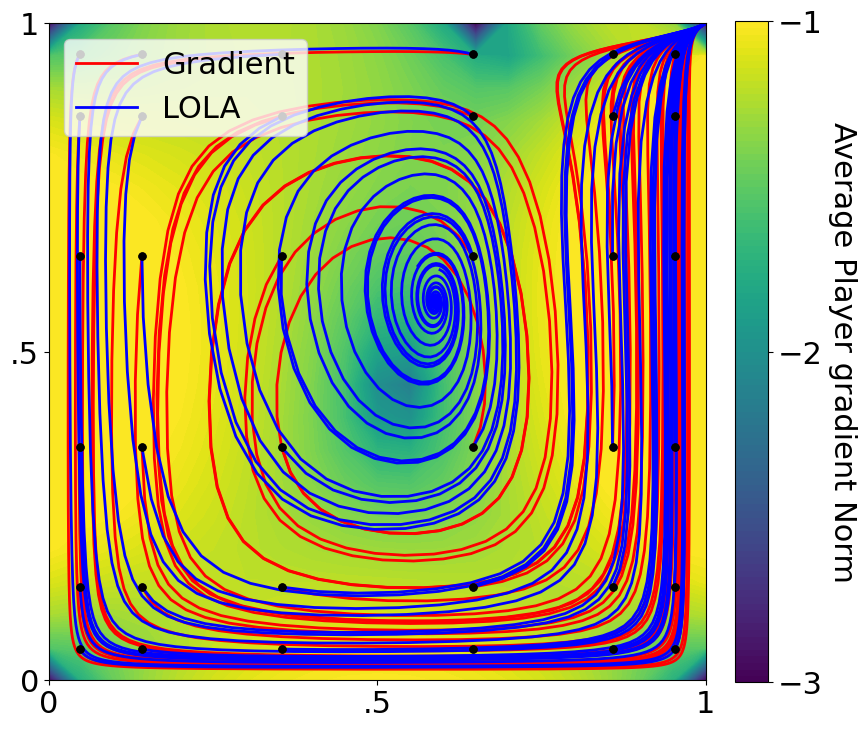}};
                \node[left=of img, node distance=0cm, rotate=90, xshift=1.5cm, yshift=-.9cm, font=\color{black}] {Player 1 Strategy};
                \node[below=of img, node distance=0cm, xshift=-.0cm, yshift=1.25cm,font=\color{black}] {Player 2 Strategy};
                \node[right=of img, node distance=0cm, rotate=270, xshift=-2.75cm, yshift=-.9cm, font=\color{black}] {Joint-player grad. log-norm $\log\left(\|\bothGrad\|\right)$};
            \end{tikzpicture}
            \caption[Phase portrait of common methods on the IPD]{
                The phase portrait for two standard optimization algorithms on the mixed small IPD and Matching Pennies problem.
                We show the trajectories following the gradient with \ref{lyapunov_eq:multi_objective_gd_long} in {\color{red}red} and LOLA~\citep{foerster2018learning}  -- a method for learning in games -- in {\color{blue}blue}.
                All \ref{lyapunov_eq:multi_objective_gd_long} initializations find only the solution on the top right because the center solution has imaginary eigenvalues, while LOLA finds all solutions.
                For comparisons on more test problems, see Appendix Figure~\ref{lyapunov_fig:fig_new_problems_baseline_full}.
            }\label{lyapunov_fig:fig_new_problems_baseline}
        \end{figure}

            Using tools from dynamical systems research, we look at how to find general bifurcation points. 
            Our objectives are inspired by the Lyapunov exponent, which measures asymptotic separation rates of optimization trajectories for small perturbations.
            We use a similar quantity but for finite-length trajectories.
            Given a $k$-step trajectory generated by our fixed-point operator $\lyapunovfixedPointOp$ -- i.e., optimizer -- with initialization $\bothParam_0$ and Jacobian at iteration $j$ of $\jacFixedPointOp^{j}$, we measure the separation rate for an initial normalized displacement $\displacement$ with:\vspace{-0.01\textheight}
            \begin{align}\label{lyapunov_eq:lyap_calc2}
                \lyap_k\left(\bothParam_0, \displacement\right) &= \frac{1}{k}\sum_{j = 1}^{k} \gamma_j\left(\bothParam_0, \displacement\right),\\
                \textnormal{where }\gamma_j\left(\bothParam_0, \displacement\right) &\defeq \log\left(\displacement^\transpose \left(\jacFixedPointOp^{j-1}\left(\bothParam_0\right)\right)^\transpose \jacFixedPointOp^{j-1}\left(\bothParam_0\right) \displacement\right)\label{lyapunov_eq:lyap_calc}
            \end{align}
            We call $\gamma_j$ the \emph{$j^{\textnormal{th}}$ Lyapunov term}.
            When $k=0$, the $\lyap$ are called the \emph{local Lyapunov exponents}, while as $k \to \infty$ they are called the \emph{(global) Lyapunov exponents}~\citep{wolff1992local}.
            We denote this as the \emph{$k$-step or truncated Lyapunov exponent}.
            Figure~\ref{lyapunov_fig:lyap_calc} visualizes the exponents' calculation, providing additional intuition.
            In the future, we suppress the dependency of $\jacFixedPointOp^{j-1}$ on $\bothParam_0$ to simplify notation.
            
            Within a basin of attraction to a given fixed-point, the global Lyapunov exponent is constant~\citep{katok1997introduction}.
            Intuitively, this is because an arbitrarily high number of Lyapunov terms near the fixed-point dominate the average defining the exponent in Equation~\ref{lyapunov_eq:lyap_calc2}.
            This property prevents us from optimizing the global exponent using gradients, making it a poor bifurcation objective.
            As such, our interest in the truncated exponent is motivated from multiple directions:
            \begin{enumerate}
                \item Non-zero gradient signals for finding bifurcations
                \item Computationally tractability
                \item A better separation rate description for the finite trajectories used in practice
            \end{enumerate}
            However, unlike the global exponent, the truncated version has less theoretical results.
            In more than one dimension, the $k$-step Lyapunov exponent depends on the perturbation direction $\displacement$~\citep{tabor1989chaos}.
            We look at using the direction for maximal separation --- i.e., the \emph{max $k$-step Lyapunov exponent}:
            \begin{equation}\label{lyapunov_eq:lyap_max}
                \lyap_k^{\textnormal{max}}(\bothParam_0) = \max_{\displacement, \|\displacement\| = 1} \lyap_k\left(\bothParam_0, \displacement\right)
            \end{equation}
            For dynamical systems with basins of attraction to fixed points, common in optimization, the max exponent is the largest on the boundary between basins, motivating maximizing the max exponent to find bifurcations.
            The max exponent can be evaluated by finding the largest eigenvalue of an average of Jacobians over the optimization steps~\citep{katok1997introduction}:
            \begin{align}\label{lyapunov_eq:jac_sum}
                 \jacFixedPointOp^{\dagger} = \frac{1}{k} \sum_{j = 1}^{k} \left(\jacFixedPointOp^{j-1}\right)^{\transpose}\left(\jacFixedPointOp^{j-1}\right),\\
                 \lyap_k^{\textnormal{max}}\left(\bothParam_0\right) = \max_{\lambda \in \spectrum\left(\jacFixedPointOp^{\dagger}\right)} \left|\lambda\right|
            \end{align}
            Importantly, in higher dimensions, an eigenvalue dominates the spectrum of $\jacFixedPointOp$ after many steps \citep{loreto1996characterization,kachman2017numerical}.
            
            We note some practical points for computing these exponents:
            When $k=1$, the maximum exponent is the maximum eigenvalue of $\jacFixedPointOp^0$.
            As $k \to \infty$ and our fixed-point operator converges to a fixed-point $\bothParam^*$, the maximum exponent is the maximum eigenvalue of $\jacFixedPointOp$ at $\bothParam^*$.
            Calculating $\lyap_k^{\textnormal{max}}$ is easiest when $k=1$ or $k \to \infty,$ e.g., by power iteration on the relevant $\jacFixedPointOp$.
            For intermediary $k$, directly using leading eigenvalues of $\jacFixedPointOp^{\dagger}$ can often involve re-evaluating the entire optimization trajectory.
            Instead, it can be easier to work with bounds.
            A simple lower bound is formed by using the leading eigenvector in any single step, or an upper bound by using the leading eigenvector in each step, which are tight as $k \to \infty$~\citep{loreto1996characterization}.
            We investigate these strategies in Appendix Figure~\ref{lyapunov_fig:lyap_dir_mix}.
            
            We aim to obtain many qualitatively different solutions from a single starting point, which motivates the simultaneous optimization of the exponents corresponding to multiple different directions.
            Relatedly, the sum of positive global Lyapunov exponents estimates the Kolmogorov–Sinai or metric entropy by Pesin's theorem~\citep{pesin1977characteristic}.
            We use this to motivate different performance metrics in Section~\ref{lyapunov_sec:grr}.
            
    \section{Proposed Algorithms}\label{lyapunov_sec:algs}
        Having given an overview of the key mathematical concepts, we now present our overall algorithm.
        First, we introduce a general branching-tree search framework to find diverse solutions in differentiable games.
        Next, we present our method -- Generalized Ridge Rider (GRR) -- which implements this framework using truncated Lyapunov exponents (Equation~\ref{lyapunov_eq:lyap_max}) as the branching criterion.
        Lastly, we highlight the differences between GRR and RR.
        
        \subsection{Branching Optimization Tree Searches}
\begin{figure}[t!]
                \centering
                \begin{tikzpicture}
                    \centering
                    \node (img){\includegraphics[trim={.5cm 2.75cm 4.5cm 1.9cm},clip, width=.99\linewidth]{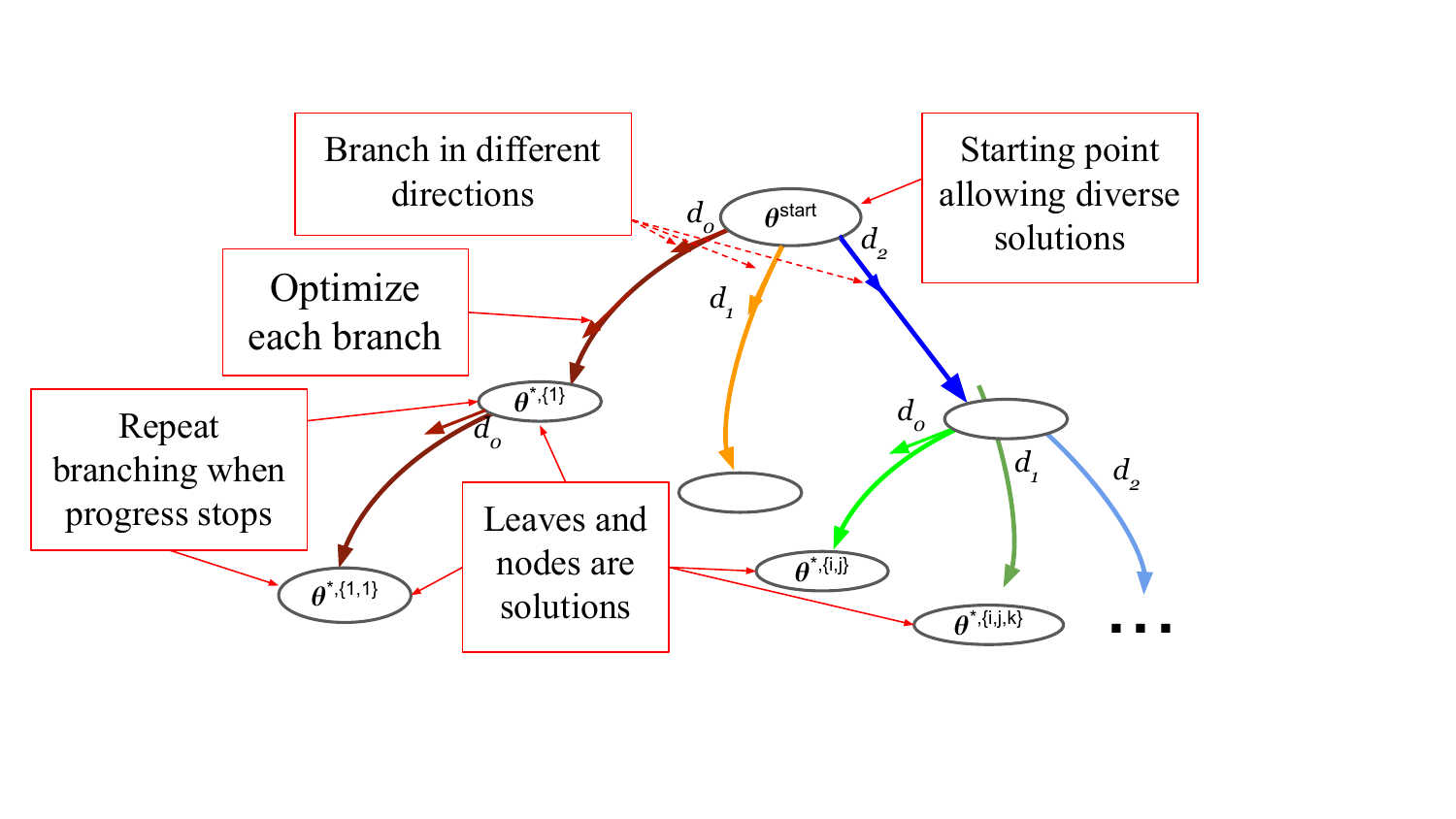}};
                \end{tikzpicture}
                \caption[Visualization of branching optimization tree search]{
                    A visualization of branching optimization tree search with components: (1) selecting the starting point, (2) creating branches, (3) optimizing each branch, and (4) choosing when to re-branch.
                }\label{lyapunov_fig:branching}
            \end{figure}

\begin{figure}[t!]
                \centering
                \begin{tikzpicture}
                    \centering
                    \node (img){\includegraphics[trim={.25cm .25cm 3.5cm .25cm},clip, width=.45\linewidth]{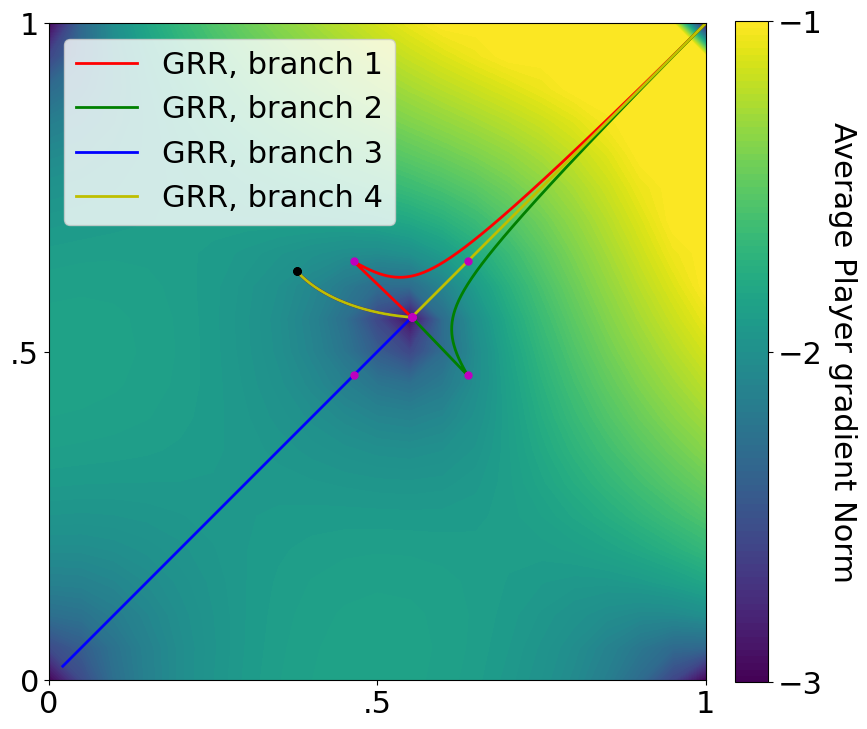}};
                    \node[left=of img, node distance=0cm, rotate=90, xshift=1.5cm, yshift=-.9cm, font=\color{black}] {Player 1 Strategy};
                    \node[right=of img2, node distance=0cm, rotate=270, xshift=-2.75cm, yshift=-.4cm, font=\color{black}] {\footnotesize{Joint-player gradient log-norm $\log(\|\bothGrad\|)$}};
                    \node[above=of img, node distance=0cm, yshift=-1.1cm,font=\color{black}] {\footnotesize{GRR -- Our Method -- at a Saddle Bifurcation}};
                    
                    \node [right=of img, yshift=0.0cm, xshift=-1.2cm] (img2){\includegraphics[trim={1.0cm .25cm 1.05cm .25cm},clip, width=.49\linewidth]{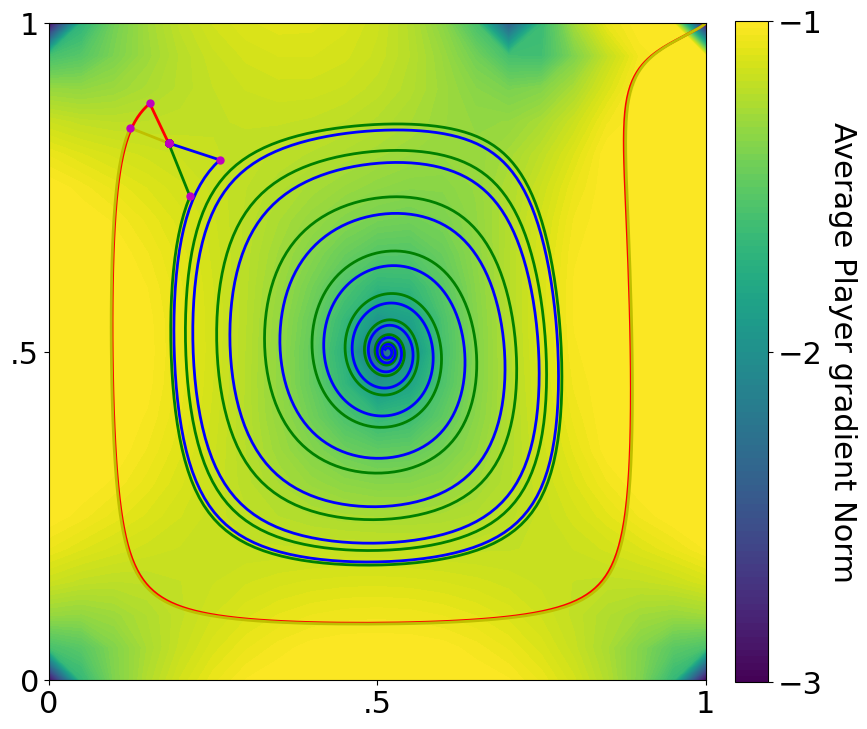}};
                    \node[below=of img1, node distance=0cm, xshift=.1cm, yshift=3.3cm,font=\color{black}] {Player 2 Strategy};
                    \node[below=of img2, node distance=0cm, xshift=-.3cm, yshift=1.2cm,font=\color{black}] {Player 2 Strategy};
                    \node[above=of img2, node distance=0cm, xshift=-.1cm,yshift=-1.1cm,font=\color{black}] {\footnotesize{GRR at a Hopf Bifurcation}};
                \end{tikzpicture}
                \caption[Branching at different bifurcation types]{
                    We show branching at different bifurcations obtained by optimizing a Lyapunov exponent as shown in Figure~\ref{lyapunov_fig:fig_mix}.
                    In each setup, we have two eigenvectors that branch in opposite directions, giving four paths displayed in different colors.
                    The steps with the eigenvector have {\color{magenta}magenta} circles marking boundaries.
                    \emph{Left}: In the small IPD, finding, then branching at a saddle -- where the joint-player gradient log-norm $\log\left(\|\bothGrad\|\right)$ is $0$ -- allows us to find defect-defect and tit-for-tat solutions.
                    \emph{Right:} In the Mixed Problem of Small IPD and Matching Pennies, branching at the Hopf bifurcation allows us to find both solutions.
                    There are no saddle points near the bifurcation, so RR's starting point does not allow branching to find both solutions.
                }\label{lyapunov_fig:branch_toy}
            \end{figure}
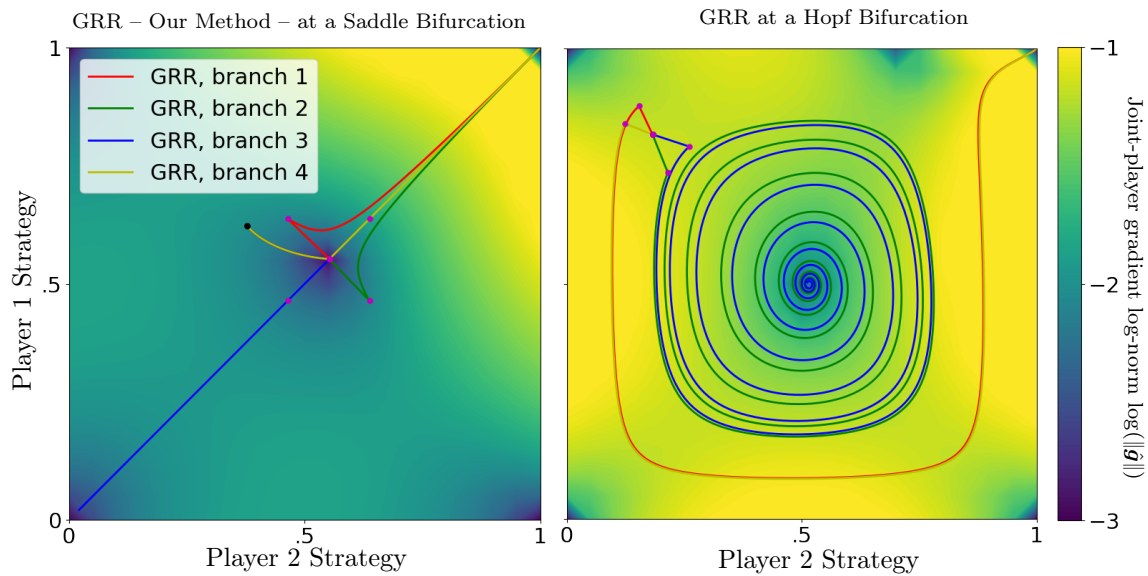

            Our framework is a generalized version of RR and contains the following components:
            \begin{enumerate}
                \item A method for finding a suitable \emph{starting point} for our branching process - see Figure~\ref{lyapunov_fig:fig_mix}.
                \item A process for selecting \emph{branching directions} (or perturbations) from a given branching point - see Figure~\ref{lyapunov_fig:branch_toy}.
                \item A prescription for how to \emph{continue the optimization process} along a given branch after the initial perturbation.
                \item A re-branching decision rule, i.e., when to return to step 2.
                This was important in RR because optimizers in high-dimensional non-convex machine learning problems often finish at saddle points~\citep{yao2020pyhessian}.
                \item Lastly, a metric to rank the different solutions.
            \end{enumerate}
            We visualize this process in Figure~\ref{lyapunov_fig:branching}.
            RR is an instance of this general process where each component is suitable for single-objective optimization.
            In the next section, we present another instance of this method, designed for game optimization.
            We include a more detailed description of branching tree searches in Appendix Algorithm~\ref{lyapunov_algorithm:branching_tree}, highlighting the important changes compared to RR.
            
        \hphantom{a}\newline
        \hphantom{a}\newline
        \hphantom{a}\newline
        \hphantom{a}\newline
        \hphantom{a}\newline
        \subsection{Generalized Ridge Rider (GRR)}\label{lyapunov_sec:grr}
            \textbf{Starting point:}\label{lyapunov_sec:method_start_obj}
                Motivated by Section~\ref{lyapunov_sec:lyap_for_bifurcation}, we look at optimizing the maximal $k$-step Lyapunov exponent from Equation~\ref{lyapunov_eq:lyap_max} to obtain our starting point:\vspace{-0.01\textheight}
                \begin{equation}\label{lyapunov_eq:1d_obj}
                    \loss\left(\bothParam_0\right) = -\lyap_k^{\textnormal{max}}\left(\bothParam_0\right) = -\max_{\displacement, \|\displacement\| = 1} \lyap_k\left(\bothParam_0, \displacement\right)
                \end{equation}
                However, using a single exponent only guarantees trajectory separation in a single direction.
                If we want to branch across multiple bifurcations in different directions, we need an objective using exponents in multiple directions.
                We look at the simple objective choice by adding exponents:\vspace{-0.01\textheight}
                \begin{align}
                    &\loss_{n}^{\textnormal{sum}}(\bothParam_0) = -\max_{\displacement_1, \dots, \displacement_n} \sum_{l=1}^{n} \lyap_k(\bothParam_0, \displacement_l),\\
                    \textnormal{such that } \|\displacement_l\| &= 1, \displacement_l^\transpose \displacement_m = 0 \textnormal{ for all $l,m \in {1, \dots, n}$, $l \neq m$}
                    \label{lyapunov_eq:constraint}
                \end{align}
                Intuitively, the constraint guarantees that we have different directions to separate in by making them orthogonal.
                It is straightforward to evaluate this objective by evaluating the top-$n$ eigenvalues of the matrix from Equation~\ref{lyapunov_eq:jac_sum}.
                More generally, various functions of the $k$-step exponents in different directions form reasonable objectives more amenable to optimization.
                Specifically, we also look at:\vspace{-0.01\textheight}
                \begin{align}
                    &\loss_{n}^{\textnormal{min}}\left(\bothParam_0\right) = -\max_{\displacement_1, \dots, \displacement_n} \min_{l=1\dots n} \lyap_k\left(\bothParam_0, \displacement_l\right)\\
                    \textnormal{such that } \|\displacement_l\| &= 1, \displacement_l^\transpose \displacement_m = 0 \textnormal{ for all $l,m \in {1, \dots, n}$, $l \neq m$}
                \end{align}

            \textbf{Branching the parameter optimization:}
                We must choose which direction to branch in; our procedure for evaluating Lyapunov exponent objectives creates natural candidates.
                Specifically, evaluating the maximum exponent involves finding the direction maximizing trajectory separation, which we re-use for branching.
                Notably, this is the most negative eigenvalue of the Hessian if we start at a saddle point and use SGD when calculating the trajectories, generalizing the choice from RR.
                We can move in a positive and negative direction for each eigenvector, giving two branches.
                
                Also, we must choose how far to move in each direction.
                If we move too far, we may leap into entirely different parts of the parameter space -- e.g., missing interesting regions and recovering similar solution modes.
                If we are not exactly at a bifurcation -- only near it -- then we may need to move some minimum distance to cross the separatrix and find a new solution.
                We look at two simple strategies to move sufficiently far.
                First, we try to take a single step with the normalized exponent direction.
                Second, we look at taking small steps in the exponent direction until the alignment with the joint-gradient flips, generalizing RR's ``riding a ridge'' (following an eigenvector of the Hessian) while it is a descent direction.
            
            \textbf{Optimizing each branch:}
                For game optimization, the stability properties of the solutions can crucially depend on the optimizer choice~\citep{gidel2018negative}.
                One should choose an optimizer suited to the problem.
                In our experiments, we use Learning with Opponent Learning Awareness (LOLA)~\citep{foerster2018learning}, which can converge to periodic solutions and is attracted to high-welfare solutions in the IPD.
                In Section~\ref{lyapunov_sec:toy_opt}, we compare the finding of diverse solutions using LOLA with simultaneous SGD (\ref{lyapunov_eq:multi_objective_gd_long}) -- a method that works well for single-objective optimization but cannot find periodic solutions.
                Appendix~\ref{lyapunov_sec:related-work} summarizes other optimizer choices.
                \newline

            \textbf{Rebranching:}
                In single-objective optimization in machine learning, our optimizer often finishes at a high-dimensional saddle~\citep{yao2020pyhessian}, making rebranching important.
                RR rebranches at the saddle in negative eigenvector directions to find critical points until we have fewer negative eigenvectors.
                In our setup, we are interested in rebranching if our optimizer finishes at a point where the eigenvalues of the Jacobian of the fixed-point operator $\jacFixedPointOp$ are greater than $1$.
                Figure~\ref{cm_fig:gan_spectrum_main} shows eigenvalues of $\jacFixedPointOp$ larger than $1$ at the end of GAN training.
        %
        
        \subsection{Comparing GRR and RR}
            RR is a branching optimization search specifically for single-objective optimization -- which is less general than optimization in games -- so it can simplify GRR.
            For example, nonconservative systems have more bifurcation types than conservative ones.
            If we are only concerned with saddle bifurcations, we can just find a saddle stationary point by minimizing the gradient norm.
            We know this (relatively) easy-to-find stationary point lies on the separatrix.
            However, Hopf bifurcations are not necessarily near stationary points.
            Thus, minimizing gradient norms generally does not work, while optimizing a Lyapunov exponent does (Figure~\ref{lyapunov_fig:fig_new_problems_baseline}).
            
            Although finding a separatrix with Lyapunov exponents in a single-objective setting might be overkill, we take some lessons from GRR back to RR.
            It is useful to view RR as a method to find bifurcations and branch across them.
            This motivates ways to sort between different stationary points to start at -- an open problem from RR.
            For example, we use the point with the highest Kolmogorov-Sinai entropy~\citep{pesin1977characteristic}.
            This is roughly the (negative) sum of negative eigenvalues at stationary points.
            Another limitation of RR is effectively estimating the most negative eigenvalues of the Hessian.
            It is often simpler -- in computation and implementation -- to estimate the leading eigenvalues of the Jacobian of the fixed-point operator $\jacFixedPointOp$ instead of the most negative eigenvalues of the Hessian $\bothHess$.
            Section~\ref{lyapunov_subsec:exp_coop} shows that our method reduces Hessian-vector product evaluations when estimating eigenvectors in setups from RR.
            
        \begin{figure*}[h!]
            \centering
            \begin{tikzpicture}
                \centering
                \node (img){\includegraphics[trim={1.0cm 1.0cm 55.0cm 1.15cm},clip, width=.3\linewidth]{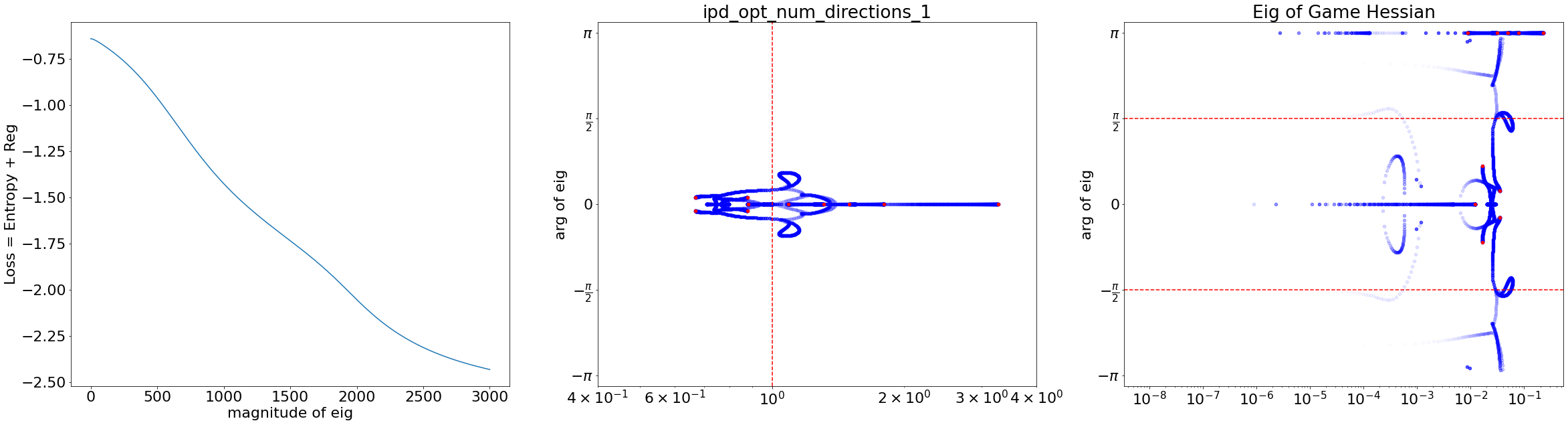}};
                \node[left=of img, node distance=0cm, rotate=90, xshift=2.0cm, yshift=-.8cm, font=\color{black}] {loss $\loss\left(\bothParam_0\right) = -\lyap_k^{\textnormal{max}}\left(\bothParam_0\right)$};
                \node[below=of img, node distance=0cm, xshift=-.0cm, yshift=1.0cm,font=\color{black}] {optimization iteration};
                
                \node [right=of img, xshift=-.6cm](img2){\includegraphics[trim={30.0cm 1.0cm 27.5cm 1.15cm},clip, width=.31\linewidth]{latex/lyapunovForDiversity/images/experiments/ipd/tune_lyap_ipd_main.png}};
                \node[left=of img2, node distance=0cm, rotate=90, xshift=2.0cm, yshift=-.85cm, font=\color{black}] {argument of EVal $\arg\left(\eigval\right)$};
                \node[below=of img2, node distance=0cm, xshift=2.5cm, yshift=1.0cm,font=\color{black}] {log-norm of EVal $\log\left(|\eigval|\right)$};
                \node[above=of img2, node distance=0cm, xshift=-.0cm, yshift=-1.15cm,font=\color{black}] {\tiny{EVals of Jac. of fixed-point operator $\spectrum\left(\jacFixedPointOp\right)$}};
                
                \node [right=of img2, xshift=-1.1cm](img3){\includegraphics[trim={59.1cm 1.0cm 0.5cm 1.15cm},clip, width=.28\linewidth]{latex/lyapunovForDiversity/images/experiments/ipd/tune_lyap_ipd_main.png}};
                \node[above=of img3, node distance=0cm, xshift=-.0cm, yshift=-1.15cm,font=\color{black}] {\footnotesize{EVals of game Hessian $\spectrum\left(\bothHess\right)$}};
            \end{tikzpicture}
            \caption[Gradient descent on the 1-step maximum Lyapunov exponent]{
                We display gradient descent optimization on the $1$-step maximum Lyapunov exponent objective (Equation~\ref{lyapunov_eq:1d_obj}) on the IPD.
                \textbf{\emph{Takeaway:}} We effectively reduce our loss and correspondingly increase the maximum eigenvalue of $\jacFixedPointOp$.
                \emph{Left:} We display our loss -- i.e., the negative Lyapunov exponent objective-- as optimization progresses.
                \emph{Middle:} We visualize the spectrum of the Jacobian of our fixed-point operator in log-polar coordinates as optimization progresses.
                The spectrum is shown with a scatterplot in {\color{blue}blue}, with a progressively larger alpha at each iteration.
                The final spectrum is shown in {\color{red}red}.
                A vertical {\color{red}red} line is shown where the eigenvalue norm is equal to $1$, signifying the cutoff between the (locally) convergent and divergent eigenspaces.
                We effectively maximize the norm of the largest eigenvalue.
                \emph{Right:} We display the spectrum of the game Hessian.
                A horizontal {\color{red}red} line is shown where the real part of the eigenvalue transitions from negative to positive, signifying the cutoff between (locally) convergent and divergent eigenspaces under gradient flow.
                Log-polar coordinates are required to see the structure in the spectrum.
            }\label{lyapunov_fig:tune_lyap_ipd}
        \end{figure*}
    \section{Experimental Setting}\label{lyapunov_subsec:new_problems}
        We experimentally investigate GRR on various problems summarized in this section and described in detail in Appendix~\ref{lyapunov_sec:app_test_problems}, chosen because they co various dynamics and bifurcation types.
        Some are standard benchmarks, while others -- i.e., Random Subspaces -- are novel.
        We also summarize our gradient computation for these problems.
        
        The \textbf{Iterated Prisoners' Dilemma (IPD)} is the discounted, infinitely iterated Prisoner's Dilemma~\citep{poundstone1993prisoner}.  
        Each agent's policy conditions on the actions in the prior time step, so there are $5$ parameters for each agent -- the probability of cooperating initially and given both agents' preceding actions.
        There are multiple relevant equilibria in the IPD, including \emph{unconditional} defection (DD), leading to the worst-case joint outcome, and \emph{tit-for-tat} (TT), where agents initially cooperate, then copy the opponents' action, giving a higher reward.
        We turn the IPD into a differentiable game by calculating the analytical expected return as a function of the two agents' joint policy.
        
        The \textbf{Small IPD} is a two-parameter simplification of IPD, which has both the DD and TT equilibria, allowing us to visualize the optimization difficulties of the full-scale IPD.
        However, unlike the full-scale IPD, the game Hessian has strictly real eigenvalues and, thus, conservative dynamics.

        \textbf{Matching Pennies} is a 2-parameter version of rock-paper-scissors.
        This problem's game Hessian has purely imaginary eigenvalues, unlike the small IPD, but only a single solution.
        Thus, it is a poor diagnostic for diverse solutions but a useful test for probing nonconservative behavior.
        
        \textbf{Mixing Small IPD and Matching Pennies} interpolates between the Small IPD and Matching Pennies games with an interpolation factor $\tau \in [0, 1]$ as in Equation~\ref{eq_app:mixed_obj}.
            This problem has two solutions -- one where both players cooperate and one where both players select actions uniformly, with a Hopf bifurcation separating these.
        
        We use a \textbf{Generative Adversarial Network (GAN)} setup from \citet{metz2016unrolled, balduzzi2018mechanics, letcher2018stable}, where the task is to learn a Gaussian mixture distribution using GANs.
            The data is sampled from a multimodal distribution to investigate the tendency to collapse on a subset of modes during training.
        
        We construct \textbf{Random Subspace IPD/GANs} for more complicated diagnostics by optimizing higher-dimensional problems in a random subspace to see how robustly we can find bifurcations with the exponents.
            For each player, we select a random direction to optimize by sampling a vector $\mathbf{v}$ with entries from $U[0, 1]$ and normalizing it.
            We also select a random offset $\mathbf{b}$ from an appropriate initialization for the higher-dimensional problem.
            So, the first player controls the $x$ coordinate and optimizes the loss $\outLoss\left(\mathbf{v}_{\outSymbol}x + \mathbf{b}_{\outSymbol}, \mathbf{v}_{\inSymbol}y + \mathbf{b}_{\inSymbol}\right)$, while the second player controls the $y$ coordinate and optimizes $\inLoss\left(\mathbf{v}_{\outSymbol}x + \mathbf{b}_{\outSymbol}, \mathbf{v}_{\inSymbol}y + \mathbf{b}_{\inSymbol}\right)$.
            
        We apply our method in \textbf{single-objective problems} to find bifurcations.
            There are various relevant problems in machine learning, but we focus on comparisons with RRs eigenvector estimation in MNIST classification.
            \newline
            
        To \textbf{optimizing the starting point objective}, we use automatic differentiation libraries (such as Jax~\citep{jax2018github} or PyTorch~\citep{paszke2017automatic}) to compute gradients through methods that calculate our Lyapunov exponent-inspired objectives.
            The scalability of this approach depends on the implementation of our exponent calculation, which can depend on estimating the top eigenvalues of the positive semi-definite (PSD) symmetric matrix in Equation~\ref{lyapunov_eq:jac_sum}.
            In simple settings, we can differentiate by calculating the full spectrum via \texttt{jax.linalg.numpy.eigh}; we investigate this in the toy experiment in Section~\ref{lyapunov_sec:toy_find_start} and IPD in Section~\ref{lyapunov_sec:opt_lyap_ipd}.
            However, in machine learning, the matrix of Equation~\ref{lyapunov_eq:jac_sum} is typically too large for the full spectrum.
            Directly estimating the top eigenvalues with an iterative method allows us to differentiate through them.
            We differentiate through \texttt{jax.numpy.linalg.eigh} in Figure~\ref{lyapunov_fig:fig_mix} and investigate using power iteration with Hessian-vector products in Appendix Figure~\ref{lyapunov_fig:lyap_dir_mix}.

    \vspace{-0.01\textheight}
    \section{Experimental Results}
        First, in Section~\ref{lyapunov_sec:diagnostic}, we use diagnostic problems to demonstrate and ablate the key parts of our algorithms -- i.e., optimizers, starting points, and branching.
        Next, in Section~\ref{lyapunov_sec:scaling}, we scale GRR to larger-scale problem settings by (a) demonstrating that we improve RR's eigenvector estimation for neural network classifiers and (b) calculating Lyapunov exponents for GANs.

        \vspace{-0.01\textheight}
        \subsection{Diagnostic Experiments}\label{lyapunov_sec:diagnostic}
        \subsubsection{Optimizer Choice}\label{lyapunov_sec:toy_opt}
            Here, we show a system with complex eigenvectors showing (a) the importance of selecting a convergent optimizer in GRR and (b) an example task where RR cannot be applied.
            Figure~\ref{lyapunov_fig:fig_new_problems_baseline} shows the phase portrait for the baseline methods in our Mixed Problem.
            LOLA (and other game optimizers) can find both solutions while na\"ively following the gradient always finds a single solution.
        
        \subsubsection{Starting Point Selection}\label{lyapunov_sec:toy_find_start}
\begin{figure}
                \vspace{-.02\textheight}
                \centering
                \begin{tikzpicture}
                    \centering
                    \node (img){\includegraphics[trim={.25cm .25cm 5.5cm 1.25cm},clip, width=.41\linewidth]{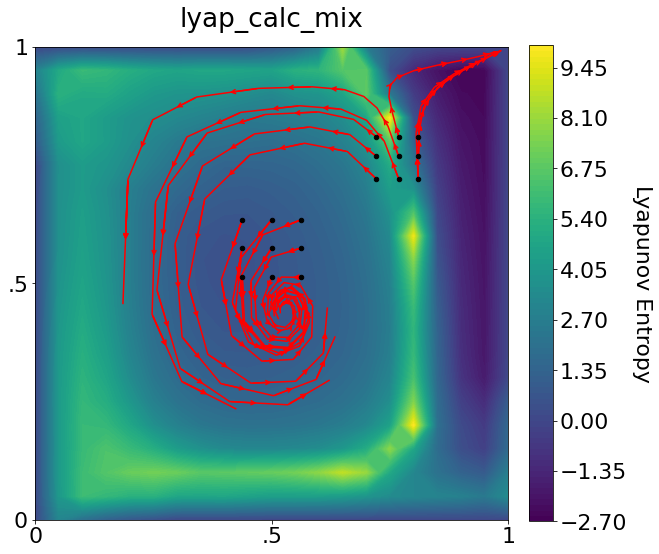}};
                    \node[left=of img, node distance=0cm, rotate=90, xshift=1.5cm, yshift=-.9cm, font=\color{black}] {Player 1 Strategy};
                    \node[above=of img, node distance=0cm, xshift=-.0cm, yshift=-1.3cm,font=\color{black}] {Calculating the exponent};
                    \node[below=of img, node distance=0cm, xshift=.25cm, yshift=1.15cm,font=\color{black}] {Player 2 Strategy};
                    
                    \node [right=of img, xshift=-1.0cm] (img2){\includegraphics[trim={1.0cm .25cm 1.05cm 1.25cm},clip, width=.50\linewidth]{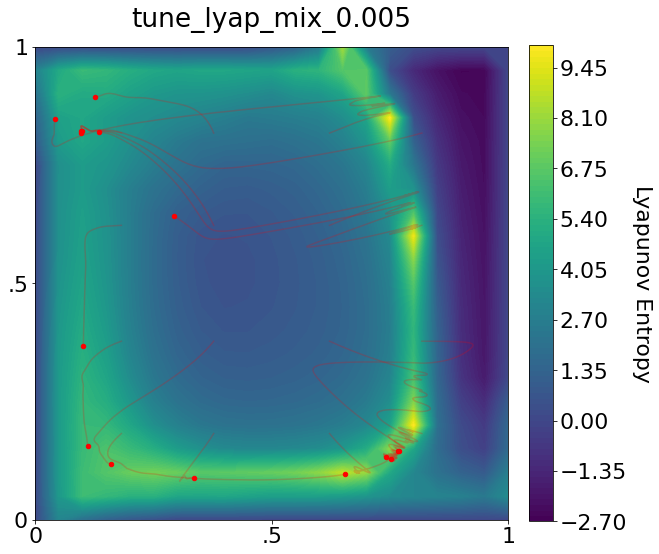}};
                    \node[below=of img2, node distance=0cm, xshift=-.6cm, yshift=1.15cm,font=\color{black}] {Player 2 Strategy};
                    \node[above=of img2, node distance=0cm, xshift=-.75cm, yshift=-1.3cm,font=\color{black}] {Optimizing the exponent};
                    \node[right=of img2, node distance=0cm, rotate=270, xshift=-2.75cm, yshift=-.9cm, font=\color{black}] {Max $10$-step Lyapunov Exponent};
                    
                \end{tikzpicture}
                \vspace{-.01\textheight}
                \caption[Optimization of a maximum 10-step Lyapunov exponent]{
                    We show the calculation and optimization of a maximum $10$-step Lyapunov exponent from Equation~\ref{lyapunov_eq:lyap_max} on the mixed small IPD and Matching Pennies problem.
                    Gradient-based optimization of this objective effectively finds the bifurcation. 
                    \emph{Left:} We show a heatmap of the exponent and visualize the calculation of each exponent in two regions.
                    This involves simulating the $10$-step trajectories shown in red, starting at the black points, and then finding a direction that maximizes trajectory separation.
                    We use the exponent to find bifurcations -- in this case, between the solution in the top right and the center.
                    \emph{Right:} We show optimization trajectories for gradient ascent on the exponent for a grid of initializations.
                    The optimization procedure finds large exponent locations for various starting points, with the final iterate shown by {\color{red}red} circles.
                }\label{lyapunov_fig:fig_mix}
                \vspace{-0.02\textheight}
            \end{figure}
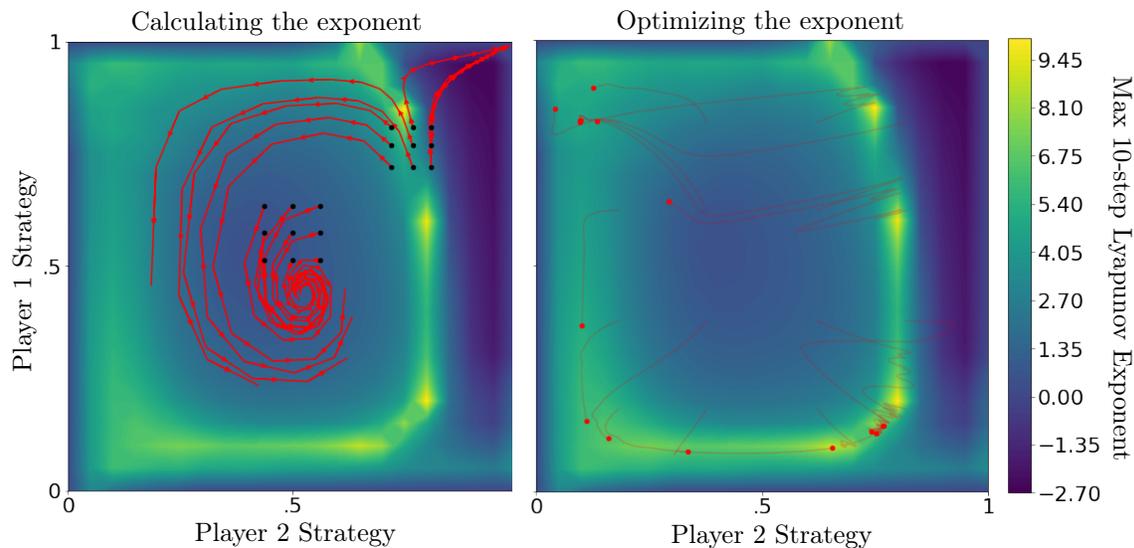

            Figure~\ref{lyapunov_fig:fig_mix} shows optimization of the maximum $10$-step Lyapunov exponent on the Mixed Problem, where we find bifurcations with gradient-based optimization.
            Figure~\ref{lyapunov_fig:lyap_dir_mix} contrasts different direction choices for the exponent calculation.
            Re-estimating the top eigenvectors at each iteration performs best, although simpler methods can also work. 
            Appendix Figure~\ref{lyapunov_fig:lyap_step_mix} shows the maximum $k$-step exponent for multiple numbers of steps $k$, showing that a moderate number of steps -- e.g., $10$ -- allows us to find bifurcations.
            Appendix Figure~\ref{lyapunov_fig:entropy_mix} shows different Lyapunov exponent objectives that seek trajectory divergence in multiple directions.
            We find bifurcations while guaranteeing trajectory separation in every direction.
            
            \textbf{Impact of inner optimizer choices on bifurcation structure:}
                Appendix Figure~\ref{lyapunov_fig:lyap_opt_mix} contrasts the exponents for LOLA and \ref{lyapunov_eq:multi_objective_gd_long}, showing we find optimizer-dependent bifurcations.
                Appendix Figure~\ref{lyapunov_fig:lyap_opt_param_mix} investigates the impact of optimization-algorithm parameter choices on the bifurcation structure.
                If the step size is too large, the optimizer does not converge, resulting in bifurcations between complicated limit cycle trajectories~\citep{strogatz2018nonlinear}, making GRR difficult to apply.
            
            \textbf{Starting points on single-objective problems:}
                Appendix Figure~\ref{lyapunov_fig:lyap_coop_toy} shows our method in single-objective problems and finds bifurcations in the same setup as RR.
                Appendix Figure~\ref{lyapunov_fig:lyap_oneDim_toy} shows the logistic map, giving intuition for our method on a canonical example for bifurcations.

        \vspace{-0.01\textheight}
        \subsubsection{Branching at Bifurcations}\label{lyapunov_sec:toy_branch}
        \vspace{-0.0075\textheight}
            In Figure~\ref{lyapunov_fig:branch_toy}, we demonstrate branching at bifurcations to find multiple solutions to toy problems.
            This shows an explicit example of where RR's starting point does not work, but GRR's does.
            The small IPD has a saddle bifurcation, while the Mixed Problem has a Hopf bifurcation.
        
        \vspace{-0.01\textheight}
        \subsubsection{A Range of Complicated Toy Problems}
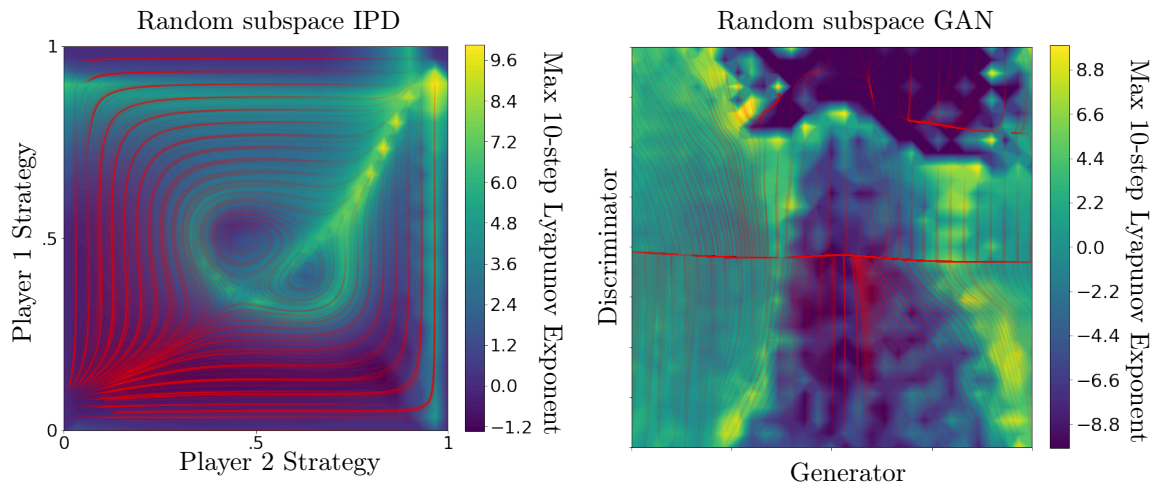
\begin{figure}
                \centering
                \begin{tikzpicture}
                    \centering
                    \node (img){\includegraphics[trim={.25cm .25cm 1.05cm 1.25cm},clip, width=.43\linewidth]{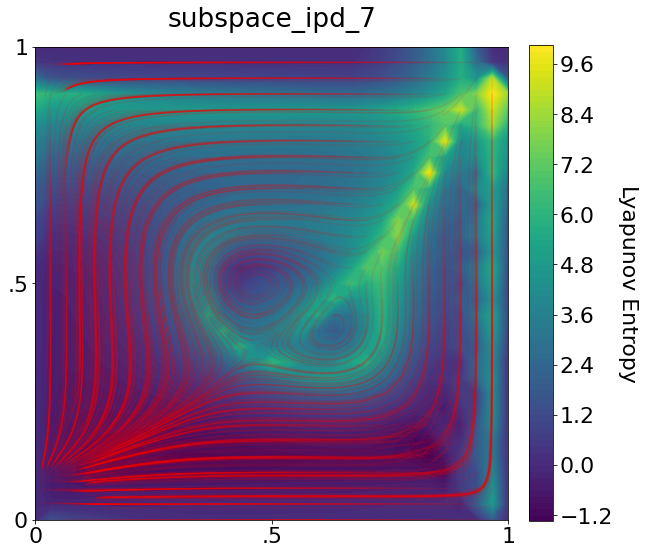}};
                    \node[left=of img, node distance=0cm, rotate=90, xshift=1.5cm, yshift=-.9cm, font=\color{black}] {Player 1 Strategy};
                    \node[right=of img, node distance=0cm, rotate=270, xshift=-2.7cm, yshift=-.9cm, font=\color{black}] {Max $10$-step Lyapunov Exponent};
                    \node[below=of img, node distance=0cm, xshift=-.1cm,yshift=1.25cm,font=\color{black}] {Player 2 Strategy};
                    \node[above=of img, node distance=0cm, xshift=-.25cm, yshift=-1.2cm,font=\color{black}] {Random subspace IPD};
                    
                    \node [right=of img, xshift=-.0cm, yshift=0.0cm] (img2){\includegraphics[trim={2.75cm 1.0cm 1.05cm 1.25cm},clip, width=.43\linewidth]{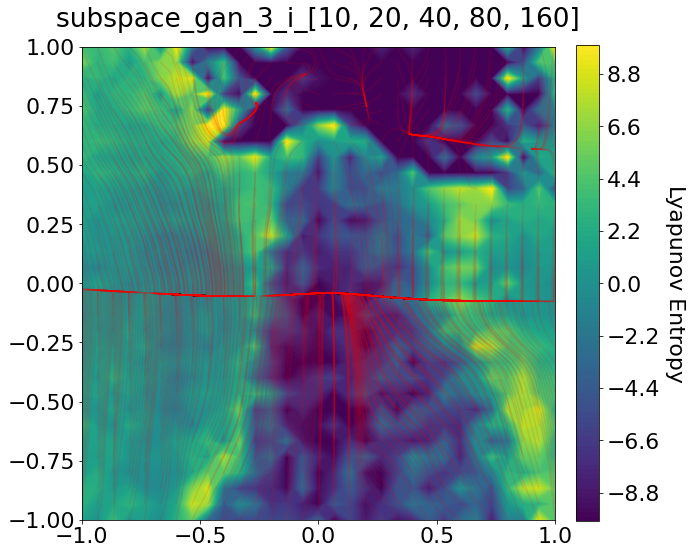}};
                    \node[left=of img2, node distance=0cm, rotate=90, xshift=1.15cm, yshift=-.85cm, font=\color{black}] {Discriminator};
                    \node[below=of img2, node distance=0cm, xshift=-.35cm, yshift=1.1cm,font=\color{black}] {Generator};
                    \node[right=of img2, node distance=0cm, rotate=270, xshift=-2.6cm, yshift=-.9cm, font=\color{black}] {Max $10$-step Lyapunov Exponent};
                    \node[above=of img2, node distance=0cm, xshift=-.25cm, yshift=-1.2cm,font=\color{black}] {Random subspace GAN};
                \end{tikzpicture}
                \caption[Lyapunov exponent heatmap on toy problems]{
                    We show the Lyapunov exponent heatmap (as in Figure~\ref{lyapunov_fig:fig_mix}) on more complicated toy problems to see how robustly we can find different bifurcations.
                    The exponent peaks near where the trajectories (shown in {\color{red}red}) separate, showing that we find various bifurcations.
                    See Appendix Figure~\ref{lyapunov_fig:complicated_toy_moreSeeds} for other sampled subspaces.
                    Section~\ref{lyapunov_subsec:new_problems} describes how we construct these examples by taking higher-dimensional problems and optimizing them in a random subspace.
                    \emph{Left:} An IPD subspace with multiple Hopf bifurcations.
                    \emph{Right:} A GAN subspace with various bifurcations.
                }\label{lyapunov_fig:complicated_toy}
            \end{figure}

        \vspace{-0.0075\textheight}
            In Figure~\ref{lyapunov_fig:complicated_toy}, we calculate Lyapunov exponents on problems with more complicated dynamics by taking the high-dimensional IPD and GAN problems and selecting a random subspace to optimize.
            We effectively highlight bifurcations in this setup.
            %
            
        \vspace{-0.01\textheight}
        \subsection{Scaling the Results}\label{lyapunov_sec:scaling}
        
        \vspace{-0.005\textheight}
        \subsubsection{Optimizing Lyapunov Exponents on IPD}\label{lyapunov_sec:opt_lyap_ipd}
        \vspace{-0.0075\textheight}
            Here, we investigate our ability to use gradient-based optimizers on Lyapunov exponents in the IPD.
            Figure~\ref{lyapunov_fig:tune_lyap_ipd} shows the feasibility of using gradient descent to tune the $1$-step maximum Lyapunov exponent.
            Appendix Figure~\ref{lyapunov_fig:tune_lyap_ipd_moreDirections} optimizes an objective using multiple exponents, showing that we effectively optimize multiple exponents, which gives trajectory separation in multiple directions.
            Appendix Figure~\ref{lyapunov_fig:tune_lyap_ipd_difObjectives} compares objectives using multiple exponents, showing that using the minimum of the top $n$ exponents gives trajectory separation in all $n$ directions, unlike the na\"ive choice of optimizing their sum.
            The sum of exponents finds solutions that separate extremely fast in the top directions while (slowly) converging in the bottom directions.
            In contrast, the minimum of the exponents does not allow for convergence in the bottom directions.
            Appendix Figure~\ref{lyapunov_fig:tune_lyap_ipd_difNumLyap} compares optimizing the $k$-step maximum Lyapunov exponent for variable $k$, showing that we effectively minimize multi-step exponents in higher-dimensional problems if required. 

        \subsubsection{GRR Applied to the IPD}\label{lyapunov_subsec:exp_ipd}
            Here, we use our method on the IPD, where existing methods have difficulty finding diverse solutions.
            There are two solution modes: ones where both agents defect and cooperate, respectively.
            Table~\ref{lyapunov_tab:results} compares our method with the baselines of the following gradients and LOLA, each run with random initializations.
            Our method finds both solution modes, unlike existing approaches.
            Using the maximum Lyapunov exponent as our objective was sufficient, which only guarantees separation in $1$ direction.
            Similarly, we found that using a $1$-step or local Lyapunov exponent objective was sufficient, though we may require more steps to find bifurcations in other problems.
            \begin{table*}[t!]
            	\centering
                    \footnotesize{
            	\begin{tabular}{lllllll}
            		                                & P1 loss $\outLoss$ & \multicolumn{5}{c}{(P)layer 1 Strategy Distribution: min, max} \\
            		                                        \cmidrule(lr){3-7}
            		Search Strategy                  & {\,\,\,min, max} & {$p\left(C_0\right)$}           & {$p\left(C | CC\right)$}      & {$p(C | CD)$}      & {$p(C | DC)$}      & {$p(C | DD)$} \\
            		\midrule
            		\!\!Us: Lyap init, EVec branch, SimSGD\!\! & $1.000, 2.000$ & $.003,.999$  & $.032,.999$  & $.004, .884$  & $.001, .912$  & $.000, .013$\\
            		\!\!Us: Lyap init, EVec branch, LOLA & $1.000, 2.000$ & $.002,.999$  & $.063,.993$  & $.001, .910$  & $.000, .922$  & $.005, .103$\\
            		\!\!$20$ Random init, SimSGD           & $1.997, 1.998$ & $.043, .194$  & $.142, .480$  & $.041, .143$  & $.055, .134$  & $.001, .001$\\
            		\!\!$20$ Random init,  LOLA           & $1.000,1.396$ & $.000, 1.00$  & $.093, 1.00$  & $.000, .966$  & $.057, 1.00$  & $.000, .947$\\
            		\!\!$1$ Random init, EVec branch, SimSGD\!\!\!      & $2.000, 2.000$ & $.001, .003$  & $.027, .030$  & $.003, .007$  & $.008, .009$  & $.000, .000$\\
            	\end{tabular}
             }
            	\caption[Comparing strategies for finding diverse solutions to the iterated prisoner's dilemma]{
            	    We show strategies for finding diverse solutions to the iterated prisoner's dilemma (IPD).
            	    \textbf{\emph{Takeaway:}} Our method finds solutions in both loss modes, while existing approaches of using random initializations, then following the gradient or using LOLA do not find diverse solutions.
            	    The IPD has two solution modes -- i.e., solutions where both agents defect with a loss of $2$ and where both agents cooperate with a loss of $1$ (like tit-for-tat).
            	    We assess which modes were found by showing (P)layer 1's strategy, which is the chance of cooperating given both players' last action -- ex., $p(C|DC)$ is the chance if previously P1 ($D$)efected and P2 ($C$)ooperated.
            	    We compare GRR flavors with just following gradients via \ref{lyapunov_eq:multi_objective_gd_long} and LOLA~\citep{foerster2018learning} from random (init)ializations.
            	    We compare with $20$ random initializations because GRR follows at most $20$ branches due to having $10$ eigenvectors in $2$ directions (+/-).
            	    GRR only branches in directions where eigenvalues of the Jacobian of the fixed-point operator are greater than $1$ (i.e., trajectories locally diverge) as visualized in Figure~\ref{lyapunov_fig:tune_lyap_ipd} (middle).
            	    We look at the impact of starting at an approximate bifurcation in GRR by branching on the eigenvectors at a random init.
            	    Each branch finds the same solution if the maximum Lyapunov exponent is not tuned.
            	}
            	\label{lyapunov_tab:results}
            \end{table*}

        \subsubsection{Improving RR's Eigenvector Estimation}\label{lyapunov_subsec:exp_coop}
            We investigate efficiently finding the most negative eigenvectors in RR by estimating the largest eigenvalues/vectors of the Jacobian of our fixed-point operator, where RR measures the most negative eigenvalues/vectors of the Hessian.
            We measure efficiency by comparing the number of Hessian-vector product (HVP) evaluations, which dominate the cost of eigenvector estimation.
            Table~\ref{lyapunov_tab:hvp_evals} shows how many HVP evaluations to reach different MNIST classifier accuracies by following eigenvectors.
            Our method can more efficiently use HVP evaluations than the RR method because we simply do power iteration on the fixed-point operator Jacobian.
            
            We stress that this problem is not designed to train a single strong classifier; it is easy to just train our network by following the gradient to $100\%$ train accuracy.
            This problem was selected from RR's experiments because it requires us to estimate negative eigenvectors accurately and efficiently many times.
            A downstream use of this is training an ensemble of classifiers for generalization. 
            
            \begin{table}
            	\centering
            	\begin{tabular}{ccc}
            		    &\multicolumn{2}{c}{MNIST Accuracy} \\
            		\cmidrule(lr){2-3}
            		Number of HVP Evaluations & Our Method & RR's Method\\
            		\midrule
            		\num{10000} & $\num{19}\%({\color{green}+8\%})$ & $\num{11}\%$ \\
            		\num{100000} & $\num{89}\%({\color{green}+6\%})$ & $\num{83}\%$ \\
            		\num{1000000} & $\num{93}\%({\color{green}+2\%})$ & $\num{91}\%$ \\
            	\end{tabular}
            	\caption[Comparing HVP evaluations required to reach various accuracies]{
            	    We show how many HVP evaluations to reach different MNIST classifier accuracies by following eigenvectors, repeating the experiment in RR's Figure~4.
            	    We aim not to train a single strong classifier but to test our ability to follow negative eigenvectors efficiently -- see Section~\ref{lyapunov_subsec:exp_coop}.
            	 }
            	\label{lyapunov_tab:hvp_evals}
            \end{table}
        
        \subsubsection{Calculating Lyapunov Exponents for GANs}\label{lyapunov_subsec:exp_gan}
            Here, we scale our exponent calculations to machine learning models where the (game) Hessian is so large that we cannot materialize it and only use Hessian-vector products.
            Specifically, we use the GAN described in Section~\ref{lyapunov_subsec:new_problems}.
            We look at calculating our exponent for various hyperparameters and random re-starts.
            We evaluate the quality of using our exponent to find diverse solutions with the log probability of samples from an ensemble of GANs from the top $5$ optimization branches.
            Table~\ref{lyapunov_tab:gan_lyap_calc} shows the mean and standard deviation (over $10$ random restarts) of the maximum $10$-step Lyapunov exponent and the resulting ensemble's log probability.
            Each GAN was trained for $\num{10000}$ updates, so evaluating ensembles costs approximately $\num{50000}$ evaluations of both players' gradients.
            In contrast, each exponent costs less than $\num{1000}$ evaluations of both gradients to compute.
            
            This shows that we effectively scale our exponent calculation to larger models of interest from machine learning. We find that a large (mean) exponent may align with regions where we can branch to train the strongest ensemble of GANs.
            
            \begin{table}
            	\centering
            	\begin{tabular}{ccc}
            		Initial scale, step size & Max Lyapunov Coefficient & Ensemble log prob\\
            		\midrule
            		\num{0.001}, 1.0 & $\num{0.952} \pm  \num{0.834}$& $\num{-16342} \pm \num{817}$ \\
            		\num{0.1}, 1.0 & $\num{6.485} \pm  \num{1.155}$& $\num{-13691} \pm \num{1317}$ \\
            		\num{10.0}, 1.0 & $\num{0.053} \pm  \num{0.128}$ & $\num{-46659} \pm \num{26793}$ \\
            		\num{0.001}, 0.1 & $\num{0.849} \pm  \num{0.765}$ & $\num{-12321} \pm \num{126}$ \\
            		\num{0.1}, 0.1 & $\num{6.571} \pm  \num{0.953}$ & $\num{-10846} \pm \num{256}$\\
            		\num{10.0}, 0.1 & $\num{-0.012} \pm  \num{0.014}$ & $\num{-23459} \pm \num{12693}$ \\
            	\end{tabular}
            	\caption[Quantitative results for Lyapunov exponent calculation on GANs]{
            	    We display the mean and standard deviation (over $10$ random restarts) of the maximum $10$-step Lyapunov exponent and the log probability of an ensemble of $5$ GANs obtained by branching in the top $5$ directions at the initialization.
            	    We show that the better-performing ensembles also have higher Lyapunov coefficients and demonstrate that our exponent calculation is scalable to larger problems.
            	    The best GANs log probability from the best ensemble was $\num{-12861} \pm \num{356}$, which is worse than the ensemble's performance of $\num{-10846} \pm \num{256}$.
            	    This indicates that each GAN may be learning a different part of the data distribution, with samples shown in Appendix Figure~\ref{lyapunov_fig:gan_2d_samples}.
            	 }
            	\label{lyapunov_tab:gan_lyap_calc}
            \end{table}
            
    \section{Conclusion}\label{lyapunov_sec:conclusion}
        This chapter introduces Generalized Ridge Rider (GRR), an extension of the Ridge Rider (RR) algorithm to settings with multiple losses.
        In these settings, we showed that a broader class of bifurcation points needs to be considered and that GRR indeed discovers them in various problems.
        Experimentally, we isolate each component of GRR, demonstrating their effectiveness, and show that -- in contrast to baseline methods -- GRR can obtain a diversity of qualitatively different solutions in multi-agent settings such as the iterated prisoner's dilemma.
        We also provide empirical justification for our method using tools from the dynamical systems literature, allowing us to find arbitrary bifurcations.
        This hints at a multitude of approaches and tools from dynamical systems that can be used for understanding game dynamics and learning diversity.




\section*{Acknowledgements}
    The Province of Ontario, the Canadian government through CIFAR, and companies sponsoring the Vector Institute provided resources used in preparing this chapter.
    We would also like to thank C. Daniel Freeman, H\'erve J\'egou, Noam Brown, and David Acuna for feedback on this work and acknowledge the Python community ~\citep{van1995python, oliphant2007python} for developing the tools that enabled this work, including NumPy~\citep{oliphant2006guide, van2011numpy, harris2020array}, Matplotlib~\citep{hunter2007matplotlib} and SciPy~\citep{jones2001scipy}.

\subsection*{My Contributions Towards this Paper As it Pertains to the Thesis}
    I contributed the initial idea to this project, ran all the experiments, and did most of the writing.

    \chapter{Discussion and Concluding Remarks}
This thesis introduces new methods for scalable nested optimization in various relevant problems within machine learning.
First, we give methods based on hypernetworks and implicit differentiation for computing the best-response Jacobian, which we used to calculate approximate hypergradients for large-scale hyperparameter optimization.
Next, we augmented classical first-order optimization methods to reduce rotational dynamics when applied to nested optimization problems.
Finally, we generalize an algorithm that branches gradient-based optimization at bifurcations to work with nonconservative dynamics often present in nested optimization.

\section{Summary of Chapters}
    In Chapter~\ref{chap:hyperparamWithHypernets}, we address tuning hyperparameters using gradient-based optimization by replacing the training optimization loop with a differentiable hypernetwork.
    We presented a simple and scalable method that jointly optimizes both hyperparameters and hypernetwork weights.

    In Chapter~\ref{chap:hyperparamWithIFT}, we present a gradient-based hyperparameter optimization algorithm that scales to high-dimensional hyperparameters the size of modern, deep \nns\!\!.
    We use the implicit function theorem to formulate the hypergradient whose bottleneck is inverting the training loss Hessian with respect to the \nn parameters.
    We approximate inverse-Hessian-vector products using a relationship with unrolled differentiation.

    In Chapter~\ref{chap:complexMomentum}, we generalize existing momentum methods for learning in differentiable games by allowing complex momentum with real-valued updates.
    Our method robustly converges in games with a different range of eigenspace mixtures than existing methods.
    We also presented a practical generalization of our method to the Adam optimizer, which we used to improve BigGAN training.

    In Chapter~\ref{chap:lyapunovExponent}, we introduce Generalized Ridge Rider, an extension of the Ridge Rider algorithm to settings with multiple losses.
    In these settings, we showed that a broader class of bifurcation points needs to be considered and that GRR indeed discovers them in various problems.
    We also provide an empirical justification for our method using tools from the dynamical systems literature, allowing us to find arbitrary bifurcations.
    Experimentally, GRR obtains various solutions in multi-agent settings, such as the iterated prisoner's dilemma.

\section{Limitations and Future Directions}\label{conclusion_sec:future_and_limit}
    \subsection{Hyperparameter Optimization Through Hypernetworks}
        \subsubsection{Limitations}
            The most scalable architecture proposed in this work is a linear hypernetwork, which will not scale when the output weights are the size of modern neural networks. 
            We look at smarter factorized networks in non-thesis research \citet{mackay2019self} that scale to large modern networks.
            Another limitation is that it is difficult to set the scale of the hyperparameters sampled during training, which is again partially resolved by adaptive scales in non-thesis research \citet{mackay2019self}.
            Unlike Bayesian optimization, this method does not facilitate uncertainty-aware exploration and requires a differentiable validation loss.
            A further limitation is that this method relies on sampling hyperparameters, which will not scale for ultra-large hyperparameter regimes, i.e., hundreds of thousands to millions of hyperparameters.

        \subsubsection{Future directions}
            Future directions could include combining this amortized optimization framework into other nested optimization setups besides hyperparameter optimization -- ex., meta-learning schemes like MAML~\citep{finn2017model} or (small-scale) GANs.
            Alternatively, we could combine these methods with other strategies for approximating the best-response, like unrolled or implicit differentiation.
            Furthermore, optimizing thousands of hyperparameters raises the question of \emph{hyper-regularization}, or regularization of hyperparameters.
            We could also look for improved ways to formulate hypernetwork training, as in \citet{bae2020delta}.

    \subsection{Optimizing Millions of Hyperparameters by Implicit Differentiation}
        \subsubsection{Limitations}
            For implicit differentiation, we can only optimize hyperparameters that change the loss manifold, so our approach is not straightforwardly applicable to optimizer hyperparameters.
            Also, we need continuous hyperparameters to use gradient-based optimization, but many discrete hyperparameters (e.g., number of hidden units) have continuous relaxations~\citep{maddison2016concrete, jang2016categorical}.
            We cannot exactly compute the hypergradients since we must find $\left(\hp',\!\ep'\right)$ such that $\left. \smash{\trainGrad}\!\right|_{\hp'\!,\! \ep'}\!=\!0$, which we can only solve to a tolerance with an approximate solution denoted $\smash{\widehat{\ep^*}(\hp)}$.
            Some hyperparameters fail when we retrain the model with the final fixed value.
            Furthermore, this method does not facilitate uncertainty-aware exploration.
            Unlike Bayesian optimization, we also require a differentiable validation loss, which can directly optimize validation objectives such as accuracy.
            We also observed that the introduction of millions of hyperparameters can easily overfit the validation dataset.
            Some noisy hyperparameters, such as the continuous relaxation of discrete parameters as in \citet{mackay2019self}, did not work robustly with this method -- perhaps due to a higher sensitivity to noise.
            It would be better to investigate the effects of stochasticity in our results, for example, when we should compare to CG or re-sample batches in our hypergradient evaluation.
            As such, this does not scale to settings such as generating the code for our model as a hyperparameter as in \citet{zhang2023using}.

        \subsubsection{Future directions}
            \vspace{-0.005\textheight}
            We believe algorithms of this nature provide a path for practical nested optimization, where we have Hessians with a known structure.
            We seek to understand why some hyperparameters fail to retrain, which is partially solved in the non-thesis work of \cite{vicol2020implicit}.
            Since na\"ively introducing hyperparameters causes validation overfitting, we should find useful ways to introduce many hyperparameters.
            Further, we should investigate useful cross-validation data splits when using many hyperparameters and when we want to retrain with the final hyperparameters.
            Investigating other matrix-inversion methods besides the Neumann series could also be fruitful.

    \vspace{-0.005\textheight}
    \subsection{Complex Momentum for Optimization in Games}
        \vspace{-0.005\textheight}
        \subsubsection{Limitations}
        \vspace{-0.005\textheight}
            Our method stores real and imaginary components in the momentum buffer $\momVal$, so we use more memory than standard momentum.
            Our method is more challenging to analyze, as our augmented dynamics requires wielding the roots of a cubic instead of a quadratic.
            Our method introduces a new hyperparameter -- the imaginary components or $\arg$ of $\momCoeff$ --  that needs to be tuned.
            For some games, we need higher-order information than first-order to converge -- ex., pure-response games ~\citep{lorraine2019optimizing} -- because the first-order information for a player is identically zero.
            So, momentum methods that only use first-order info, like the direct gradient, will not generally converge.

        \vspace{-0.005\textheight}
        \subsubsection{Future directions}
        \vspace{-0.005\textheight}
            We should consider which kind of spectrum gradient optimization induces in nested optimization setups of interest, such as GANs, to guide our algorithm design.
            Furthermore, we should find useful approximations of these empirical spectrums that are tractable to analyze yet still provide useful results.
            We should also look to get a closed-form solution for the optimal learning rate and momentum as a function of the spectrum so that we can tightly bound the convergence rates for our method.
            Additionally, we should explore other, more general recurrently linked momentum setups to see which achieves the best convergence rates on spectrums of interest.

    \vspace{-0.005\textheight}
    \subsection{Lyapunov Exponents for Diversity in Differentiable Games}
        \vspace{-0.005\textheight}
        \subsubsection{Limitations}
        \vspace{-0.005\textheight}
            Our method does not have guarantees if our optimization process does not converge -- ex., when the learning rate is too high, when there is stochasticity, or when we have solutions as the parameters tend to infinity.
            Furthermore, our approaches to estimating the exponent involve simulating trajectories for a finite horizon, which we must tune.
            If the horizon is too long, our gradients become too sharp and have a limited signal, while if the horizon is too short, we may not find bifurcations.
            Further, instead of approximating the maximum eigenvalue of the sum of Jacobians in optimization trajectories, we tractably approximate the maximum eigenvalue for each Jacobian.
            We approximate the maximum eigenvalue via a power iteration.
            We also do gradient descent on our approximate Lyapunov exponent, which finds approximate solutions, and it is unclear how far we have to move in each direction to move across the bifurcation.
            Further, scaling to large-scale setups like GANs proves challenging, especially when optimizing the exponent.
            We should find strong use cases for which we want to find diverse solutions in nested optimization setups.

        \subsubsection{Future directions}
            This work hints at many approaches and tools from dynamical systems that can be used to understand game dynamics and learning diversity.
            We should also look for more use cases where we care about finding multiple solutions to the nested optimization.
            Using objectives wielding more elements of the Lyapunov spectrum may also be useful for finding regions where we can branch across bifurcations in multiple directions.
            Finding uses of Lyapunov exponent optimization may be helpful in other problems, e.g., augmented influence functions, where removing a data point changes our optimization trajectory instead of just the converged value.

  \appendix
    
    \chapter{Optimizing Millions of Hyperparameters by Implicit Differentiation}\label{chap:app_hyperparamWithIFT}
        
        \section{Extended Background}\label{ift_app:background}
            This section outlines our notation (Table~\ref{ift_tab:TableOfNotation}) and the proposed algorithm.
            Here, we assume that we have access to a finite dataset $\dataset = \{\left(\vx_i, \vy_i\right) | i = 1 \dots n \}$, with $n$ examples drawn from the distribution $p\left(\vx, \vy\right)$ with support $\mathcal{P}$.
            We denote the input and target domains by $\mathcal{X}$ and $\mathcal{Y}$, respectively.
            Assume $\vy: \xdom \to \tdom$ is a function and we wish to learn $\hat{\vy}:\xdom \times \edom \to \tdom$ with a \nn parameterized by $\ep \in \edom$, so that $\hat{\vy}$ is close to $\vy$.
            We measure how close a predicted value is to a target with the prediction loss $\loss: \tdom \times \tdom \to \Real$.
            Our goal is to minimize the expected prediction loss or population risk: $\argmin_{\ep} \mathbb{E}_{\vx \sim p(\vx)}\left[\loss\left(\hat{\vy}\left(\vx, \ep\right), \vy\left(\vx\right)\right)\right]$.
            Since we only have access to a finite number of samples, we minimize the empirical risk: $\argmin_{\ep} \nicefrac{1}{n}\sum_{\vx, \vy \in \dataset}\loss\left(\hat{\vy}\left(\vx, \ep\right), \vy\left(\vx\right)\right)$.
            
            Due to a limited size dataset $\dataset$, there may be a significant difference between the minimizer of the empirical risk and the population risk.
            We can estimate this difference by partitioning our dataset into {training} and {validation} datasets--- $\dataset_{train}, \dataset_{valid}$.
            We find the minimizer on the training dataset $\dataset_{train}$ and estimate its performance on the population risk by evaluating the empirical risk over the validation dataset $\dataset_{valid}$.
            We modify the empirical training risk to decrease our population risk, parameterized by $\hp \in \hdom$.
            These parameters for generalization are called hyperparameters.
            We call the modified empirical training risk our training loss for simplicity and denote it $\Ltr \left(\hp, \ep\right)$.
            Our empirical validation risk is called validation loss for simplicity and is denoted by $\Lval \left(\hp, \ep\right)$.
            Often, the validation loss does not directly depend on the hyperparameters, and we have $\Lval \left(\ep\right)$.

            The population risk is estimated by plugging the training loss minimizer $\response = \argmin_{\ep} \Ltr \left(\hp, \ep\right)$ into the validation loss for the estimated population risk $\Lval^*(\hp) = \LvalResponse$.
            We want our hyperparameters to minimize the estimated population risk: $\hp^* = \argmin_{\hp} \Lval^*\left(\hp\right)$.
            We can create a third partition of our dataset $\dataset_{test}$ to assess if we overfit the validation dataset $\dataset_{valid}$ with our hyperparameters $\hp$.
            A citation for this thesis is included at \citet{lorraine2024scalable}.

        \vspace{-0.01\textheight}
        \section{Extended Related Work}
        \label{ift_app:related_work}
             Black-box optimization methods such as random search \citep{bergstra2012random} or Bayesian optimization \citep{mockus1998application,shahriari2015taking} have been deployed for hyperparameter optimization (HPO).
             Beyond black-box methods, other works further use the problem structure, e.g., are compute-environment-aware~\cite{ginsbourger2010kriging}, compute-budget-aware~\citep{lam2016bayesian}, multi-task~\citep{swersky2013multi}, like transfer-learning~\citep{golovin2017google} or multi-fidelity~\citep{klein2017fast}, iterative-optimization-aware~\citep{li2017hyperband}, online~\citep{jaderberg2017population}, or multi-objective~\citep{daulton2022multi}.
             However, these methods (a) still rely on practitioners to design a search space, which includes selecting which parameters can be optimized and specifying bounds on these parameters, and (b) typically struggle in the initial search phase ($< 2^{d}$ queries for $d$-dimensional hyperparameters).
             We further detail a few notable cases:
            
            \textbf{Independent \ho\!\!:}
                A simple class of \ho algorithms involves making several independent hyperparameter selections and training the model to completion.
                Popular examples include grid search and random search~\citep{bergstra2012random}.
                Since each hyperparameter selection is independent, these algorithms are trivial to parallelize.

            \textbf{Global \ho\!\!:}
                Some \ho algorithms try to find a globally optimal hyperparameter setting, which can be important if the loss is non-convex.
                A simple example is random search, while a more sophisticated example is Bayesian optimization~\citep{movckus1975bayesian, snoek2012practical, kandasamy2019tuning}.
                These HO algorithms often involve reinitializing the hyperparameter and weights on each optimization iteration.
                This allows global optimization at the cost of expensive retraining weights or hyperparameters.

            \textbf{Local \ho\!\!:}
                Other \ho algorithms only try to find a locally optimal hyperparameter setting.
                Often, these algorithms maintain a current estimate of the best combination of hyperparameters and weights.
                The hyperparameter is adjusted by a small amount on each optimization iteration, which allows us to avoid excessive retraining of the weights on each update.
                This is because the new optimal weights are close to the old optimal weights because of a small change in the hyperparameters.
            
            \textbf{Learned proxy function based \ho\!\!:}
                Many \ho algorithms try to learn a proxy function for optimization.
                The proxy function is used to estimate the loss for a hyperparameter selection.
                We could learn a proxy function for global or local \ho.
                We can learn a useful proxy function on any node in our computational graph, including the optimized weights.
                For example, we could learn how optimized weights change with respect to hyperparameters ~\citep{lorraine2018stochastic}, how the optimized predictions change with respect to the hyperparameters~\citep{mackay2019self}, or how the optimized validation loss changes with respect to the hyperparameters as in Bayesian optimization.
                As in Bayesian optimization, it is possible to do gradient descent on the proxy function to find new hyperparameters to query.
                Alternatively, we could use a non-differentiable proxy function to get cheap estimates of the validation loss like SMASH~\citep{brock2017smash} for architecture choices.
                \newline
                \newline
                
         \begin{table*}\caption{Notation For Optimizing Millions of Hyperparameters by Implicit Differentiation}
            \begin{center}
                \begin{tabular}{c | c}
                    \toprule
                    HO & Hyperparameter optimization\\
                    NN & Neural network\\
                    IFT & Implicit Function Theorem\\
                    HVP / JVP & Hessian/Jacobian-vector product\\
                    $\hp, \ep$ & Hyperparameters and \nn parameters/weights\\
                    $\hdim, \edim$ & Hyperparameter and \nn parameter dimensionality\\
                    $\hdom  \subseteq  \Real^\hdim, \edom  \subseteq  \Real^\edim$ & Hyperparameters and \nn parameter domains\\
                    $\hp', \ep'$ & Arbitrary, fixed hyperparameters and weights\\
                    $\Ltr(\hp, \ep),\Lval(\hp, \ep)$& Training loss \& validation loss\\
                    ${\color{blue}\response}$ & Best-response of the weights to the hyperparameters\\
                    ${\color{blue}\widehat{\ep^*}(\hp)}$ & Approximate best-response of the weights to the hyperparameters\\
                    ${\color{red}\Lval^{*}\left(\hp\right)} \!=\! {\color{red}\Lval\left(\hp, \response\right)}$& The validation loss with best-responding weights\\
                    {\color{red}Red}& (Approximate) Validation loss with best-responding weights\\
                    $\edom^* = \ep^*(\hdom)$ & The domain of best-responding weights\\
                    $\hp^*$ & The optimal hyperparameters\\
                    $\vx, \vy$ & An input and its associated target\\
                    $\xdom, \tdom$ & The input and target domains, respectively.\\
                    $\dataset$ & A data matrix consisting of tuples of inputs and targets\\
                    $\vy(\vx, \ep)$ & A predicted target for a input data and weights\\
                    ${\color{mydarkgreen}\hpderiv{\Lval}}, \epderiv{\Lval}$ & The (validation loss hyperparameter / parameter) direct gradient\\
                    {\color{mydarkgreen}Green} & (Approximations to) The validation loss direct gradient.\\
                    ${\color{blue}\hpderiv{\ep^*}}$ & The best-response Jacobian\\
                    {\color{blue}Blue} & (Approximate) (Jacobian of the) best-response of the weights\\
                    & to the hyperparameters\\
                    $\pd{\Lval}{\ep} {\color{blue}\hpderiv{\ep^*}}$& The indirect gradient\\
                    $\hpderiv{{\color{red}\Lval^*}}$& Hypergradient: sum of validation losses direct and indirect gradient\\
                    ${\color{magenta}\left[ \trainHess{\ep} \right]^{-1}}$ & The training Hessian inverse\\
                    {\color{magenta}Magenta} & (Approximations to) The training Hessian inverse\\
                    ${\color{orange} \epderiv{\Lval} \left[ \trainHess{\ep} \right]^{-1}}$ & The vector - Inverse Hessian product.\\
                    {\color{orange}Orange} & (Approximations to) The vector - Inverse Hessian product.\\
                    $\trainMixed{\ep}$ & The training mixed partial derivatives\\
                    $\identity$ & The identity matrix\\
                    \bottomrule
                \end{tabular}
            \end{center}
            \label{ift_tab:TableOfNotation}
        \end{table*}
        
        \vspace{-0.01\textheight}
        \section{Implicit Function Theorem}
            \begin{thm*}[Augustin-Louis Cauchy, Implicit Function Theorem]\label{ift_app:IFTComplete}
                \textnormal{
                    Let $\iftf\left(\iftx, \ifty\right): \ifthdom \times \iftedom \to \iftedom$ be a continuously differentiable function.
                    Fix a point $\left(\ifta, \iftb\right)$ with $\iftf\left(\ifta, \iftb\right) = 0$.
                    If the Jacobian $J^{\iftf}_{\ifty}\left(\ifta, \iftb\right)$ is invertible, there exists an open set $U \subseteq \ifthdom$ containing $\ifta$ such that there exists a continuously differentiable function $\iftg: U \to \iftedom$ such that:\vspace{-0.01\textheight}
                    \begin{equation*}
                        \iftg\left(\ifta\right) = \iftb \textnormal{ and } \forall \iftx \in U, \iftf\left(\iftx, \iftg\left(\iftx\right)\right) = 0
                    \end{equation*}
                    Moreover, the partial derivatives of $\iftg$ in $U$ are given by the matrix product:\vspace{-0.01\textheight}
                    \begin{equation*}
                        \pd{\iftg}{\iftx} \left(\iftx\right) = -\left[ J^{\iftf}_{\ifty}\left(\iftx, \iftg\left(\iftx\right)\right) \right]^{-1} J^{\iftf}_{\iftx} \left(\iftx, \iftg\left(\iftx\right)\right)
                    \end{equation*}
                }
            \end{thm*}
            \noindent
            Typically the IFT is presented with $\iftf \!=\! f$, $\iftg \!=\! g$, $\ifthdom \!=\! \Real^{m}$, $\iftedom \!=\! \Real^{n}$, $\iftx \!=\! x$, $\ifty \!=\! y$, $\ifta \!=\! a$, $\iftb \!=\! b$.
        
        \section{Proofs}\label{ift_app:proofs}
            \begin{lemma*}[1]
                \textnormal{
                    If the recurrence given by unrolling SGD optimization in Equation~\ref{ift_eqn:sgd_recurrence} has a fixed point $\epA$ (i.e., $0 = \epderiv{\Ltr} |_{\hp, \epA(\hp)}$), then:
                    \begin{equation*}
                        \hpderiv{\epA} = \left. -\left[ \trainHess{\epA(\hp)} \right]^{-1} \trainMixed{\epA(\hp)} \right|_{\epA(\hp)}
                    \end{equation*}
                }
            \end{lemma*}
            \begin{proof}
                \begin{align*}
                    &\Rightarrow \hpderiv{}\left( \left. \epderiv{\Ltr} \right|_{\hp, \epA(\hp)} \right) = 0
                        && \text{given}\\
                    &\Rightarrow \left. \left( \trainMixed{\epA(\hp)} I + \trainHess{\epA(\hp)} \hpderiv{\epA} \right) \right|_{\hp, \epA(\hp)} = 0
                        && \left. \text{chain rule through } \right|_{\hp, \epA(\hp)}\\
                    &\Rightarrow \left. \trainHess{\epA(\hp)} \hpderiv{\epA} \right|_{\hp, \epA(\hp)} = \left. -\trainMixed{\epA(\hp)} \right|_{\hp, \epA(\hp)}
                        && \text{re-arrange terms}\\
                    &\Rightarrow \left. \hpderiv{\epA} \right|_{\hp} = \left. -\left[ \trainHess{\epA(\hp)} \right]^{-1} \trainMixed{\epA(\hp)} \right|_{\hp}
                        && \text{left-multiply by } \left. \left[ \trainHess{\epA(\hp)} \right]^{-1} \right|_{\hp, \epA(\hp)}
                \end{align*}
            \end{proof}
            
            \begin{lemma*}[2]
                \textnormal{
                    Given the recurrence from unrolling SGD optimization in Equation~\ref{ift_eqn:sgd_recurrence} we have:
                    \begin{equation*}
                        \hpderiv{\ep_{i+1}} = -\sum_{j\leq i} \left( \prod_{k<j} I - \left. \trainHess{\ep_{i-k}(\hp)} \right|_{\hp, \ep_{i-k}(\hp)} \right) \left. \trainMixed{\ep_{i-j}(\hp)} \right|_{\hp, \ep_{i-j}(\hp)}
                    \end{equation*}
                }
            \end{lemma*}
            \begin{proof}
                \begin{align*}
                    &\left. \hpderiv{\ep_{i+1}} \right|_{\hp}
                    = \hpderiv{} \left( \left. \ep_{i}(\hp) - \epderiv{\Ltr} \right|_{\hp, \ep_{i}(\hp)} \right)
                        && \text{take derivative with respect to } \hp\\
                    &\smallSpace= \left. \hpderiv{\ep_{i}} \right|_{\hp} - \hpderiv{}\left( \left. \epderiv{\Ltr} \right|_{\hp, \ep_{i}(\hp)} \right)
                        && \text{chain rule}\\
                    &\smallSpace= \left. \hpderiv{\ep_{i}} \right|_{\hp} - \left. \left( \trainHess{\ep_{i}(\hp)} \hpderiv{\ep_{i}} + \trainMixed{\ep_{i}(\hp)}\right) \right|_{\hp, \ep_{i}(\hp)}
                        && \left. \text{chain rule through } \right|_{\hp, \ep_{i}(\hp)}\\
                    &\smallSpace= \left. -\trainMixed{\ep_{i}(\hp)} \right|_{\hp, \ep_{i}(\hp)} + \left. \left(I - \trainHess{\ep_{i}(\hp)}\right)\hpderiv{\ep_{i}} \right|_{\hp, \ep_{i}(\hp)}
                        && \text{re-arrange terms}\\
                    &\smallSpace= \left. -\trainMixed{\ep_{i}(\hp)} \right|_{\hp, \ep_{i}(\hp)} + \left. \left(I - \trainHess{\ep_{i}(\hp)}\right) \right|_{\hp, \ep_{i}(\hp)}\cdot\\
                        &\medSpace \left. \left( \!\! \left(\! I - \trainHess{\ep_{i-1}(\hp)} \!\right)\! \hpderiv{\ep_{i-1}} - \trainMixed{\ep_{i-1}(\hp)} \!\right) \right|_{\hp, \ep_{i-1}(\hp)}
                        && \text{expand } \hpderiv{\ep_{i}}\\
                    &\smallSpace= \left. -\trainMixed{\ep_{i}(\hp)} \right|_{\hp, \ep_{i}(\hp)} - \left( \!I - \left. \trainHess{\ep_{i}(\hp)} \right|_{\hp, \ep_{i}(\hp)} \! \right) \! \left. \trainMixed{\ep_{i-1}(\hp)} \right|_{\hp, \ep_{i-1}(\hp)} +\\
                        &\medSpace \left[ \prod_{k<2}I - \left. \trainHess{\ep_{i-k}(\hp)} \right|_{\hp, \ep_{i-k}(\hp)} \right] \left. \hpderiv{\ep_{i-1}} \right|_{\hp}
                        && \text{re-arrange terms}\\
                    &\smallSpace= \cdots 
                        && \text{}\\
                    &\textnormal{So, }\hpderiv{\ep_{i+1}} = -\sum_{j\leq i} \left[ \prod_{k<j} I - \left. \trainHess{\ep_{i-k}(\hp)} \right|_{\hp, \ep_{i-k}(\hp)} \right] \left. \trainMixed{\ep_{i-j}(\hp)} \right|_{\hp, \ep_{i-j}(\hp)}
                        && \text{telescope the recurrence}
                \end{align*}
            \end{proof}
            \begin{thm*}[Neumann-SGD]
                \textnormal{
                    Given the recurrence from unrolling SGD optimization in Equation~\ref{ift_eqn:sgd_recurrence}, if $\ep_0 = \ep^*(\hp)$:
                    \begin{equation*}
                        \hpderiv{\ep_{i+1}} = -\left. \left( \sum_{j \leq i} \left[I - \trainHess{\ep(\hp)}\right]^j \right) \trainMixed{\ep(\hp)} \right|_{\ep^*(\hp)}
                    \end{equation*}
                }
                and if \textnormal{$I - \trainHess{}$} is contractive:
                \textnormal{
                    \begin{equation*}
                        \lim_{i\to\infty} \hpderiv{\ep_{i+1}} = -\left. \left[ \trainHess{\ep(\hp)} \right]^{-1} \trainMixed{\ep(\hp)} \right|_{\ep^*(\hp)}
                    \end{equation*}
                }
            \end{thm*}
            \begin{proof}
                \begin{align*}
                    &\lim_{i\to\infty} \left. \hpderiv{\ep_{i+1}} \right|_{\hp}
                        && \text{take} \lim_{i\to\infty} \\
                    &\smallSpace= \lim_{i\to\infty} \left( -\sum_{j\leq i} \left[ \prod_{k<j} I - \left. \trainHess{\ep_{i-k}(\hp)} \right|_{\hp, \ep_{i-k}(\hp)} \right] \left. \trainMixed{\ep_{i-j}(\hp)} \right|_{\hp, \ep_{i-j}(\hp)} \right)
                        && \text{by Lemma 2}\\
                    &\smallSpace= -\lim_{i\to\infty} \left. \left( \sum_{j\leq i} \left[ \prod_{k<j} I - \trainHess{\ep(\hp)} \right] \trainMixed{\ep(\hp)} \right) \right|_{\hp, \ep^*(\hp)}
                        && \ep_0 = \ep^*(\hparam) = \ep_i \\
                    &\smallSpace= -\lim_{i\to\infty} \left. \left( \sum_{j \leq i} \left[I - \trainHess{\ep(\hp)}\right]^j \right) \trainMixed{\ep(\hp)} \right|_{\hp, \ep^*(\hp)}
                        && \text{simplify} \\
                    &\smallSpace= -\left. \left[I - \left( I - \trainHess{\ep(\hp)} \right) \right]^{-1} \trainMixed{\ep(\hp)} \right|_{\hp, \ep^*(\hp)}
                        && \text{contractive \& Neumann series} \\
                    &\smallSpace= -\left. \left[ \trainHess{\ep(\hp)} \right]^{-1} \trainMixed{\ep(\hp)} \right|_{\hp, \ep^*(\hp)}
                        && \text{simplify}
                \end{align*}
            \end{proof}
        
            
        \section{Experiments}\label{ift_app:experiments}
            We use PyTorch~\citep{paszke2017automatic} as our computational framework.
            All experiments were performed on NVIDIA TITAN Xp GPUs.
        
            For all CNN experiments, we use the following optimization setup:
            for the \nn weights we use Adam~\citep{kingma2014adam} with a learning rate of $1$e$-4$.
            For the hyperparameters, we use RMSprop~\citep{hinton2012neural} with a learning rate of $1$e$-2$.
            \subsection{Overfitting a Small Validation Set}
                We see our algorithm's ability to overfit the validation data (see Fig.~\ref{ift_fig:overfit_valid_all}).
                We use $\numOverfitTrain$ training input and $\numOverfitValid$ validation input with the standard testing partition for both MNIST and CIFAR-\num{10}.
                We check performance with logistic regression (Linear), a $1$-layer fully-connected \nn with as many hidden units as input size (ex., $\num{28} \times \num{28} = \num{784}$, or $\num{32} \times \num{32} \times \num{3} = \num{3072}$), LeNet~\citep{lecun1998gradient}, AlexNet~\citep{krizhevsky2012imagenet}, and ResNet\num{44}~\citep{he2016deep}.
                In all examples, we can achieve \SI{100}{\percent} training and validation accuracy, while the testing accuracy is significantly lower.
                \begin{figure}
                    \centering
                    \begin{tikzpicture}
                        \centering
                        \node (img){\includegraphics[trim={0.15cm 0cm 0.15cm 0cm},clip, width=.48\linewidth]{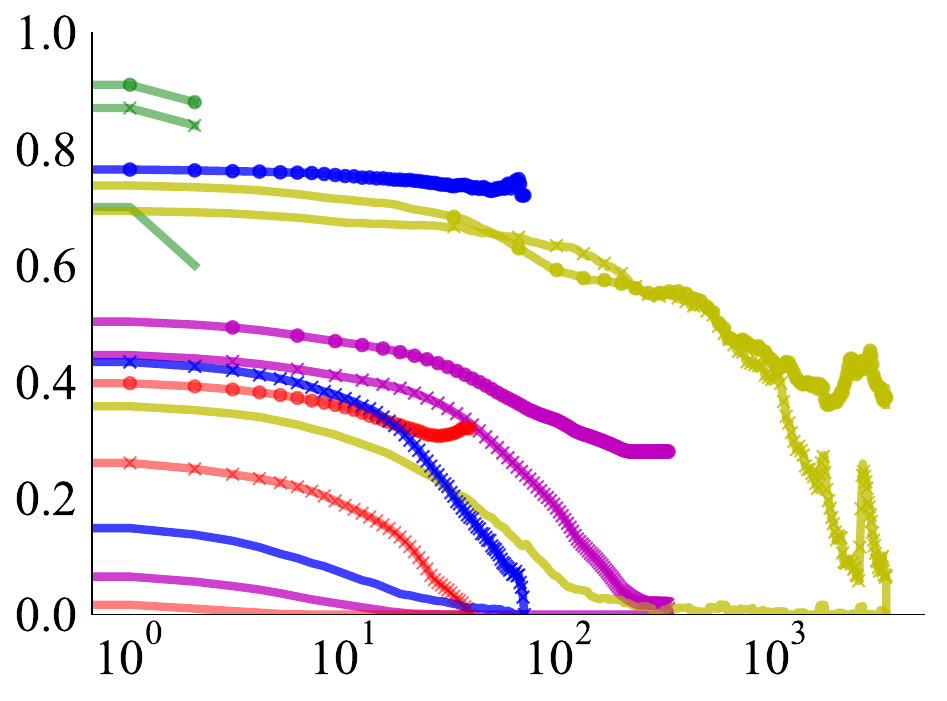}};
                        \node[left=of img, node distance=0cm, rotate=90, xshift=1.75cm, yshift=-.90cm, font=\color{black}] {Classification Error};
                        \node[above=of img, node distance=0cm, yshift=-1.25cm,font=\color{black}] {MNIST};
                        \node[below=of img, node distance=0cm, yshift=1.25cm,font=\color{black}] {Iteration};
                        \node (img2)[right=of img, xshift=-1.3cm]{\includegraphics[trim={1.3cm 0cm 0.15cm 0cm},clip, width=.455\linewidth]{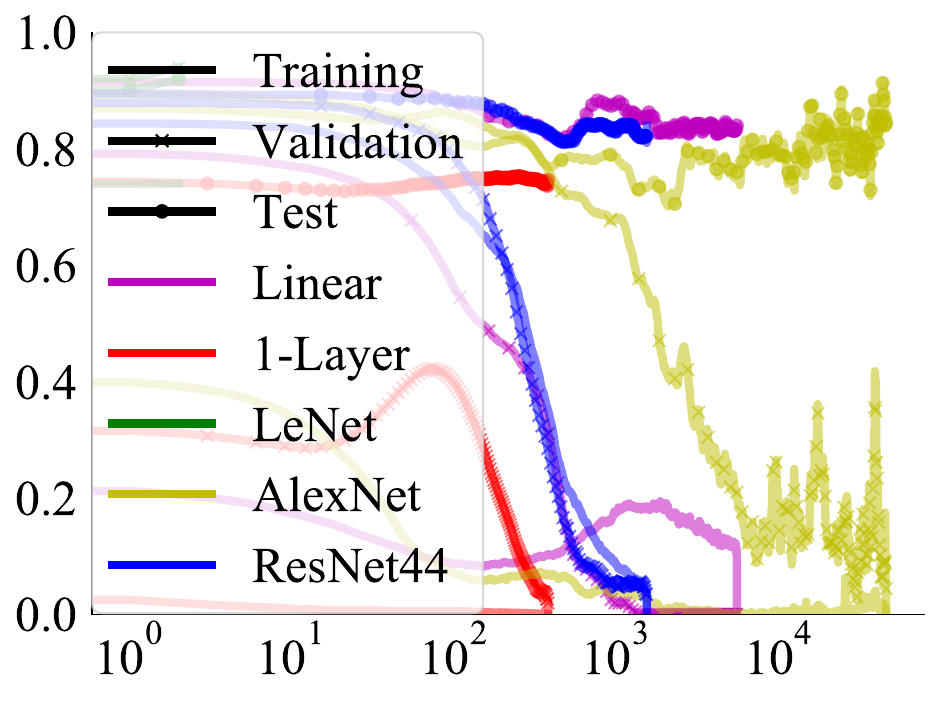}};
                        \node[above=of img2, node distance=0cm, yshift=-1.25cm,font=\color{black}] {CIFAR-\num{10}};
                        \node[below=of img2, node distance=0cm, yshift=1.25cm,font=\color{black}] {Iteration};
                    \end{tikzpicture}
                    \caption[Overfitting validation data]{{Overfitting validation data.}
                        {Algorithm~\ref{ift_alg:joint_train} can overfit the validation dataset.}
                        We use $\numOverfitTrain$ training input and $\numOverfitValid$ validation input with the standard testing partition for both MNIST and CIFAR-\num{10}.
                        We check the performance with logistic regression (Linear), a fully connected $1$-layer \nn with as many hidden units as input size (ex., $\num{28} \times \num{28} = \num{784}$, or $\num{32} \times \num{32} \times \num{3} = \num{3072}$), LeNet~\citep{lecun1998gradient}, AlexNet~\citep{krizhevsky2012imagenet}, and ResNet\num{44}~\citep{he2016deep}.
                        Separate lines are plotted for the training, validation, and testing errors.
                        In all examples, we achieve \SI{100}{\percent} training and validation accuracy, while testing accuracy is significantly lower.
                    }
                    \label{ift_fig:overfit_valid_all}
                \end{figure}

            \subsection{Dataset Distillation}
                With MNIST, we use the entire validation dataset, while for CIFAR, we use $\numDistillValid$ data points.
                \newpage

                \begin{figure*}
                    \centering
                    \begin{tikzpicture}
                        \centering
                        \node (img){\includegraphics[trim={3cm 2.5cm 2.5cm 3cm},clip, width=.98\linewidth]{latex/hyperparamsWithIFT/learned_images_CIFAR100_100_mlp.pdf}};
                        \node[above=of img, node distance=0cm, yshift=-1.cm,font=\color{black}] {\textbf{CIFAR-100 Distillation}};
                    \end{tikzpicture}
                    \caption[The complete dataset distillation for CIFAR-100]{
                        The complete dataset distillation for CIFAR-100.
                        Referenced in Fig.~\ref{ift_fig:distilled_cifar}.
                    }
                    \label{ift_app:cifar-100-distillation}
                \end{figure*}
                
            \subsection{Learned Data Augmentation}
                \textbf{Augmentation Network Details:}
                    Data augmentation can be framed as an image-to-image transformation problem.
                    Inspired by this, we use a U-Net~\citep{ronneberger2015u} as the data augmentation network.
                    To allow stochastic transformations, we provide random noise by concatenating a noise channel into the input image so that the resulting input has $4$ channels.
                    We evaluated the accuracy for training, validation, and testing with an average of $10$ augmented samples.

            \subsection{RNN Hyperparameter Optimization}
                \label{ift_app:exp_rnn}
                We base our implementation on the AWD-LSTM codebase.\footnote{\url{https://github.com/salesforce/awd-lstm-lm}}
                Similar to~\citet{gal2016theoretically}, we used a \num{2}-layer LSTM with \num{650} hidden units per layer and \num{650}-dimensional word embeddings.

                \paragraph{Overfitting Validation Data:}
                We used a subset of \num{10} training sequences and $10$ validation sequences and tuned separate weight decays per parameter.
                The LSTM architecture we use has \num{13280400} weights and thus an equal number of weight decay hyperparameters.

                \paragraph{Optimization Details:}
                For the large-scale experiments, we follow the training setup proposed in \citet{merity2017regularizing}: for the \nn weights, we use SGD with learning rate \num{30} and gradient clipping to magnitude \num{0.25}.
                The learning rate was decayed by a factor of $4$ based on the non-monotonic criterion introduced by~\citet{merity2017regularizing} (i.e., when the validation loss fails to decrease for $5$ epochs).
                We used Adam with learning rate \num{0.001} to optimize the hyperparameters.
                We trained on sequences of length $70$ in mini-batches of size $40$.

    \chapter{Complex Momentum for Optimization in Games}
            \begin{table}[h!]
                \vspace{-0.035\textheight}
                \caption{Notation for Complex Momentum for Optimization in Games}
                \vspace{-0.02\textheight}
                \begin{center}
                    \begin{tabular}{c c}
                        \toprule
                        SGD & Stochastic Gradient Descent\\
                        CM & Complex Momentum\\
                        SGDm, SimSGDm, \dots & \dots with momentum\\
                        SimSGD, SimCM & Simultaneous \dots\\
                        AltSGD, AltCM & Alternating \dots\\
                        GAN & Generative Adversarial Network\\
                        EG, OG & Extragradient and Optimistic Gradient\\
                        IS & Inception Score~\citep{salimans2016improved}\\
                        $\defeq$ & Defined to be equal to\\
                        $x, y, z, \dots \in \mathbb{C}$ & Scalars\\
                        $\boldsymbol{x}, \boldsymbol{y}, \boldsymbol{z}, \dots \in \mathbb{C}^{n}$ & Vectors\\
                        $\boldsymbol{X}, \boldsymbol{Y}, \boldsymbol{Z}, \dots \in \mathbb{C}^{n \times n}$ & Matrices\\
                        $\boldsymbol{X}^\transpose$ & The transpose of matrix $\boldsymbol{X}$\\
                        $\identity$ & The identity matrix\\
                        $\Re(z), \Im(z)$ & The real or imaginary component of $z \in \mathbb{C}$\\
                        $i$ & The imaginary unit. $z \in \mathbb{C} \implies z = \Re(z) + i \Im(z)$\\
                        $\widebar{z}$ & The complex conjugate of $z \in \mathbb{C}$\\
                        $|z| \defeq \sqrt{z \widebar{z}}$ & The magnitude or modulus of $z \in \mathbb{C}$\\
                        $\arg(z)$ & The argument or phase of $z \in \mathbb{C} \implies z = |z| \exp(i\arg(z))$\\
                        $z \!\in\! \mathbb{C}$ is \emph{almost-positive}& $\arg(z) \!=\! \epsilon$ for small $\epsilon$ respectively\\
                        $\outSymbol, \inSymbol$ & A symbol for the outer/inner players\\
                        $\outDim, \inDim \in \mathbb{N}$ & The number of weights for the outer/inner players\\
                        $\paramSymbol$ & A symbol for the parameters or weights of a player\\
                        $\outParam \in \mathbb{R}^{\outDim}, \inParam \in \mathbb{R}^{\inDim}$ & The outer/inner parameters or weights\\
                        $\loss: \mathbb{R}^{n} \to \mathbb{R}$ & A symbol for a loss\\
                        $\outLoss(\outParam, \inParam), \inLoss(\outParam, \inParam)$ & The outer/inner losses -- $\mathbb{R}^{\outDim + \inDim} \mapsto \mathbb{R}$\\
                        $\outGrad(\outParam, \inParam), \inGrad(\outParam, \inParam)$ & Gradient of outer/inner losses w.r.t. their weights in $\mathbb{R}^{\outDim/\inDim}$\\
                        $\inParam^*\!(\outParam\!) \!\defeq\! \argmin\limits_{\inParam} \!\!\inLoss\!(\!\outParam \!, \!\inParam\!)$&The best-response of the inner player to the outer player\\
                        $\outLoss^*(\outParam) \!\defeq\! \outLoss\!(\!\outParam \!, \!\inParam^*(\outParam)\!)$ & The outer loss with a best-responding inner player\\
                        $\outParam^* \!\defeq\! \argmin\limits_{\outParam} \outLoss^*(\outParam\!) $ & Outer optimal weights with a best-responding inner player\\
                        $\bothDim \defeq \outDim + \inDim$ & The combined number of weights for both players\\
                        $\bothParam \defeq [\outParam, \inParam] \in \mathbb{R}^{\bothDim}$ & A concatenation of the outer/inner weights\\
                        $\bothGrad(\bothParam) \!\defeq\! [\outGrad(\bothParam), \inGrad(\bothParam)] \in \mathbb{R}^{\bothDim}$ & A concatenation of the outer/inner gradients\\
                        $\bothParam^{\num{0}} = [\outParam^{\num{0}}, \inParam^{\num{0}}] \in \mathbb{R}^{\bothDim}$ & The initial parameter values\\
                        $j$ & An iteration number\\
                        $\bothGrad^j \defeq \bothGrad(\bothParam^j) \in \mathbb{R}^{\bothDim}$ & The joint-gradient vector field at weights $\bothParam^j$\\
                        $\nabla_{\bothParam} \bothGrad^j \defeq \nabla_{\bothParam}\bothGrad |_{\bothParam^j} \in \mathbb{R}^{\bothDim \times \bothDim}$ & The Jacobian of the joint-gradient $\bothGrad$ at weights $\bothParam^j$\\
                        $\lr \in \mathbb{C}$ & The step size or learning rate\\
                        $\momCoeff \in \mathbb{C}$ & The momentum coefficient\\
                        $\momCoeff_1 \in \mathbb{C}$ & The first momentum parameter for Adam\\
                        $\momVal \in \mathbb{C}^{\bothDim}$ & The momentum buffer\\
                        $\eigval \in \mathbb{C}$ & Notation for an arbitrary eigenvalue\\
                        $\spectrum(\boldsymbol{M}) \in \mathbb{C}^{n}$ & The spectrum -- or set of eigenvalues -- of $\boldsymbol{M} \in \mathbb{R}^{n \times n}$\\
                        \emph{Purely adversarial/cooperative} game& $\spectrum(\nabla_{\bothParam} \bothGrad)$ is purely real/imaginary\\
                        $\spectralRadius(\boldsymbol{M}) \!\defeq\! \max_{z \in \spectrum(\boldsymbol{M})} |z|$ & The spectral radius in $\mathbb{R}^{+}$ of $\boldsymbol{M} \in \mathbb{R}^{n \times n}$\\
                        $\fixedPointOp([\momVal, \bothParam])$ & Fixed point op. for CM, or augmented learning dynamics\\
                        $\dynamics \defeq \nabla_{[\momVal, \bothParam]}\fixedPointOp \in \mathbb{R}^{\num{3}\bothDim \times \num{3}\bothDim}$ & Jacobian of the augmented learning dynamics in \resultName~\ref{cm_thm:theorem}\\
                        $\lr^*\!, \momCoeff^* \!\defeq\! \argmin\limits_{\lr, \momCoeff} \spectralRadius(\dynamics(\lr,\! \momCoeff)\!)$ & The optimal step size and momentum coefficient\\
                        $\spectralRadius^* \defeq \spectralRadius(R(\lr^*, \momCoeff^*))$ & The optimal spectral radius or convergence rate\\
                        $\condition \defeq \frac{\max \spectrum(\nabla_{\bothParam} \gradSymbol)}{\min \spectrum(\nabla_{\bothParam} \gradSymbol)}$ & Condition number, for convex single-objective optimization\\
                        $\sigma_{min}^2(\boldsymbol{M}) \!\defeq\! \max \spectrum(\boldsymbol{M}^\transpose\boldsymbol{M}) $ & The minimum singular value of a matrix $\boldsymbol{M}$\\
                        \bottomrule
                    \end{tabular}
                \end{center}
                \label{cm_tab:TableOfNotation}
            \end{table}
        
    \section{Supporting Results}
        First, some basic results about complex numbers that are used:
        \begin{equation}
            z = \Re\left(z\right) + i \Im\left(z\right) = |z| \exp\left(i \arg\left(z\right)\right)
        \end{equation}
        \begin{equation}
             \widebar{z} =  \Re\left(z\right) - i \Im\left(z\right) = |z| \exp\left(- i \arg\left(z\right)\right)
        \end{equation}
        \begin{equation}
            \exp\left(i z\right) + \exp\left(-i z\right) = 2 \cos\left(z\right)
        \end{equation}
        \begin{equation}
            \widebar{z_1 z_2} = \widebar{z}_1 \widebar{z}_2
        \end{equation}
        \begin{equation}
            \nicefrac{1}{2}\left(z + \widebar{z}\right) = \Re\left(z\right)
        \end{equation}
        \begin{equation}
            \Re\left(z_1 z_2\right) = \Re\left(z_1\right) \Re\left(z_2\right) - \Im\left(z_1\right) \Im\left(z_2\right)
        \end{equation}
        \begin{equation}\label{cm_eq:complex_add}
            z_1 + z_2 = \left(\Re\left(z_1\right) + \Re\left(z_2\right)\right) + i\left(\Im\left(z_1\right) + \Im\left(z_2\right)\right)
        \end{equation}
        \begin{equation}\label{cm_eq:complex_mult}
            z_1 z_2 = \left(\Re\left(z_1\right) \Re\left(z_2\right) - \Im\left(z_1\right)\Im\left(z_2\right)\right) + i\left(\Im\left(z_1\right)\Re\left(z_2\right) + \Re\left(z_1\right)\Im\left(z_2\right)\right)
        \end{equation}
        \begin{equation}\label{cm_eq:complex_mult_polar}
            z_1 z_2 = |z_1| |z_2| \exp\left(i \left(\arg\left(z_1\right) + \arg\left(z_2\right)\right)\right)
        \end{equation}
        \begin{equation}\label{cm_eq:euler_formula}
            z^k = |z|^k \exp\left(i \arg\left(z\right)k\right) = |z|^k \left(\cos\left(k\arg\left(z\right)\right) + i \sin\left(k\arg\left(z\right)\right)\right)
        \end{equation}
        \new{
        This Lemma shows how we expand the complex-valued momentum buffer $\momVal$ into its Cartesian components as in Equation~\ref{cm_eq:cartesian_complex_update}.
        }
        \begin{lemma}\label{cm_lemma:complex_momentum_cartesian_buffer}
            \begin{align*}
                \momVal^{j\!+\!1} &= \momCoeff \momVal^{j} - \bothGrad^j \iff\\
                \Re\left(\momVal^{j\!+\!1}\right) &= \Re\left(\momCoeff\right) \Re\left(\momVal^j\right) - \Im\left(\momCoeff\right) \Im\left(\momVal^j\right)  - \Re\left(\bothGrad^j\right), \Im\left(\momVal^{j\!+\!1}\right) = \Im\left(\momCoeff\right) \Re\left(\momVal^{j}\right) + \Re\left(\momCoeff\right) \Im\left(\momVal^{j}\right) - \Im\left(\bothGrad^j\right)
            \end{align*}
        \end{lemma}
        \begin{proof}
            \begin{align*}
                \momVal^{j\!+\!1} &= \momCoeff \momVal^{j} - \bothGrad^j\\
                \iff \momVal^{j\!+\!1} &= \left(\Re\left(\momCoeff\right) + i\Im\left(\momCoeff\right) \right) \left( \Re\left(\momVal^{j}\right) + i\Im\left(\momVal^{j}\right) \right) - \left( \Re\left(\bothGrad^j\right) + i\Im\left(\bothGrad^j\right) \right)\\
                \iff \momVal^{j\!+\!1} &= \left(\Re\left(\momCoeff\right) \Re\left(\momVal^j\right) - \Im\left(\momCoeff\right) \Im\left(\momVal^j\right) \right) +\\
                &\hspace{0.25\linewidth}i \left(\Im\left(\momCoeff\right) \Re\left(\momVal^{j}\right) + \Re\left(\momCoeff\right) \Im\left(\momVal^{j}\right) \right) - \left( \Re\left(\bothGrad^j\right) + i\Im\left(\bothGrad^j\right) \right)\\
                \iff \momVal^{j\!+\!1} &= \left(\Re\left(\momCoeff\right) \Re\left(\momVal^j\right) - \Im\left(\momCoeff\right) \Im\left(\momVal^j\right)  - \Re\left(\bothGrad^j\right) \right) +\\
                &\hspace{0.25\linewidth}i \left(\Im\left(\momCoeff\right) \Re\left(\momVal^{j}\right) + \Re\left(\momCoeff\right) \Im\left(\momVal^{j}\right) - \Im\left(\bothGrad^j\right)\right)\\
                \iff \Re\left(\momVal^{j\!+\!1}\right) \!&=\! \Re\left(\momCoeff\right) \Re\left(\momVal^j\right) \!-\! \Im\left(\momCoeff\right) \Im\left(\momVal^j\right)  \!-\! \Re\left(\bothGrad^j\right), \\
                \Im\left(\momVal^{j\!+\!1}\right) \!&=\! \Im\left(\momCoeff\right) \Re\left(\momVal^{j}\right) \!+\! \Re\left(\momCoeff\right) \Im\left(\momVal^{j}\right) \!-\! \Im\left(\bothGrad^j\right)\\
            \end{align*}
        \end{proof}
        
        \noindent
        We further assume $\Im\left(\bothGrad^j\right)$ is $\num{0}$ - i.e., our gradients are real-valued.
        \new{
        This Lemma decomposes the joint-parameters $\bothParam$ at the next iterate as a linear combination of the joint-parameters, joint-gradient, and Cartesian components of the momentum-buffer at the current iterate as in Equation~\ref{cm_eq:parameter_dynamics}.
        }
        \begin{lemma}\label{cm_lemma:complex_momentum_cartesian_parameters}
            \begin{align*}
                \bothParam^{j\!+\!1} = \bothParam^j + \Re\left(\lr \momVal^{j\!+\!1}\right)
                \iff \bothParam^{j\!+\!1} = \bothParam^j - \Re\left(\lr\right)\bothGrad^j + \Re\left(\lr \momCoeff\right) \Re\left(\momVal^j\right) - \Im\left(\lr \momCoeff\right)\Im\left(\momVal^j\right)
            \end{align*}
        \end{lemma}
        \begin{proof}
            \begin{align*}
                & \Re\left(\lr \momVal^{j\!+\!1}\right)\\
                =& \left(\Re\left(\lr\right) \Re\left(\momVal^{j\!+\!1}\right) - \Im\left(\lr\right) \Im\left(\momVal^{j\!+\!1}\right)\right)\\
                =& \left(\Re\left(\lr\right) \left(\Re\left(\momCoeff\right) \Re\left(\momVal^j\right) - \Im\left(\momCoeff\right) \Im\left(\momVal^j\right) - \bothGrad^j\right) - \Im\left(\lr\right) \left( \Im\left(\momCoeff\right) \Re\left(\momVal^{j}\right) + \Re\left(\momCoeff\right) \Im\left(\momVal^{j}\right) \right)\right)\\
                =& -\Re\left(\lr\right)\bothGrad^j +  \left(\Re\left(\lr\right) \left(\Re\left(\momCoeff\right) \Re\left(\momVal^j\right) - \Im\left(\momCoeff\right) \Im\left(\momVal^j\right) \right) - \Im\left(\lr\right) \left( \Im\left(\momCoeff\right) \Re\left(\momVal^{j}\right) + \Re\left(\momCoeff\right) \Im\left(\momVal^{j}\right) \right)\right)\\
                =& -\Re\left(\lr\right)\bothGrad^j +  \left(\Re\left(\lr\right)\Re\left(\momCoeff\right) - \Im\left(\lr\right)\Im\left(\momCoeff\right) \right)\Re\left(\momVal^j\right) - \left(\Re\left(\lr\right) \Im\left(\momCoeff\right) + \Im\left(\lr\right)\Re\left(\momCoeff\right)\right)\Im\left(\momVal^j\right)\\
                =& -\Re\left(\lr\right)\bothGrad^j +  \Re\left(\lr \momCoeff\right) \Re\left(\momVal^j\right) - \Im\left(\lr \momCoeff\right)\Im\left(\momVal^j\right)
            \end{align*}
            Thus,
            \begin{align*}
                \bothParam^{j\!+\!1} = \bothParam^j + \Re\left(\lr \momVal^{j\!+\!1}\right)
                \iff \bothParam^{j\!+\!1} = \bothParam^j - \Re\left(\lr\right)\bothGrad^j + \Re\left(\lr \momCoeff\right) \Re\left(\momVal^j\right) - \Im\left(\lr \momCoeff\right)\Im\left(\momVal^j\right)
            \end{align*}
        \end{proof}
        
        \subsection{Theorem~\ref{cm_thm:theorem} Proof Sketch}\label{cm_pf:thm1}
            \mainThm*
            \begin{proof}
                We reproduce the proof quadratic games, which is simple case of the well-known method from \cite{polyak1964some} for analyzing the convergence of iterative methods.
                \cite{bertsekas2008nonlinear} generalizes this result from quadratic games to when we are sufficiently close to any stationary point.
                
                For quadratic games, we have that $\bothGrad^j = \left(\nabla_{\bothParam} \bothGrad\right)^\transpose \bothParam^j$.
                By Lemma~\ref{cm_lemma:complex_momentum_cartesian_buffer} and \ref{cm_lemma:complex_momentum_cartesian_parameters} we have:
            	\begin{equation}
            	    \begin{pmatrix}
            	        \Re\left(\momVal^{j\!+\!1}\right)\\
            	        \Im\left(\momVal^{j\!+\!1}\right)\\
            	        \bothParam^{j\!+\!1}
            	    \end{pmatrix}
            	    = \dynamics 
            	    \begin{pmatrix}
            	        \Re\left(\momVal^{j}\right)\\
            	        \Im\left(\momVal^{j}\right)\\
            	        \bothParam^{j}
            	    \end{pmatrix}
            	\end{equation}
            	By telescoping the recurrence for the $j^{\textnormal{th}}$ augmented parameters: 
            	\begin{equation}
            	    \begin{pmatrix}
            	        \Re\left(\momVal^{j}\right)\\
            	        \Im\left(\momVal^{j}\right)\\
            	        \bothParam^{j}
            	    \end{pmatrix}
            	    = \dynamics^j
            	    \begin{pmatrix}
            	        \Re\left(\momVal^{0}\right)\\
            	        \Im\left(\momVal^{0}\right)\\
            	        \bothParam^{0}
            	    \end{pmatrix}
            	\end{equation}
            	We can compare $\momVal^j$ with the value it converges to $\momVal^*$ which exists if $\dynamics$ is contractive.
            	We do the same with $\bothParam$.
            	Because $\momVal^* = \dynamics \momVal^* = \dynamics^j \momVal^*$:
            	\begin{equation}
            	    \begin{pmatrix}
            	        \Re\left(\momVal^{j}\right) - \Re\left(\momVal^{*}\right)\\
            	        \Im\left(\momVal^{j}\right) - \Im\left(\momVal^{*}\right)\\
            	        \bothParam^{j} - \bothParam^{*}
            	    \end{pmatrix}
            	    = \dynamics^j
            	    \begin{pmatrix}
            	        \Re\left(\momVal^{0}\right)  - \Re\left(\momVal^{*}\right)\\
            	        \Im\left(\momVal^{0}\right)  - \Im\left(\momVal^{*}\right)\\
            	        \bothParam^{0} - \bothParam^{*}
            	    \end{pmatrix}
            	\end{equation}
            	By taking norms:
            	\begin{align}
            	    \left\|
                	    \begin{pmatrix}
                	        \Re\left(\momVal^{j}\right) - \Re\left(\momVal^{*}\right)\\
                	        \Im\left(\momVal^{j}\right) - \Im\left(\momVal^{*}\right)\\
                	        \bothParam^{j} - \bothParam^{*}
                	    \end{pmatrix}
            	    \right\|_2
            	    &= 
            	    \left\|
                	    \dynamics^j
                	    \begin{pmatrix}
                	        \Re\left(\momVal^{0}\right)  - \Re\left(\momVal^{*}\right)\\
                	        \Im\left(\momVal^{0}\right)  - \Im\left(\momVal^{*}\right)\\
                	        \bothParam^{0} - \bothParam^{*}
                	    \end{pmatrix}
            	    \right\|_2
            	    \\
            	    \implies
            	    \left\|
                	    \begin{pmatrix}
                	        \Re\left(\momVal^{j}\right) - \Re\left(\momVal^{*}\right)\\
                	        \Im\left(\momVal^{j}\right) - \Im\left(\momVal^{*}\right)\\
                	        \bothParam^{j} - \bothParam^{*}
                	    \end{pmatrix}
            	    \right\|_2
            	    &\leq
            	    \left\|
                	    \dynamics^j
                	\right\|_2
                	\left\|
                	    \begin{pmatrix}
                	        \Re\left(\momVal^{0}\right)  - \Re\left(\momVal^{*}\right)\\
                	        \Im\left(\momVal^{0}\right)  - \Im\left(\momVal^{*}\right)\\
                	        \bothParam^{0} - \bothParam^{*}
                	    \end{pmatrix}
            	    \right\|_2
            	\end{align}
            	With Lemma 11 from \citet{foucart2012}, we have there exists a matrix norm $\forall \epsilon > 0$ such that:
            	\begin{equation}\label{cm_eq:foucart}
            	    \| \dynamics^j \| \leq \left(\spectralRadius \left(\dynamics \right) + \epsilon\right)^j
            	\end{equation}
            	We have a finite-dimensional space norm equivalence.
            	So, for all norms $\| \cdot \|$, $\exists C\! \geq\! B\! >\! 0$ such that:
            	\begin{equation}\label{cm_eq:norm_eq}
            	    B \| \dynamics^j \| \leq \| \dynamics^j \|_2 \leq C \| \dynamics^j \|
            	\end{equation}
            	Combining Equations~\ref{cm_eq:foucart}) and \ref{cm_eq:norm_eq} we have:
            	\begin{equation}
            	    \left\|
                	    \begin{pmatrix}
                	        \Re\left(\momVal^{j}\right) - \Re\left(\momVal^{*}\right)\\
                	        \Im\left(\momVal^{j}\right) - \Im\left(\momVal^{*}\right)\\
                	        \bothParam^{j} - \bothParam^{*}
                	    \end{pmatrix}
            	    \right\|_2
            	    \leq
            	    C \left(\spectralRadius \left(\dynamics \right) + \epsilon\right)^j
                	\left\|
                	    \begin{pmatrix}
                	        \Re\left(\momVal^{0}\right)  - \Re\left(\momVal^{*}\right)\\
                	        \Im\left(\momVal^{0}\right)  - \Im\left(\momVal^{*}\right)\\
                	        \bothParam^{0} - \bothParam^{*}
                	    \end{pmatrix}
            	    \right\|_2
            	\end{equation}
            	So, we have:
            	\begin{equation}
            	    \left\|
                	    \begin{pmatrix}
                	        \Re\left(\momVal^{j}\right) - \Re\left(\momVal^{*}\right)\\
                	        \Im\left(\momVal^{j}\right) - \Im\left(\momVal^{*}\right)\\
                	        \bothParam^{j} - \bothParam^{*}
                	    \end{pmatrix}
            	    \right\|_2
            	    = \mathcal{O}\left(\left(\spectralRadius\left(\dynamics\right) + \epsilon\right)^j\right)
            	\end{equation}
            	Thus, we converge linearly with a rate of $\mathcal{O}\left(\spectralRadius\left(\dynamics\right) + \epsilon\right)$.
            \end{proof}
        \subsection{Characterizing the Augmented Dynamics Eigenvalues}\label{cm_sec:poly}
            Here, we present polynomials whose roots are the eigenvalues of the Jacobian of our augmented dynamics $\spectrum\left(\dynamics\right)$, given the eigenvalues of the Jacobian of the joint-gradient vector field $\spectrum\left(\nabla_{\bothParam} \bothGrad\right)$.
            We use a similar decomposition as \citet{gidel2018negative}.
            
            We expand $\nabla_{\bothParam} \bothGrad = \boldsymbol{P} \boldsymbol{T} \boldsymbol{P}^{-1}$ where $\boldsymbol{T}$ is upper-triangular and $\eigval_i$ is an eigenvalue of $\nabla_{\bothParam} \bothGrad$.
            \begin{equation}
                \boldsymbol{T} = \begin{bmatrix}
                    \eigval_1 & * & \dots & *\\
                    \num{0}  & \dots & \dots & \dots\\
                    \dots & \dots & \dots & *\\
                    \num{0} & \dots & \num{0} & \eigval_{\bothDim}\\
                    \end{bmatrix}
            \end{equation}
            We then break up into components for each eigenvalue, giving us submatrices $\mathbf{R}_k \in \mathbb{C}^{\num{3} \times \num{3}}$:
            \begin{equation}\label{cm_eq:R-decomp}
        	    \dynamics_{k}  \defeq \begin{bmatrix}
                    \Re\left(\momCoeff\right) & -\Im\left(\momCoeff\right) & -\eigval_k\\
                    \Im\left(\momCoeff\right) & \Re\left(\momCoeff\right) & 0\\
                    \Re\left(\lr \momCoeff\right) & -\Im\left(\lr \momCoeff\right) & 1 - \Re\left(\lr\right)\eigval_k\\
                    \end{bmatrix}
        	\end{equation}
        	We can get the characteristic polynomial of $\dynamics_k$ with the following Mathematica command, where we use substitute the symbols $r + i u = \eigval_k$, $a = \Re\left(\momCoeff\right)$, $b = \Im\left(\momCoeff\right)$, $c = \Re\left(\lr\right)$, and $d = \Im\left(\lr\right)$.
        	
        	\texttt{CharacteristicPolynomial[\{\{a, -b, -(r + u I)\}, \{b, a, 0\}, \{a c - b d, -(b c + a d), 1 - c (r + u I)\}\}, x]}
        	
        	The command gives us the polynomial associated with the eigenvalue $\eigval_k = r + i u$:
        	\begin{equation}\label{cm_eq:R-poly}
        	    p_k(x) = -a^2 x + a^2 + a c r x + i a c u x + 2 a x^2 - 2 a x - b^2 x + b^2 + b d r x + i b d u x - c r x^2 - i c u x^2 - x^3 + x^2
        	\end{equation}
        	Consider the case where $\eigval_k$ is imaginary -- i.e., $r = 0$ -- which is true in all purely adversarial and bilinear zero-sum games.
        	Then (\ref{cm_eq:R-poly}) simplifies to:
        	\begin{equation}\label{cm_eq:R-poly-imEig}
        	    p_k(x) = -a^2 x + a^2 + i a c u x + 2 a x^2 - 2 a x - b^2 x + b^2 + i b d u x - i c u x^2 - x^3 + x^2
        	\end{equation}
                \newpage
        	Our complex $\eigval_k$ come in conjugate pairs where $\eigval_k = u_k i$ and $\bar{\eigval}_k = - u_k i$.
        	Equation~\ref{cm_eq:R-poly-imEig} has the same roots for $\eigval_k$ and $\bar{\eigval}_k$, which can be verified by writing the roots with the cubic formula.
        	This corresponds to spiraling around the solution in either a clockwise or counterclockwise direction.
        	Thus, we restrict ourselves to analyzing $\eigval_k$ where $u_k$ is positive without loss of generality.
        	
        	If we make the step size $\lr$ real -- i.e., $d = 0$ -- then Equation~\ref{cm_eq:R-poly-imEig} simplifies to:
        	\begin{equation}\label{cm_eq:R-poly-imEig-realLr}
        	    p_k(x) = x \left(-a^2 + i a c u - 2 a - b^2\right) + a^2 + x^2 \left(2 a - i c u + 1\right) + b^2 - x^3
        	\end{equation}
        	Using a heuristic from single-objective optimization, we look at making step size proportional to the inverse of the magnitude of eigenvalue $k$ -- i.e., $\lr_k = \frac{\lr'}{|\eigval_k|} = \frac{\lr'}{u_k}$.
        	With this, (\ref{cm_eq:R-poly-imEig-realLr}) simplifies to:
        	\begin{equation}\label{cm_eq:R-poly-imEig-realLr-prop}
        	    p_k(x) = x \left(-a^2 + i a \lr' - 2 a - b^2\right) + a^2 + x^2 \left(2 a - i \lr' + 1\right) + b^2 - x^3
        	\end{equation}
        	Notably, in Equation~\ref{cm_eq:R-poly-imEig-realLr-prop} there is no dependence on the components of imaginary eigenvalue $\eigval_k = r + i u = \num{0} + i u$, by selecting a $\lr$ that is proportional to the eigenvalues inverse magnitude.
        	We can further simplify with $a^2 + b^2 = |\momCoeff|^2$:
        	\begin{equation}\label{cm_eq:R-poly-imEig-realLr-prop-s1}
        	    p_k(x) = x \left(\Re\left(\momCoeff\right) \left(i \lr' - 2\right) - |\momCoeff|^2\right) + x^2 \left(2 \Re\left(\momCoeff\right) - i \lr' + 1\right) + |\momCoeff|^2 - x^3
        	\end{equation}
        	We could expand this in polar form for $\momCoeff$ by noting $\Re(\momCoeff) = |\momCoeff| \cos(\arg(\momCoeff))$:
        	\begin{equation}\label{cm_eq:R-poly-imEig-realLr-prop-s2}
        	    p_k(x) = x \left(|\momCoeff| \cos\left(\arg\left(\momCoeff\right)\right)\left(i  \lr' - 2\right) - |\momCoeff|^2\right) + x^2 \left(2 |\momCoeff| \cos\left(\arg\left(\momCoeff\right)\right) - i \lr' + 1\right) + |\momCoeff|^2 - x^3
        	\end{equation}
        	We can simplify further by considering an imaginary $\momCoeff$ -- i.e., $\Re\left(\momCoeff\right) = 0$ or $\cos\left(\arg\left(\momCoeff\right)\right) = 0$:
        	\begin{equation}\label{cm_eq:R-poly-imEig-realLr-prop-s3}
        	    p_k(x) = |\momCoeff|^2 - x |\momCoeff|^2 - x^2 \left(i \lr' - 1\right)  - x^3
        	\end{equation}

        	
        	
        	The roots of these polynomials can be trivially evaluated numerically or symbolically with the by plugging in $\momCoeff, \lr,$ and $\eigval_k$ then using the cubic formula.
        	This section can be easily modified for the eigenvalues of the augmented dynamics for variants of complex momentum by defining the appropriate $\dynamics$ and modifying the Mathematica command to get the characteristic polynomial for each component, which can be evaluated if it is a sufficiently low degree using known formulas.
        	
        \subsection{Convergence Bounds}\label{cm_sec:thm2}
                \resultExist*
                \begin{proof}
                    \new{
                    Note that Theorem~\ref{cm_thm:theorem} bounds the convergence rate of Algorithm~\ref{cm_alg:simultaneous_complex} by $\spectrum(\dynamics)$.
                    Also, Equation~\ref{cm_eq:R-poly-imEig-realLr} gives a formula for 3 eigenvalues in $\spectrum\left(\dynamics\right)$ given $\lr, \momCoeff,$ and an eigenvalue $\eigval \in \spectrum\left(\nabla_{\bothParam}\bothGrad\right)$.
                    The formula outputs a cubic polynomial whose roots are eigenvalues of $\spectrum(\dynamics)$, which can be trivially evaluated with the cubic formula.
                    
                    We denote the $k^{\textnormal{th}}$ eigenspace of $\spectrum\left(\nabla_{\bothParam}\bothGrad\right)$ with eigenvalue $\eigval_k = i c_k$ and $|c_1| \leq \dots \leq |c_n|$, because bilinear zero-sum games have purely imaginary eigenvalues due to $\nabla_{\bothParam}\bothGrad$ being antisymmetric.
                    Eigenvalues come in a conjugate pairs, where $\bar{\eigval_k} = i \left(-c_k\right)$
                    
                    If we select momentum coefficient $\momCoeff = |\momCoeff| \exp\left(i \arg\left(\momCoeff\right)\right)$ and step size $\lr_k = \frac{\lr_k'}{|c_k|}$, and use that $ \eigval \in \spectrum\left(\nabla_{\bothParam}\bothGrad\right)$ are imaginary, then -- as shown in Appendix Section~\ref{cm_sec:poly} -- (\ref{cm_eq:R-poly-imEig-realLr}) simplifies to:
                	\begin{equation}\label{cm_eq:R-poly-imEig-realLr-prop-pf}
                	    p_k\left(x\right) = x \left(|\momCoeff| \cos\left(\arg\left(\momCoeff\right)\right)\left(i  \lr_k' - 2\right) - |\momCoeff|^2\right) + x^2 \left(2 |\momCoeff| \cos\left(\arg\left(\momCoeff\right)\right) - i \lr_k' + 1\right) + |\momCoeff|^2 - x^3
                	\end{equation}
                	So, with these parameter selections, the convergence rate of Algorithm~\ref{cm_alg:simultaneous_complex} in the $k^{\textnormal{th}}$ eigenspace is bounded by the largest root of Equation~\ref{cm_eq:R-poly-imEig-realLr-prop-pf}.
                	
                	Note that if we find a separate $\alpha'_k$ that converges in each eigenspace $k$, then selected the smallest $\alpha'_k$ converges in every eigenspace because the convergence rate in eigenspace $k$ is a convex function of $\alpha$ that equals $1$ when $\alpha = 0$ and is minimized when $\alpha > 0$.
                	
                	First, consider $\arg\left(\momCoeff\right) = \pi - \epsilon$, where $\epsilon = \frac{\pi}{16}$.
                	We select $\lr_k' = \almostNegMomentumLR$ (equivalently, $\lr_k = \frac{\almostNegMomentumLR}{|c_k|}$) and $| \momCoeff | = \almostNegMomentumMag$ via grid search.
                    Using the cubic formula on the associated $p(x)$ from Equation~\ref{cm_eq:R-poly-imEig-realLr-prop-pf}, the maximum magnitude root has size $ \approx \almostNegRate < 1$, so this selection converges in the $k^{\textnormal{th}}$ eigenspace.
                    So, selecting:
                    \begin{align}
                        \hat{\lr} &\leq \min_k \lr_k\\
                        &= \min_k \frac{\almostNegMomentumLR}{c_k}\\
                        &= \frac{\almostNegMomentumLR}{\max_k c_k}\\
                        &= \frac{\almostNegMomentumLR}{\|\nabla_{\bothParam}\bothGrad\|_2}
                    \end{align}
                     with $\momCoeff = \almostNegMomentumMag \exp\left(i \left(\pi - \epsilon\right)\right)$ will converge in each eigenspace.
                    
                    Now, consider $\arg\left(\momCoeff\right) = \epsilon = \frac{\pi}{16}$ with $\lr_k' = \almostPosMomentumLR$ and $| \momCoeff | = \almostPosMomentumMag$.
                    Using the cubic formula on the associated $p(x)$ from Equation~\ref{cm_eq:R-poly-imEig-realLr-prop-pf} the maximum magnitude root has size $ \approx \almostPosRate < 1$, so this selection converges in the $k^{\textnormal{th}}$ eigenspace.
                    So, selecting:
                    \begin{align}
                        \hat{\lr} &\leq \min_k \lr_k\\
                        &= \min_k \frac{\almostPosMomentumLR}{c_k}\\
                        &= \frac{\almostPosMomentumLR}{\max_k c_k}\\
                        &= \frac{\almostPosMomentumLR}{\|\nabla_{\bothParam}\bothGrad\|_2}
                    \end{align}
                    with $\momCoeff = \almostPosMomentumMag \exp\left(i \epsilon\right)$ will converge in each eigenspace.
                    
                    Thus, for any of the $\arg(\momCoeff)$ choices, we can select $\hat{\lr}, |\momCoeff|$ that converges in every eigenspace and thus converges.

                    }
                \end{proof}
                In the preceding proof, our prescribed $\hat{\lr}$ depends on the largest norm eigenvalue of $\spectrum\left(\nabla_{\bothParam}\bothGrad\right)$, due to our selections of $\hat{\lr} \,\,\propto\,\, \frac{1}{\|\nabla_{\bothParam}\bothGrad\|_2}$.
                In practice, we may not have access to the largest norm eigenvalue of $\spectrum\left(\nabla_{\bothParam}\bothGrad\right)$.
                Nevertheless, this shows that a parameter selection converges, even if it may be difficult to find.
                Often, in convex optimization we describe the choices of $\lr, \momCoeff$ in terms of the largest and smallest norm eigenvalues of $\spectrum(\nabla_{\bothParam}\bothGrad)$ (i.e., the Hessian of the loss)~\citep{boyd2004convex}.

    \begin{figure*}
                \centering
                \begin{subfigure}{.16\linewidth}
                    \begin{tikzpicture}[
                            > = stealth, 
                            shorten > = 1pt, 
                            auto,
                            node distance = 1.8cm, 
                            semithick 
                        ]
                        \tikzstyle{state}=[
                            draw = black,
                            thick,
                            fill = white,
                            minimum size = 4mm
                        ]
                
                        \node[state] (gNext) {gradient};
                        \node[state] (more) [yshift=.6cm, below of=gNext] {$\momVal$};
                        \node[state] (wNext) [yshift=.6cm, below of=more] {update};
                        
                        \path[->]
                            (gNext) 
                                    edge                node                [xshift=-0cm,]{}              (more)
                            (more)    
                                    edge                node                [xshift=0cm, yshift=0cm]{$\lr$}          (wNext)
                                    edge[loop left]    node                {$\momCoeff$}                             (re)
                          ;
                    \end{tikzpicture}
                    \caption[Classical momentum]{
                        Classical
                        \newline
                        \citep{polyak1964some}
                    }
                \end{subfigure}
                \begin{subfigure}{.27\linewidth}
                    \begin{tikzpicture}[
                            > = stealth, 
                            shorten > = 1pt, 
                            auto,
                            node distance = 1.8cm, 
                            semithick 
                        ]
                        \tikzstyle{state}=[
                            draw = black,
                            thick,
                            fill = white,
                            minimum size = 4mm
                        ]
                
                        \node[state] (gNext) {gradient};
                        \node[state] (re) [below left of=gNext] {$\momVal_{(1)}$};
                        \node[state] (more) [xshift=-.55cm, right of=re] {\dots};
                        \node[state] (im) [below right of=gNext] {$\momVal_{(n)}$};
                        \node[state] (wNext) [below right of=re] {update};
                        
                        \path[->]
                            (re)    
                                    edge                node                [xshift=-.7cm, yshift=-.45cm]{$\lr_{(1)}$}          (wNext)
                                    edge[loop above]    node                {$\momCoeff_{(1)}$}                             (re)
                            (im)    
                                    edge                node                {$\lr_{(n)}$}                                       (wNext)
                                    edge[loop above]    node                {$\momCoeff_{(n)}$}                             (im)
                            (gNext) edge                node                [xshift=-.6cm, yshift=.65cm]{} (re)
                                    edge                node                [xshift=-.2cm,]{}              (im)
                                    edge                node                [xshift=-0cm,]{}              (more)
                            (more)    
                                    edge                node                [xshift=0cm, yshift=-.25cm]{}          (wNext)
                                    edge[loop right]    node                [xshift=0cm]{}                             (re)
                          ;
                    \end{tikzpicture}
                    \caption[Aggregated momentum]{
                        Aggregated
                        \newline
                        \citep{lucas2018aggregated}
                    }
                    \label{cm_fig:aggmo_comp_graph}
                \end{subfigure}
                \begin{subfigure}{.27\linewidth}
                    \begin{tikzpicture}[
                            > = stealth, 
                            shorten > = 1pt, 
                            auto,
                            node distance = 1.8cm, 
                            semithick 
                        ]
                        \tikzstyle{state}=[
                            draw = black,
                            thick,
                            fill = white,
                            minimum size = 4mm
                        ]
                
                        \node[state] (gNext) {gradient};
                        \node[state] (re) [below left of=gNext] {$\momVal_{(1)}$};
                        \node[state] (im) [below right of=gNext] {$\momVal_{(2)}$};
                        \node[state] (wNext) [below right of=re] {update};
                        
                        \path[->]
                            (re)    edge[bend left]     node                [yshift=-.1cm]{$\momCoeff_{(1,2)}$}                             (im)
                                    edge                node                [xshift=-.7cm, yshift=-.45cm]{$\lr_{(1)}$}          (wNext)
                                    edge[loop above]    node                {$\momCoeff_{(1,1)}$}                             (re)
                            (im)    edge[bend left]     node                [xshift=0cm, yshift=.65cm]{$\momCoeff_{(2,1)}$}   (re)
                                    edge                node                {$\lr_{(2)}$}                                       (wNext)
                                    edge[loop above]    node                {$\momCoeff_{(2,2)}$}                             (im)
                            (gNext) edge                node                [xshift=-.6cm, yshift=.65cm]{} (re)
                                    edge                node                [xshift=-.2cm,]{}              (im)
                          ;
                    \end{tikzpicture}
                    \caption[Recurrently linked momentum]{
                        Recurrently linked
                        \newline
                        (new)
                    }
                    \label{cm_fig:recurrent_comp_graph}
                \end{subfigure}
                \begin{subfigure}{.27\linewidth}
                    \begin{tikzpicture}[
                            > = stealth, 
                            shorten > = 1pt, 
                            auto,
                            node distance = 1.8cm, 
                            semithick 
                        ]
                        \tikzstyle{state}=[
                            draw = black,
                            thick,
                            fill = white,
                            minimum size = 4mm
                        ]
                
                        \node[state] (gNext) {gradient};
                        \node[state] (re) [below left of=gNext] {$\Re(\momVal$)};
                        \node[state] (im) [below right of=gNext] {$\Im(\momVal$)};
                        \node[state] (wNext) [below right of=re] {update};
                        
                        \path[->]
                            (re)    edge[bend left]     node                [yshift=-.1cm]{${\color{red}\Im(\momCoeff)}$}                 (im)
                                    edge                node                [xshift=-.5cm, yshift=-.45cm]{$\lr$}     (wNext)
                                    edge[loop above]    node                {{\color{blue}$\Re(\momCoeff)$}}                (re)
                            (im)    edge[bend left]     node                 [xshift=0cm, yshift=.65cm]{$-{\color{red}\Im(\momCoeff)}$}  (re)
                                    edge[loop above]    node                {{\color{blue}$\Re(\momCoeff)$}}                (im)
                            (gNext) edge                node                {}                                              (re)
                          ;
                    \end{tikzpicture}
                    \caption[Complex momentum]{
                        Complex
                        \newline
                        (ours)
                    }
                    \label{cm_fig:cm_comp_graph_complex}
                \end{subfigure}
                \caption[Compute graphs for momentum variants]{
                        We show computational diagrams for momentum variants simultaneously updating all players parameters, which update the momentum buffers $\momVal$ at iteration $j\!+\!1$ with coefficient $\momCoeff$ via $\momVal^{j\!+\!1} \!=\! \left(\momCoeff \momVal^{j} - \textnormal{gradient}\right)$.
                        Our parameter update is a linear combination of momentum buffers weighted by step sizes $\lr$.
                        (a) Classical momentum~\citep{polyak1964some, sutskever2013importance}, with a single buffer and coefficient $\momCoeff \in [0, 1)$.
                        (b) Aggregated momentum~\citep{lucas2018aggregated} adds multiple buffers with different coefficients.
                        (c) Recurrently linked momentum adds cross-buffer coefficients and updates the buffers with $\momVal_{(k)}^{j+1} \!=\! \left(\sum_l \momCoeff_{(l, k)}\momVal_{(l)}^{j} - \textnormal{gradient}\right)$.
                        We allow $\momCoeff_{(l, k)}$ to be negative as negative momentum~\citep{gidel2018negative} for solutions with simultaneous updates in adversarial games.
                        (d) Complex momentum is a special case of recurrently linked momentum with two buffers and \smash{${\color{blue}\momCoeff_{(1,1)}}\!=\!{\color{blue}\momCoeff_{(2,2)}} \!=\! {\color{blue}\Re(\momCoeff)}$, ${\color{red}\momCoeff_{(1,2)}} \!=\! -{\color{red}\momCoeff_{(2,1)}} \!=\! {\color{red}\Im(\momCoeff)}$}.
                        Analyzing other recurrently linked momentum setups is an open problem.
                }\label{cm_fig:cm_comp_graph}
            \end{figure*}
    \section{Algorithms}
        Here, we include additional algorithms that may be useful to some readers.
        Algorithm~\ref{cm_alg:aggmo} show aggregated momentum~\citep{lucas2018aggregated}.
        Algorithm~\ref{cm_alg:recurrent_momentum} shows the recurrently linked momentum that generalizes and unifies aggregated momentum with negative momentum~\citep{gidel2018negative}.
        Algorithm~\ref{cm_alg:alternating_complex} shows our algorithm with alternating updates, which we use for training GANs.
        Algorithm~\ref{cm_alg:simultaneous_expanded} shows our method with all real-valued objects if one wants to implement complex momentum in a library that does not support complex arithmetic.
        \begin{figure}[h!]
            \centering
                \begin{algorithm}[H]
                    \caption{Aggregated momentum} \label{cm_alg:aggmo}
                    \begin{algorithmic}[1]
                        \State Select number of buffers $K \in \mathbb{N}$
                        \State Select $\momCoeff_{(k)} \in [0, 1)$ for $k = 1 \dots K$
                        \State Select $\lr_{(k)} \in \mathbb{R}^{+}$ for $k = 1 \dots K$
                        \State Initialize $\momVal_{(k)}^{0}$ for $k = 1 \dots K$
                        \For{$j = \num{1} \dots N$}
                            \For{$k = \num{1} \dots K$}
                                \State $\momVal_{(k)}^{j+\num{1}} = \momCoeff_{(k)} \momVal_{(k)}^{j} - \bothGrad^{j}$
                            \EndFor
                            \State $\bothParam^{j+\num{1}} = \bothParam^j + \sum_{k=1}^{K}  \lr_{(k)} \momVal ^{j+1}_{(k)}$
                        \EndFor
                        \State \Return $\bothParam_N$
                    \end{algorithmic}
                \end{algorithm}
        \end{figure}
        \begin{figure}
            \vspace{-0.02\textheight}
                \begin{algorithm}[H]
                    \caption{Recurrently linked momentum} \label{cm_alg:recurrent_momentum}
                    \begin{algorithmic}[1]
                        \State Select number of buffers $K \in \mathbb{N}$
                        \State Select $\momCoeff_{(l,k)} \in \mathbb{R}$ for $l=1 \dots K$ and $k = 1 \dots K$
                        \State Select $\lr_{(k)} \in \mathbb{R}^{+}$ for $k = 1 \dots K$
                        \State Initialize $\momVal_{(k)}^{0}$ for $k = 1 \dots K$
                        \For{$j = \num{1} \dots N$}
                            \For{$k = \num{1} \dots K$}
                                \State $\momVal_{(k)}^{j+1} = \sum_l \momCoeff_{(l, k)}\momVal_{(l)}^{j} - \bothGrad^{j}$
                            \EndFor
                            \State $\bothParam^{j+\num{1}} = \bothParam^j + \sum_{k=1}^{K} \lr_{(k)} \momVal ^{j+1}_{(k)}$
                        \EndFor
                        \State \Return $\bothParam_N$
                    \end{algorithmic}
                \end{algorithm}
        \end{figure}
        \begin{figure}[h!]
            \vspace{-0.03\textheight}
            \centering
                \begin{algorithm}[H]
                    \caption[Complex momentum]{(\ref{cm_eq:alt_update}) Momentum} \label{cm_alg:alternating_complex}
                    \begin{algorithmic}[1]
                        \State Select $\momCoeff \in \mathbb{C}, \lr \in \mathbb{R}^{+}$
                        \State Initialize $\momVal_{\outSymbol}^{0}, \momVal_{\inSymbol}^{0}$
                        \For{$j = \num{1} \dots N$}
                            \State $\momVal_{\outSymbol}^{j+\num{1}} = \momCoeff \momVal_{\outSymbol}^{j} - \outGrad^j$
                            \State $\outParam^{j+\num{1}} = \outParam^j + \Re\left(\lr \momVal^{j+\num{1}}_{\outSymbol}\right)$
                            \State $\momVal_{\inSymbol}^{j+\num{1}} = \momCoeff \momVal_{\inSymbol}^{j} - \inGrad\left(\outParam^{j+1}, \inParam^{j}\right)$
                            \State $\inParam^{j+\num{1}} = \inParam^j + \Re\left(\lr \momVal^{j+1}_{\inSymbol}\right)$
                        \EndFor
                        \State \Return $\bothParam_N$
                    \end{algorithmic}
                \end{algorithm}
            %
            %
            %
        \end{figure}
        \begin{figure}[h!]
            \vspace{-0.03\textheight}
                \centering
                \begin{algorithm}[H]
                    \caption[Complex momentum with only real values]{(\ref{cm_eq:simul_update}) Complex Momentum - $\mathbb{R}$ valued} \label{cm_alg:simultaneous_expanded}
                    \begin{algorithmic}[1]
                        \State Select $\Re(\momCoeff), \Im(\momCoeff), \Re(\lr), \Im(\lr) \in \mathbb{R}$
                        \State Select $\Re(\momCoeff), \Im(\momCoeff), \Re(\lr), \Im(\lr) \in \mathbb{R}$
                        \State Initialize $\Re(\momVal^{0}), \Im(\momVal^{0})$
                        \For{$j = 1 \dots N$}
                            \State $\Re\left(\momVal^{j+1}\right) = \Re\left(\momCoeff\right) \Re\left(\momVal^j\right) - \Im\left(\momCoeff\right)\Im\left(\momVal^j\right) - \bothGrad^j$
                            \State $\Im\left(\momVal^{j+1}\right) = \Re\left(\momCoeff\right) \Im\left(\momVal^j\right) + \Im\left(\momCoeff\right)\Re\left(\momVal^j\right)$
                            \State $\bothParam^{j+1} = \bothParam^{j} - \Re\left(\lr\right) \bothGrad^j + \Re\left(\lr \momCoeff\right)\Re\left(\momVal^{j}\right) - \Im\left(\lr \momCoeff\right)\Im\left(\momVal^{j}\right)$
                        \EndFor
                        \State \Return $\bothParam_N$
                    \end{algorithmic}
                \end{algorithm}
        \end{figure}
    \subsection{Complex Momentum in PyTorch}
        Our method can be easily implemented in PyTorch 1.6+ by using complex tensors.
        The only necessary change to the SGD with momentum optimizer is extracting the real-component from momentum buffer as with JAX -- see \href{https://pytorch.org/docs/stable/_modules/torch/optim/sgd.html#SGD}{{\color{blue}here}}.
        
        In older versions of Pytorch, we can use a tensor to represent the momentum buffer $\momVal$, step size $\lr$, and momentum coefficient $\momCoeff$.
        Specifically, we represent the real and imaginary components of the complex number independently.
        Then, we redefine the operations \texttt{\_\_add\_\_} and \texttt{\_\_mult\_\_} to satisfy the rules of complex arithmetic -- i.e., Equations~\ref{cm_eq:complex_add} and \ref{cm_eq:complex_mult}.
    \vspace{-0.01\textheight}
    \section{Experiments}\label{cm_app:experiments}
        \vspace{-0.01\textheight}
        \subsection{Computing Infrastructure and Runtime}
            We do our computing in CPU for the purely adversarial experiments in Sections \ref{cm_sec:single_param_det} and \ref{cm_sec:exp_spectrum}.
            Training each $\num{2}$D GAN in Section~\ref{cm_sec:train_small_gan} takes $\smallGANTime$ hours and we can train $\num{10}$ simultaneously on an NVIDIA T$\num{4}$ GPU.
            Training each CIFAR GAN in Section~\ref{cm_sec:train_big_gan} takes $\bigGANTime$ hours, and we can only train $\num{1}$ model per NVIDIA T4 GPU.
        \vspace{-0.01\textheight}
        \subsection{Optimization in Purely Adversarial Games}\label{cm_sec:app_bilinear}
            We include the alternating update version of Figure~\ref{cm_fig:det_minmax_expAvg_simul_phase_tune} in Appendix Figure~\ref{cm_fig:det_minmax_expAvg_alt_phase_tune}, which allows us to contrast simultaneous and alternating updates.
            With alternating updates on a Dirac-GAN for $\lr = \num{0.1}$ the best momentum coefficient $\momCoeff$ was complex, but we could converge with real, negative momentum.
            Simultaneous updates may be a competitive choice with alternating updates, if alternating updates cost two gradient evaluations per step, which is common in deep learning setups.
        \vspace{-0.01\textheight}
        \subsection{Adversarialnesses Effect on Convergence}\label{cm_sec:app_spectrum}
            We include the extragradient (EG) update with extrapolation parameter $\lr'$ and step size $\lr$:
            \begin{align*}\label{cm_eq:eg_update}\tag{EG}
                \bothParam^{j+\frac{1}{2}} = \bothParam^j - \lr' \bothGrad^{j}\\
                \bothParam^{j+1} = \bothParam^j - \lr \bothGrad^{j+\frac{1}{2}}
            \end{align*}
            and the optimistic gradient (OG) update with extrapolation parameter $\lr'$ and step size $\lr$:
            \begin{align*}\label{cm_eq:og_update}\tag{OG}
                \bothParam^{j+1} = \bothParam^j - 2 \lr \bothGrad^{j} + \lr' \bothGrad^{j-1}
            \end{align*}
            
            EG and OG are often used with $\lr = \lr'$.
            However, we found that this constraint crippled these methods in cooperative games (that is, minimization).
            As such, we tune the extrapolation parameter $\lr'$ separately from the step size $\lr$, so EG and OG were competitive baselines.
            
            We include Figure~\ref{cm_fig:gan_decomp}, which shows a GANs spectrum throughout training and elaborates on Figure~\ref{cm_fig:gan_spectrum_main}.
            This shows many real and imaginary eigenvalues, so GAN training is neither purely cooperative nor adversarial.
            In addition, the structure of the set of eigenvalues for the discriminator differs from the generator, which may motivate separate optimizer choices.
            The structure between the players persists through training, but the eigenvalues grow in magnitude and spread out their phases.
            This indicates how adversarial the game is can change during training.

            \begin{figure*}
                \centering
                \begin{tikzpicture}
                    \centering
                    \node (img0){\includegraphics[trim={1.2cm 1.05cm 23.5cm .7cm},clip,width=\spectrumAxisWidth]{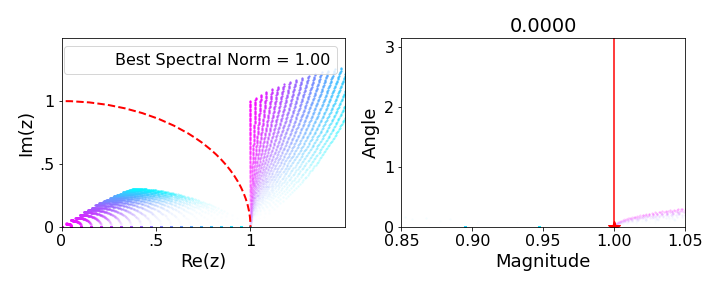}};
                    \node[left=of img1, node distance=0cm, rotate=90, xshift=.9cm, yshift=-7.35cm, font=\color{black}] {$\Im\left(\spectrum\left(\dynamics\right)\right)$};
                    
                    \node (img1)[right=of img0, node distance=0cm, xshift=-1.265cm]{\includegraphics[trim={1.9cm 1.25cm 12.5cm .7cm},clip,width=\spectrumVaryWidth]{latex/complexMomentum/images/spectrum/phase=0.png}};
                    \node[above=of img1, node distance=0cm, xshift=-0cm, yshift=-1.25cm,font=\color{black}] {$\arg(\momCoeff) = \num{0}$};
                    
                    \node (img2)[right=of img1, node distance=0cm, xshift=\spectrumSpacer]{\includegraphics[trim={1.9cm 1.25cm 12.5cm .7cm},clip,width=\spectrumVaryWidth]{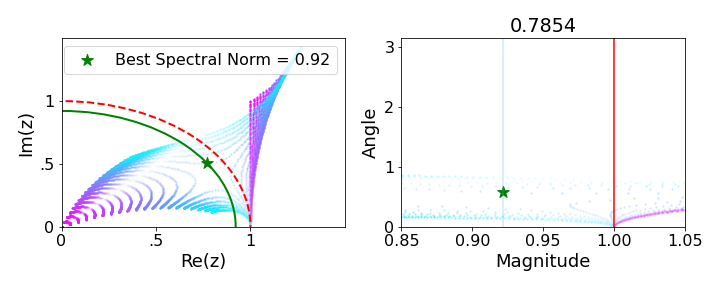}};
                    \node[above=of img2, node distance=0cm, xshift=0cm, yshift=-1.25cm,font=\color{black}] {$\arg(\momCoeff) = \frac{\pi}{\num{4}}$};
                     
                    \node (img3)[right=of img2, node distance=0cm, xshift=\spectrumSpacer]{\includegraphics[trim={1.9cm 1.25cm 12.5cm .7cm},clip,width=\spectrumVaryWidth]{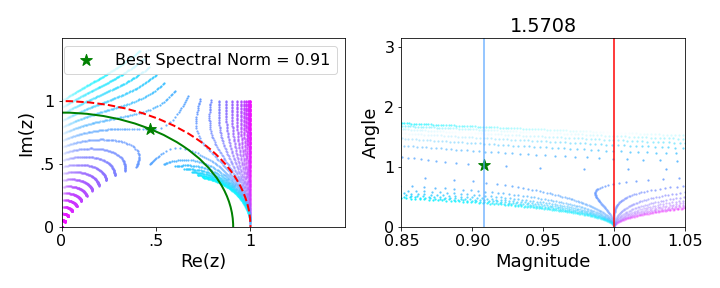}};
                    \node[below=of img3, node distance=0cm, xshift=-.1cm, yshift=1.2cm,font=\color{black}] {The real component of the spectrum of the augmented learning dynamics $\Re(\spectrum(\dynamics))$};
                    \node[above=of img3, node distance=0cm, xshift=0cm, yshift=-1.25cm,font=\color{black}] {$\arg(\momCoeff) = \frac{\pi}{\num{2}}$};
                     
                    \node (img4)[right=of img3, node distance=0cm, xshift=\spectrumSpacer]{\includegraphics[trim={1.9cm 1.25cm 12.5cm .7cm},clip,width=\spectrumVaryWidth]{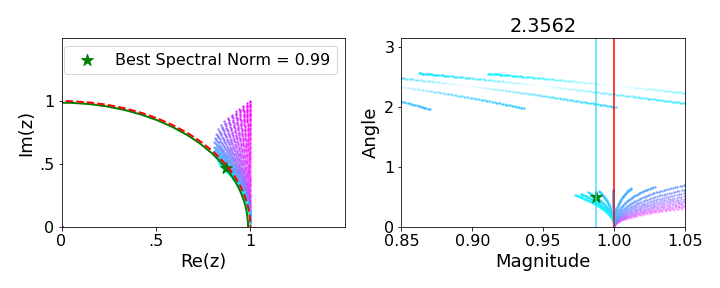}};
                    \node[above=of img4, node distance=0cm, xshift=0cm, yshift=-1.25cm,font=\color{black}] {$\arg(\momCoeff) =\ \frac{\num{3}\pi}{\num{4}}$};
                     
                    \node (img5)[right=of img4, node distance=0cm, xshift=\spectrumSpacer]{\includegraphics[trim={1.9cm 1.25cm 12.5cm .7cm},clip,width=\spectrumVaryWidth]{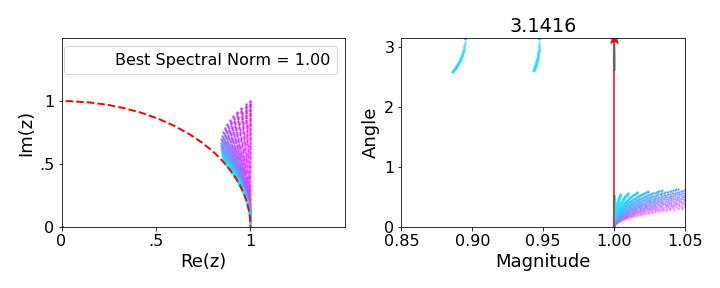}};
                    \node[above=of img5, node distance=0cm, xshift=0cm, yshift=-1.25cm,font=\color{black}] {$\arg(\momCoeff) = \pi$};
                    
                    \node (img6)[right=of img5, node distance=0cm, xshift=\spectrumSpacer]{\includegraphics[trim={.55cm .75cm .75cm .55cm},clip,width=\spectrumCbarWidth]{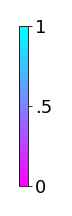}};
                    \node[right=of img6, node distance=0cm, rotate=270, xshift=0.0cm, yshift=-.75cm, font=\color{black}] {$| \momCoeff |$};
                \end{tikzpicture}
                \caption[Spectrum of the augmented learning dynamics for complex momentum]{
                    The spectrum of the augmented learning dynamics $\dynamics$ is shown, whose spectral norm is the convergence rate in Theorem~\ref{cm_thm:theorem}.
                    Each image is a different momentum phase $\arg(\momCoeff)$ for a range of $\lr, |\momCoeff| \!\in\! [\num{0},\! \num{1}]$.
                    The opacity of an eigenvalue is the step size $\lr$, and the color corresponds to the magnitude of the momentum $|\momCoeff|$.
                    A {\color{red}red} unit circle shows where all eigenvalues must lie to converge for a fixed $\lr, \momCoeff$.
                    If the max eigenvalue norm $<\! \num{1}$, we draw a {\color{green}green} circle whose radius is our convergence rate and a {\color{green}green} star at the associated eigenvalue.
                    In particular, at every non-real $\momCoeff$ we can select $\lr,\! |\momCoeff|$ for convergence.
                    The eigenvalues are symmetric on the $x$-axis, and the eigenvalues near $\Re(\eigval) \!=\! \num{1}$ dictate convergence rate.
                    The eigenvalues near the center are due to state augmentation, are of small magnitude, and do not impact the convergence rate.
                    Simultaneous gradient descent corresponds to the {\color{magenta}magenta} values where $|\momCoeff| \!=\! \num{0}$.
                }
                \label{cm_fig:spectrum_vary_phase}
            \end{figure*}
            
            \begin{figure}
                \centering
                \begin{tikzpicture}
                    \centering
                    \node (img1){\includegraphics[trim={1.0cm .9cm 2.7cm .9cm},clip,width=.28\linewidth]{latex/complexMomentum/images/bilinear/ExpAvg_Simul_Deterministic_beta_tune.png}};
                    \node[left=of img1, node distance=0cm, rotate=90, xshift=1.75cm, yshift=-.9cm, font=\color{black}] {\footnotesize Momentum phase $\arg(\momCoeff)$};
                    \node[above=of img1, node distance=0cm, xshift=-.1cm, yshift=-1.3cm,font=\color{black}] {\footnotesize \ref{cm_eq:simul_update} for \# grad. eval. = \# steps};
                    
                    \node [right=of img1, xshift=-1.1cm](img2){\includegraphics[trim={1.85cm .9cm .97cm .9cm},clip,width=.3\linewidth]{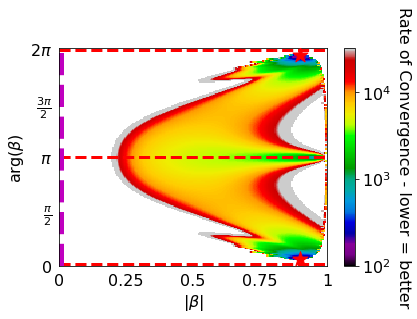}};
                    \node[below=of img2, node distance=0cm, xshift=.0cm, yshift=1.2cm,font=\color{black}] {Momentum Magnitude $| \momCoeff |$};
                    \node[above=of img2, node distance=0cm, xshift=-.3cm, yshift=-1.3cm,font=\color{black}] {\footnotesize \ref{cm_eq:alt_update} for \# grad. eval.};
                    \node[right=of img2, node distance=0cm, rotate=270, xshift=-1.9cm, yshift=-.9cm, font=\color{black}] {\footnotesize \# grad. eval. to converge};
                    
                    \node [right=of img2, xshift=-.65cm](img3){\includegraphics[trim={1.85cm .9cm .97cm .9cm},clip,width=.3\linewidth]{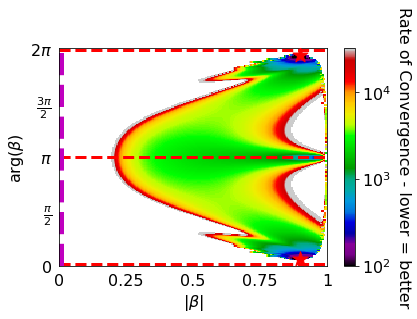}};
                    \node[above=of img3, node distance=0cm, xshift=-.3cm, yshift=-1.3cm,font=\color{black}] {\small \ref{cm_eq:alt_update} for \# steps};
                    \node[right=of img3, node distance=0cm, rotate=270, xshift=-1.5cm, yshift=-.9cm, font=\color{black}] {\footnotesize \# steps to converge};
                \end{tikzpicture}
                \caption[Dirac GAN convergence rates for optimizer parameters]{
                    We show many steps and gradient evaluations both simultaneous and alternating complex momentum take on a Dirac-GAN take for a set solution distance.
                    We fix the step size $\lr = \num{0.1}$ as in Figure~\ref{cm_fig:trajectories_limited}, while varying the phase and magnitude of our momentum $\momCoeff  = |\momCoeff| \exp\left(i \arg\left(\momCoeff\right)\right)$.
                    There is a {\color{red}red} star at the optima, dashed {\color{red}red} lines at real $\momCoeff$, and a dashed {\color{magenta}magenta} line for simultaneous or alternating gradient descent.
                    We only display color for convergent setups.
                    \emph{Left:} Simultaneous complex momentum (\ref{cm_eq:simul_update}).
                    This is the same as in Figure~\ref{cm_fig:det_minmax_expAvg_simul_phase_tune}, which we repeat to contrast with alternating updates.
                    No real-valued $\momCoeff$ converges for this -- or any -- $\lr$ with simultaneous updates ~\citep{gidel2018negative}.
                    Simultaneous updates can parallelize gradient computation for all players at each step, thus costing only one gradient evaluation per step for many deep learning setups.
                    The best convergence rate per step and gradient evaluation is $\approx {\color{red}\num{.955}}$.
                    \emph{Middle:} Alternating complex momentum (\ref{cm_eq:alt_update}), where we show how many gradient evaluations -- as opposed to steps -- to reach a set solution distance.
                    Alternating updates are bottlenecked by waiting for the first player's update to compute the second player's update, effectively costing two gradient evaluations per step for many deep learning setups.
                    As shown by \cite{gidel2018negative}, negative momentum can converge here, but the best momentum is still complex.
                    Also, alternating updates can make the momentum phase $\arg(\momCoeff)$ choice less sensitive to our convergence.
                    The best convergence rate per gradient evaluation is $\approx {\color{red}\num{.965}}$.
                    \emph{Right:} \ref{cm_eq:alt_update}, where we show how many steps it takes to reach a set solution distance.
                    The best convergence rate per step is $\approx {\color{red}\num{.931}}$.
                    \textbf{Takeaway:} If we can parallelize the computation of the gradients of both players, we can benefit from \ref{cm_eq:simul_update}.
                    However, if we cannot, then \ref{cm_eq:alt_update} can converge more quickly for a wider set of optimizer parameters.
                    In any case, the best solution uses a complex momentum $\momCoeff$ for this $\lr$.
                    %
                }
                \label{cm_fig:det_minmax_expAvg_alt_phase_tune}
            \end{figure}
            \begin{figure}
                \vspace{-0.02\textheight}
                \centering
                \begin{tikzpicture}
                    \centering
                    \node (img1)[]{\hspace{-1.5cm}\includegraphics[trim={.5cm .5cm .37cm .475cm},clip,width=.39\linewidth]{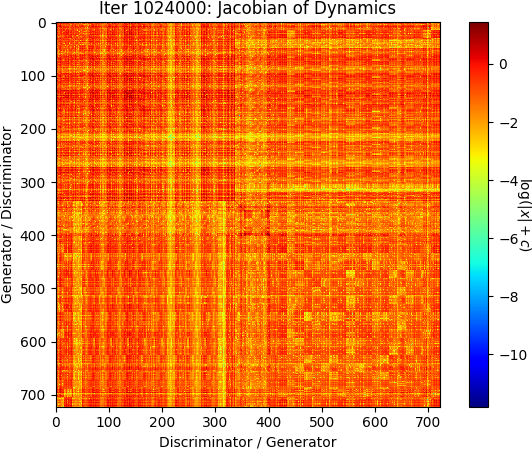}};
                    \node[left=of img1, node distance=0cm, rotate=90, xshift=2.0cm, yshift=.675cm, font=\color{black}] {Joint-gradient $\bothGrad$ index $l$};
                    \node[right=of img1, node distance=0cm, rotate=270, xshift=-2.4cm, yshift=-.75cm, font=\color{black}] {$\log(|a_{lk}| + \epsilon)$ for $a_{lk}$ in $\nabla_{\!\bothParam} \bothGrad^{j}$};
                    \node[below=of img1, node distance=0cm, xshift=-1.1cm, yshift=1.2cm,font=\color{black}] {Joint-parameter $\bothParam$ index $k$};
                    \node[above=of img1, node distance=0cm, xshift=-.8cm, yshift=-1.2cm,font=\color{black}] {Jacobian of joint-gradient $\nabla_{\!\bothParam} \bothGrad^{j}$ for GAN};
                    
                    \node (img2)[right=of img1, xshift=0.25cm]{\includegraphics[trim={.5cm .5cm .0cm .475cm},clip,width=.43\linewidth]{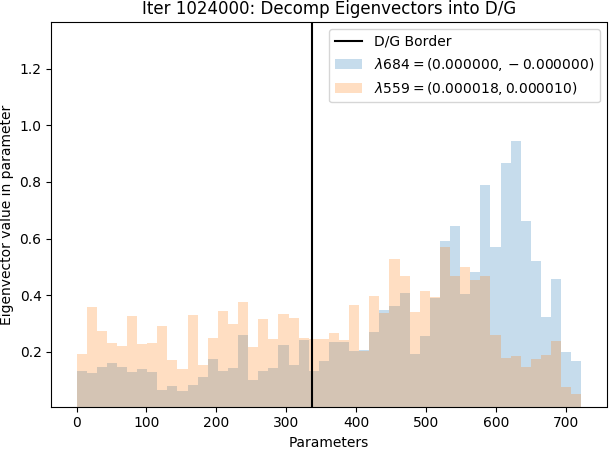}};
                    \node[left=of img2, node distance=0cm, rotate=90, xshift=2.25cm, yshift=-0.825cm, font=\color{black}] {Abs of EVec component};
                    \node[below=of img2, node distance=0cm, xshift=-.1cm, yshift=1.2cm,font=\color{black}] {Index into $\mathbf{v}$ -- first $\num{337}$ are for D's params};
                    \node[above=of img2, node distance=0cm, xshift=-.1cm, yshift=-1.0cm,font=\color{black}] {EVec $\mathbf{v}$ components for eigenvalues $\eigval \in \spectrum\left(\nabla_{\!\bothParam} \bothGrad\right)$};
                    
                    
                    \node (img3)[below=of img1, yshift=-.3cm, xshift=0]{\includegraphics[trim={1.3cm .5cm 1.8cm .475cm},clip,width=.41\linewidth]{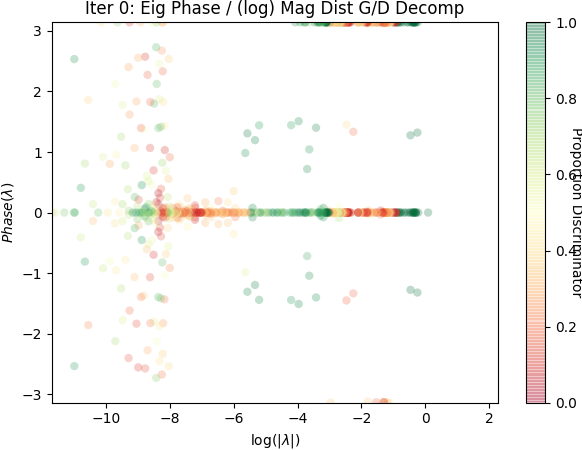}};
                    \node (img4)[right=of img3, xshift=\shiftHeatmapx]{\includegraphics[trim={1.3cm .5cm .9cm .475cm},clip,width=.44\linewidth]{latex/complexMomentum/images/16_gan_example/eig_mag_phase_distribution_decomposition_1024000.png}};
                    
                    \node (img3ytick)[left=of img3, node distance=0cm, rotate=90, xshift=2.9cm, yshift=-0.825cm, font=\color{black}] {-$\pi$ \hspace{\ytickspaceApp} -$\frac{\pi}{2}$ \hspace{\ytickspaceApp} $\num{0}$ \hspace{\ytickspaceApp} \hphantom{-}$\frac{\pi}{2}$ \hspace{\ytickspaceApp} \hphantom{-}$\pi$};
                    \node[left=of img3ytick, node distance=0cm, rotate=90, xshift=5.05cm, yshift=-0.43cm, font=\color{black}] {Phase of eigenvalue $\arg(\eigval)$};
                    
                    \node (img4colortick)[right=of img4, node distance=0cm, rotate=270, xshift=-2.8cm, yshift=-0.83cm, font=\color{black}] {disc. \hspace{0.08\textwidth} unsure \hspace{0.09\textwidth} gen.};
                    \node[right=of img4colortick, node distance=0cm, rotate=270, xshift=-5.3cm, yshift=-.55cm, font=\color{black}] {Does EVec point at a player?};

                    \node[below=of img3, node distance=0cm, xshift=3.65cm, yshift=1.2cm, font=\color{black}] {Log-magnitude of eigenvalue $\log\left(|\eigval|\right)$};
                    
                    \node[above=of img3, node distance=0cm, xshift=0.0cm, yshift=-1.25cm, font=\color{black}] {Start of training};
                    \node[above=of img4, node distance=0cm, xshift=0.0cm, yshift=-1.25cm, font=\color{black}] {End of training};
                    
                    \node[above=of img3, node distance=0cm, xshift=3.2cm, yshift=-.8cm, font=\color{black}] {Log-polar graph of the spectrum of Jacobian of the joint-gradient $\spectrum\left(\nabla_{\!\bothParam} \bothGrad^{j}\right)$ throughout training};
                \end{tikzpicture}
                \caption[GAN eigenvalue spectrum during training]{
                    These plots investigate the spectrum of the Jacobian of the joint-gradient for the GAN in Figure~\ref{cm_fig:gan_spectrum_main} through training.
                    The spectrum is key for bounding convergence rates in optimization.
                    \newline
                    \emph{Top left:} The Jacobian $\nabla_{\bothParam} \bothGrad$ for a GAN on a $\num{2}$D mixture of Gaussians with a two-layer, fully-connected $\ganSize$ hidden unit discriminator (D) and generator (G) at the end of training.
                    In the concatenated parameters $\bothParam \in \mathbb{R}^{723}$, the first $\num{337}$ are for D, while the last $\num{386}$ are for G.
                    We display the $\log$ of the absolute value of each component plus $\epsilon = 10^{-10}$.
                    The upper left and lower right quadrants are the Hessian of D's and G's losses, respectively.
                    \newline
                    \emph{Top Right:} We visualize two randomly sampled eigenvectors from $\nabla_{\bothParam} \bothGrad$.
                    The first part of the parameters is for the discriminator, while the second is for the generator.
                    Given an eigenvalue with eigenvector $\mathbf{v}$, we approximate the attribution of eigenvectors to players by calculating how much of it lies in the parameter space of D with $\frac{\| \mathbf{v}_{1:|\textnormal{D}|} \|_1}{\| \mathbf{v} \|_1} = \frac{\| \mathbf{v}_{1:337} \|_1}{\| \mathbf{v} \|_1}$.
                    If this ratio is close to $\num{1}$ (or $\num{0}$) and say \emph{the eigenvector mostly points at D (or G)}.
                    The blue eigenvector mostly points to G, while the orange eigenvector is unclear.
                    Finding useful ways to attribute eigenvalues to players is an open problem.
                    \newline
                    \emph{Bottom:} The spectrum of the Jacobian of the joint-gradient $\spectrum\left(\nabla_{\bothParam} \bothGrad^{j}\right)$ is shown in log-polar coordinates because it is difficult to see structure when graphing in Cartesian (i.e., $\Re$ and $\Im$) coordinates, due to eigenvalues spanning orders of magnitude, while being positive and negative.
                    The end of training is when we stop making progress on the log-likelihood.
                    We have imaginary eigenvalues at $\arg(\eigval) = \pm \nicefrac{\pi}{\num{2}}$, positive eigenvalues at $\arg(\eigval) = \num{0}$, and negative eigenvalues at $\arg(\eigval) = \pm \pi$.
                    \newline
                    \emph{Takeaway:}
                    A banded structure for the coloring of the eigenvalues persists through training.
                    We may want different optimizer parameters for the discriminator and generator due to asymmetry in their associated eigenvalues.
                    Furthermore, the magnitude of the eigenvalues increases during training, and the $\arg$s spread out, indicating that the game can change the eigenstructure near the solutions.
                }\label{cm_fig:gan_decomp}
            \end{figure}

        \vspace{-0.01\textheight}
        \subsection{Training GANs on 2D Distributions}\label{cm_sec:app_2d_gan}
            For $\num{2}$D distributions, the data is generated by sampling from a mixture of $\mixtureNum$ Gaussian distributions distributed uniformly around the unit circle.
            
            
            For the GAN, we use a fully connected network with $\num{4}$ ReLU~\citep{hahnloser2000digital} layers with $\numHiddenUnits$ hidden units.
            We chose this architecture to be the same as \cite{gidel2018negative}.
            Our noise source for the generator is a $\num{4}$D Gaussian.
            We trained the models for $\numIterations$ iterations.
            The performance of the optimizer settings is evaluated by computing the negative log-likelihood of a batch of $\batchSizeSmallGAN$ generated $\dataDim$D samples.

            \begin{figure}
                \centering
                \begin{tikzpicture}
                    \centering
                    \node (img1){\includegraphics[trim={4.2cm 1.75cm 8.85cm 2.57cm},clip, width=\ganHeatmapWidth]{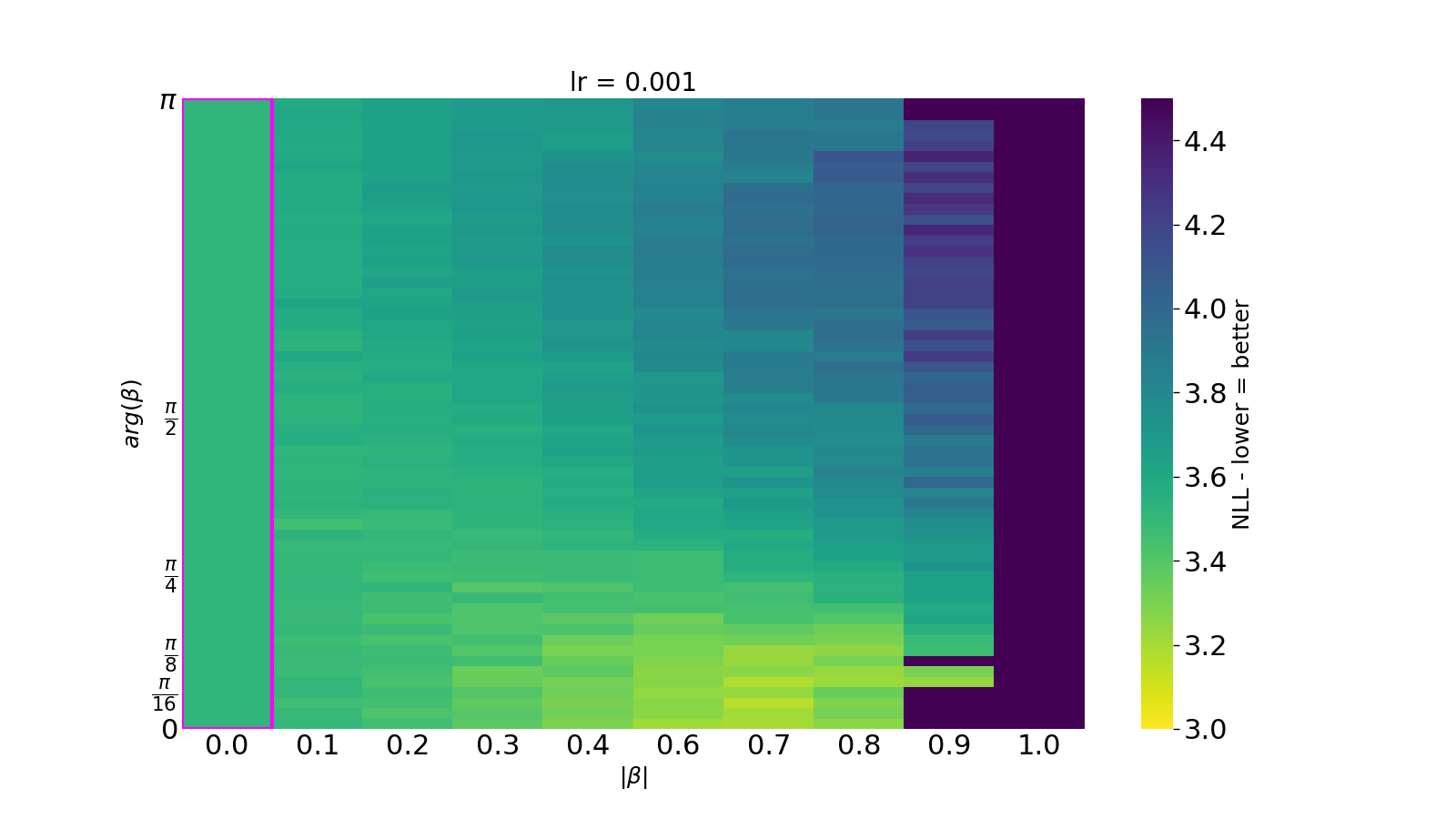}};
                    \node[left=of img1, node distance=0cm, rotate=90, xshift=2.25cm, yshift=-.9cm, font=\color{black}] {Momentum phase $\arg(\momCoeff)$};
                    \node[above=of img1, node distance=0cm, xshift=0cm, yshift=-1.1cm,font=\color{black}] {Step size $\lr = \num{0.001}$};
                    \node(img2cap)[below=of img1, node distance=0cm, xshift=3.3cm, yshift=1.1cm,font=\color{black}] {Momentum Magnitude $| \momCoeff |$};
                    
                    \node (img2)[right=of img1, node distance=0cm, xshift=-1.4cm]{\includegraphics[trim={5.1cm 1.75cm 6.4cm 2.57cm},clip,width=\ganHeatmapWidth + 0.02\textwidth]{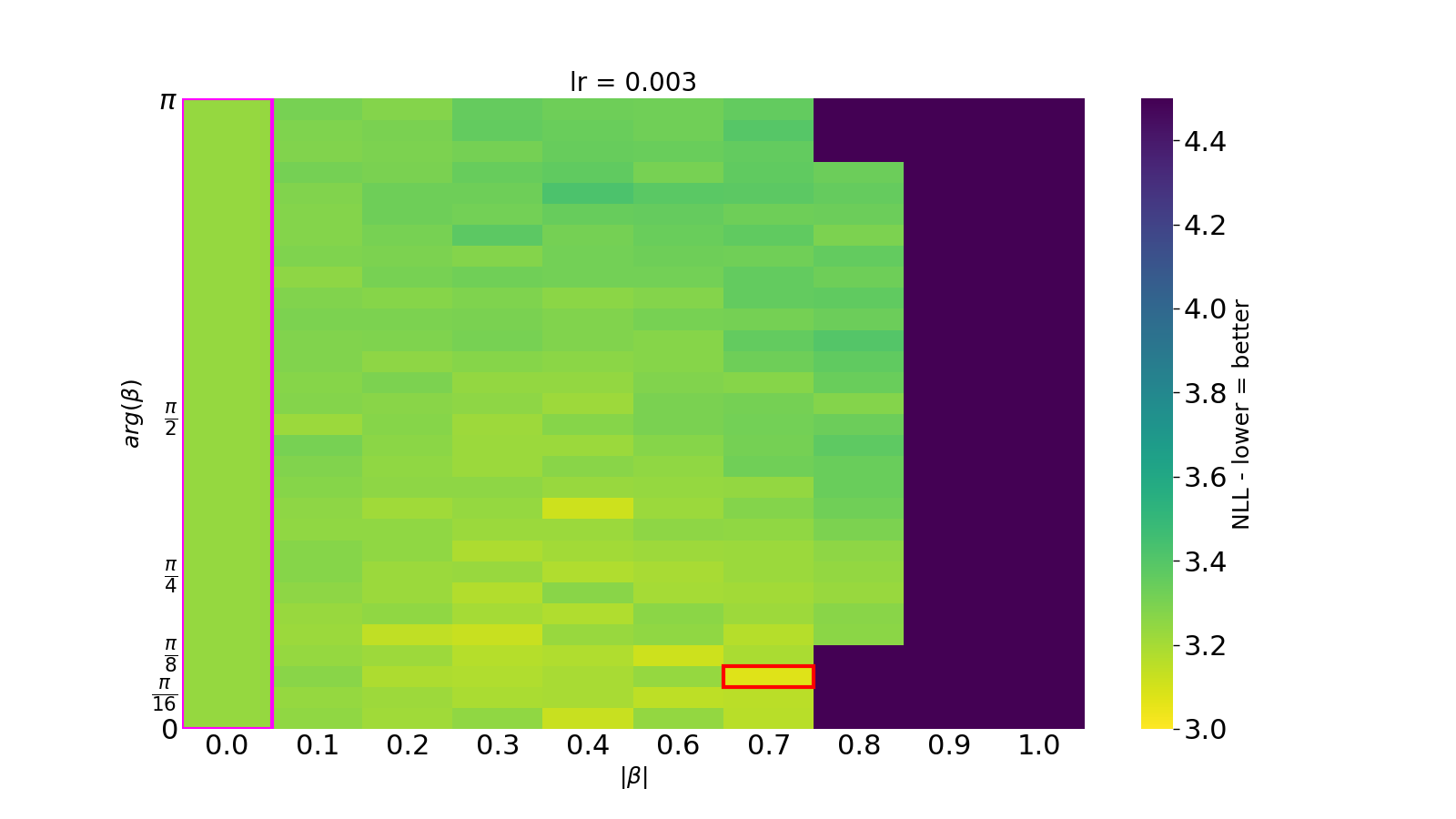}};
                    \node[above=of img2, node distance=0cm, xshift=0.0cm, yshift=-1.1cm,font=\color{black}] {Step Size $\lr = \num{0.003}$};
                    \node[right=of img2, node distance=0cm, rotate=270, xshift=-1.9cm, yshift=-.85cm, font=\color{black}] {NLL, lower = better};
                \end{tikzpicture}
                \caption[BigGAN heatmaps of performance with complex momentum]{
                    Heatmaps of the negative log-likelihood (NLL) for tuning $\arg(\momCoeff), |\momCoeff|$ with various fixed $\lr$ on a $\num{2}$D mixture of Gaussians GAN.
                    We highlight the best performing cell in {\color{red}red}, which had $\arg(\momCoeff) \approx \nicefrac{\pi}{\num{8}}$.
                    Runs equivalent to alternating SGD are shown in a {\color{magenta}magenta} box.
                    We compare to negative momentum with alternating updates as in \cite{gidel2018negative} in the top row with $\arg(\momCoeff) = \pi$.
                    \emph{Left}: Tuning the momentum with $\lr = \num{0.001}$.
                    \emph{Right}: Tuning the momentum with $\lr = \num{0.003}$.
                }
                \label{cm_fig:gan_heatmaps}
            \end{figure}
        \begin{figure}
            \centering
                \includegraphics[trim={1.25cm 1.25cm 1.25cm 1.25cm},clip,width=.9\textwidth]{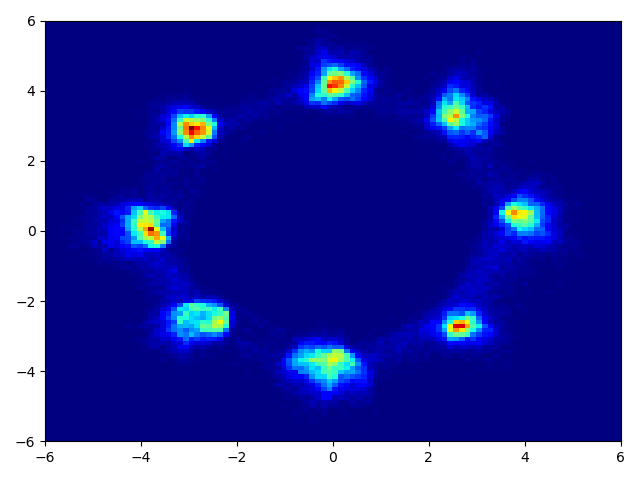}
                \caption[Learned mixture of Gaussians samples]{
                    The mixture of Gaussian samples from the GAN with the best hyperparameters from Figure~\ref{cm_fig:gan_heatmaps}.
                }
                \label{cm_fig:2d_samples}
        \end{figure}
        \begin{figure}
            \centering
                \includegraphics[trim={0cm 0cm 3.6cm 6cm},clip,width=.75\textwidth]{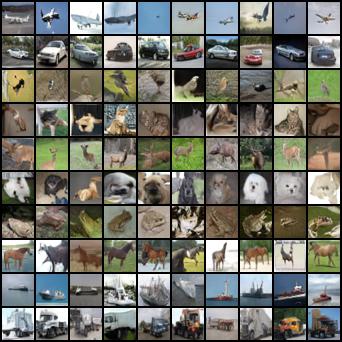}
                \caption[Class-conditional BigGAN samples]{
                    Class-conditional CIFAR-10 samples from the GAN with the best hyperparameters from Figure~\ref{cm_fig:biggan_inception_heatmap}.
                }
                \label{cm_fig:biggan_samples}
            %
                    
        \end{figure}
        \begin{figure}
            \centering
                \centering
                \begin{tikzpicture}
                    \centering
                    \node (img1){\includegraphics[trim={.875cm .9cm 0cm 0cm},clip,width=.66\linewidth]{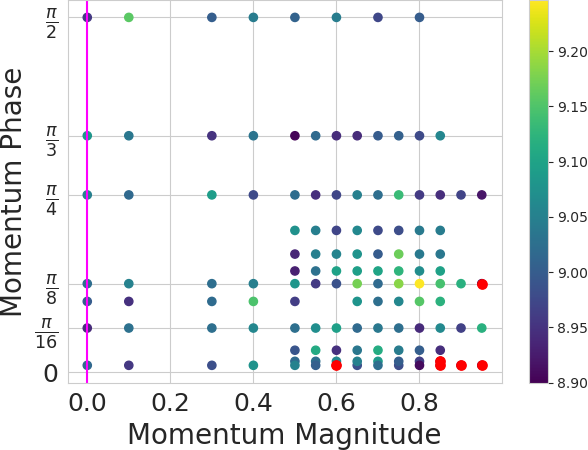}};
                    \node[left=of img1, node distance=0cm, rotate=90, xshift=2.2cm, yshift=-.9cm, font=\color{black}] {Momentum phase $\arg\left(\momCoeff_1\right)$};
                    \node[below=of img1, node distance=0cm, xshift=-.1cm, yshift=1.0cm,font=\color{black}] {Momentum magnitude $| \momCoeff_1 |$};
                    \node[right=of img1, node distance=0cm, rotate=270, xshift=-1.5cm, yshift=-.8cm, font=\color{black}] {Inception score $\uparrow$};
                \end{tikzpicture}
                \caption[BigGAN grid search results]{
                    The inception score for a grid search on $\arg\left(\momCoeff_1\right)$ and $|\momCoeff_1|$ for training BigGAN on CIFAR-10 with the Adam variant in Algorithm~\ref{cm_alg:complex_adam}.
                    The $\momCoeff_1$ is complex for the discriminator, while the generator's optimizer is fixed to the defaults supplied by the author.
                    {\color{red}Red} points are runs that failed to train to the minimum inception score in the color bar.
                    The vertical {\color{magenta}magenta} line denotes runs equivalent to alternating SGD.
                    Negative momentum did not train for any magnitude of momentum $|\momCoeff_1| > .5$, so we did not display it for more resolution near the values of interest.
                }
                \label{cm_fig:biggan_inception_heatmap}
        \end{figure}
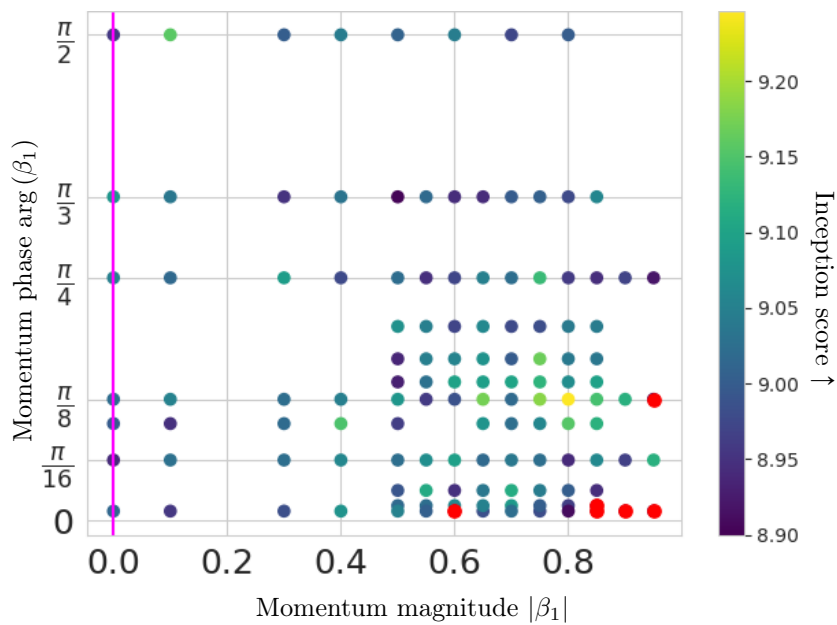
            %
        \begin{figure}
                \centering
                \begin{tikzpicture}
                    \centering
                    \node (img1){\includegraphics[trim={.875cm .9cm 0cm 0cm},clip,width=.66\linewidth]{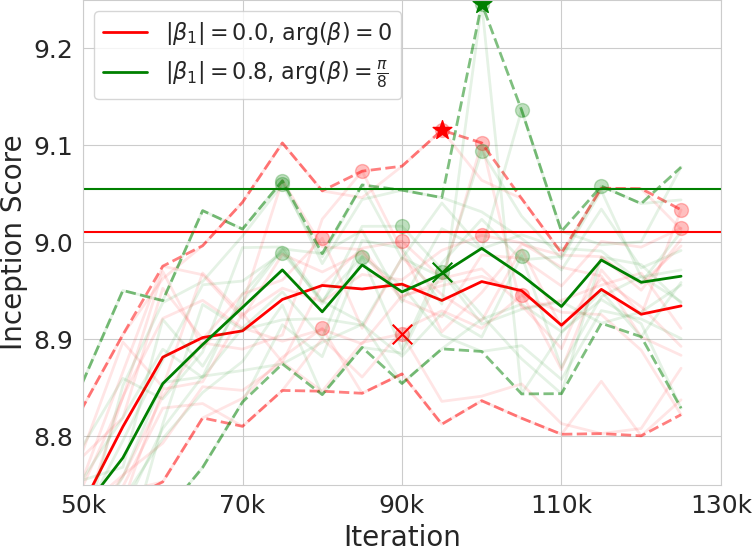}};
                    \node[left=of img1, node distance=0cm, rotate=90, xshift=1.5cm, yshift=-.8cm, font=\color{black}] {Inception score $\uparrow$};
                    \node[below=of img1, node distance=0cm, xshift=-.1cm, yshift=1.0cm,font=\color{black}] {Iteration};
                \end{tikzpicture}
                \caption[BigGAN inception scores during training]{
                    We compare the best optimization parameters from grid search Figure~\ref{cm_fig:biggan_inception_heatmap} for our complex Adam variant (i.e., Algorithm~\ref{cm_alg:complex_adam}) shown in {\color{green}green}, with the author provided values shown in {\color{red}red} for the CIFAR-10 BigGAN over $\num{10}$ seeds.
                    A star is displayed at the best inception score for all runs, a cross is at the worst inception score for all runs, and a circle is at the best inception score for each run.
                    Dashed lines are shown at the max/min inception score over all runs at each iteration, low-alpha lines for each runs inception score, and solid lines for the average inception score over all seeds at each iteration.
                    The results are summarized in Table~\ref{cm_tab:biggan}.
                }
                    
                \label{cm_fig:biggan_inception_runs}
        \end{figure}

    \chapter{Lyapunov Exponents for Diversity in Differentiable Games}

\begin{table*}[htbp]\caption{Notation for Lyapunov Exponents for Diversity in Differentiable Games}
        \vspace{-0.02\textheight}
        \begin{center}
            \begin{tabular}{c c}
                \toprule
                RR & Ridge Rider\\
                IPD & Iterated Prisoners' Dilemma\\
                GAN & Generative Adversarial Network\\
                LOLA & Learning with Opponent Learning Awareness\\
                EVec, EVal & Shorthand for Eigenvector or Eigenvalue\\
                SGD & Stochastic Gradient Descent\\
                SimSGD & Simultaneous SGD\\
                $\defeq$ & Defined to be equal to\\
                $x, y, z, \dots \in \mathbb{C}$ & Scalars\\
                $\boldsymbol{x}, \boldsymbol{y}, \boldsymbol{z}, \dots \in \mathbb{C}^{n}$ & Vectors\\
                $\boldsymbol{X}, \boldsymbol{Y}, \boldsymbol{Z}, \dots \in \mathbb{C}^{n \times n}$ & Matrices\\
                $\boldsymbol{X}^\transpose$ & The transpose of matrix $\boldsymbol{X}$\\
                $\identity$ & The identity matrix\\
                $\Re(z), \Im(z)$ & The real or imaginary component of $z \in \mathbb{C}$\\
                $i$ & The imaginary unit. $z \in \mathbb{C} \implies z = \Re(z) + i \Im(z)$\\
                $\conj{z}$ & The complex conjugate of $z \in \mathbb{C}$\\
                $|z| \defeq \sqrt{z \conj{z}}$ & The magnitude or modulus of $z \in \mathbb{C}$ \\
                $\arg(z)$ & The argument or phase of $z \in \mathbb{C} \implies z = |z| \exp(i\arg(z))$ \\
                $\outSymbol, \inSymbol$ & A symbol for the outer/inner players\\
                $\outDim, \inDim \in \mathbb{N}$ & The number of weights for the outer/inner players\\
                $\paramSymbol$ & A symbol for the parameters or weights of a player\\
                $\outParam \in \mathbb{R}^{\outDim}, \inParam \in \mathbb{R}^{\inDim}$ & The outer/inner parameters or weights\\
                $\loss: \mathbb{R}^{n} \to \mathbb{R}$ & A symbol for a loss\\
                $\outLoss(\outParam, \inParam), \inLoss(\outParam, \inParam)$ & The outer/inner losses -- $\mathbb{R}^{\outDim + \inDim} \mapsto \mathbb{R}$ \\
                $\outGrad(\outParam, \inParam), \inGrad(\outParam, \inParam)$ & Gradient of outer/inner losses w.r.t. their weights in $\mathbb{R}^{\outDim/\inDim}$ \\
                $\inParam^*\!(\outParam\!) \!\defeq\! \argmin\limits_{\inParam} \!\!\inLoss\!(\!\outParam \!, \!\inParam\!)$&The best-response of the inner player to the outer player\\
                $\outLoss^*(\outParam) \!\defeq\! \outLoss\!(\!\outParam \!, \!\inParam^*(\outParam)\!)$ & The outer loss with a best-responding inner player\\
                $\outParam^* \!\defeq\! \argmin\limits_{\outParam} \outLoss^*(\outParam\!) $ & Outer optimal weights with a best-responding inner player\\
                $\bothDim \defeq \outDim + \inDim$ & The combined number of weights for both players\\
                $\bothParam \defeq [\outParam, \inParam] \in \mathbb{R}^{\bothDim}$ & A concatenation of the outer/inner weights\\
                $\bothGrad(\bothParam) \!\defeq\! [\outGrad(\bothParam), \inGrad(\bothParam)] \in \mathbb{R}^{\bothDim}$ & A concatenation of the outer/inner gradients\\
                $\bothParam^{\num{0}} = [\outParam^{\num{0}}, \inParam^{\num{0}}] \in \mathbb{R}^{\bothDim}$ & The initial parameter values\\
                $j$ & An iteration number\\
                $\bothGrad^j \defeq \bothGrad(\bothParam^j) \in \mathbb{R}^{\bothDim}$ & The joint-gradient vector field at weights $\bothParam^j$\\
                $\nabla_{\bothParam} \bothGrad^j \defeq \nabla_{\bothParam}\bothGrad |_{\bothParam^j} \in \mathbb{R}^{\bothDim \times \bothDim}$ & The Jacobian of the joint-gradient $\bothGrad$ at weights $\bothParam^j$\\
                $\bothHess$ & The game Hessian \\
                $\lr \in \mathbb{C}$ & The step size or learning rate\\
                $\eigval \in \mathbb{C}, e$ & Notation for an arbitrary Eval \\
                $\spectrum(\boldsymbol{M}) \in \mathbb{C}^{n}$ & The spectrum -- or set of eigenvalues -- of $\boldsymbol{M} \in \mathbb{R}^{n \times n}$ \\
                $\spectralRadius(\boldsymbol{M}) \!\defeq\! \max_{z \in \spectrum(\boldsymbol{M})} |z|$ & The spectral radius in $\mathbb{R}^{+}$ of $\boldsymbol{M} \in \mathbb{R}^{n \times n}$ \\
                $\lyapunovfixedPointOp(\bothParam)$ & Fixed-point operator for our optimization \\
                $\jacFixedPointOp$ & The Jacobian of the fixed-point operator \\
                $\displacement$ & A displacement for a Lyapunov exponent \\
                $\gamma_j(\bothParam_0, \displacement) \defeq \log(\displacement^\transpose (\jacFixedPointOp^{j}(\bothParam_0))^\transpose \jacFixedPointOp^{j}(\bothParam_0) \displacement)$ & A Lyapunov term for a Lyapunov exponent \\
                $\lyap_k(\bothParam_0, \displacement) = \frac{1}{k}\smash{\sum_{j = 0}^{k}} \gamma_j(\bothParam_0, \displacement)$ & A $k$-step Lyapunov exponent \\
                $\lyap_k^{max}(\bothParam_0) = \max_{\displacement, \|\displacement\| = 1} \lyap_k(\bothParam_0, \displacement)$ & The max $k$-step Lyapunov exponent \\
                $\loss(\bothParam_0)$ & A starting point loss using Lyapunov exponents \\
                $\loss_{n}^{\textnormal{sum}}(\bothParam_0)$ & A loss using the sum of top $n$ exponents \\
                $\loss_{n}^{\textnormal{min}}(\bothParam_0)$ & A loss using the min of top $n$ exponents \\
                \bottomrule
            \end{tabular}
        \end{center}
        \label{lyapunov_tab:TableOfNotation}
    \end{table*}

    \vspace{-0.03\textheight}
    \section{Related Work}\label{lyapunov_sec:related-work}
        The material from Chapter~\ref{chap:lyapunovExponent} is primarily from \citet{lorraine2021using}, with a prior workshop submission of \citet{lorraine2021usingB} and an arXiv submission of \citet{lorraine2021usingARXIV}.
        
        \textbf{Diversity in machine learning:}
            Finding diverse solutions is often desirable in machine learning, for example, improving performance for model ensembles~\citep{1990_ensembles}, with canonical approaches directly optimizing for negative correlation amongst model predictions~\citep{NCL}.
            In recent times, these ideas have begun to re-emerge, improving ensemble performance \citep{dibs, lit2020, marSra16}, robustness~\citep{pang2019improving, cully2015robots} and boosting exploration in reinforcement learning~\citep{Lehman08exploitingopen, eysenbach2018diversity, parkerholder2020effective, qdnature}.
        
            Many of these approaches seek diverse solutions by following gradients of an altered, typically multi-objective loss function.
            In contrast, the recent \emph{Ridge Rider} (RR)~\citep{parker2020ridge} algorithm searches for diverse solutions by following eigenvectors of the Hessian with respect to the original loss function, producing orthogonal (loss reducing) search directions.
        
        \textbf{Finding solutions in games:}
            There are first-order methods for finding solutions in games~\citep{zhang2021unified} including alternating updates~\citep{zhangnear}, extragradient~\citep{korpelevich1976extragradient,azizian2020tight}, optimistic gradient~\citep{rakhlin2013optimization,daskalakis2017training}, negative momentum~\citep{gidel2018negative, zhang2021suboptimality}, complex momentum~\citep{lorraine2021complex}, and iterate averaging~\citep{gidel2018variational}.
            There are also higher-order methods such as consensus optimization~\citep{mescheder2017numerics}, symplectic gradient adjustment (SGA)~\citep{letcher2019differentiable}, local symplectic surgery (LSS)~\citep{mazumdar2019finding}, competitive gradient descent (CGD)~\citep{schafer2019competitive}, follow-the-ridge~\citep{wang2019solving}.
            Recently, \citet{vicol2020implicit} looked at the effects of overparameterization while doing online learning in games.
        
        \textbf{Diversity in multi-agent RL:}
            In recent times, a series of works have explored the benefit of diversity in competitive \citep{balduzzi2019open, garnelo2021pick, vinyals2019grandmaster} and cooperative \citep{yang2020multi, lupu2021trajectory} multi-agent RL.
            However, once again, these approaches all consider augmented loss functions. Instead, we take inspiration from RR and extend it to the multi-agent setting.
            \newline
        
        \textbf{Tree searches in MDP:}
            Tree searches are a classic technique for planning and control in discrete Markov Decision Processes (MDP).
            The branch locations and the possible branch choices are provided in these settings.
            All that remains is to choose which to explore.
            There is much interest in performing these searches ranging from depth/breadth-first search, A* \citep{Hart1968}, to more sophisticated methods such as Monte Carlo Tree Search \citep{abramson2014expected} potentially with learned value functions \citep{silver2016mastering}.
            Searching in continuous action spaces was explored in \citet{moerland2018a0c, kim2020monte, mao2020poly}.
            All of these methods search in the agents' action space, and we employ them to find agent parameters.
        
            Rapidly-exploring random tree builds a space-filling tree to cover a continuous optimization space \citep{lavalle1998rapidly, lavalle2001randomized}.
            Unlike RR-based methods, the branching points are not determined by the underlying properties of the loss landscape (e.g., only at saddles or bifurcation).
            This technique is commonly used in robotic motion planning \citep{rodriguez2006obstacle}.
        
        \textbf{Branching in evolutionary optimization:}
            Designing optimization algorithms in a multimodal loss landscape is a focus of the evolutionary optimization community \citep{singh2006comparison, wong2015evolutionary}, which implicitly build a tree of candidate solutions.
            Evolutionary algorithms designed to encourage diversity have been explored in the context of multi-player games \citep{balduzzi2019open, vinyals2019grandmaster, arulkumaran2019alphastar} by encouraging agent populations to learn different strategies that perform well against other agents, resulting in a growing collection of strategies.
        
        \textbf{Bifurcations:}
            This work builds on the previous workshop submission of \citet{lorraine2021using}.
            The works of \citet{zeeman1980population, yang2018bifurcation, chotibut2020route, piliouras2020catastrophe, leonardos2020exploration, bielawski2021follow} leverage the bifurcations for learning in games.
            \citet{yang2018bifurcation} introduces hysteresis and optimal control mechanisms to control equilibrium selection and improve social welfare.
            We do not focus on maximizing social welfare but on finding multiple solutions.
            \citet{chotibut2020route} studies the properties of bifurcations in routing games, showing various analyses for social cost.
            \citet{piliouras2020catastrophe} combines catastrophe theory with mechanism design to destabilize inefficient solutions.
            \citet{leonardos2020exploration} studies how changing an exploration parameter in multi-agent learning can affect equilibrium selection.
            Our work focuses on connecting bifurcations with finding diverse solutions in optimization with Ridge Rider and finding those bifurcations - potentially with gradient-based methods.
        
        \textbf{Lyapunov exponents:}
            The works of \citet{cheung2019vortices, cheung2020chaosb, sato2002chaos} use Lyapunov exponents when learning in games.
            \citet{cheung2019vortices} shows that some learning algorithms are Lyapunov chaotic in the payoff space.
            Our work focuses on using Lyapunov exponent variants to identify bifurcations (which are used to find diverse solutions in optimization) and potentially using optimization methods on the exponent.

        \textbf{Separatrices:}
            The works of ~\citet{panageas2016average, zhang2015equilibrium, nagarajan2020chaos} look at computing separatrices in games used for various purposes.
            \citet{nagarajan2020chaos} finds the shape of separatrices using invariant functions with online learning dynamics.

    \vspace{-0.01\textheight}
    \section{Proposed Algorithms}
        \subsection{Branching Optimization Tree Searches}
        \vspace{-0.01\textheight}
            \begin{algorithm}[H]
                \caption[Branching optimization tree search]{Branching Optimization Tree Search -- important changes from RR to GRR in \textcolor{red}{red}}
                \label{lyapunov_algorithm:branching_tree}
                \begin{algorithmic}[1]
                    \State \textbf{Input:}  \textcolor{red}{$\mathrm{FindStartingPoint}$}, $\mathrm{ChooseBranch}$, \textcolor{red}{$\mathrm{SplitBranch}$}, \textcolor{red}{$\mathrm{EndRide}$}, \textcolor{red}{$\mathrm{Optimize}$}, $\mathrm{VerifySolution}$, $\mathrm{ContinueBranching}$
                    \State Select optimization parameters $\alpha$
                    \State Find starting parameters $\bothParam^{\textnormal{start}} = \textcolor{red}{\mathrm{FindStartingPoint}(\alpha)}$
                    \State Initialize a branch $\psi^{\textnormal{init}} = \mathrm{InitBranch}(\bothParam^{\textnormal{start}}, \alpha)$
                    \State Initialize the set of branches $\mathcal{B} = \textcolor{red}{\mathrm{SplitBranch}}(\psi^{\textnormal{init}})$
                    \State Initialize the set of solutions $\mathcal{S} = \emptyset$
                    \While{$\textnormal{Branches } \mathcal{B} \textnormal{ non-empty}$}
                        \State $\psi, \mathcal{B}  = \mathrm{ChooseBranch}(\mathcal{B})$
                        \State $\bothParam^* = \textcolor{red}{\mathrm{Optimize}}(\psi.\bothParam, \psi.\alpha)$ \# Optimize our parameters
                        \If{$\mathrm{VerifySolution}(\bothParam^{*})$}
                            \State $\mathcal{S} = \mathcal{S} \cup \{ \bothParam^{*} \}$
                        \EndIf
                        \State Make new branch to split $\psi' = \mathrm{copy}(\psi)$
                        \State Store the optimized parameters $\psi'\textnormal{.parameters} = \bothParam^*$
                        \If{$\mathrm{ContinueBranching}(\psi')$}
                            \State $\mathcal{B} = \mathcal{B} \cup \textcolor{red}{\mathrm{SplitBranch}}(\psi')$
                        \EndIf
                    \EndWhile
                    \State \textbf{return} $\mathcal{S}$
                \end{algorithmic}
            \end{algorithm}

    \vspace{-0.01\textheight}
    \section{Experiments}
        \subsection{Test Problems}\label{lyapunov_sec:app_test_problems}
            \textbf{Iterated Prisoners' Dilemma (IPD):} This game is an infinite sequence of the Prisoner's Dilemma, where the future payoff is discounted by $\gamma \in [0, 1)$.
            In other words, the rewards sequences $r_j$ for each player were summed through $\sum_{j=0}^{\infty} \gamma^j r_j$.
            Each agent is conditioned on the actions in the prior state ($s$).
            Thus, there are 5 parameters for each agent $i$: $P^{i}(C | s)$ is the probability of cooperating in the start state $s_0 = \emptyset$ or in the state $s_t = \left(a^{1}_{t-1}, a^{2}_{t-1}\right)$ for $t > 0$.
            There are two Nash equilibria we care about: Defect-Defect (DD), where agents are selfish (resulting in poor reward), and tit-for-tat (TT), where agents initially cooperate, then copy the opponents' action (resulting in higher reward).
            
            \textbf{Small IPD}: This is a 2-parameter 
             simplification of IPD, which allows DD and TT Nash equilibria.
            We fix the strategy if our opponent defects, to defect with high probability.
            We also constrain the probability of cooperating to depend only on whether the opponent cooperates, and in the initial state, we assume our opponent cooperated.
            This game allows us to visualize some optimization difficulties for the full-scale IPD.
            However, the game Hessian has strictly real eigenvalues, unlike the full-scale IPD.
            See Figure~\ref{lyapunov_fig:fig_new_problems_baseline} top for a visualization of the strategy space.

            \newpage
            \textbf{Matching Pennies:} This is a simplified 2-parameter version of rock-paper-scissors, where each player selects to cooperate or defect.
            This game has a Nash equilibrium, in which players select their actions uniformly.
            Notably, this problem's game Hessian has purely imaginary eigenvalues, so following the gradient does not converge to solutions, and we need a method for learning in games like LOLA.
            Also, this game only has one solution.
            Thus, it is a poor fit for evaluating RR, which finds diverse solutions.
            See Figure~\ref{lyapunov_fig:fig_new_problems_baseline}, bottom for a visualization of the strategy space.
            
            \textbf{Mixing Small IPD and Matching Pennies:} This game interpolates between the Small IPD and Matching Pennies games, with the loss for player $j$:
            \begin{equation}\label{eq_app:mixed_obj}
                \mathcal{L}_{\textnormal{mix}, P_{j}, \tau} = \tau \mathcal{L}_{\textnormal{smallIPD}, P_{j}} + (1 - \tau)\mathcal{L}_{\textnormal{matchingPennies}, P_{j}}
            \end{equation}
            This problem has two solutions: one where both players cooperate and one where both players select actions uniformly.
            The uniform action solution has imaginary eigenvalues, so it is only stable under a method for learning in games, while the both cooperate solution has real eigenvalues.
            There is a Hopf bifurcation that separates these solutions.
            See Figure~\ref{lyapunov_fig:fig_new_problems_baseline} for standard methods on this problem and Appendix Figure~\ref{lyapunov_fig:fig_new_problems_baseline_full} to contrast this problem with Small IPD or Matching Pennies.
        
            \begin{figure*}
                \centering
                \begin{tikzpicture}
                    \centering
                    \node (img){\includegraphics[trim={.25cm .25cm 3.5cm .25cm},clip, width=.31\linewidth]{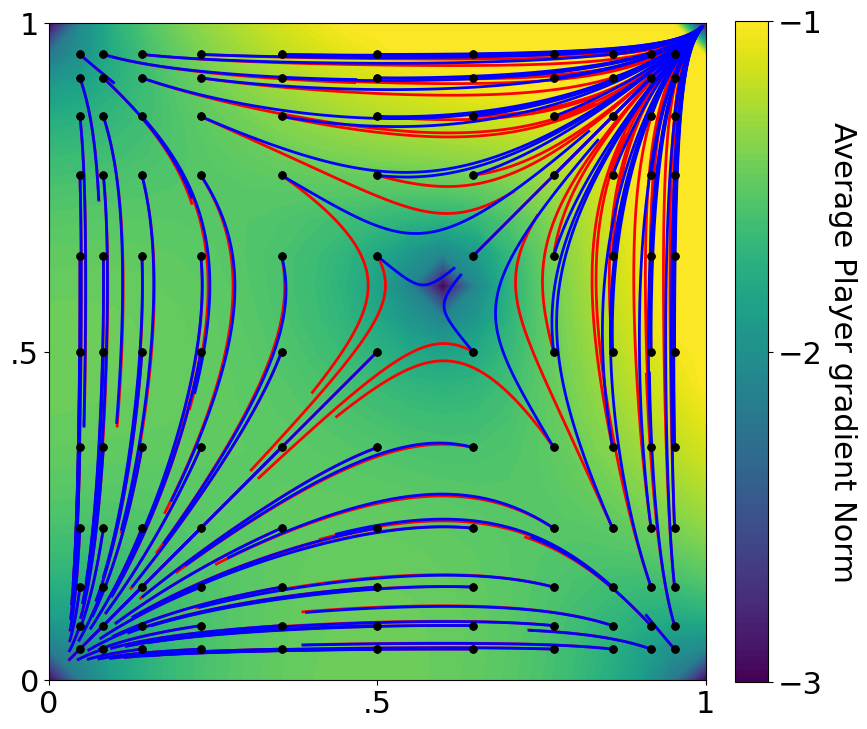}};
                    \node[left=of img, node distance=0cm, rotate=90, xshift=1.5cm, yshift=-.9cm, font=\color{black}] {Player 1 Strategy};
                    \node[below=of img, node distance=0cm, xshift=-.0cm, yshift=1.25cm,font=\color{black}] {Player 2 Strategy};
                    \node[above=of img, node distance=0cm, xshift=-.0cm, yshift=-1.25cm,font=\color{black}] {Small IPD};
                    
                    \node [right=of img, xshift=-1.25cm] (img2){\includegraphics[trim={1.0cm .25cm 3.5cm .25cm},clip, width=.295\linewidth]{latex/lyapunovForDiversity/images/standard_methods/small_mixture_grid_standard_mix=0.25.png}};
                    \node[below=of img2, node distance=0cm, xshift=-.0cm, yshift=1.25cm,font=\color{black}] {Player 2 Strategy};
                    \node[above=of img2, node distance=0cm, xshift=-.0cm, yshift=-1.25cm,font=\color{black}] {Mixed Problem};
                    
                    \node [right=of img2, xshift=-1.25cm] (img3){\includegraphics[trim={1.0cm .25cm 1.05cm .25cm},clip, width=.335\linewidth]{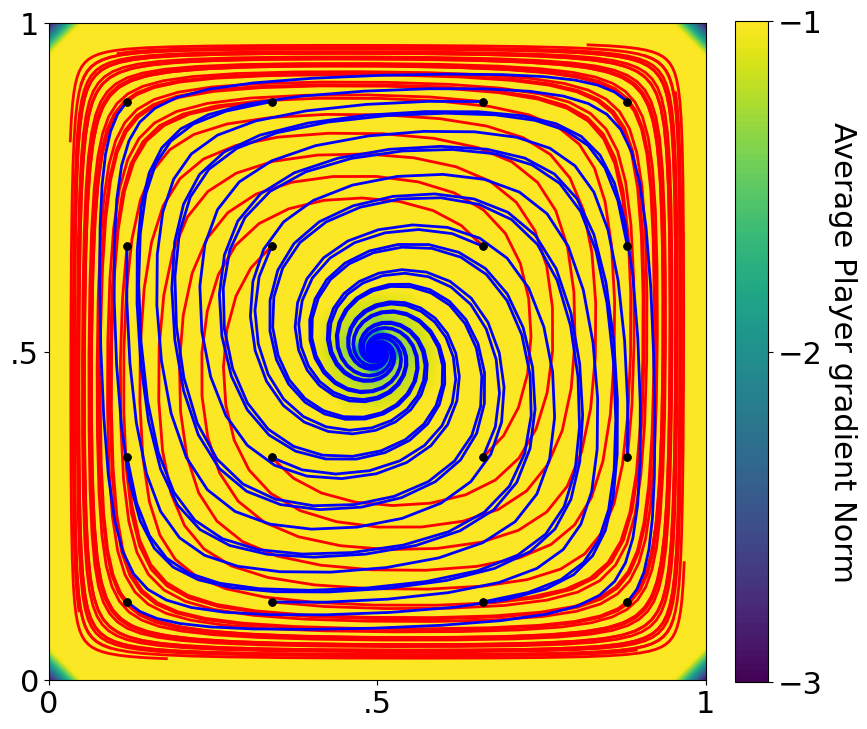}};
                    \node[right=of img3, node distance=0cm, rotate=270, xshift=-2.5cm, yshift=-.9cm, font=\color{black}] {\footnotesize Joint-player grad. log-norm $\log(\|\bothGrad\|)$};
                    \node[below=of img3, node distance=0cm, xshift=-.35cm, yshift=1.25cm,font=\color{black}] {Player 2 Strategy};
                    \node[above=of img3, node distance=0cm, xshift=-.35cm, yshift=-1.25cm,font=\color{black}] {Matching Pennies};
                \end{tikzpicture}
                \caption[Phase portrait for standard methods on various problems]{
                    This shows the phase portrait for two standard optimization algorithms on various problems.
                    Following the gradient is shown in {\color{red}red}, while LOLA -- a method for learning in games -- is shown in {\color{blue}blue}.
                    \emph{Left}: The small IPD has solutions in the top right and bottom left.
                    \emph{Middle}: Matching Pennies, which has a single solution in the middle.
                    Following the gradient does not find this solution because it has imaginary eigenvalues, so we must use a method like LOLA.
                    \emph{Right}: A mixture of small IPD and Matching Pennies.
                    Following the gradient only finds the solution in the top right because the center solution has imaginary eigenvalues;
                    LOLA can find either solution.
                    \textbf{Takeaway}: The mixture game has a range of phenomena, including an imaginary eigenvalue solution, a real eigenvalue solution, and a Hopf bifurcation.
                    We may want to use a method for learning in games to converge robustly to different solutions.
                }\label{lyapunov_fig:fig_new_problems_baseline_full}
            \end{figure*}
        
            \begin{figure*}
                \vspace{-0.01\textheight}
                \centering
                \begin{tikzpicture}
                    \centering
                    \node (img11){\includegraphics[trim={.25cm .25cm 5.0cm 1.25cm},clip, width=.22\linewidth]{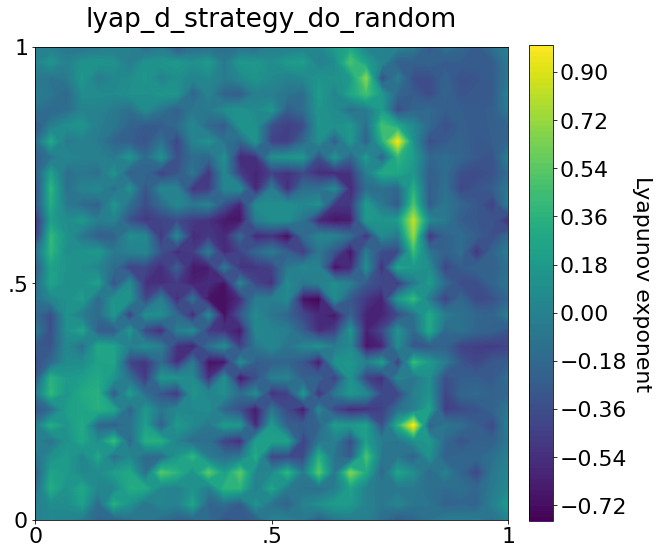}};
                        \node[left=of img11, node distance=0cm, rotate=90, xshift=1.5cm, yshift=-.9cm, font=\color{black}] {Player 1 Strategy};
                        \node[below=of img11, node distance=0cm, xshift=6.0cm,yshift=1.0cm,font=\color{black}] {Player 2 Strategy};
                        \node[above=of img11, node distance=0cm, xshift=6.0cm,yshift=-.75cm,font=\color{black}] {Different direction estimation strategies in $10$-step Lyapunov exponent calculation};
                        \node[above=of img11, node distance=0cm, xshift=-.15cm, yshift=-1.2cm,font=\color{black}] {\footnotesize Random Direction};
                    
                    \node [right=of img11, xshift=-1.1cm](img12){\includegraphics[trim={.25cm .25cm 5.2cm 1.25cm},clip, width=.22\linewidth]{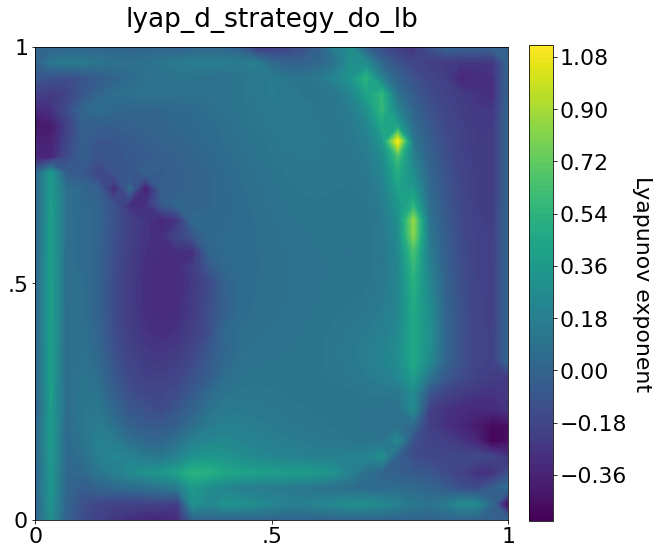}};
                        \node[above=of img12, node distance=0cm, xshift=-.1cm, yshift=-1.2cm,font=\color{black}] {\footnotesize Power Iteration, first step};
                    
                    \node [right=of img12, xshift=-1.1cm](img13){\includegraphics[trim={.25cm .25cm 5.0cm 1.25cm},clip, width=.22\linewidth]{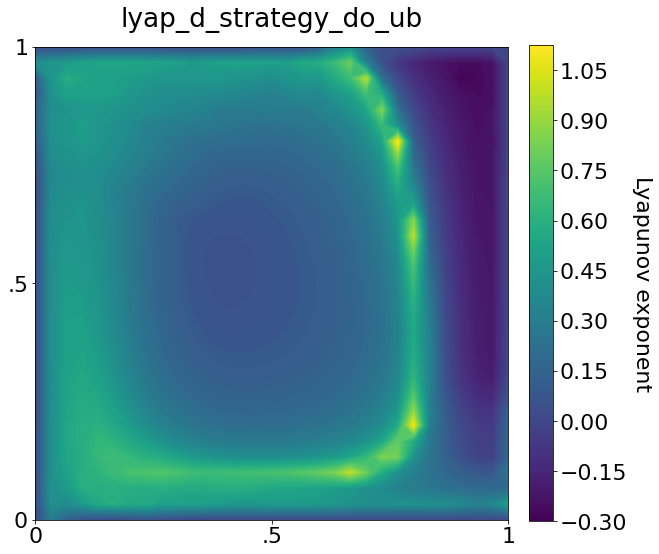}};
                        \node[above=of img13, node distance=0cm, xshift=-.05cm, yshift=-1.2cm,font=\color{black}] {\footnotesize Power Iteration, every step};
                    
                    \node [right=of img13, xshift=-1.1cm](img14){\includegraphics[trim={.25cm .25cm 1.05cm 1.25cm},clip, width=.26\linewidth]{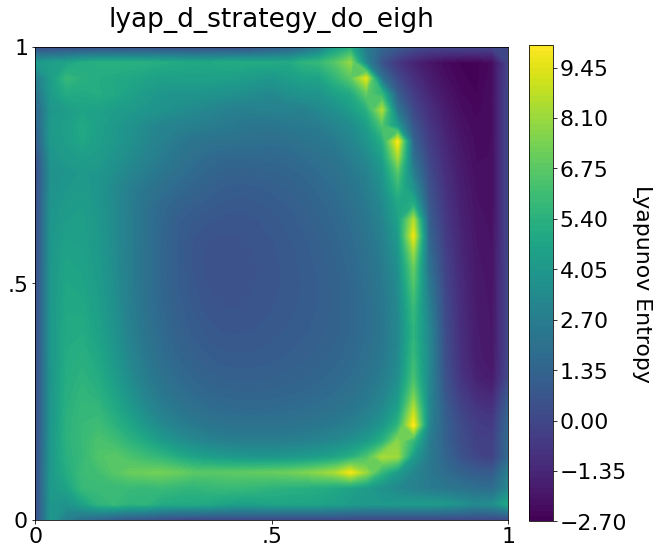}};
                        \node[above=of img14, node distance=0cm, xshift=-.25cm, yshift=-1.2cm,font=\color{black}] {\tiny{\texttt{jax.numpy.linalg.eigh}} \footnotesize{every step}};
                    \node[right=of img14, node distance=0cm, rotate=270, xshift=-.25cm, xshift=-2cm, yshift=-.9cm, font=\color{black}] {\scriptsize Max $10$-step Lyapunov Exponent};
                \end{tikzpicture}
                \caption[Different direction estimation strategies in 10-step Lyapunov exponent calculation]{
                    We compare different methods for choosing a direction in the max $10$-step Lyapunov exponent calculation on the Mixed Objective.
                    \textbf{\emph{Takeaway:}} Re-estimating the top eigenvectors at each iteration performs best but is most expensive.
                    \emph{Left}: We uniformly sample a random normalized direction for our displacement $\displacement$ at each exponent calculation, which does not clearly show the bifurcation.
                    \emph{Middle}: We perform $10$ steps of power iteration to tune $\displacement$.
                    First, we tune the displacement only at the first iteration, which shows the bifurcation.
                    Next, we tune the displacement at every iteration in the exponent calculation, which shows the bifurcation more clearly.
                    \emph{Right}: We use \texttt{jax.numpy.linalg.eigh} to tune $\displacement$, which also clearly shows the bifurcation.
                }\label{lyapunov_fig:lyap_dir_mix}
            \end{figure*}
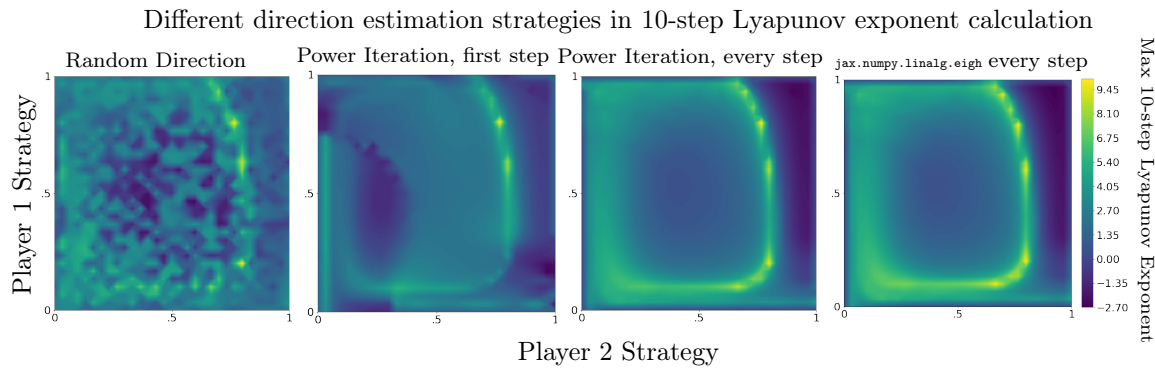
            
            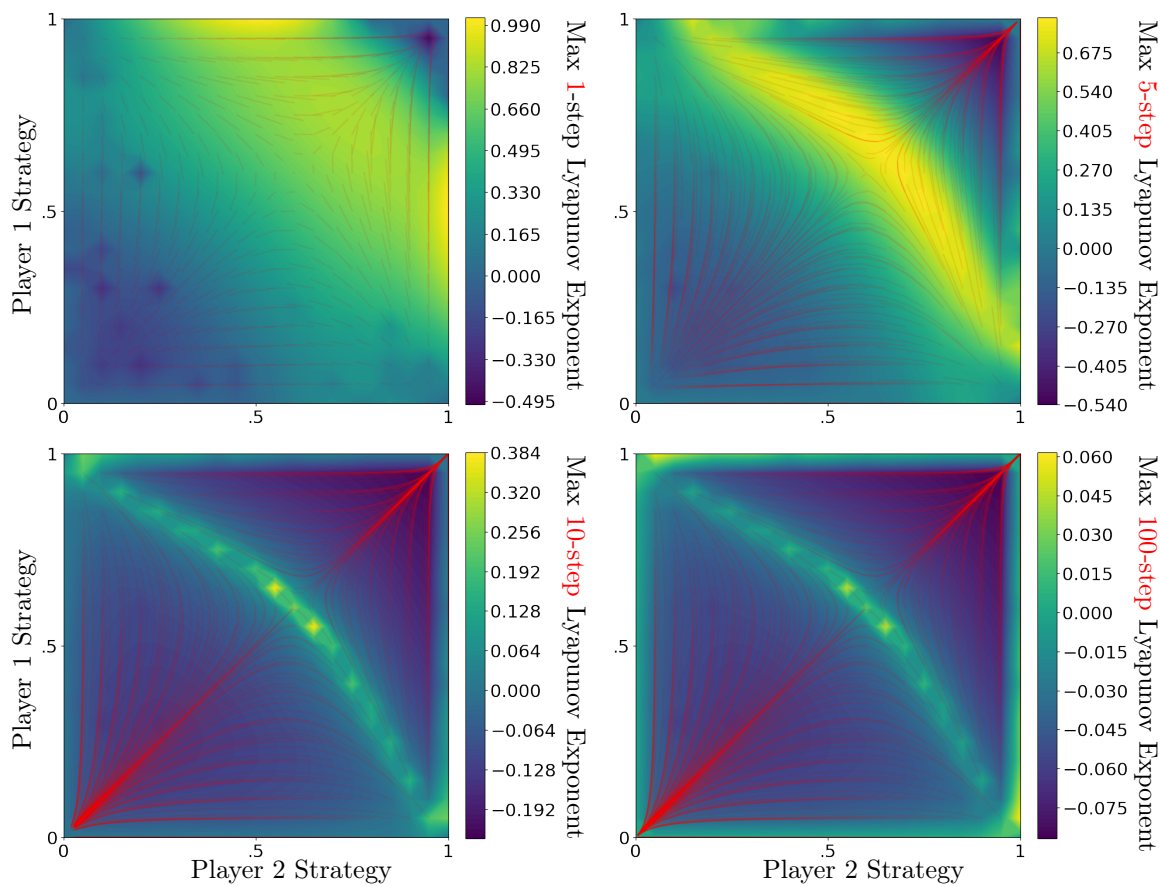
\begin{figure*}
                \vspace{-0.02\textheight}
                \centering
                \begin{tikzpicture}
                    \centering
                    \node (img11){\includegraphics[trim={.25cm .25cm 1.05cm 1.25cm},clip, width=.45\linewidth]{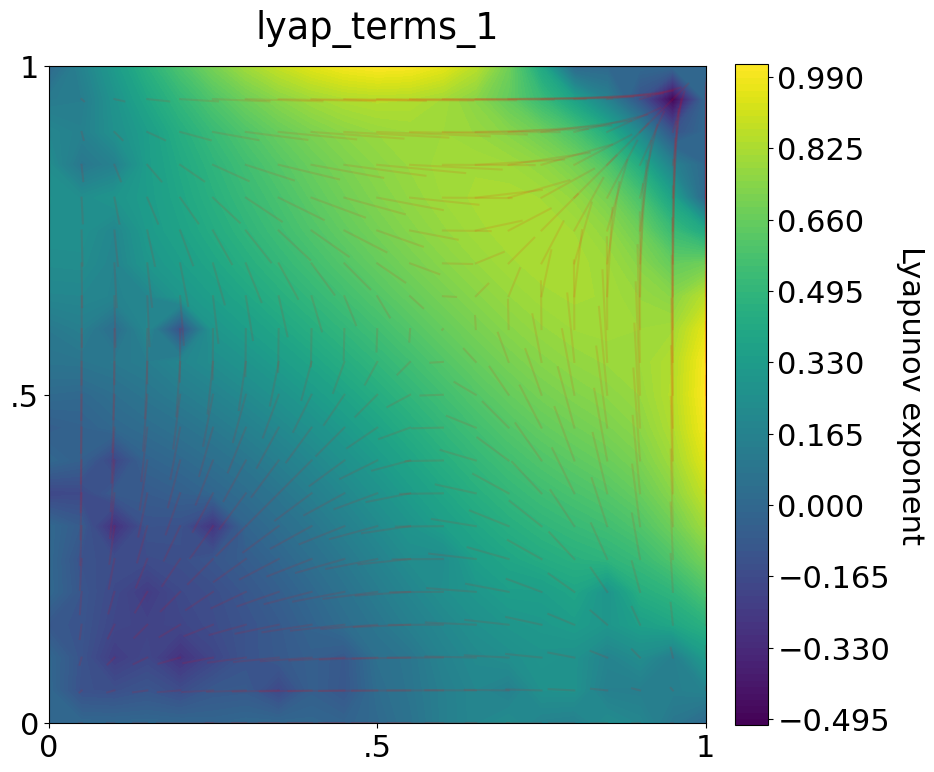}};
                        \node[left=of img11, node distance=0cm, rotate=90, xshift=1.5cm, yshift=-.9cm, font=\color{black}] {Player 1 Strategy};
                        \node[right=of img11, node distance=0cm, rotate=270, xshift=-2.75cm, yshift=-.9cm, font=\color{black}] {Max {\color{red}$1$}-step Lyapunov Exponent};
                    
                    \node [right=of img11, xshift=-.5cm](img12){\includegraphics[trim={.25cm .25cm 1.05cm 1.25cm},clip, width=.45\linewidth]{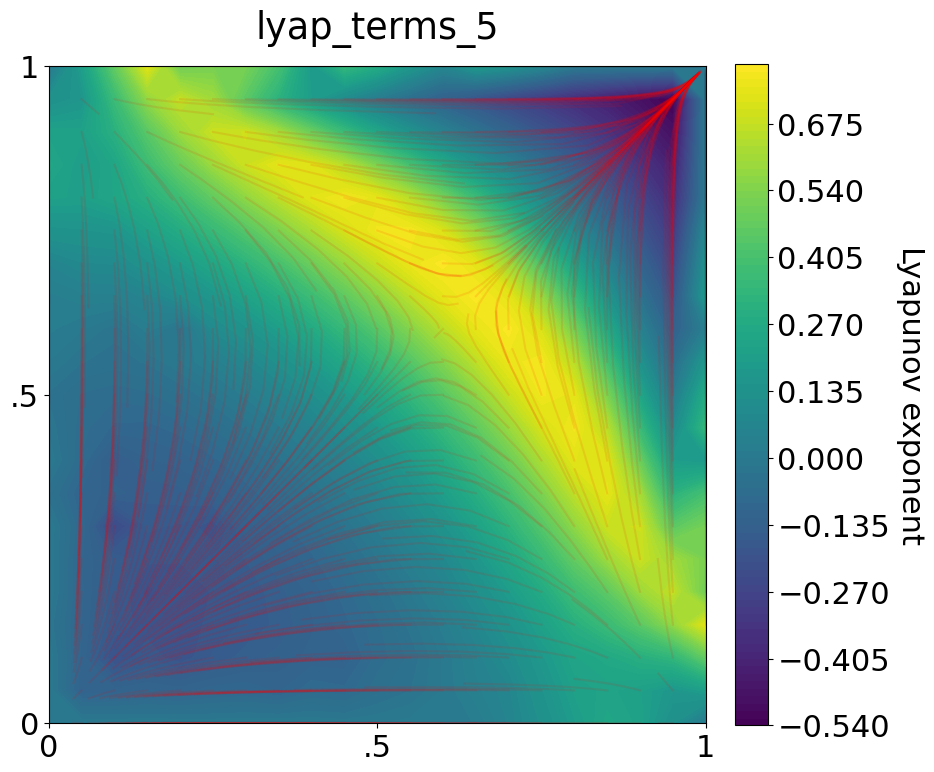}};
                        \node[right=of img12, node distance=0cm, rotate=270, xshift=-2.75cm, yshift=-.9cm, font=\color{black}] {Max {\color{red}$5$-step} Lyapunov Exponent};
                    
                    \node [below=of img11, xshift=-0cm, yshift=1cm](img21){\includegraphics[trim={.25cm .25cm 1.05cm 1.25cm},clip, width=.45\linewidth]{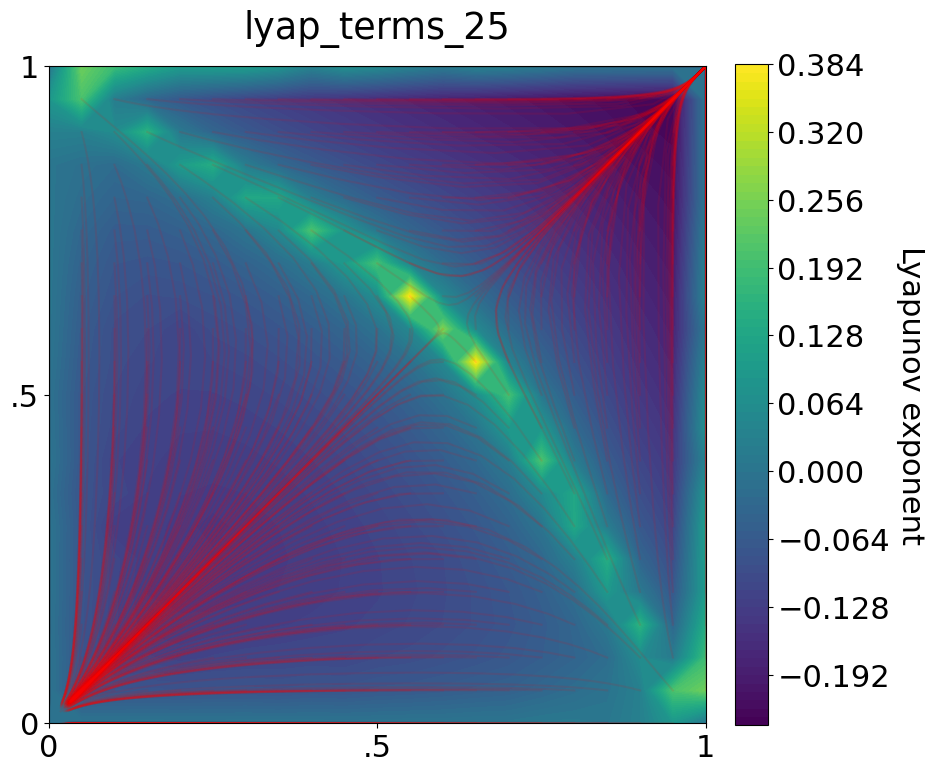}};
                        \node[right=of img21, node distance=0cm, rotate=270, xshift=-2.75cm, yshift=-.9cm, font=\color{black}] {Max {\color{red}$10$-step} Lyapunov Exponent};
                        \node[left=of img21, node distance=0cm, rotate=90, xshift=1.5cm, yshift=-.9cm, font=\color{black}] {Player 1 Strategy};
                        \node[below=of img21, node distance=0cm, xshift=-.1cm,yshift=1.25cm,font=\color{black}] {Player 2 Strategy};
                        
                    \node [right=of img21, xshift=-.5cm](img22){\includegraphics[trim={.25cm .25cm 1.05cm 1.25cm},clip, width=.45\linewidth]{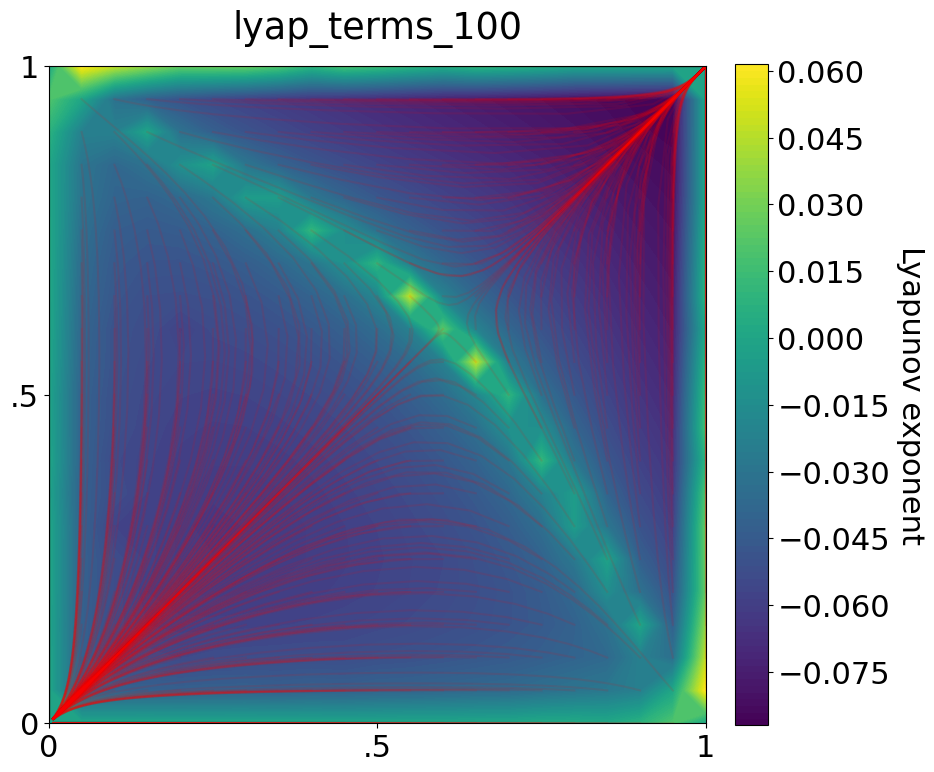}};
                        \node[right=of img22, node distance=0cm, rotate=270, xshift=-2.75cm, yshift=-.9cm, font=\color{black}] {Max {\color{red}$100$-step} Lyapunov Exponent};
                        \node[below=of img22, node distance=0cm, xshift=-.1cm,yshift=1.25cm,font=\color{black}] {Player 2 Strategy};
                \end{tikzpicture}
                \caption[Visualizing Lyapunov exponent calculations for various step numbers]{
                    We show the $k$-step Lyapunov exponent calculation for various $k$ on the small IPD.
                    The $k$-step optimization trajectories used for calculating the exponent are shown in {\color{red}red}, allowing us to see the horizon over which the associated exponent measures separation.
                    \textbf{\emph{Takeaway:}} We can effectively find bifurcations with a $k$-step exponent for various $k$.
                    Using only $1$-step does not show the bifurcation, while using more does.
                    As $k$ gets larger, the scale of the gradients changes, causing difficult optimization in the limit --  note the changing color bar scale.
                }\label{lyapunov_fig:lyap_step_mix}
                \vspace{-0.03\textheight}
            \end{figure*}
            
            \begin{figure*}
                \centering
                \begin{tikzpicture}
                    \centering
                    \node (img){\includegraphics[trim={.25cm .25cm 1.05cm 1.25cm},clip, width=.67\linewidth]{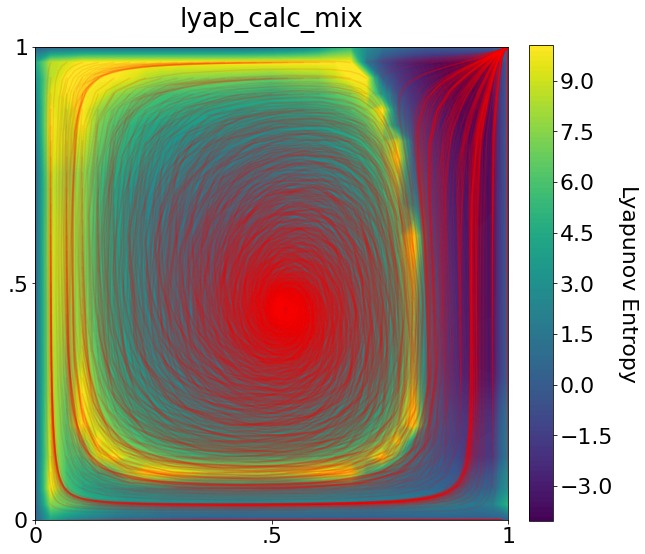}};
                    \node[left=of img, node distance=0cm, rotate=90, xshift=1.5cm, yshift=-.9cm, font=\color{black}] {Player 1 Strategy};
                    \node[right=of img, node distance=0cm, rotate=270, xshift=3cm, yshift=-.9cm, font=\color{black}] {Max $10$-step Lyapunov Exponent};
                    \node[above=of img, node distance=0cm, xshift=-.35cm, yshift=-1.2cm,font=\color{black}] {LOLA};
                    
                    \node [below=of img, yshift=1.25cm] (img2){\includegraphics[trim={.25cm .25cm 1.05cm 1.25cm},clip, width=.67\linewidth]{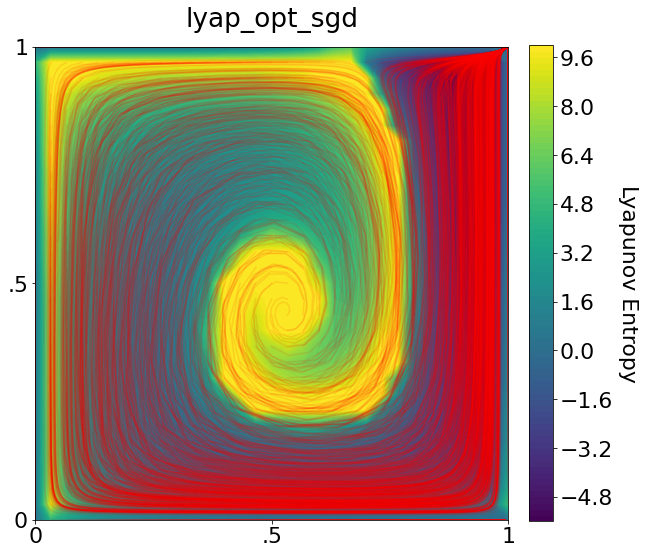}};
                    \node[left=of img2, node distance=0cm, rotate=90, xshift=1.5cm, yshift=-.9cm, font=\color{black}] {Player 1 Strategy};
                    \node[below=of img2, node distance=0cm, xshift=-.25cm, yshift=1.25cm,font=\color{black}] {Player 2 Strategy};
                    \node[below=of img2, node distance=0cm, xshift=-.25cm, yshift=.75cm,font=\color{black}] {SimSGD};
                \end{tikzpicture}
                \caption[Contrasting Lyapunov exponent calculations between LOLA and simSGD]{
                    We contrast the max $20$-step Lyapunov exponent calculations between LOLA and \ref{lyapunov_eq:multi_objective_gd_long}.
                    The $20$-step optimization trajectories used for calculating the exponent are shown in {\color{red}red}, allowing us to see the trajectories in which the associated exponent measures separation.
                    \textbf{Takeaway:} We can find optimizer-dependent bifurcations with our method.
                    \emph{Top:} $10$ steps from the optimization with LOLA at each point where we calculate an exponent for the heatmap.
                    \emph{Bottom:} The same visualization as the top, except with a \ref{lyapunov_eq:multi_objective_gd_long} optimizer.
                }\label{lyapunov_fig:lyap_opt_mix}
            \end{figure*}
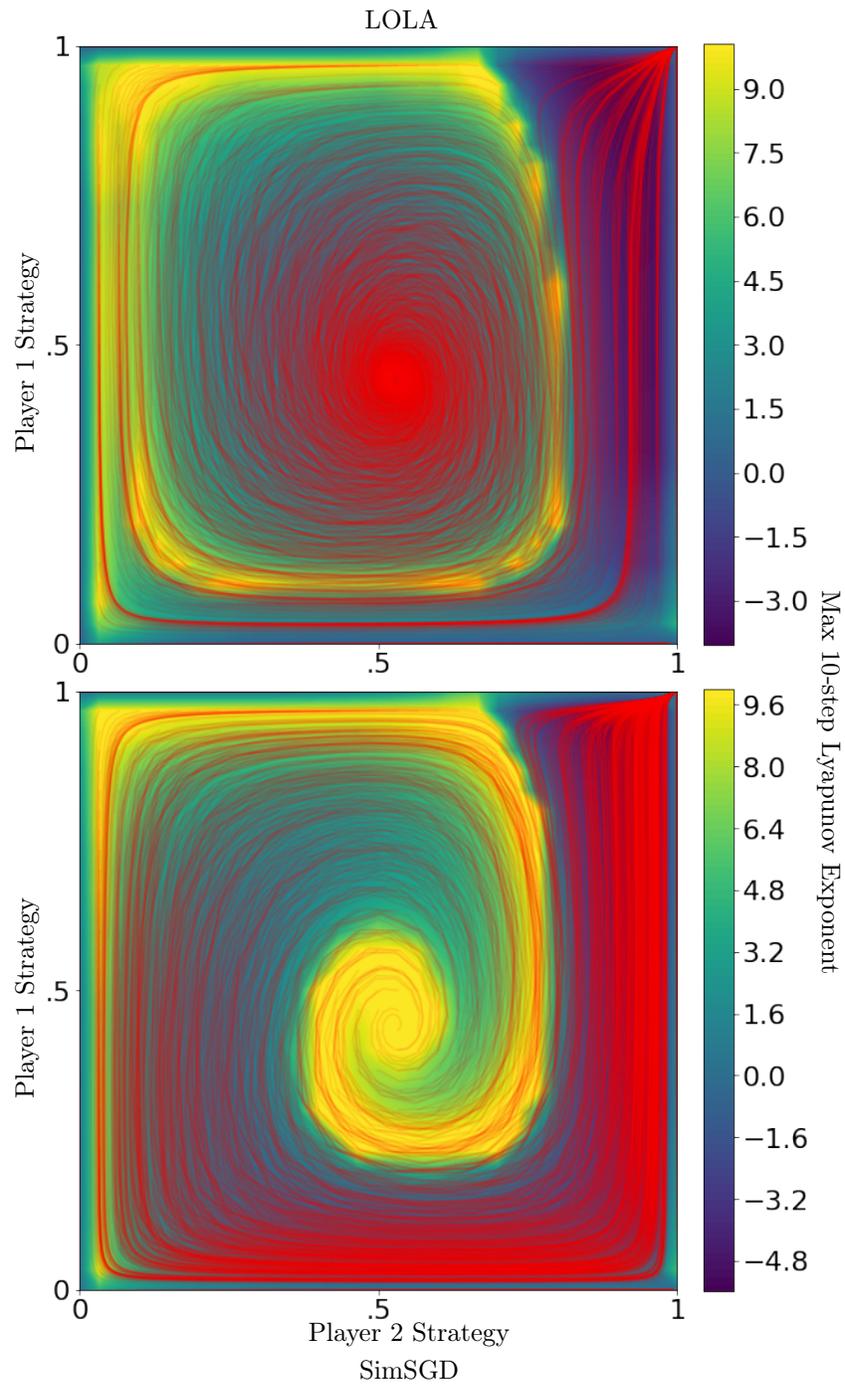
            
            \begin{figure*}
                \centering
                \begin{tikzpicture}
                    \centering
                    \node (img){\includegraphics[trim={.25cm .25cm 1.05cm 1.25cm},clip, width=.63\linewidth]{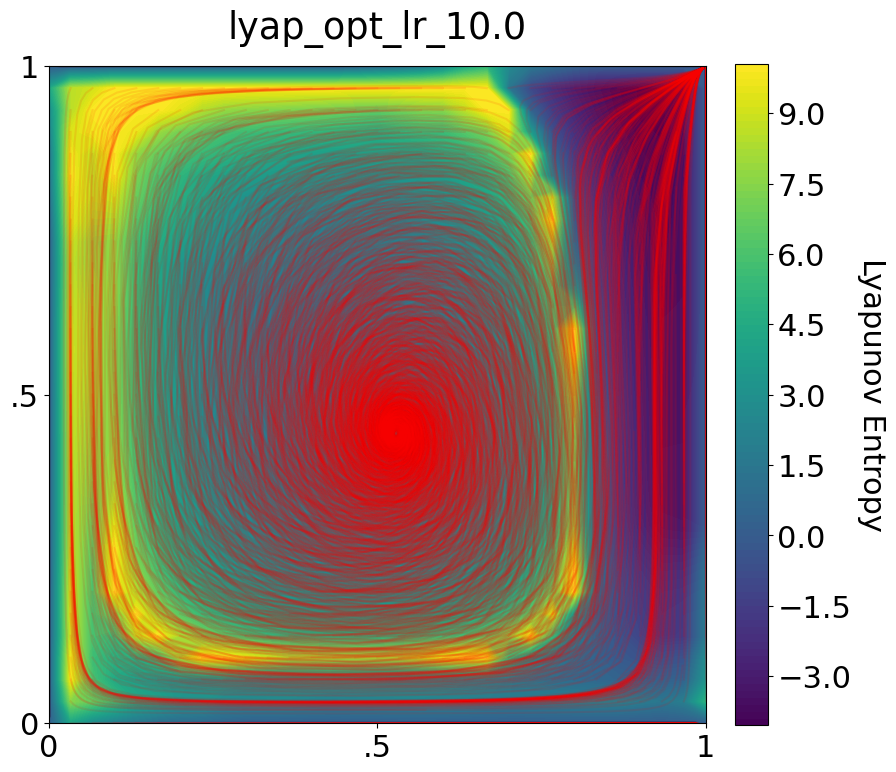}};
                    \node[left=of img, node distance=0cm, rotate=90, xshift=1.5cm, yshift=-.9cm, font=\color{black}] {Player 1 Strategy};
                    \node[right=of img, node distance=0cm, rotate=270, xshift=1.5cm, yshift=-.9cm, font=\color{black}] {Max $10$-step Lyapunov Exponent};
                    \node[above=of img, node distance=0cm, xshift=-.35cm, yshift=-1.2cm,font=\color{black}] {Step size {\color{red}$\alpha = 10$}};
                    
                    \node [below=of img, yshift=1.25cm] (img2){\includegraphics[trim={.25cm .25cm 1.05cm 1.25cm},clip, width=.63\linewidth]{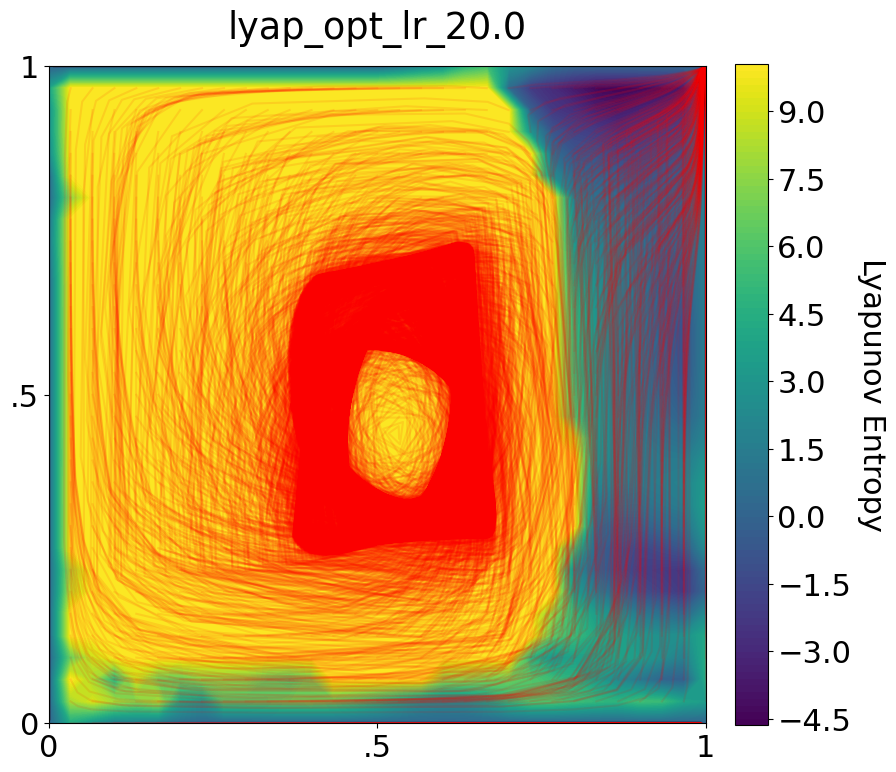}};
                    \node[left=of img2, node distance=0cm, rotate=90, xshift=1.5cm, yshift=-.9cm, font=\color{black}] {Player 1 Strategy};
                    \node[below=of img2, node distance=0cm, xshift=-.25cm, yshift=1.25cm,font=\color{black}] {Player 2 Strategy};
                    \node[below=of img2, node distance=0cm, xshift=-.25cm, yshift=.75cm,font=\color{black}] {Step size {\color{red}$\alpha = 20$}};
                \end{tikzpicture}
                \caption[Visualizing the impact of optimization algorithm parameters on the Lyapunov exponent]{
                    We investigate the impact of optimization algorithm parameters on the $10$-step max Lyapunov exponent calculation with LOLA.
                    The $10$-step optimization trajectories used for calculating the exponent are shown in {\color{red}red}, allowing us to see the trajectories in which the associated exponent measures separation.
                    \textbf{Takeaway:} If the step size is too large, the optimizer does not converge to solutions, resulting in limit cycles.
                    \emph{Top:} $10$ steps from the optimization with a step size of $\alpha = 10$ at each point where we calculate an exponent for the heatmap.
                    \emph{Bottom:} The same visualization as the top, except with a step size of $\alpha = 20$.
                    Optimization trajectories only converge at the solution in the top right.
                    In the center, the trajectories accumulate in a limit cycle, rotating around the solution.
                }\label{lyapunov_fig:lyap_opt_param_mix}
            \end{figure*}
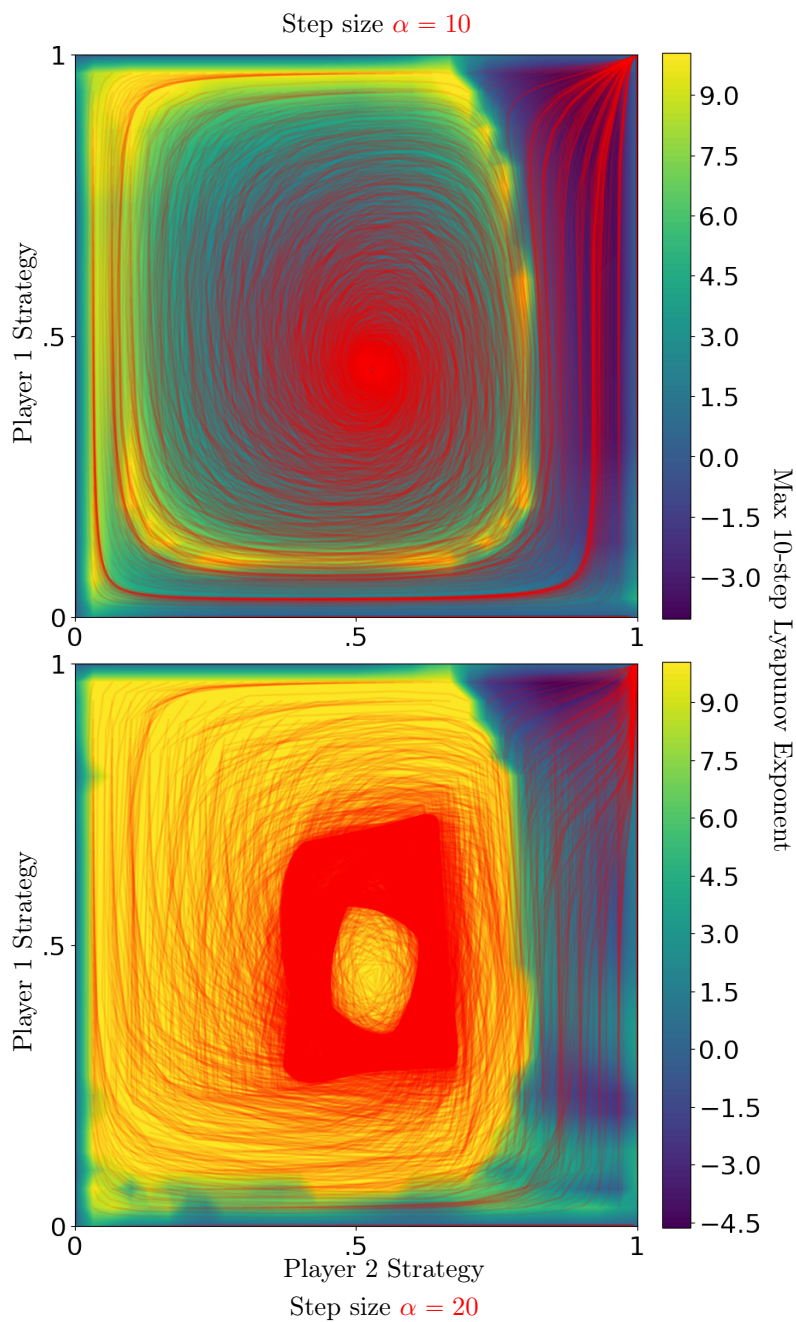
            
            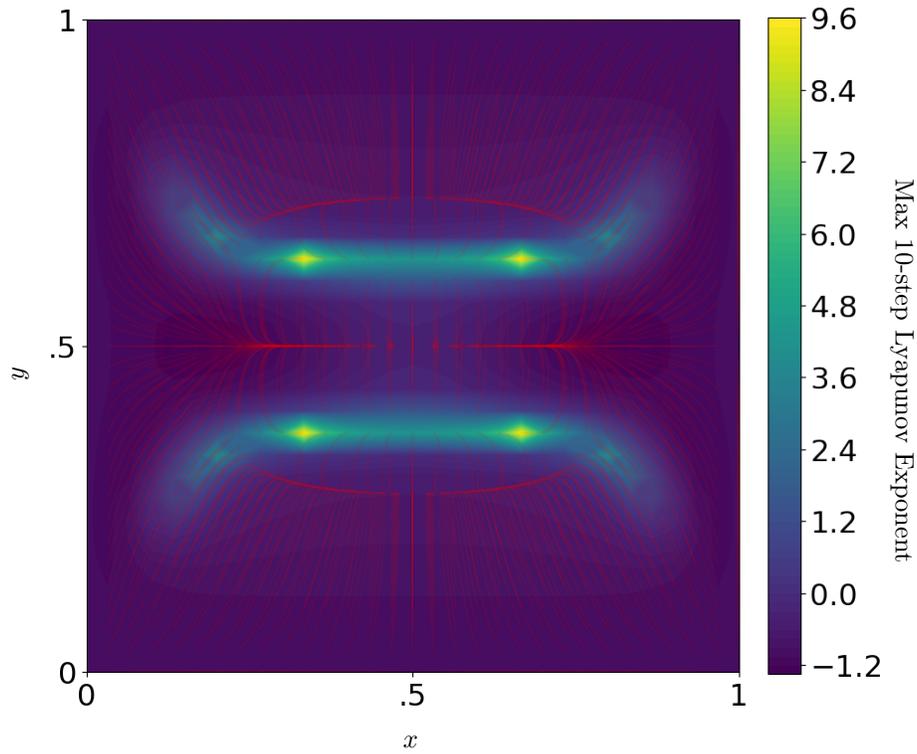
\begin{figure*}
                \centering
                \begin{tikzpicture}
                    \centering
                    \node (img){\includegraphics[trim={.25cm .25cm 1.05cm 1.25cm},clip, width=.73\linewidth]{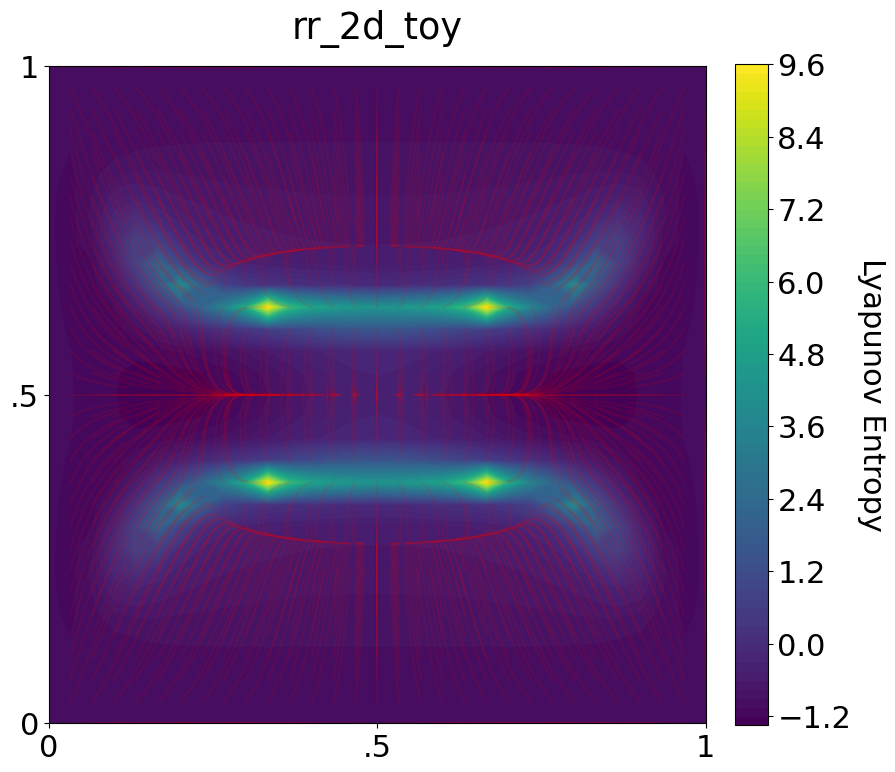}};
                    \node[left=of img, node distance=0cm, rotate=90, xshift=0cm, yshift=-.75cm, font=\color{black}] {$y$};
                    \node[right=of img, node distance=0cm, rotate=270, xshift=-2.5cm, yshift=-.9cm, font=\color{black}] {Max $10$-step Lyapunov Exponent};
                    \node[below=of img, node distance=0cm, xshift=-.75cm, yshift=.9cm,font=\color{black}] {$x$};
                \end{tikzpicture}
                \caption[Using Lyapunov exponents on single objective optimization problems]{
                    We investigate finding bifurcations with a $10$-step max Lyapunov exponent on a single-objective optimization problem from RR.
                    The optimization trajectories used for the calculation of the exponent are shown in {\color{red}red}, allowing us to verify where the bifurcations are.
                    \textbf{Takeaway:} Our method effectively highlights bifurcations in RRs example.
                }\label{lyapunov_fig:lyap_coop_toy}
            \end{figure*}
            
            \begin{figure*}
                \centering
                \begin{tikzpicture}
                    \centering
                    \node (img){\includegraphics[trim={.25cm 1.0cm 1.05cm 1.14cm},clip, width=.8\linewidth]{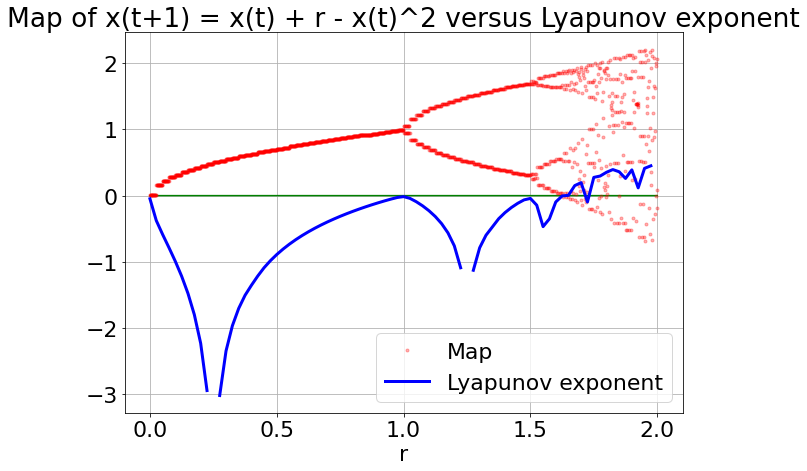}};
                    \node[left=of img, node distance=0cm, rotate=90, xshift=1.0cm, yshift=-2.2cm, font=\color{black}] {$x_0$};
                    \node[below=of img, node distance=0cm, xshift=-.35cm, yshift=1.0cm,font=\color{black}] {$r$};
                \end{tikzpicture}
                \caption[Lyapunov exponent of the logistic map]{
                    We display the Lyapunov exponent on the logistic map: $x(t+1) = x(t) + r + x(t)^2$.
                    \textbf{Takeaway:} Intuition for Lyapunov exponents on a canonical 1-dimensional example for bifurcations.
                }\label{lyapunov_fig:lyap_oneDim_toy}
            \end{figure*}
            
            \begin{figure*}
                \centering
                \begin{tikzpicture}
                    \centering
                    \node (img){\includegraphics[trim={.25cm .25cm 1.05cm 1.25cm},clip, width=.69\linewidth]{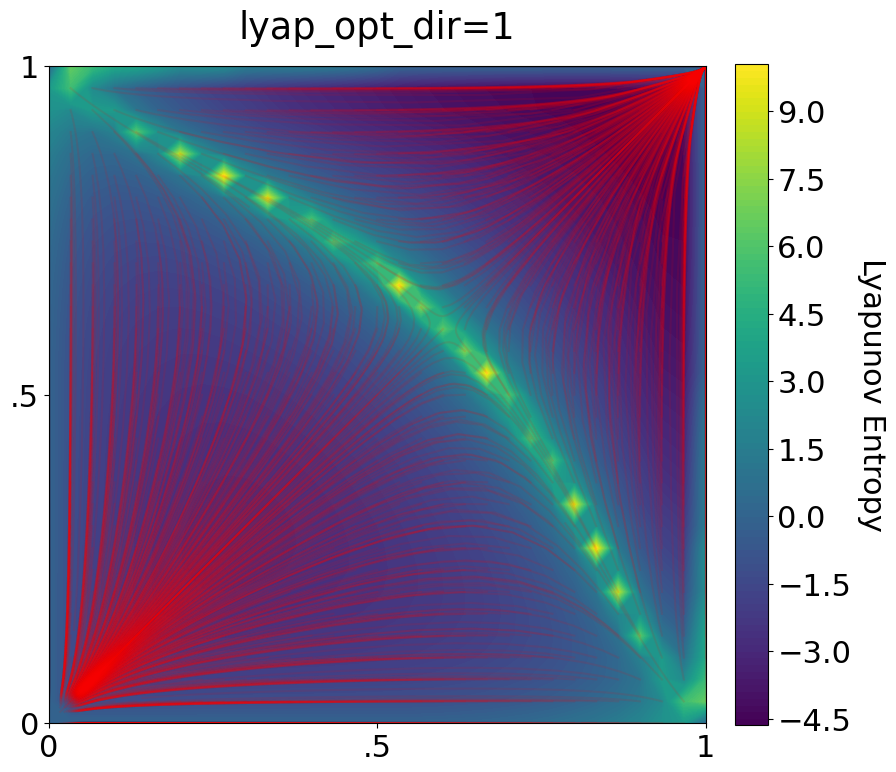}};
                        \node[left=of img, node distance=0cm, rotate=90, xshift=1.5cm, yshift=-.9cm, font=\color{black}] {Player 1 Strategy};
                        \node[right=of img, node distance=0cm, rotate=270, xshift=-3.0cm, yshift=-.9cm, font=\color{black}] {{\color{red}Max = 1-direction} Lyapunov Exponent};
                    
                    \node [below=of img, yshift=1.25cm] (img2){\includegraphics[trim={.25cm .25cm 1.05cm 1.25cm},clip, width=.69\linewidth]{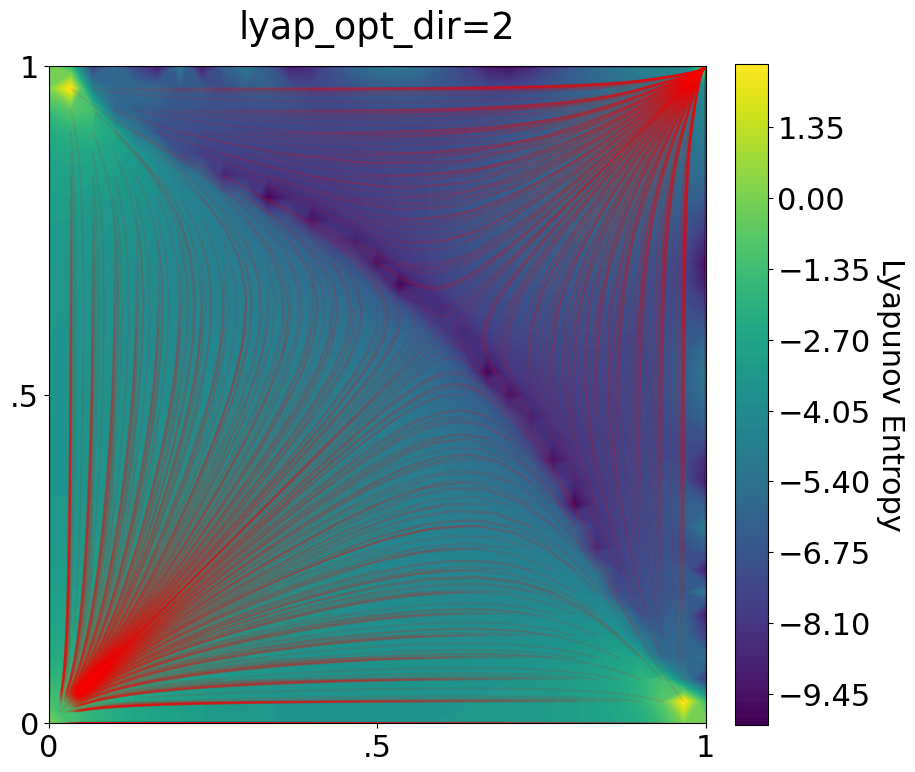}};
                        \node[left=of img2, node distance=0cm, rotate=90, xshift=1.5cm, yshift=-.9cm, font=\color{black}] {Player 1 Strategy};
                        \node[below=of img2, node distance=0cm, xshift=-.25cm, yshift=1.25cm,font=\color{black}] {Player 2 Strategy};
                        \node[right=of img2, node distance=0cm, rotate=270, xshift=-2.75cm, yshift=-.9cm, font=\color{black}] { {\color{red}Sum of top 2} Lyapunov Exponents};
                \end{tikzpicture}
                \caption[Lyapunov exponents and Lyapunov entropy visualization]{
                    We compare different $10$-step Lyapunov exponent objectives trying to guarantee divergence in multiple directions on the small IPD.
                    The optimization trajectories used for the exponent calculation are shown in {\color{red}red}, allowing us to see the trajectories for the associated objective.
                    \textbf{Takeaway:} We can find bifurcations while guaranteeing trajectory separation in every direction.
                    Local maxima -- not saddles -- allow trajectory separation in all directions here.
                    \emph{Top:} The max $10$-step Lyapunov exponent.
                    \emph{Bottom:} The sum of the top two $10$-step Lyapunov exponents.
                }\label{lyapunov_fig:entropy_mix}
            \end{figure*}
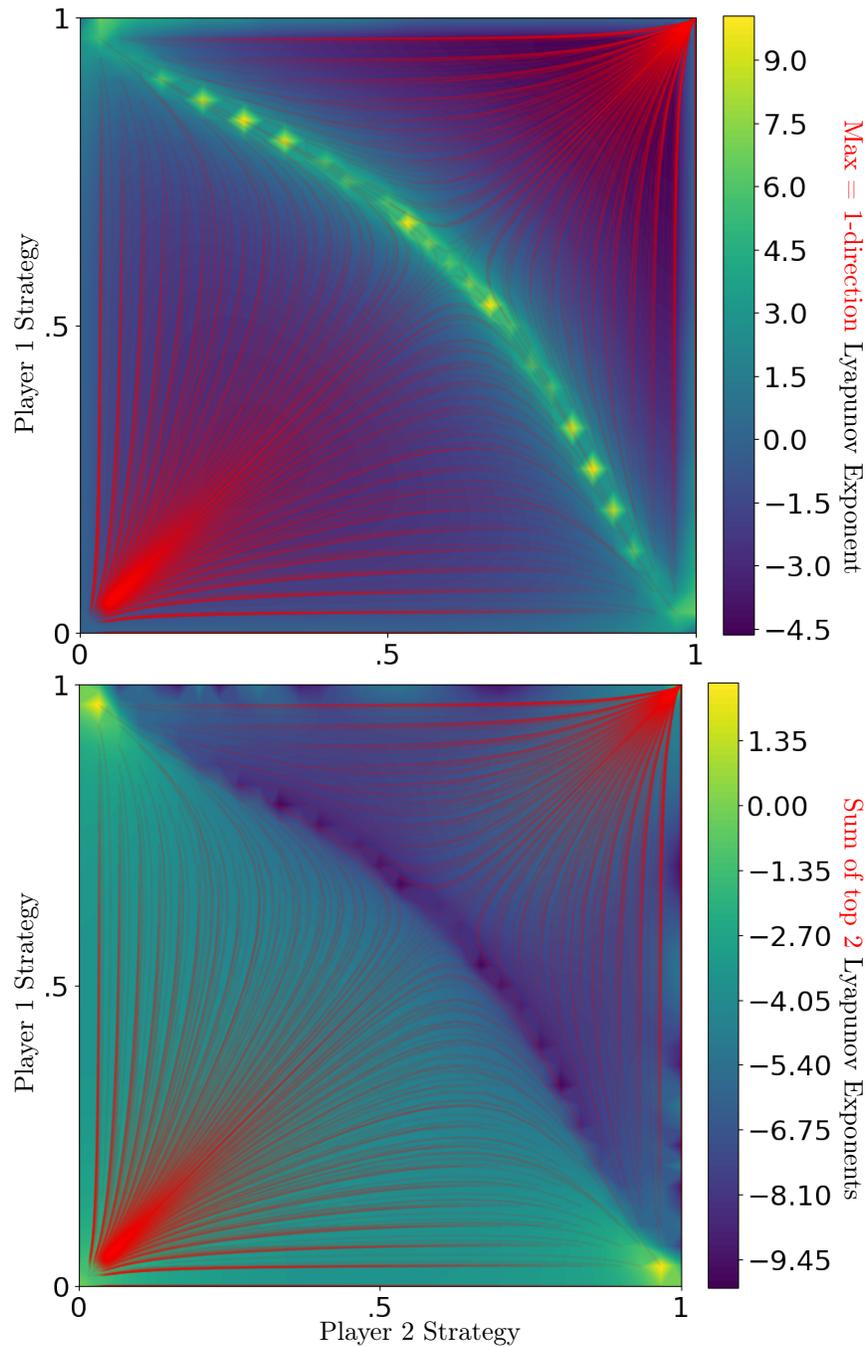
            
            \begin{figure*}
                \centering
                \includegraphics[trim={2.3cm 1.75cm .0cm 1.0cm},clip,width=.3\linewidth]{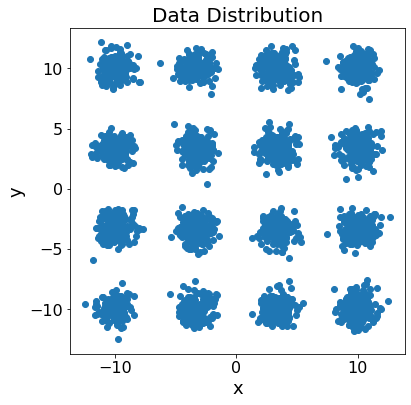}
                \caption[Ground truth samples for our mixture of Gaussians]{
                    Ground truth samples for our Mixture of Gaussian experiments.
                }
                \label{lyapunov_fig:gan_2d_samples}
            \end{figure*}
            
            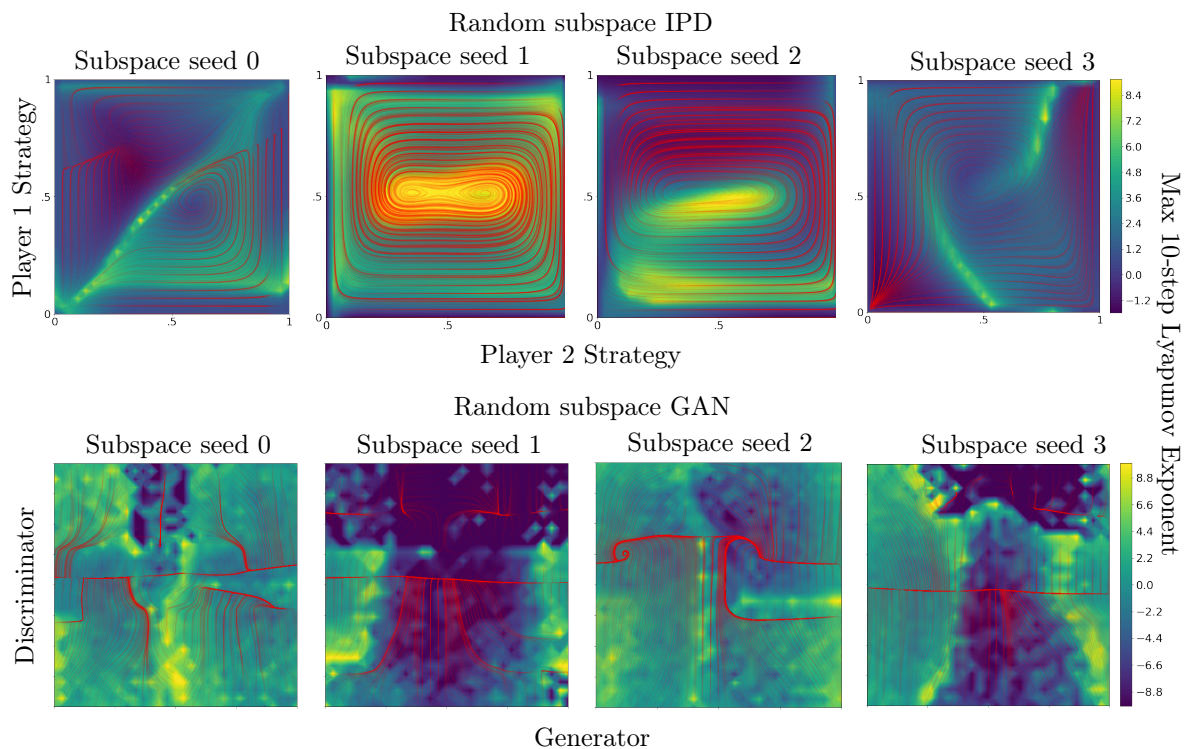
\begin{figure*}
                \centering
                \begin{tikzpicture}
                    \centering
                    \node (img11){\includegraphics[trim={.25cm .25cm 4.5cm 1.25cm},clip, width=.22\linewidth]{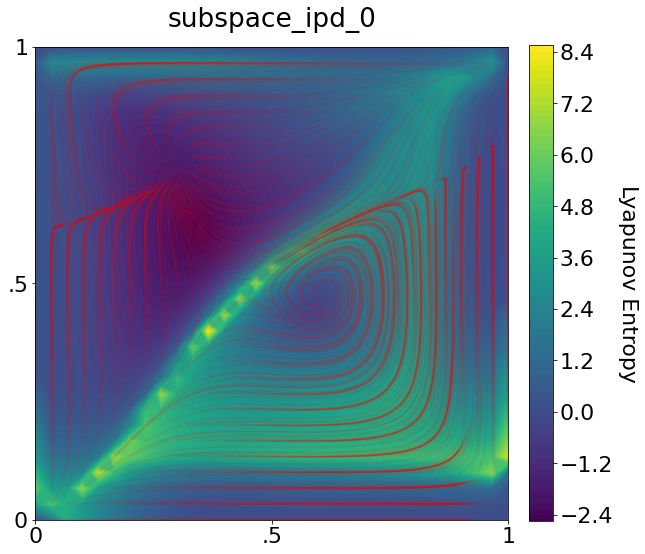}};
                        \node[left=of img11, node distance=0cm, rotate=90, xshift=1.5cm, yshift=-.9cm, font=\color{black}] {Player 1 Strategy};
                        \node[above=of img11, node distance=0cm, xshift=-.0cm, yshift=-1.25cm,font=\color{black}] {Subspace seed 0};
                    
                        \node[below=of img11, node distance=0cm, xshift=5.5cm,yshift=1.0cm,font=\color{black}] {Player 2 Strategy};
                        \node[above=of img11, node distance=0cm, xshift=5.5cm, yshift=-.75cm,font=\color{black}] {Random subspace IPD};
                    
                    \node [right=of img11, xshift=-1cm](img12){\includegraphics[trim={.25cm .25cm 4.5cm 1.25cm},clip, width=.22\linewidth]{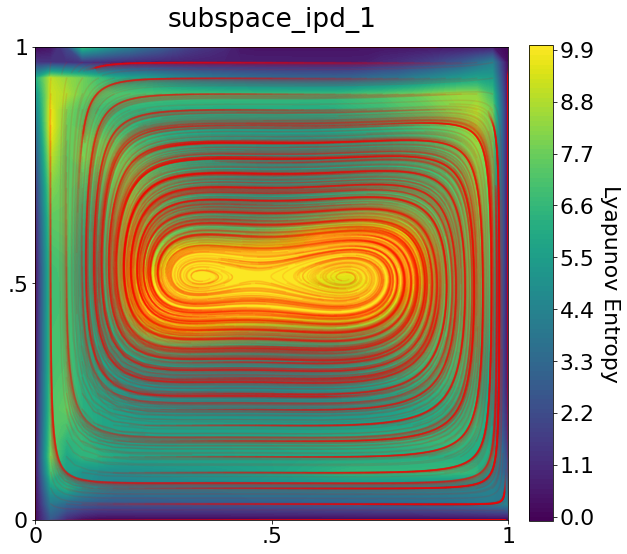}};
                        \node[above=of img12, node distance=0cm, xshift=-.0cm, yshift=-1.25cm,font=\color{black}] {Subspace seed 1};
                    
                    \node [right=of img12, xshift=-1cm](img13){\includegraphics[trim={.25cm .25cm 4.5cm 1.25cm},clip, width=.22\linewidth]{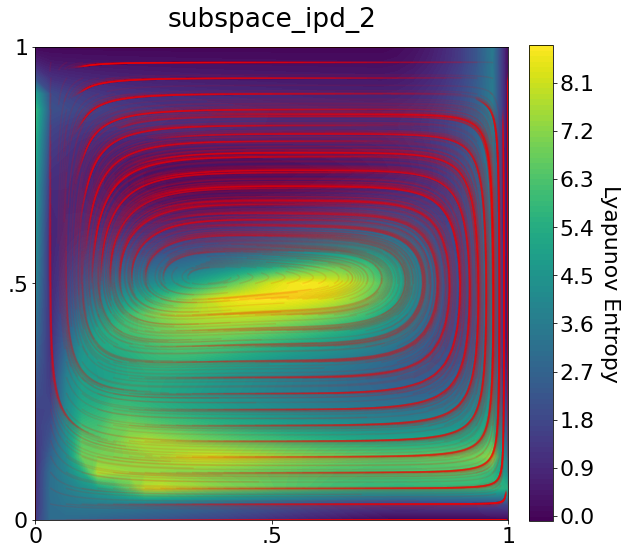}};
                        \node[above=of img13, node distance=0cm, xshift=-.0cm, yshift=-1.25cm,font=\color{black}] {Subspace seed 2};
                    
                    \node [right=of img13, xshift=-1cm](img14){\includegraphics[trim={.25cm .25cm 1.05cm 1.25cm},clip, width=.26\linewidth]{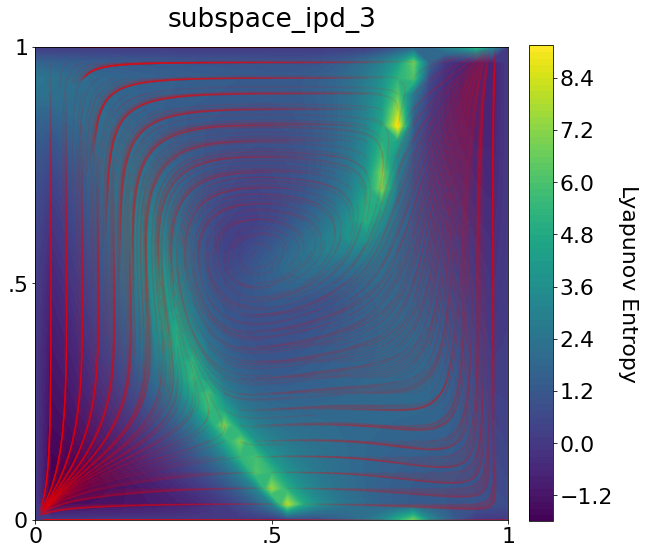}};
                        \node[above=of img14, node distance=0cm, xshift=-.0cm, yshift=-1.25cm,font=\color{black}] {Subspace seed 3};
                        
                    \node [below=of img11, xshift=.15cm, yshift=-.5cm] (img21){\includegraphics[trim={2.75cm 1.0cm 4.25cm 1.25cm},clip, width=.22\linewidth]{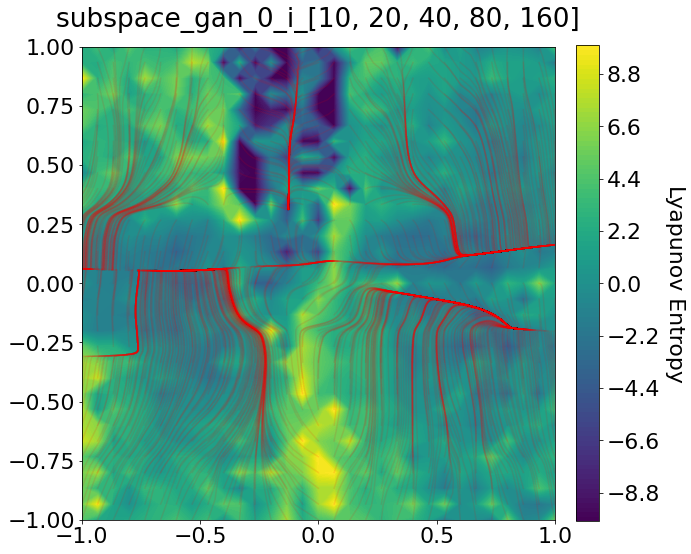}};
                        \node[left=of img21, node distance=0cm, rotate=90, xshift=1.15cm, yshift=-.75cm, font=\color{black}] {Discriminator};
                        \node[above=of img21, node distance=0cm, xshift=-.0cm, yshift=-1.25cm,font=\color{black}] {Subspace seed 0};
                        \node[below=of img21, node distance=0cm, xshift=5.5cm,yshift=1.0cm,font=\color{black}] {Generator};
                        \node[above=of img21, node distance=0cm, xshift=5.5cm, yshift=-.75cm,font=\color{black}] {Random subspace GAN};
                    
                    \node [right=of img21, xshift=-1cm](img22){\includegraphics[trim={2.75cm 1.0cm 4.25cm 1.25cm},clip, width=.22\linewidth]{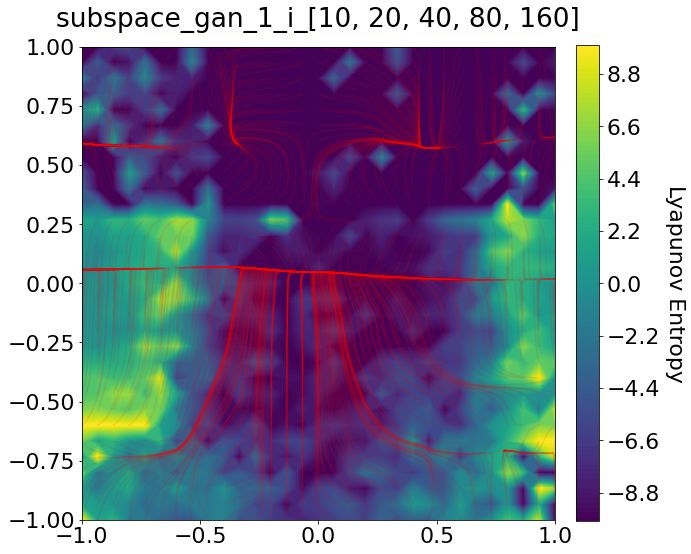}};
                        \node[above=of img22, node distance=0cm, xshift=-.0cm, yshift=-1.25cm,font=\color{black}] {Subspace seed 1};
                    
                    \node [right=of img22, xshift=-1cm](img23){\includegraphics[trim={2.75cm 1.0cm 4.25cm 1.25cm},clip, width=.22\linewidth]{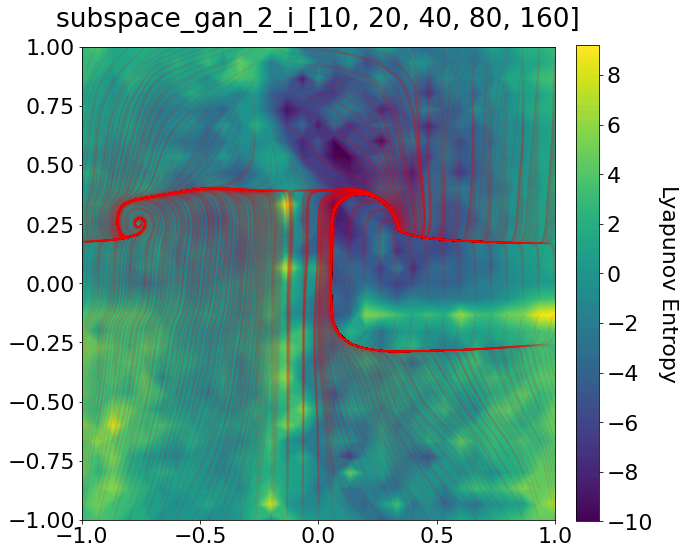}};
                        \node[above=of img23, node distance=0cm, xshift=-.0cm, yshift=-1.25cm,font=\color{black}] {Subspace seed 2};
                    
                    \node [right=of img23, xshift=-1cm](img24){\includegraphics[trim={2.75cm 1.0cm 1.05cm 1.25cm},clip, width=.26\linewidth]{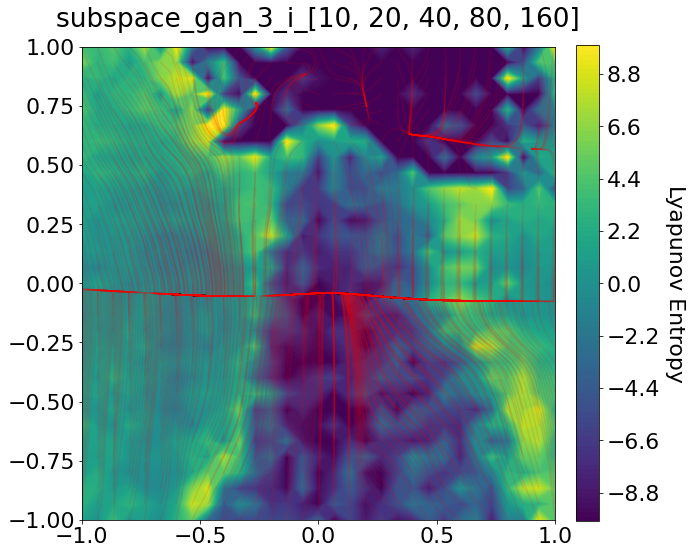}};
                        \node[above=of img24, node distance=0cm, xshift=-.0cm, yshift=-1.25cm,font=\color{black}] {Subspace seed 3};
                    
                        \node[right=of img14, node distance=0cm, rotate=270, xshift=-0.5cm, yshift=-.9cm, font=\color{black}] {Max $10$-step Lyapunov Exponent};
                \end{tikzpicture}
                \caption[Lyapunov exponent heatmaps on various toy problems]{
                    This reproduces Figure~\ref{lyapunov_fig:complicated_toy} with more random subspaces.
                    The optimization trajectories used for the exponent calculation are shown in {\color{red}red}, allowing us to see the trajectories the associated exponent measures separation over.
                    \textbf{\emph{Takeaway:}} The exponent is peaked near where trajectories separate for each subspace, showing that these strategies can find various bifurcations.
                    We display the Lyapunov exponent calculation -- as in Figure~\ref{lyapunov_fig:fig_mix} --  on more complicated toy problems to see how robustly we can find different bifurcations.
                    Section~\ref{lyapunov_subsec:new_problems} describes how we construct these examples by taking higher-dimensional problems and optimizing them in a random subspace.
                    \emph{Top:} Different IPD subspaces: we effectively highlight bifurcations -- i.e., regions where trajectory behavior qualitatively changes.
                    \emph{Bottom:} Different GAN subspaces: we can find bifurcations, but the highlighted structure is less crisp in this more complicated and stochastic toy example.
                }\label{lyapunov_fig:complicated_toy_moreSeeds}
            \end{figure*}
        
            \begin{figure*}
                \centering
                \begin{tikzpicture}
                    \centering
                    \node (img11){\includegraphics[trim={1.0cm 1.85cm 55.0cm 1.15cm},clip, width=.45\linewidth]{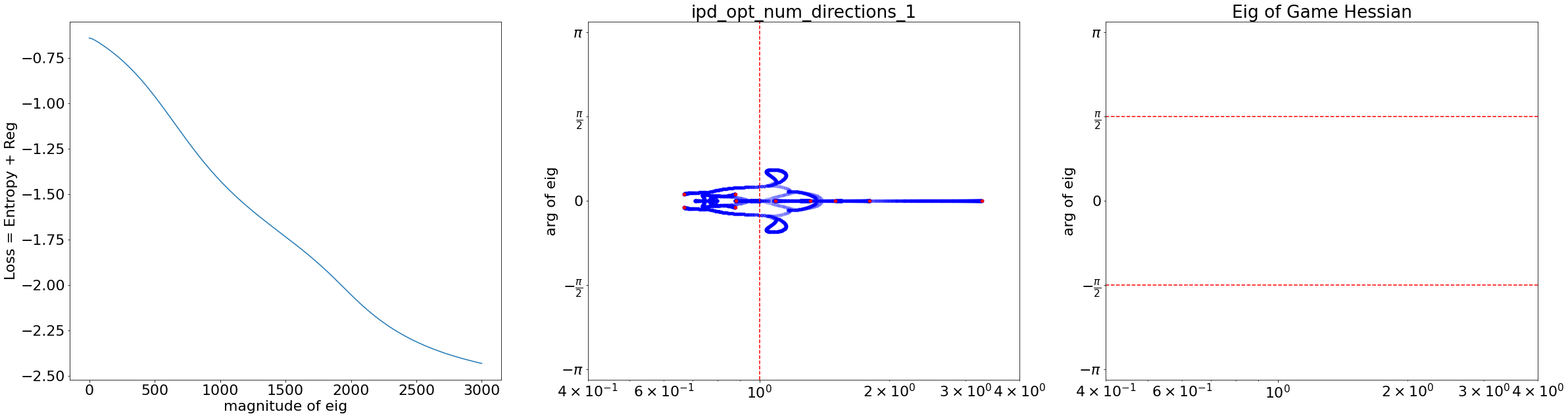}};
                        \node[left=of img11, node distance=0cm, rotate=90, xshift=2.0cm, yshift=-.75cm, font=\color{black}] {\footnotesize loss $\loss_{{\color{red}1}}^{\textnormal{min}}(\bothParam_0)$= -{\color{red}Max} exp loss};
                        
                        \node [right=of img11, xshift=-.5cm](img12){\includegraphics[trim={30.0cm 1.85cm 27.5cm 1.15cm},clip, width=.44\linewidth]{latex/lyapunovForDiversity/images/experiments/ipd/tune_lyap_ipd_1dir.png}};
                        \node[left=of img12, node distance=0cm, rotate=90, xshift=2.0cm, yshift=-.75cm, font=\color{black}] {argument of EVal $\arg(\eigval)$};
                        \node[above=of img12, node distance=0cm, xshift=-.0cm, yshift=-1.15cm,font=\color{black}] {EVals of Jac. of fixed-point operator $\spectrum(\jacFixedPointOp)$};
                    
                    \node [below=of img11, yshift=1cm](img21){\includegraphics[trim={1.0cm 1.85cm 55.0cm 1.15cm},clip, width=.45\linewidth]{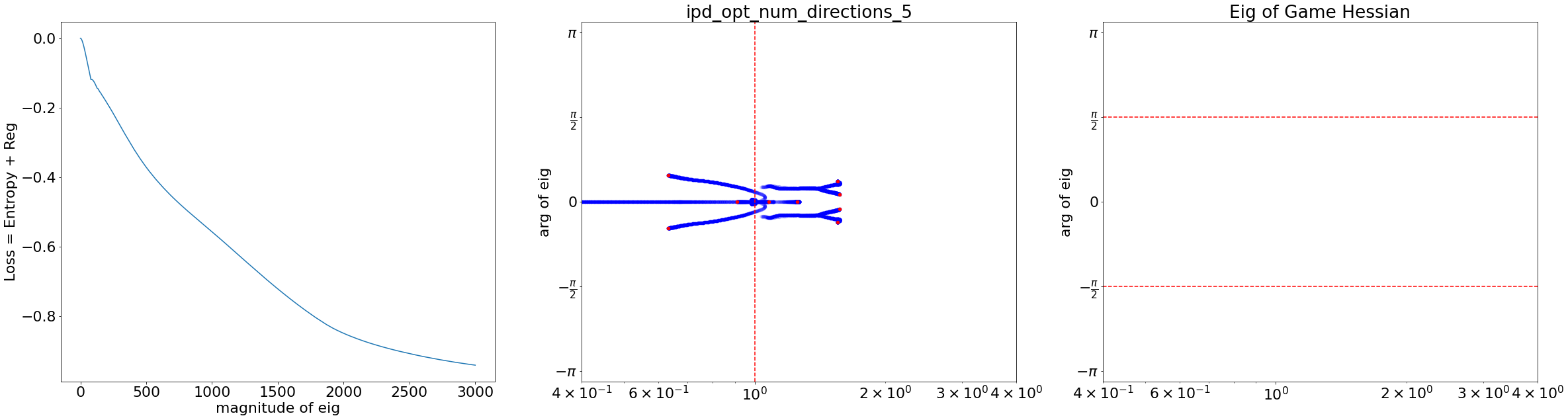}};
                        \node[left=of img21, node distance=0cm, rotate=90, xshift=2.25cm, yshift=-.75cm, font=\color{black}] {\footnotesize loss $\loss_{{\color{red}5}}^{\textnormal{min}}(\bothParam_0)$= -{\color{red}Min top 5} exps};
                        
                        \node [right=of img21, xshift=-.5cm](img22){\includegraphics[trim={29.5cm 1.85cm 27.5cm 1.15cm},clip, width=.44\linewidth]{latex/lyapunovForDiversity/images/experiments/ipd/tune_lyap_ipd_5dir.png}};
                        \node[left=of img22, node distance=0cm, rotate=90, xshift=2.0cm, yshift=-.75cm, font=\color{black}] {argument of EVal $\arg(\eigval)$};
                    
                    \node [below=of img21, yshift=1cm](img31){\includegraphics[trim={1.0cm 1.0cm 55.0cm 1.15cm},clip, width=.45\linewidth]{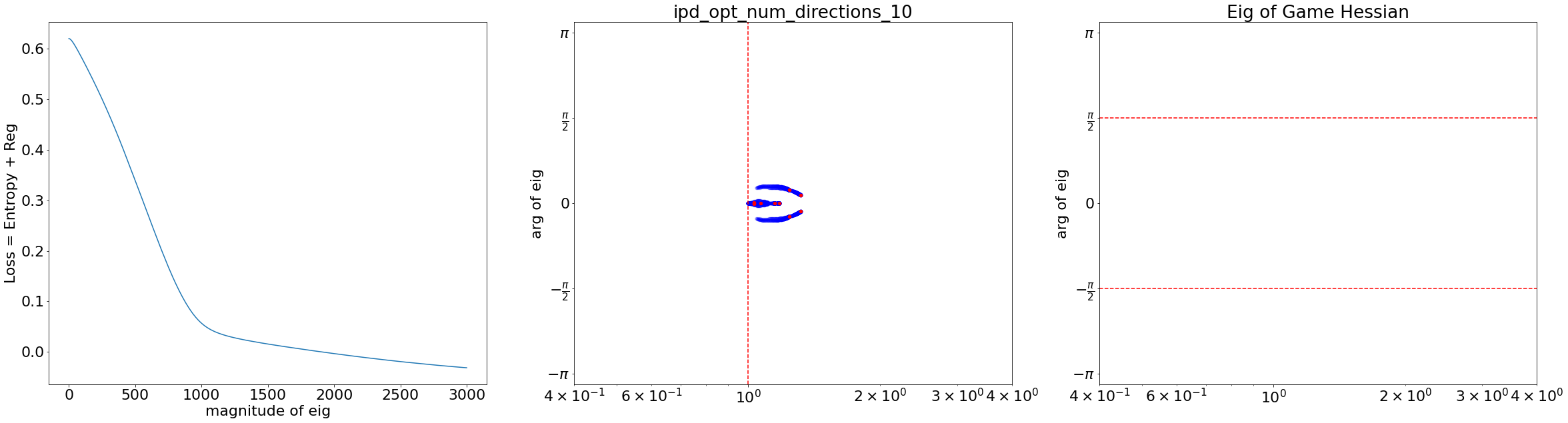}};
                        \node[left=of img31, node distance=0cm, rotate=90, xshift=2.5cm, yshift=-.75cm, font=\color{black}] {\footnotesize loss $\loss_{{\color{red}10}}^{\textnormal{min}}(\bothParam_0)$ = -{\color{red}Min top 10} exps};
                        \node[below=of img31, node distance=0cm, xshift=-.0cm, yshift=1.0cm,font=\color{black}] {optimization iteration};
                        
                        \node [right=of img31, xshift=-.5cm](img32){\includegraphics[trim={28.85cm 1.0cm 27.5cm 1.15cm},clip, width=.44\linewidth]{latex/lyapunovForDiversity/images/experiments/ipd/tune_lyap_ipd_10dir.png}};
                        \node[left=of img32, node distance=0cm, rotate=90, xshift=2.0cm, yshift=-.75cm, font=\color{black}] {argument of EVal $\arg(\eigval)$};
                        \node[below=of img32, node distance=0cm, xshift=0.0cm, yshift=1.0cm,font=\color{black}] {log-norm of EVal $\log(|\eigval|)$};
                \end{tikzpicture}
                \caption[Comparing optimization for various numbers of Lyapunov exponents]{
                    We compare the optimization for different losses using the local Lyapunov exponent with varying numbers of exponents.
                    \textbf{\emph{Takeaway:}} We can effectively optimize multiple exponents, which give trajectory separation in multiple directions.
                    Our objective is the minimum of the top $n$ local Lyapunov exponents, with a different $n$'s optimization trajectory shown in each row.
                    To review this visualization also in Figure~\ref{lyapunov_fig:tune_lyap_ipd}:
                    The spectrum is shown with a scatterplot in {\color{blue}blue} and a progressively larger alpha at each iteration.
                    The final spectrum is shown in {\color{red}red}.
                    For the Jacobian of the fixed-point operator $\jacFixedPointOp$, a vertical {\color{red}red} line is shown where the eigenvalue norm equals $1$, signifying the cutoff between (locally) convergent and divergent eigenspaces.
                    \emph{Top}: Optimizing the max local Lyapunov exponent from Figure~\ref{lyapunov_fig:tune_lyap_ipd}, where the largest eigenvalue is effectively maximized.
                    \emph{Middle}: Optimizing the top $5$ local exponents, where we find $5$ moderately divergent directions instead of $1$ extremely divergent direction as in the top.
                    \emph{Bottom}:  Optimizing the top $10$ local exponents results in trajectory separation in all directions.
                    We study different objectives of the top $n$ exponents in Appendix Figure~\ref{lyapunov_fig:tune_lyap_ipd_difObjectives}, and study non-local $k$-step exponents in Appendix Figure~\ref{lyapunov_fig:tune_lyap_ipd_difNumLyap}.
                }\label{lyapunov_fig:tune_lyap_ipd_moreDirections}
            \end{figure*}
            
            \begin{figure*}
                \centering
                \begin{tikzpicture}
                    \centering
                    \node (img11){\includegraphics[trim={1.0cm 1.85cm 55.0cm 1.15cm},clip, width=.46\linewidth]{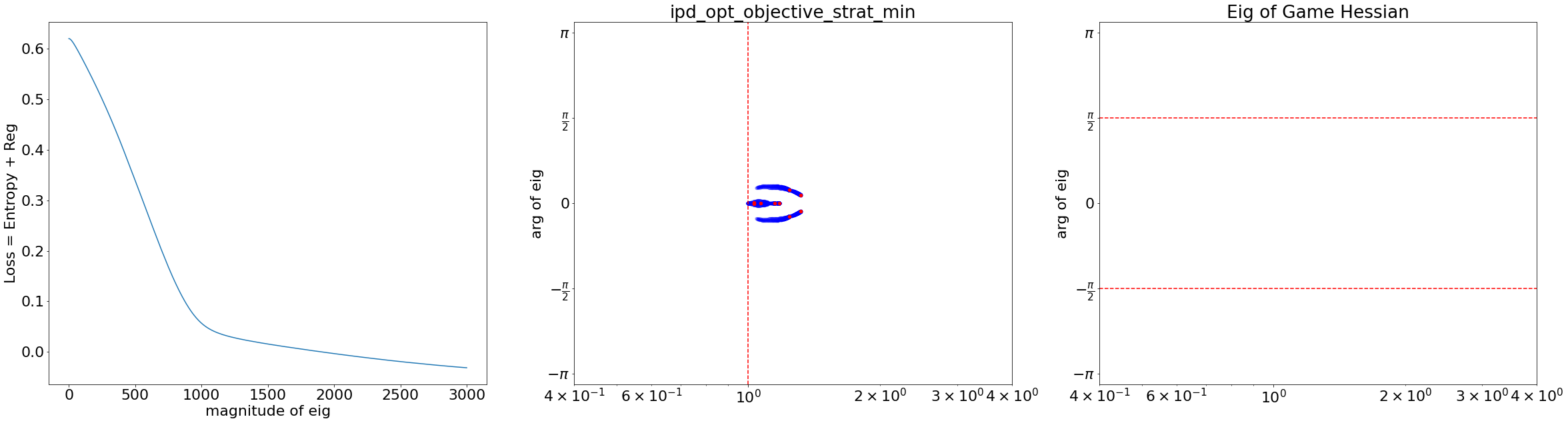}};
                        \node[left=of img11, node distance=0cm, rotate=90, xshift=2.25cm, yshift=-.9cm, font=\color{black}] {\footnotesize loss $\loss_{10}^{{\color{red}\textnormal{min}}}(\bothParam_0)$= -{\color{red}Min} top 10 exps};
                        
                        \node [right=of img11, xshift=-.5cm](img12){\includegraphics[trim={28.85cm 1.85cm 27.5cm 1.15cm},clip, width=.45\linewidth]{latex/lyapunovForDiversity/images/experiments/ipd/ipd_objective_strat/ipd_opt_objective_strat_min.png}};
                        \node[left=of img12, node distance=0cm, rotate=90, xshift=2.0cm, yshift=-.75cm, font=\color{black}] {argument of EVal $\arg(\eigval)$};
                        \node[above=of img12, node distance=0cm, xshift=-.0cm, yshift=-1.15cm,font=\color{black}] {EVals of Jac. of fixed-point operator $\spectrum(\jacFixedPointOp)$};

                    \node [below=of img11, yshift=1cm](img21){\includegraphics[trim={1.0cm 1.1cm 55.0cm 1.15cm},clip, width=.46\linewidth]{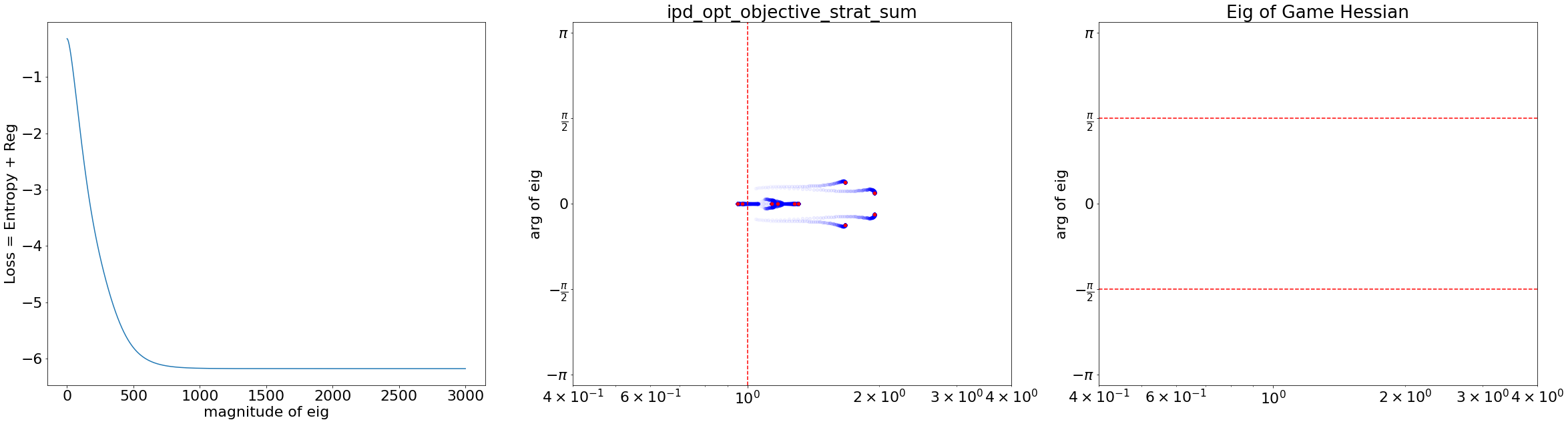}};
                        \node[left=of img21, node distance=0cm, rotate=90, xshift=2.5cm, yshift=-.9cm, font=\color{black}] {\footnotesize loss $\loss_{10}^{{\color{red}\textnormal{sum}}}(\bothParam_0)$= -{\color{red}Sum} top 10 exps};
                        \node[below=of img21, node distance=0cm, xshift=-.0cm, yshift=1.0cm,font=\color{black}] {optimization iteration};
                        
                        \node [right=of img21, xshift=-.5cm](img22){\includegraphics[trim={28.775cm 1.0cm 27.5cm 1.15cm},clip, width=.45\linewidth]{latex/lyapunovForDiversity/images/experiments/ipd/ipd_objective_strat/ipd_opt_objective_strat_sum.png}};
                        \node[left=of img22, node distance=0cm, rotate=90, xshift=2.0cm, yshift=-.75cm, font=\color{black}] {argument of EVal $\arg(\eigval)$};
                        \node[below=of img22, node distance=0cm, xshift=0.0cm, yshift=1.0cm,font=\color{black}] {log-norm of EVal $\log(|\eigval|)$};
                \end{tikzpicture}
                \caption[Comparing optimization for different Lyapunov exponent losses]{
                    We compare the optimization for different losses using the top $10$ local Lyapunov exponents.
                    \textbf{\emph{Takeaway:}} Using the minimum of the top exponents is a strong method to guarantee separation in many directions.
                    This visualizes the same aspects as Figure~\ref{lyapunov_fig:tune_lyap_ipd}:
                    The spectrum is shown with a scatterplot in {\color{blue}blue} and a progressively larger alpha at each iteration.
                    The final spectrum is shown in {\color{red}red}.
                    For the Jacobian of the fixed-point operator $\jacFixedPointOp$, a vertical {\color{red}red} line is shown where the eigenvalue norm equals $1$, signifying the cutoff between (locally) convergent and divergent eigenspaces.
                    \emph{Top:} To guarantee trajectory separation in all directions, we optimize the minimum of the top exponents.
                    At the end of the training, all eigenvalues of $\jacFixedPointOp$ are greater than $1$, signifying local trajectory separation in all directions.
                    \emph{Bottom:} We optimize the sum of the top $10$ exponents, which does not guarantee trajectory separation in every direction.
                    Our optimizer is happy with solutions diverging rapidly in the top directions while converging (slowly) in the bottom directions.
                }\label{lyapunov_fig:tune_lyap_ipd_difObjectives}
            \end{figure*}
            
            \begin{figure*}
                \centering
                \begin{tikzpicture}
                    \centering
                    \node (img){\includegraphics[trim={1.0cm 1.0cm 55.9cm 2.95cm},clip, width=.9\linewidth]{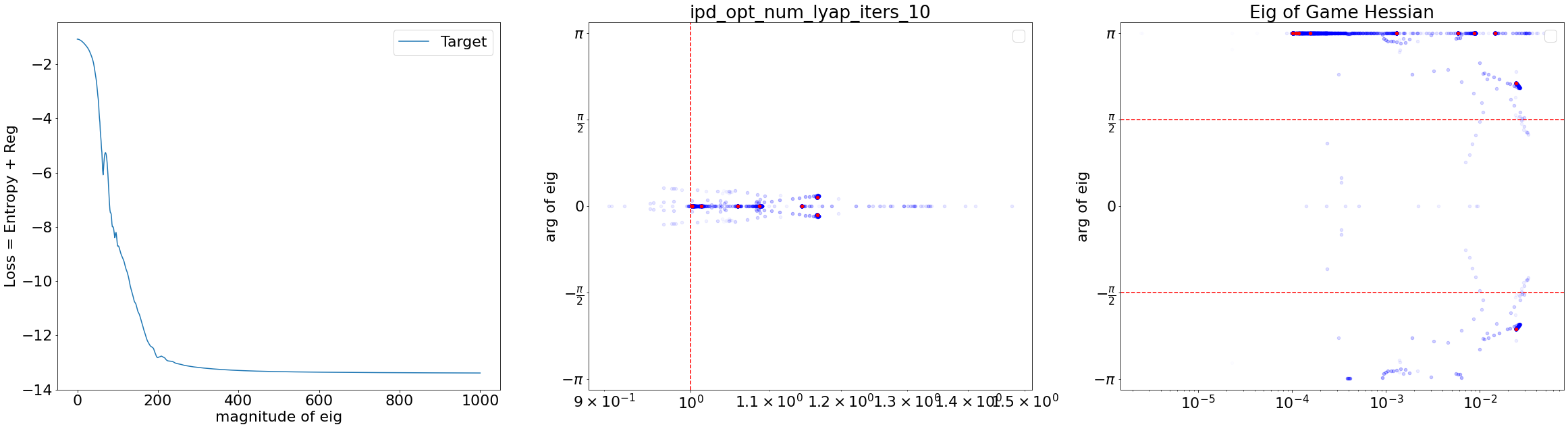}};
                    \node[left=of img, node distance=0cm, rotate=90, xshift=2.75cm, yshift=-.75cm, font=\color{black}] {{\color{red}10-step} loss $\loss^{\textnormal{min}}_{1}(\bothParam_0) = -\lyap_{{\color{red}10}}^{\textnormal{max}}(\bothParam_0)$};
                    \node[below=of img, node distance=0cm, xshift=-.0cm, yshift=1.0cm,font=\color{black}] {optimization iteration};
                \end{tikzpicture}
                \caption[Optimization with multi-step Lyapunov exponents]{
                    We investigate optimizing a loss of the max $k$-step Lyapunov exponent for $k > 1$.
                    \textbf{\emph{Takeaway:}} We effectively minimize multi-step exponents in higher-dimensional problems if required, extending the visualization in Figure~\ref{lyapunov_fig:tune_lyap_ipd}.
                }\label{lyapunov_fig:tune_lyap_ipd_difNumLyap}
            \end{figure*}

  \backmatter
  \bibliography{main}  
\end{document}